\documentclass{article}
\usepackage[T2A]{fontenc}
\usepackage{amsmath,amssymb}
\usepackage[english]{babel}
\usepackage[colorlinks, linkcolor=black]{hyperref}
\usepackage{natbib}
\usepackage[dvipsnames]{xcolor}
\usepackage{titlesec}
\usepackage{graphicx}
\usepackage{changepage}
\usepackage{booktabs}
\usepackage[top=3cm, bottom=3cm, right=3cm, left=3cm]{geometry}
\usepackage{caption}
\usepackage{subcaption}
\usepackage{tgadventor}
\usepackage[T1]{fontenc}

\makeatletter
\renewcommand{\maketitle}{\bgroup\setlength{\parindent}{0pt}
\begin{flushleft}
  \textbf{\LARGE\textcolor{Green}{\fontfamily{qag}\selectfont \@title}}
  
  \vspace{0.3cm}
  {\fontfamily{qag}\selectfont \large \@author}
  \vspace{0.3cm}
\end{flushleft}\egroup
}
\makeatother

\title{Kandinsky 5.0: A Family of Foundation Models  for Image and Video Generation}
\author{{\color{Green}\textbf{Kandinsky Lab}$^*$} \\
\small $^*$ A detailed list of the contributors can be found in the end of this paper.}

\titleformat{\section}
  {\normalfont\sffamily\Large\bfseries\color{Green}}
  {\thesection}{1em}{}
\titleformat{\subsection}
  {\normalfont\sffamily\large\bfseries\color{Green}}
  {\thesubsection}{1em}{}
\titleformat{\subsubsection}
  {\normalfont\sffamily\normalsize\bfseries\color{Green}}
  {\thesubsubsection}{1em}{}
  
\begin{document}


\maketitle

\begin{figure}[h!]
    \centering
    \includegraphics[width=\textwidth]{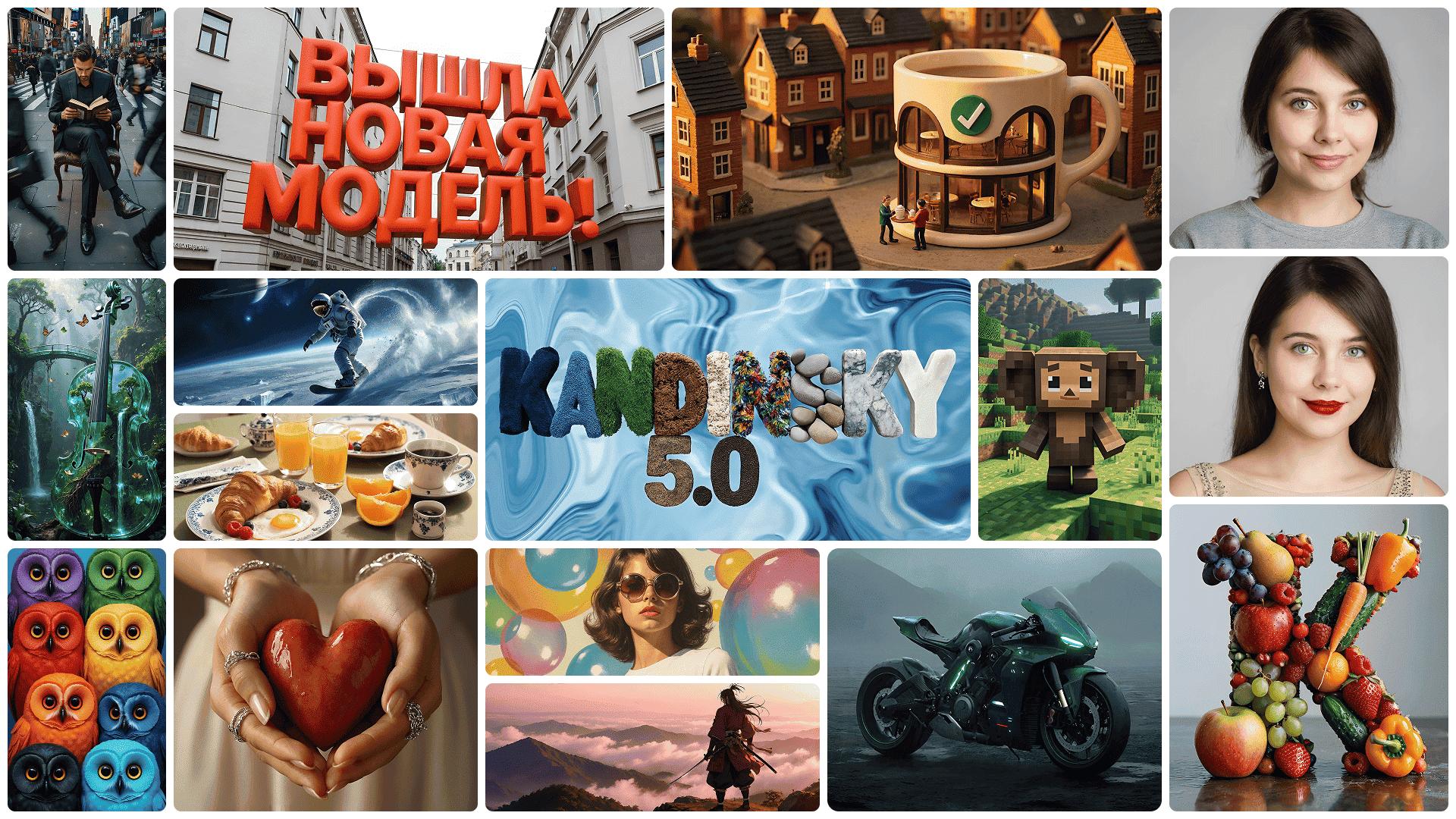}
\end{figure}

\noindent
\textbf{\textcolor{Green}{Abstract:}} This report introduces \textbf{Kandinsky 5.0}, a family of state-of-the-art foundation models for high-resolution image and 10-second video synthesis. The framework comprises three core line-up of models: \textbf{Kandinsky 5.0 Image Lite} -- a line-up of 6B parameter image generation models, \textbf{Kandinsky 5.0 Video Lite} -- a fast and lightweight 2B parameter text-to-video and image-to-video models, and \textbf{Kandinsky 5.0 Video Pro} -- 19B parameter models that achieves superior video generation quality. We provide a comprehensive review of the data curation lifecycle -- including collection, processing, filtering and clustering -- for the multi-stage training pipeline that involves extensive pre-training and incorporates quality-enhancement techniques such as self-supervised fine-tuning (SFT) and reinforcement learning (RL)-based post-training. We also present novel architectural, training, and inference optimizations that enable Kandinsky 5.0 to achieve high generation speeds and state-of-the-art performance across various tasks, as demonstrated by human evaluation. As a large-scale, publicly available generative framework, Kandinsky 5.0 leverages the full potential of its pre-training and subsequent stages to be adapted for a wide range of generative applications. We hope that this report, together with the release of our open-source code and training checkpoints, will substantially advance the development and accessibility of high-quality generative models for the research community.

\noindent
\textbf{\textcolor{Green}{Date:}} November 20, 2025


\noindent
\textbf{\textcolor{Green}{Website: }} \hyperlink {https://kandinskylab.ai/}{https://kandinskylab.ai/}

\noindent
\textbf{\textcolor{Green}{GitHub:}} \hyperlink {https://github.com/kandinskylab/kandinsky-5}{https://github.com/kandinskylab/kandinsky-5}

\noindent
\textbf{\textcolor{Green}{Hugging Face:}} \hyperlink {https://huggingface.co/kandinskylab}{https://huggingface.co/kandinskylab}


\newpage
\tableofcontents

\newpage

\vspace*{2cm}

\section[Introduction]{\color{Green}Introduction}\label{sec:introduction}

Over the past few years, diffusion models~\cite{ho2020ddpm, song2021scorebased} and the subsequent flow matching approaches~\cite{lipman2023flow} have led to a qualitative breakthrough in image generation, achieving unprecedented synthesis quality and diversity. This foundation enabled the rapid development of commercial and open-source systems that provide users with a wide range of generative capabilities, from text-to-image (T2I) synthesis to complex editing~\cite{Rombach_2022_CVPR, Midjourney, pika, vladimir-etal-2024-kandinsky}. To date, image generation models have not only achieved high quality but continue to improve actively, constantly raising the bar for realism and controllability, as demonstrated by models such as Stable Diffusion 3~\cite{esser2024scalingrectifiedflowtransformers}, Flux~\cite{flux2024}, Seedream 3 \& 4~\cite{gao2025seedream30technicalreport, seedream2025seedream40nextgenerationmultimodal}, and Hunyuan Image 3~\cite{cao2025hunyuanimage}.

A natural extension of this progress has been the growing interest in video generation, leading to numerous methods that adapt and extend architectures proven successful for images~\cite{ho2022imagenvideohighdefinition, blattmann2023align, arkhipkin2023fusionframesefficientarchitecturalaspects, 10815947}. However, the direct translation of these approaches faced fundamental scalability issues due to the exponential growth in computational complexity when working with time-dependent three-dimensional video data. Partial resolution of these limitations was achieved through the active adoption of architectures like the Diffusion Transformer (DiT)~\cite{peebles2023dit}, which provided the necessary scalability and efficiency~\cite{chen2023pixartalpha, ma2025latte}, along with a series of modifications to attention mechanisms aimed at handling video data~\cite{zhang2024fast, xi2025sparsevideogenacceleratingvideo, xia2025trainingfreeadaptivesparseattention}.

Today, a number of video generation models demonstrate a high level of quality, such as Sora~\cite{openai2024sora, openai2024sora2} and Veo~\cite{Veo}. A significant part of this progress is driven by open-source initiatives. Projects such as HunyuanVideo~\cite{kong2025hunyuanvideosystematicframeworklarge}, Mochi~\cite{genmo2024}, CogVideoX~\cite{yang2025cogvideox}, Wan~\cite{wan2025} and VACE~\cite{VACE}, democratize access to foundational architectures and pre-trained weights, accelerating research and development, and demonstrating results close to professional-grade video production. All this opens up broad opportunities for the application of video models and lays the groundwork for creating multimedia generation systems~\cite{polyak2024movie}, ``world models''~\cite{genie3}, and foundational visual models, analogous in their significance to Large Language Models (LLMs) in Natural Language Processing (NLP)~\cite{grattafiori2024llama, liu2024deepseek, yang2025qwen3technicalreport}.

Despite the rapid development, critical challenges persist in video generation. Beyond processing vast amounts of data, creating such systems requires complex, multi-stage optimizations for both the training process and subsequent inference. Therefore, the efficient creation of high-quality, consistent, and controllable video remains one of the most challenging tasks in generative AI.

In this work, we aim to address some of the key challenges in the field of video generation. We present \textbf{Kandinsky 5.0} -- a family of fundamental generative models for high-resolution image and video synthesis, designed to achieve state-of-the-art quality and operational efficiency. The Kandinsky 5.0 suite comprises three line-up of models:

\begin{itemize}
    \item \textbf{Kandinsky 5.0 Video Pro:} A high-power 19B parameter models for text-to-video and image-to-video generation, creating up to 10-second videos at high resolution .

    \item \textbf{Kandinsky 5.0 Video Lite:} A lightweight 2B parameter model for text-to-video and image-to-video generation, producing up to 10-second clips.

    \item \textbf{Kandinsky 5.0 Image Lite:} A 6B parameter models for text-to-image generation and image editing at high resolution.
\end{itemize}

\begin{figure}[h!]
    \centering
    \includegraphics[width=\textwidth]{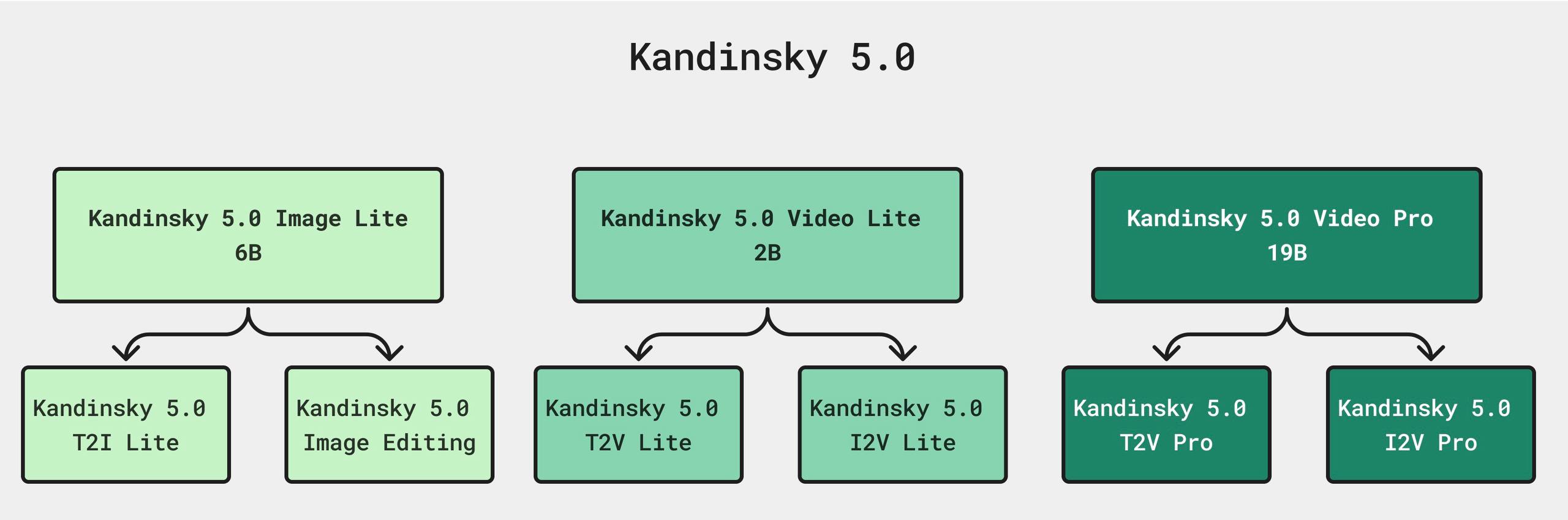}
    \caption{Kandinsky 5.0 Models Family}
    \label{fig:kandinsky5_all_models}
\end{figure}

\textbf{The key contributions} of this technical report are as follows:

\begin{enumerate}
    \item We provide a comprehensive description of the data collection and processing pipeline, including data preparation for instructive image editing tuning and self-supervised fine-tuning (SFT) for both video and image modalities.
    
    \item We detail the multi-stage training pipeline for all six models, encompassing a pretraining phase for learning general patterns of the visual world and SFT for enhancing visual quality. We also introduce our RLHF post-training adversarial method based on comparing generated images with those from the SFT dataset. This approach achieves superior realism, visual quality, and prompt alignment.
    
    \item We present the architecture of our core CrossDiT model, featuring our key optimization of the attention mechanism for high-resolution video (exceeding $512$ px) with durations longer than 5 seconds via the NABLA method~\cite{mikhailov2025nablanablaneighborhoodadaptiveblocklevel}. This overcomes the quadratic complexity of standard spatio-temporal attention, achieving a $2.7\times$ reduction in training and inference time with $90\%$ sparsity ratio while maintaining generated video quality, as confirmed by FVD~\cite{unterthiner2019accurategenerativemodelsvideo}, VBench~\cite{huang2024vbench}, CLIP-score~\cite{hessel2021clipscore} and human evaluation through side-by-side testing.
    
    \item We describe multiple optimizations implemented across the pipeline to accelerate inference, training, and reduce memory consumption. These techniques include variational autoencoder (VAE) optimization, text encoder quantization, and CrossDiT training optimizations using Fully or Hybrid Sharded Data Parallel (F/HSDP) \cite{zhao2023pytorchfsdpexperiencesscaling}, activation checkpointing~\cite{activation_checkpointing}, among others.
    
    \item For video model distillation, we employ a combined approach that integrates Classifier-Free Guidance Distillation~\cite{meng2023distillationguideddiffusionmodels}, Trajectory Segmented Consistency Distillation (TSCD)~\cite{ren2024hypersdtrajectorysegmentedconsistency}, and subsequent adversarial post-training~\cite{seawead2025seaweed7bcosteffectivetrainingvideo} to enhance visual quality. This reduces the Number of Function Evaluations (NFE) from 100 to 16 while preserving visual quality, as evidenced by side-by-side human evaluation results.
    
    \item We evaluate our final models against several state-of-the-art approaches and demonstrate superior video generation quality through human evaluation on a prompt set from MovieGen~\cite{polyak2025moviegencastmedia}.
    
    \item Finally, we open-source the code and weights for all models from various training stages, and provide access through the \texttt{diffusers} library.
\end{enumerate}

\newpage
\section[Report Overview]{\color{Green}Report Overview}

The report is structured to provide a comprehensive understanding of the model's design, training, and evaluation:

\begin{itemize}
    \item \textbf{Section~\ref{sec:background}: Background: The Evolution of Kandinsky models.} Traces the history of the Kandinsky model family, from early autoregression-based models to the current latest version of Kandinsky 5.0.
    
    \item \textbf{Section~\ref{sec:data_processing_pipeline}: Data Processing Pipeline.} Describes the large-scale, multi-stage pipeline used for curating and annotating datasets for text-to-image and text-to-video pretraining, self-supervised fine-tuning, image instruction tuning, and the collection of Russian-specific multicultural data. We emphasize quality control and scalability in our approach.
    
    \item \textbf{Section~\ref{sec:base_model_architecture}: Kandinsky 5.0 Architecture.} Presents the architecture of the Kandinsky 5.0 models, which is common for all models in this family. The core components include a Cross-Attention Diffusion Transformer (\textbf{CrossDiT}), a corresponding \textbf{CrossDiT-block} scheme, and the Neighborhood Adaptive Block-Level Attention (\textbf{NABLA}) mechanism, which is essential for optimizing both training and inference.

    \item \textbf{Section~\ref{sec:training}: Training Stages.} Outlines the multi-phase training process, from pre-training on large-scale datasets to self-supervised fine-tuning, distillation, and RL-based post-training, tailored for both image and video models.

    \item \textbf{Section~\ref{sec:optimization}: Optimizations.} Covers techniques such as VAE encoder acceleration, CrossDiT training optimization and efficient use of GPU memory.
    
    \item \textbf{Section~\ref{sec:evaluation}: Results.} Presents a growth of visual quality at different training stages and human side-by-side (SBS) evaluations, demonstrating superior performance in motion consistency, visual quality, and prompt alignment compared to existing models.
    
    \item \textbf{Section~\ref{sec:use_cases}: Use Cases.} Highlights practical applications across text-to-image, image editing, text-to-video, and image-to-video generation, supported by visual examples and technical prompts.

    \item \textbf{Section~\ref{sec:related_work}: Related Work.} Contextualizes Kandinsky 5.0 within the broader landscape of generative models, covering advancements in text-to-image and text-to-video generation, distillation, post-training techniques, and evaluation methodologies for generative models.

    \item \textbf{Section~\ref{sec:limitations}: Limitations and Further Work.} Discusses remaining challenges, that guide future research directions.
    
    \item \textbf{Section~\ref{sec:ethics}: Border Impacts and Ethical Considerations.} Details the responsible AI framework implemented, including data curation, runtime safeguards, and ethical use guidelines to ensure safe deployment.
    
    \item \textbf{Sections~\ref{sec:conclusion}--\ref{sec:contributors}: Conclusion, Contributors \& Acknowledgments.} Summarizes contributions and acknowledges the teams and collaborators involved.
\end{itemize}

\newpage
\section[Background: The Evolution of Kandinsky models]{Background: The Evolution of Kandinsky models}\label{sec:background}

\begin{figure}[h!]
    \centering
    \includegraphics[width=\textwidth]{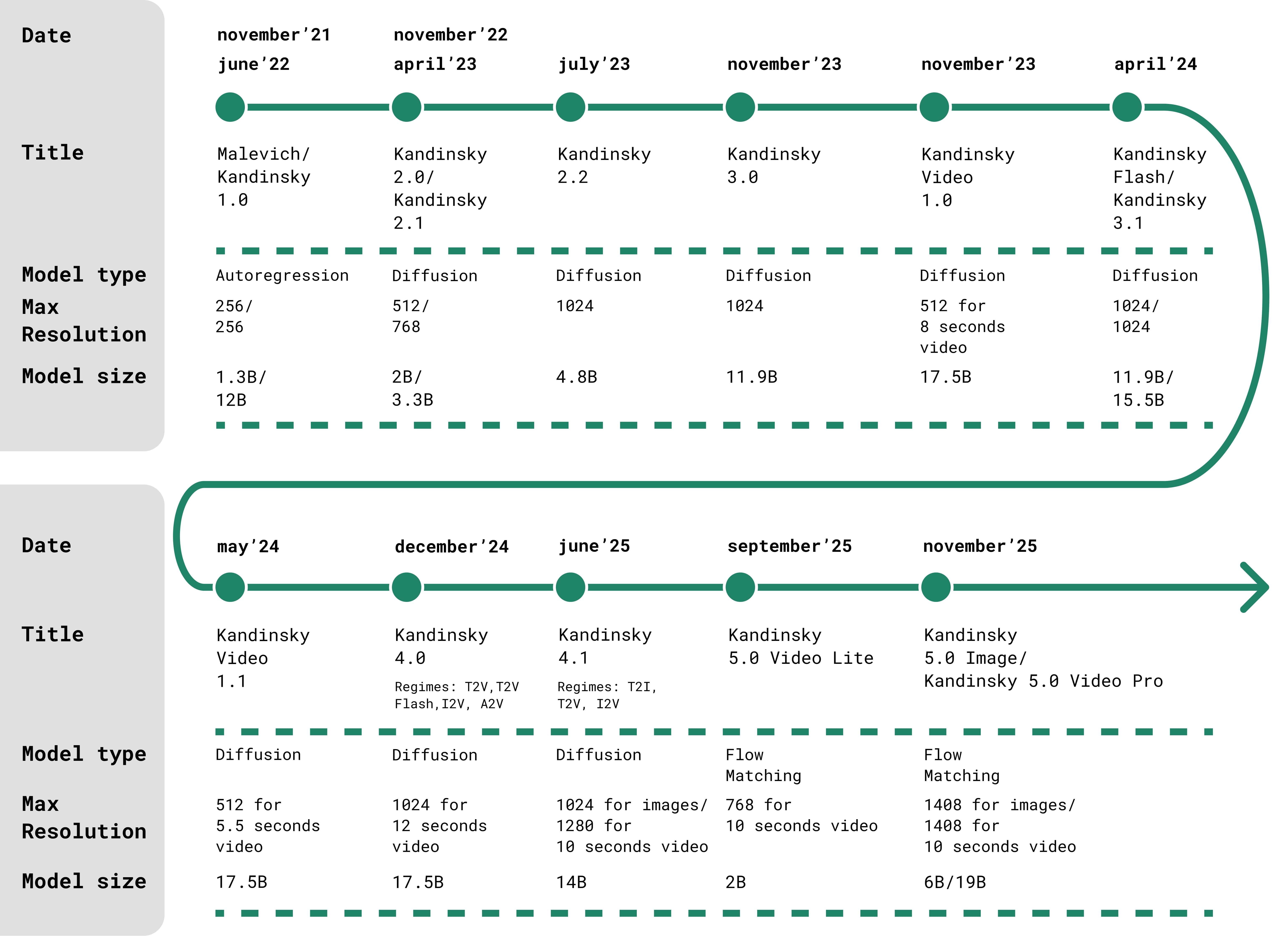}
    \caption{The evolution of Kandinsky models.}
    \label{fig:kandinsky_family}
\end{figure}

The Kandinsky family of visual generative models, named after Russian abstract artist Wassily Kandinsky (1866 -- 1944), has evolved significantly since its inception (Figure~\ref{fig:kandinsky_family}). Its history began in June 2022 with the release of \textbf{Kandinsky 1.0}, a 12B-parameter model. This model was an enlarged version of the earlier \textbf{Malevich}\footnote{\url{https://github.com/ai-forever/ru-dalle}}$^,$\footnote{\url{https://huggingface.co/ai-forever/rudalle-Malevich}} model (aka ruDALL-E XL), which had 1.3B parameters, was inspired by DALL-E~\cite{ramesh2021zeroshottexttoimagegeneration}, and was released in November 2021. Compared to Malevich model, Kandinsky 1.0 featured more layers, increased hidden space, and was trained on 120 million text-image pairs. Both models utilized an autoregressive architecture to generate 256$\times$256 resolution images.

A radical shift occurred in November 2022 with \textbf{Kandinsky 2.0}\footnote{\url{https://github.com/ai-forever/Kandinsky-2}}$^,$\footnote{\url{https://huggingface.co/ai-forever/Kandinsky_2.0}}~\cite{razzhigaev-etal-2023-kandinsky}, the family's first diffusion model. To ensure multilingual capability, it employed two encoders -- XLMR-clip~\cite{carlsson-EtAl:2022:LREC} and mT5-small~\cite{xue2021mt5massivelymultilingualpretrained} -- enabling support for queries in 101 languages. Trained on one billion text-image pairs, the 2B-parameter model could generate images at a 512$\times$512 resolution.

The next stage, \textbf{Kandinsky 2.1}\footnote{\url{https://huggingface.co/ai-forever/Kandinsky_2.1}} (April 2023), brought major architectural improvements: a single XLM-Roberta-Large-Vit-L-14\footnote{\url{XLM-Roberta-Large-Vit-L-14}} text encoder, a MoVQ image autoencoder~\cite{zheng2022movqmodulatingquantizedvectors}, and the addition of an image prior model for better text-image alignment. The diffusion mechanism was refined to use CLIP visual embeddings~\cite{radford2021clip}. After fine-tuning on an additional 170 million text-image pairs, the model surpassed DALL-E 2~\cite{ramesh2022hierarchicaltextconditionalimagegeneration}, GigaGAN~\cite{kang2023scalingganstexttoimagesynthesis}, and Stable Diffusion 2.1~\cite{Rombach_2022_CVPR} on the FID metric~\cite{heusel2018ganstrainedtimescaleupdate} on the COCO 30k dataset~\cite{li2024recapdatacomp}. The model has 3.3B parameters, a resolution of 768$\times$768, and natively supports inpainting, outpainting, image blending, synthesis of variations of an input image, and text-guided image editing.

In July 2023, \textbf{Kandinsky 2.2}\footnote{\url{https://huggingface.co/docs/diffusers/api/pipelines/kandinsky_v22}} (4.8B) further enhanced photorealism and increased resolution to 1024 pixels with support for various aspect ratios. Other innovations included  the introduction of a ControlNet mechanism~\cite{zhang2023adding} for local editing, and the capability to generate 4-second animated clips.

November 2023 saw the release of two major models. \textbf{Kandinsky 3.0}\footnote{\url{https://github.com/ai-forever/Kandinsky-3}}$^,$\footnote{\url{https://huggingface.co/docs/diffusers/api/pipelines/kandinsky3}} (11.9B)~\cite{arkhipkin2024kandinsky30technicalreport}, which focused on precise text-image alignment, used the FLAN-UL2 language encoder. Released in parallel, \textbf{Kandinsky Video 1.0}\footnote{\url{https://github.com/ai-forever/KandinskyVideo}} (17.5B)~\cite{arkhipkin2023fusionframesefficientarchitecturalaspects} was a state-of-the-art, open-source video generation model based on Kandinsky 3.0. It operated in two stages: generating keyframes and then interpolating between them. Generating an 8-second video at 512$\times$512 resolution took approximately three minutes, despite the model being trained on only 220,000 text-video pairs.

The line-up expanded again in April 2024 with two models. \textbf{Kandinsky Flash} (11.9B) was a distilled version of Kandinsky 3.0 that reduced the number of generation steps from 50 to 4 using the Adversarial Diffusion Distillation approach~\cite{sauer2023adversarialdiffusiondistillation}. \textbf{Kandinsky 3.1} (15.5B)~\cite{vladimir-etal-2024-kandinsky} combined Kandinsky 3.0 and Kandinsky Flash, using the latter as a refiner on the final steps in the reverse diffusion process, which significantly enhanced visual quality. This version also incorporated improvements to multicultural awareness, particularly for the Russian-culture domain~\cite{Vasilev_2024, vasilev-etal-2025-ruscode}. The model supports image variation, image blending, image-and-text blending, ControlNet-based editing, and inpainting. Additionally, we trained a state-of-the-art super-resolution model \textbf{KandiSuperRes}\footnote{\url{https://github.com/ai-forever/KandiSuperRes}} on the base Kandinsky 3.0 model.

A video model update, \textbf{Kandinsky Video 1.1}~\cite{10815947}, followed in May 2024. It used Kandinsky 3.0 for first frame generation and was trained on an enlarged dataset of 4.6 million text-video pairs. The model generates 5.5-second videos, and its dataset preparation leveraged automatic captioning via LLaVA-1.5~\cite{liu2023visualinstructiontuning}.

\textbf{Kandinsky 4.0}\footnote{\url{https://github.com/ai-forever/Kandinsky-4}} (17.5B) was released in December 2024 as the family's first Diffusion Transformer model. Using an MMDiT-like architecture~\cite{esser2024scalingrectifiedflowtransformers}, it generates both images and 12-second videos from text and images. In VBench benchmarks~\cite{huang2024vbench} and human evaluations, Kandinsky 4.0 demonstrated superior results against models like CogVideoX-1.5~\cite{yang2025cogvideox}, Open-Sora-Plan v1.3~\cite{lin2024opensoraplanopensourcelarge}, Mochi v1.0~\cite{genmo2024}, and Pyramid Flow~\cite{jin2024pyramidal}. This version also supported audio generation from an input video clip. Its refinement, \textbf{Kandinsky 4.1} (14B, June 2025), adopted a DiT architecture for image generation and underwent a Supervised Fine-Tuning (SFT) stage using data manually selected by a team of expert artists to enhance aesthetics.

The pinnacle of development to date is the \textbf{Kandinsky 5.0}\footnote{\url{https://github.com/ai-forever/Kandinsky-5}} model family described in this report. This is the first Kandinsky models based on the Flow Matching~\cite{lipman2023flow}, comprising six models of different sizes for various high-quality image and video generation tasks.

\section[Data Processing Pipeline]{\color{Green}Data Processing Pipeline}\label{sec:data_processing_pipeline}

The training of the video generation model leverages multiple datasets across different training stages. The primary datasets for pretraining are large Kandinsky T2I and Kandinsky T2V, while tiny Kandinsky SFT dataset is utilized during fine-tuning stages to significantly boost visual quality. The distinct Kandinsky RCC dataset is used to improve culturally specific video generation capabilities. We also invent a comprehensive routine to collect Kandinsky I2I (instruct dataset) allowing us to train a precise image editing model. Large-scale data collection and an efficient training process are impossible without properly installed and configured infrastructure.

\subsection{Data Processing Infrastructure}
A core part of the data processing pipeline is a database, which performs as a metadata storage and as a task broker for a distributed network of data processors.
This database has several indexes (including vector ones) that allow quick retrieval of required data for each training stage and prevent duplicates from appearing in the dataset.
Table partitioning and load-balancing are used to reduce the load on the database, allowing thousands of data processors to work with it simultaneously. 
Data processors run on different hardware and can utilize CPU and GPU resources for different tasks.
\subsection{Text-to-Image Dataset Processing}
The Kandinsky T2I dataset is a large-scale collection of more than 500 million general-domain images, designed to support the pretraining and fine-tuning stages of the Kandinsky model. These images originate from a diverse set of sources, including prominent open datasets (e.g., LAION \cite{schuhmann2022laion}, COYO\footnote{\url{https://github.com/kakaobrain/coyo-dataset}}) and large online image repositories. A meticulous data processing pipeline ensures the dataset's quality and suitability for its intended use.
\begin{figure}[tb]
    \centering
    \includegraphics[width=\textwidth]{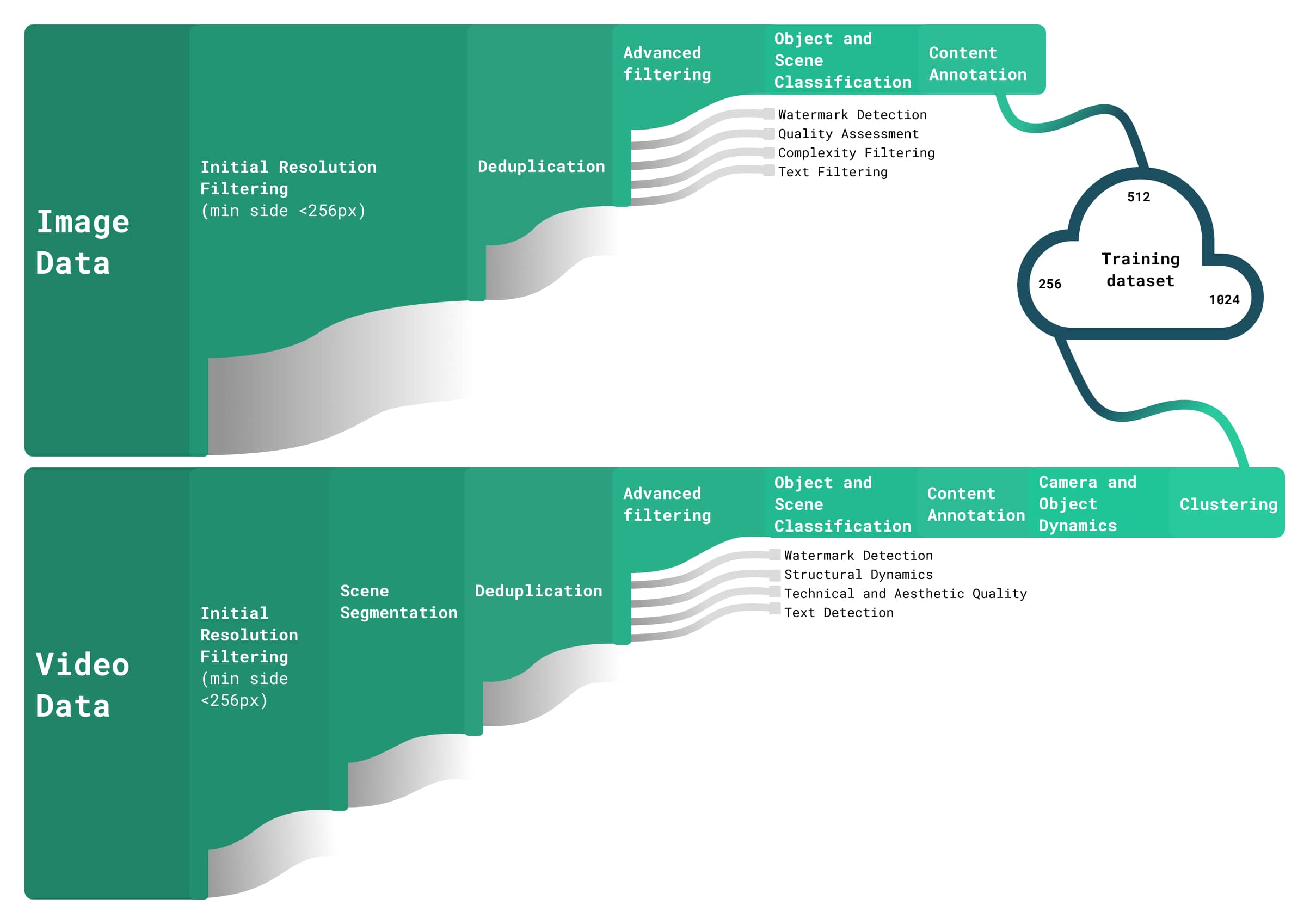}
    \caption{Data processing pipeline for Kandinsky T2V (text-to-video) and Kandinsky T2I (text-to-image) datasets. The workflow begins with raw image and video data, followed by initial filtering and deduplication it processed through advanced filtering (including watermark detection, quality assessment, complexity and text filtering), classification and content annotation stages. Final processed data is stored grouped by resolution (256, 512, and 1024 minimal side lengths) to use in correspondent pretrain stage.}
    \label{fig:data_workflow}
\end{figure}
\subsubsection*{Processing Pipeline}
The processing pipeline for the Kandinsky T2I dataset comprises the following several sequential and filtering stages:

\begin{itemize}
    \item \textbf{Initial Resolution Filtering:} Images with a shorter side measuring fewer than 256 pixels are discarded. This step ensures a baseline level of visual detail.

    \item \textbf{Deduplication:} To eliminate redundant content, an image perceptual hash \cite{zauner2010perceptualhashing} is calculated for each image. This technique identifies and removes exact duplicates and images that are visually very similar.

    \item \textbf{Advanced Filtering:} A series of models are applied to assess image quality and content, filtering out undesirable data:
    \begin{itemize}
        \item \textit{Watermark Detection:} Watermark detection is performed using a combination of two models. The system employs the \texttt{watermark\_resnext101\_32x8d-large} classifier for perceptual analysis alongside a YOLO-based detector that locates watermark-like objects. Images flagged by either model with a confidence score above a defined threshold are filtered out.
        \item \textit{Quality Assessment:} Models predict both technical and aesthetic quality to prioritize visually appealing and well-constructed images:
        \begin{itemize}
            \item The TOPIQ model \cite{tu2023topiq} provides separate scores for technical quality and aesthetic quality.
            \item The Q-Align model \cite{wu2023qalign} offers an alternative assessment of technical and aesthetic aspects.
        \end{itemize}
        \item \textit{Text Filtering:} The CRAFT text detection model \cite{baek2019craft} identifies regions of text within images. Images containing an excessive amount of text, based on the number of text boxes or the total text area, are excluded to avoid bias towards heavily annotated or subtitle-rich content.
        \item \textit{Complexity Filtering:} Visual complexity is assessed using a combined approach where the SAM 2 model \cite{ravi2024sam2} generates segmentation masks, complemented by a Sobel filter for detailed edge analysis. This pipeline effectively filters out overly simple images (e.g., plain backgrounds) based on quantified complexity metrics.
        \item \textit{Object and Scene Classification:} To enrich metadata and enable conditional generation or analysis, models classify image content:
        \begin{itemize}
            \item The YOLOv8 model\footnote{\url{https://github.com/ultralytics/ultralytics}}, trained on OpenImagesV7 \cite{kuznetsova2020openimages}, detects and classifies objects present in the image.
            \item A CLIP-based classifier \cite{radford2021clip} categorizes the image's location, style, main subject, and detailed place type based on CLIP embeddings.
        \end{itemize}
    \end{itemize}

    \item \textbf{Content Annotation (Synthetic Captions):} High-quality synthetic English captions are generated for the images using powerful multimodal models to provide textual descriptions for training:
    \begin{itemize}
        \item \textit{InternVL2-26B \cite{chen2025expandingperformanceboundariesopensource}:} Generates initial detailed captions. Variants of this caption, refined or shortened by the InternLM3-8B model \cite{cai2024internlm2technicalreport}, are also included.
        \item \textit{Qwen2.5VL-32B \cite{bai2025qwen25vltechnicalreport}:} Generates Russian captions for higher-resolution images ($width * height \geq 512^2$).
    \end{itemize}
    Post-processing is applied to these synthetic captions to improve their quality and consistency:
    \begin{itemize}
        \item Regular expressions clean common introductory phrases (e.g., "The image shows").
        \item Further filtering removes sentences containing non-English characters (e.g., Cyrillic, Chinese) to maintain language purity in English captions. This mitigates OCR errors from InternVL2, which performs poorly on scripts like Cyrillic (e.g., Russian), ensuring higher quality captions.
    \end{itemize}

    \item \textbf{Organization and Storage:} Processed images, along with their metadata, filter scores, and captions, are stored in Parquet files. These files are organized into subdirectories based on shortest image side (256, 512, and 1024) and source dataset name on the specified S3 storage endpoint.
\end{itemize}
 
\begin{figure}[tb]
    \centering
    \includegraphics[width=\textwidth]{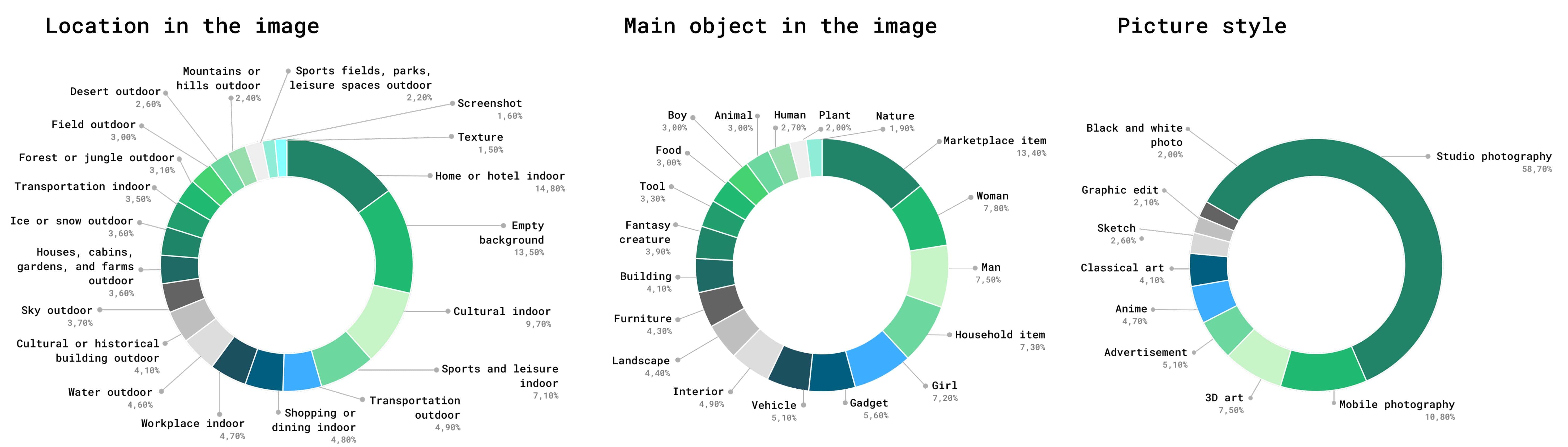} 
    \caption{Distribution of key data categories across the curated Kandinsky T2I dataset by Location, Main object and Picture style.}
    \label{fig:T2I_distribution}
\end{figure}

\subsection{Image Editing Instruct Dataset Processing}
The image editing instruct dataset (Kandinsky I2I) was constructed through a sophisticated multi-stage pipeline designed to identify high-quality image pairs suitable for editing tasks and to accomplish it with precise text instructions.

\begin{figure}[htb]
    \centering
    \includegraphics[width=\textwidth]{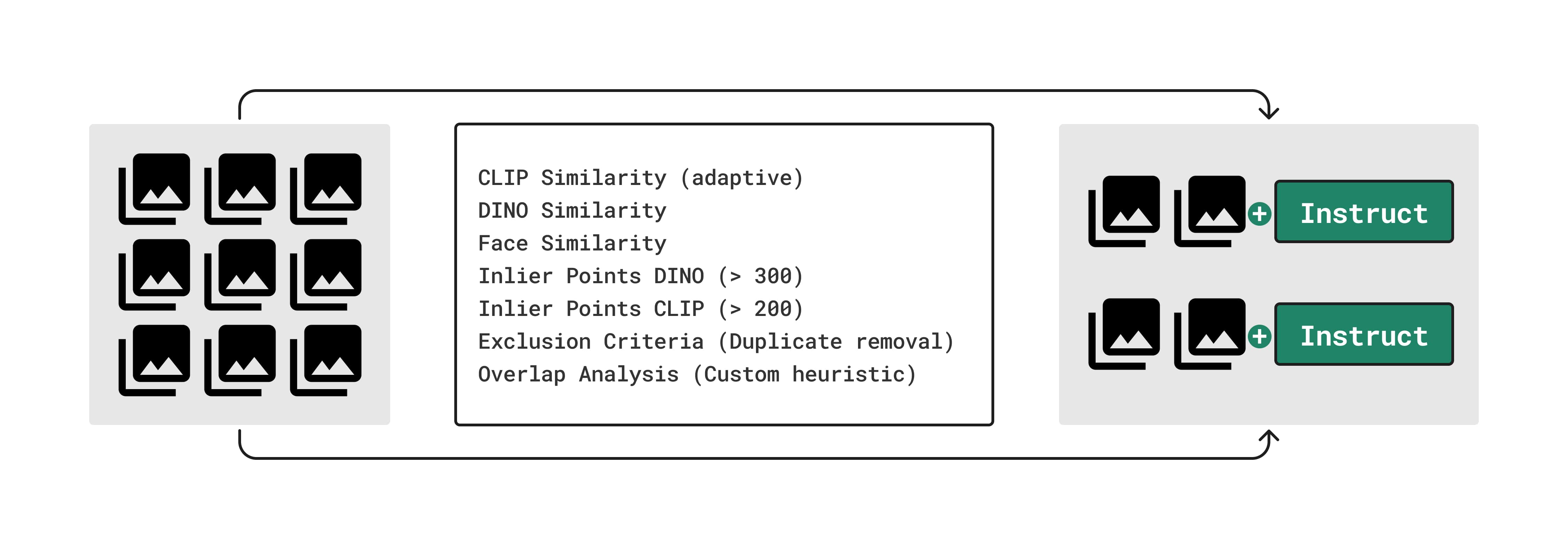} 
    \caption{Instructive dataset processing pipeline. Initial image data (left) is processed through a set of similarity and exclusion criteria — including CLIP and DINO embeddings, face similarity, duplicate removal, and custom overlap heuristics — to produce filtered image-instruction pairs (right). Each retained image is paired with an \texttt{Instruct} token, forming high-quality instruction-tuned training samples.}
    \label{fig:instructive_pipeline}
\end{figure}

\subsubsection*{Processing Pipeline}
The procedure involved the following steps:

\begin{enumerate}
    \item \textbf{Initial Image Collection:} A diverse set of about 240 million images was compiled from various sources, ensuring broad coverage of different visual content, styles, and subjects.

    \item \textbf{Similarity Matching:} To identify potential editing pairs, we employed multiple similarity metrics:
    \begin{itemize}
        \item \textbf{CLIP Score \cite{radford2021clip}:} Measured semantic similarity between images using CLIP embeddings
        \item \textbf{DINO Score \cite{oquab2023dinov2}:} Computed visual similarity using DINOv2 features
        \item \textbf{Face Recognition:} Specifically applied to images containing faces (only single-face images considered)
    \end{itemize}

    \item \textbf{Adaptive Thresholding:} Implemented an adaptive thresholding approach for CLIP similarity:
    \begin{itemize}
        \item Images were clustered into 10,000 groups
        \item Thresholds were determined per cluster based on top similarity scores
        \item Final threshold: $\text{clip\_sim} > (1-T)*\text{clip\_thr} + T$ where $T = 0.15$
    \end{itemize}

    \item \textbf{Geometric Verification:} Potential pairs underwent rigorous geometric verification using LoFTR \cite{sun2021loftr} for feature matching:
    \begin{itemize}
        \item Extracted feature points and matches between image pairs
        \item Applied RANSAC algorithm \cite{fischler1981ransac} to estimate fundamental matrix and Euclidean transformation
        \item Iteratively identified inlier groups (minimum 20 points per group)
        \item Calculated total inliers as the sum of matched points across all valid groups
    \end{itemize}

    \item \textbf{Quality Filtering:} Applied multiple filtering criteria to ensure high-quality pairs:
    \begin{itemize}
        \item \textbf{DINO inliers:} $\text{dino\_sim} > 0.8$ AND $\text{inliers} > 300$
        \item \textbf{CLIP inliers:} $\text{clip\_sim} > 0.8$ AND $\text{inliers} > 200$ AND adaptive threshold condition
        \item \textbf{Face similarity:} $\text{face\_sim} > 0.7$ for images containing faces
        \item \textbf{Exclusion criteria:} Removed pairs with $\text{dino\_sim} > 0.97$ OR $\text{dino\_aligned\_sim} > 0.97$ OR ($\text{face\_sim} < 0.5$ AND $\text{face\_sim} > 0$)
    \end{itemize}

    \item \textbf{Aligned DINO Score Calculation:} For additional verification:
    \begin{itemize}
        \item Used LoFTR \cite{sun2021loftr} to find matching points between images
        \item Applied RANSAC \cite{fischler1981ransac} to find Euclidean transformation
        \item Cropped images to overlapping regions
        \item Computed DINO similarity \cite{oquab2023dinov2} on aligned crops ($\text{dino\_aligned\_sim}$)
    \end{itemize}

    \item \textbf{Overlap Analysis:} Filtered out simple crops by ensuring significant transformation between images:
    \begin{itemize}
        \item Analyzed overlap percentage between images
        \item Removed pairs with excessive overlap (indicating simple crops rather than edits)
    \end{itemize}

   \item \textbf{Caption Generation:} Finally, for each qualified image pair, we generated descriptive captions that explicitly highlighted the visual transformations to create suitable training data for editing models. Initially, the model was selected through an extensive Side-by-Side (SBS) comparative evaluation of several state-of-the-art multimodal models, including GPT-4o, GPT-4 Mini with reasoning, Gemini 2.5 Pro, and Qwen2.5-VL-32B. This qualitative human evaluation assessed the models' ability to produce coherent, precise, and instructional captions. Results ranked Gemini 2.5 Pro first, with the GLM 4.5 model \cite{zeng2022glm} achieving competitive results. Given its favorable performance-to-cost ratio, we selected GLM 4.5 without reasoning and fine-tuned it using Low-Rank Adaptation (LoRA) \cite{hu2021lora} on a curated dataset to optimize it specifically for generating instructional captions. This adaptation resulted in a robust and cost-effective solution for our task.
\end{enumerate}

\subsubsection*{Dataset Summary}
\begin{figure}[htbp]
    \centering
    \begin{subfigure}{0.45\textwidth}
        \centering
        \includegraphics[width=\linewidth]{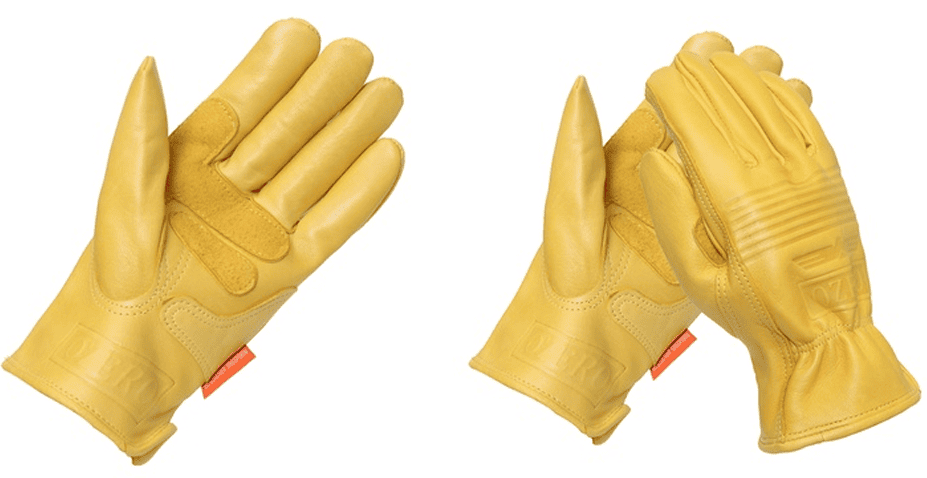}
        \caption{Instruction: Add a second, idetical glove and place it next to the first one, sligtly overlapping it.}
    \end{subfigure}    
    \hfill
    \begin{subfigure}{0.45\textwidth}
        \centering
        \includegraphics[width=\linewidth]{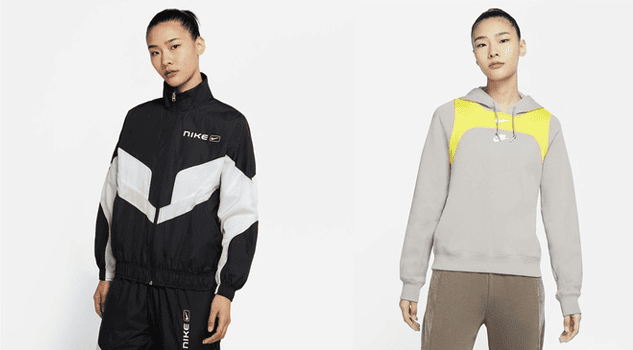}
        \caption{Instruction: Replace the black and white tracksuit with a grey hoodie with a yellow yoke and white swoosh, and brown trousers.}
    \end{subfigure}    

    \vspace{0.2cm}
    \begin{subfigure}{0.44\textwidth}
        \centering
        \includegraphics[width=\linewidth]{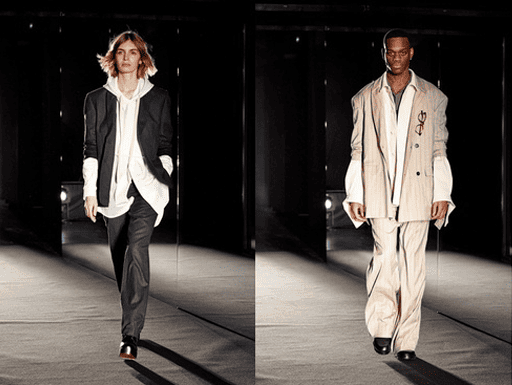}
        \caption{Instruction: Replace the model with a man wearing  a beige pinstripe suit with a white shirt underneath, and add a pair of glasses hanging from the lapel.}
    \end{subfigure}
    \hfill
    \begin{subfigure}{0.49\textwidth}
        \centering
        \includegraphics[width=\linewidth]{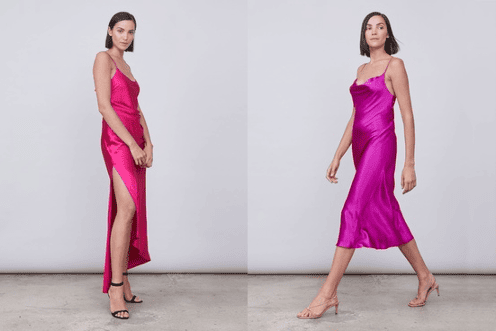}
        \caption{Instruction: Change the woman's pose to walking forward, replace her black high heels with pink ones, and change the camera view to a full-length shot.}
    \end{subfigure}    

    \vspace{0.2cm}
    \begin{subfigure}{0.67\textwidth}
        \centering
        \includegraphics[width=\linewidth]{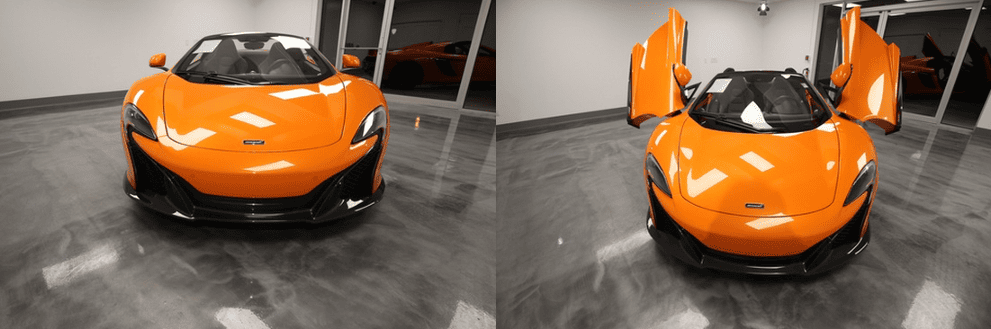}
        \caption{Instruction: Open the car's butterfly doors.}
    \end{subfigure}

    \vspace{0.2cm}
    \begin{subfigure}{\textwidth}
        \centering
        \includegraphics[width=0.67\linewidth]{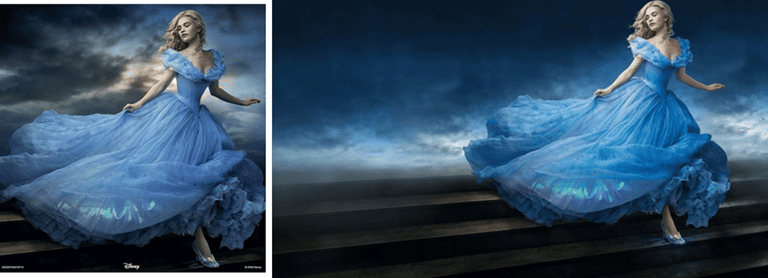}
        \caption{Instruction: Change the camera view to a wider shot, showing more of the stairs and the background, and make the lighting more dramatic with a dark, stormy sky.}
    \end{subfigure}
    
    \caption{Examples of collected pairs and generated instructions from Kandinsky I2I dataset}
    \label{fig:I2I_dataset}
\end{figure}

This meticulous multi-stage filtering and verification process ensures the creation of an extensive, high-quality dataset comprising approximately {\bf 150 million} carefully curated image pairs (see Figure \ref{fig:I2I_dataset}), each accompanied by comprehensive textual instructions that precisely describe the transformations between images, thereby providing an optimal foundation for effective training of advanced image editing models. However, to further enhance the model's aesthetic perception and instruction-following capabilities for final output refinement, we constructed an additional specialized dataset for supervised fine-tuning.

\subsubsection*{Supervised Fine-Tuning (SFT) Dataset Curation}
To create a high-quality dataset for supervised fine-tuning, a subset was curated from the Kandinsky I2I dataset used for pre-training. We applied quality filters based exclusively on the target image (the second image in the pair), as our observations indicate that this approach is sufficient for identifying high-quality transformation examples while ignoring the source image. The filters required a Q-Align score greater than 4 and a Q-Align aesthetic score greater than 2 to select visually superior results. This filtering yielded approximately 600k candidate pairs. From this pool, human annotators manually select the most appropriate and high-quality image-editing pairs for the final SFT dataset. This meticulous manual curation ensures the SFT training data consists of exemplary instances, which is critical for effective instruction tuning.

\begin{table}[htb]
\centering
\caption{Image editing dataset filtering criteria and parameters}
\label{tab:image_editing_criteria}
\begin{tabular}{p{4cm}p{7cm}p{2cm}}
\hline
\textbf{Filtering Stage} & \textbf{Description} & \textbf{Threshold} \\
\hline
CLIP Similarity & Semantic similarity with adaptive thresholding & $> 0.8$ + adaptive \\
DINO Similarity & Visual feature similarity & $> 0.8$ \\
Face Similarity & For images containing single faces & $> 0.7$ \\
Inlier Points (DINO) & Minimum matched points for DINO pairs & $> 300$ \\
Inlier Points (CLIP) & Minimum matched points for CLIP pairs & $> 200$ \\
Exclusion Criteria & Remove near-duplicates and bad face matches & Various \\
Overlap Analysis & Ensure significant transformation between images & Custom heuristic \\
\hline
\end{tabular}
\end{table}

\subsection{Text-to-Video Dataset Processing}
The Kandinsky T2V dataset comprises more than 250 million video scenes sourced from various open datasets and large video platforms. The data processing pipeline is a multi-stage process designed to ensure high-quality, diverse, and suitable data for pretraining the Kandinsky video generation model.

\subsubsection*{Processing Pipeline}
The procedure involves the following stages:

\begin{itemize}
    \item \textbf{Scene Segmentation:} The initial step involves segmenting raw videos into individual scenes using the PySceneDetect tool\footnote{\url{https://github.com/Breakthrough/PySceneDetect}}, which detects shot changes. This process isolates sequences of consecutive frames with a consistent visual perspective. Each extracted scene has a duration constrained between 2 and 60 seconds to ensure temporal coherence and manageability for subsequent processing.

    \item \textbf{Initial Filtering and Deduplication:} After segmentation, scenes undergo a series of initial filtering steps:
    \begin{itemize}
        \item \textit{Resolution Filtering:} Scenes with a shorter side of fewer than 256 pixels are removed to maintain a minimum visual quality standard.
        \item \textit{Deduplication:} To eliminate redundant content, a video perceptual hash \cite{lawto2007videoperceptualhashing} is computed for each scene. This hash is a fingerprint designed to be similar for visually alike videos, allowing for the identification and removal of identical or highly similar scenes. While the primary deduplication uses this method, some duplicates may still persist in the dataset.
    \end{itemize}

    \item \textbf{Advanced Quality and Content Filtering:} Following initial filtering, a comprehensive suite of models is applied to assess various aspects of scene quality and content:
    \begin{itemize}
        \item \textit{Watermark Detection:} Similar to the approach for images, video watermark detection employs a combination of a dedicated classifier and an object detector. Scenes containing watermarks are filtered out by averaging the models' confidence scores across five evenly spaced frames and applying specific thresholds.
        \item \textit{Structural Dynamics:} This filter evaluates scene motion by sampling frames at 2 FPS and calculating the Multi-Scale Structural Similarity (MS-SSIM) index \cite{wang2003msssim} between consecutive pairs. A low average MS-SSIM score indicates high dynamism (rapid scene changes), while a high score indicates a static scene. Scenes deemed too static or excessively dynamic are filtered out based on threshold values.
        \item \textit{Technical and Aesthetic Quality:} The DOVER model \cite{wu2023exploring} is employed to provide separate scores for technical quality (e.g., sharpness, noise) and aesthetic quality (e.g., composition, color). An overall quality score is derived as a weighted sum of these two components. The Q-Align model \cite{wu2023qalign}, which uses large multimodal models to assess visual quality, is also applied to provide an additional quality assessment.
        \item \textit{Text Detection:} The CRAFT model \cite{baek2019craft} is used to identify text regions within video frames. The number of detected text boxes and their total area are averaged over three evenly spaced frames. Scenes with an excessive amount of text are filtered to prevent the generation of videos dominated by subtitles or on-screen graphics.
        \item \textit{Object and Scene Classification:} The YOLOv8 model, trained on the OpenImagesV7 dataset \cite{kuznetsova2020openimages}, detects objects in five frames per scene. A CLIP-based classifier \cite{radford2021clip} is used to classify the scene's location, style, main subject, and more detailed place categories by analyzing frame embeddings.
        \item \textit{Camera and Object Dynamics:} A specialized model \cite{kandinskyvideotools2025} based on VideoMAE \cite{tong2023videomae} architecture was trained to  predict scores for camera movement, object movement, and dynamics of light and color changes to further refine the assessment of scene activity.
    \end{itemize}

    \item \textbf{Content Annotation:} For scenes that pass the filtering stages, synthetic English captions are generated to provide textual descriptions. This is achieved using the large multimodal model Tarsier2-7B\footnote{\url{https://github.com/netease-youdao/Tarsier}}. To improve caption quality, post-processing steps are applied:
    \begin{itemize}
        \item Regular expressions are used to remove common introductory phrases (e.g., "The video starts").
        \item Captions are filtered to exclude sentences containing non-English characters (e.g., Cyrillic or Chinese), as the models can sometimes produce incorrect or mixed-language output.
    \end{itemize}

    \item \textbf{Additional Processing (Clustering):} As an auxiliary process, scene embeddings are generated using the InternVideo2-1B model \cite{wang2024internvideo2}. K-Means clustering \cite{lloyd1982kmeans} is then performed on these embeddings to group visually or semantically similar scenes into 10,000 clusters. This information is used for balanced sampling and dataset analysis.

    \item \textbf{Organization and Storage:} The final processed scenes, along with their metadata, filter scores, and captions, are organized into Parquet files. These files are grouped by the scene's shortest frame side (256, 512, and 1024 - see Table \ref{tab:resolution} for list of supported resolutions) and stored on an S3-compatible storage endpoint for efficient access during model training.
\end{itemize}

\begin{table}[htb]
\centering
\caption{Supported Resolutions (height $\times$ width in pixels)}
\begin{tabular}{c|c}
\textbf{Lite Version} & \textbf{Pro Version} \\
\hline
512 $\times$ 512 & 1024 $\times$ 1024 \\
512 $\times$ 768 & 640 $\times$ 1408 \\
768 $\times$ 512 & 1408 $\times$ 640 \\
& 768 $\times$ 1280 \\
& 1280 $\times$ 768 \\
& 896 $\times$ 1152 \\
& 1152 $\times$ 896 \\
\end{tabular}
\label{tab:resolution}
\end{table}

\begin{figure}[tb]
    \centering
    \includegraphics[width=\textwidth]{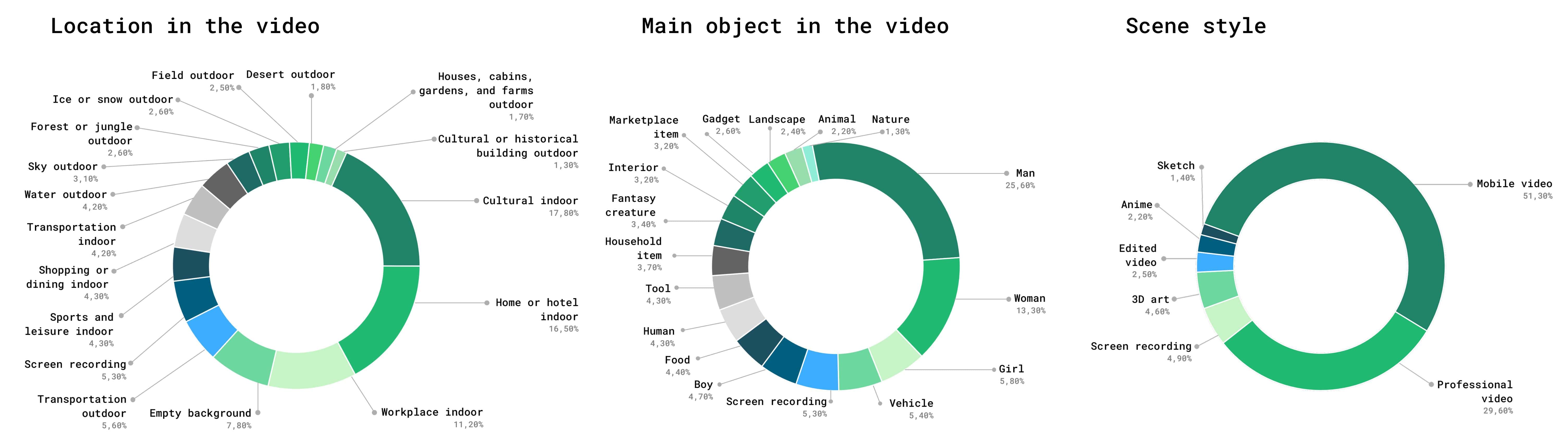} 
    \caption{Distribution of key data categories across the curated Kandinsky T2V dataset by Location, Main object and Scene style.}
    \label{fig:T2V_distribution}
\end{figure}
\subsection{Russian Cultural Code Dataset Processing}
The Kandinsky RCC dataset contains 229,504 video scenes and 768,555 images focused on the Russian cultural code (specific features linked to faith, language, historical memory, nature, architecture, personality, etc.).

\subsubsection*{Dataset Characteristics}
\begin{itemize}
    \item \textbf{Curation Method:} Unlike the T2V and T2I datasets, the RCC data is manually curated by annotators based on relevance to the Russian cultural code, visual quality, and the absence of watermarks or subtitles.
    \item \textbf{Annotations:} The images and scenes are accompanied by manually written Russian descriptions, which are then machine-translated into English with special handling for proper names.
    \item \textbf{Usage:} This dataset is used both for pretraining and for specialized fine-tuning to improve the model's performance on culturally specific content.
\end{itemize}

\subsection{Supervised Fine-Tuning Dataset Processing}\label{sec:sft_dataset_processing}

\begin{figure}[tb]
    \centering
    \includegraphics[width=\textwidth]{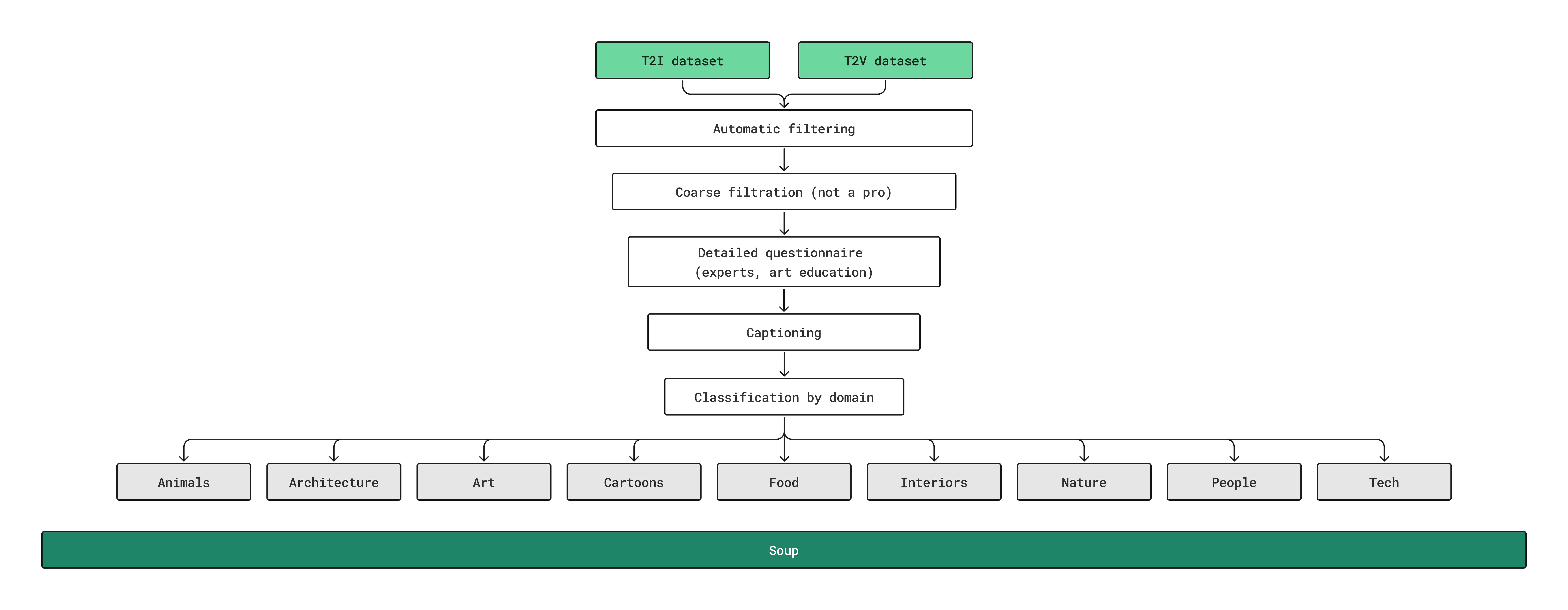}
    \caption{SFT dataset curation pipeline. Data from Kandinsky T2V and T2I datasets undergoes automatic technical quality and aesthetic filtering, followed by coarse human filtering (“not a pro”), detailed expert evaluation (including art education specialists), captioning, and domain-based classification into certain categories. All curated data is used to finetune class-specialized models and mix them into final soup SFT model.}
    \label{fig:dataset_curation}
\end{figure}

A high-quality Supervised Fine-Tuning (SFT) dataset was meticulously curated to align the
model's outputs with human preferences and significantly enhance visual and compositional
quality. This dataset is distinct from the large-scale pretraining data, as it consists of a smaller
set of high-fidelity examples manually selected by expert annotators through a rigorous multi-
stage evaluation process. The dataset comprises both images and video content. Additionally, using a video language model (VLM), all data was classified into 9 domains (see Figure \ref{fig:dataset_curation}). Moreover, images in SFT dataset were split into more detailed hierarchy of classes to obtain the best possible model.

The domain-based organization enables:
\begin{itemize}
    \item Domain-specific training approaches and curriculum learning strategies
    \item Composition of SFT-soup models by weights averaging (following the approach described in \cite{wortsman2022modelsoup})
    \item Balanced sampling across different content categories during training
    \item Specialized fine-tuning for particular content types while maintaining overall model coherence
    \item Controlled mixing ratios between video and image data within training batches
    \item Targeted improvement of model performance in specific visual domains
\end{itemize}

\subsubsection*{Dataset Construction}

\begin{itemize}
    \item \textbf{Initial Collection:} The initial pool consisted of 93,296 high-resolution video scenes and approximately 10 million images, sourced automatically for their exceptional technical quality (e.g., resolution threshold, lack of artifacts, watermarks) and aesthetic merit.

    \item \textbf{Captioning:} For all images we implemented a specialized two-stage captioning pipeline to maximize instruction following capability: first, the \textbf{Qwen2.5-VL-32B-Instruct} model generated a long textual description for the input image; second, the \textbf{Qwen3-32B} model rewrote this description into several variations: very long, long, medium length, short, and very short. Ablation study was conducted to select the optimal system prompt for each model to ensure the final textual descriptions were maximally accurate and approximated human speech. Since our models were pretrained on both English and Russian textual descriptions, we want to continue this approach during SFT, requesting the language model to return descriptions in both languages simultaneously for consistency.
    
    To provide rich textual descriptions for each video, automatic captioning was performed using the \textbf{SkyCaptioner-V1}~\cite{chen2025skyreelsv2infinitelengthfilmgenerative} and \textbf{Qwen2.5-VL 32B and 72B}~\cite{bai2025qwen25vltechnicalreport} models. This process generated multiple caption variants of differing lengths and descriptive density for each record. The captions obtained were cleaned with regular expressions to avoid useless common parts such as "the image shows".

    \item \textbf{Multi-Stage Expert Evaluation:} The data passed through a rigorous two-stage evaluation process with specialized annotator roles:
    
    \textbf{Stage 1 - Technical Screening:} Regular annotators assessed content based on fundamental quality criteria included:
    \begin{itemize}
        \item Presence and prominence of main object
        \item Proper cropping and framing
        \item Spatial depth perception
        \item Absence of visual artifacts
    \end{itemize}

    \textbf{Stage 2 - Comprehensive Quality Assessment:} Qualified experts with background in cinematography and visual arts evaluated the pre-filtered content using detailed questionnaires listed below: 
    
  \textbf{Image Evaluation Criteria}
\begin{itemize}
\item Exposure and contrast correctness
\item Horizon line positioning
\item Composition geometry
\item Object silhouette quality
\item Lighting and color scheme
\item Absence of unwanted objects
\item Overall artistic expressiveness
\item Natural, non-AI appearance
\item Organic integration
\item Need for modifications
\item Exceptional "wow factor" quality
\end{itemize}

\textbf{Video Evaluation Criteria}
\begin{itemize}
\item Video integrity and completeness
\item Clarity of content and action
\item Main object presence and highlighting
\item Dynamic properties (high/low)
\item Color solution and grading
\item Framing correctness
\item Horizon line positioning
\item Composition quality: rule of thirds, central, diagonal
\item Depth and volumetrics
\item Exposure correctness
\item Suitability for editing
\item Complexity of content
\item Aesthetic and artistic expression
\end{itemize}

    \item \textbf{Specialized Text Handling:} Images containing text underwent additional evaluation with specialized criteria:
    \begin{itemize}
        \item Text clarity and legibility
        \item Appropriate text overlay integration
        \item Absence of unusual letter distortions
        \item Harmonious text-background relationship
        \item Special categorization for poster/cover designs
    \end{itemize}

        \item \textbf{Quality Thresholds:} For the first version (v1) of the dataset, the selection was based on strict criteria requiring consensus among evaluators (minimum $\frac{3}{2}$ agreement on "good" ratings across all parameters). This rigorous approach resulted in the selection of approximately 3\% of the initial video pool and 5\% of images, prioritizing content with clearly defined subjects, balanced composition, and high production values. We conducted extensive research on question selection, acceptance thresholds, and artifact detection. This led to two dataset versions: 
    \begin{itemize}
        \item \textbf{v1 (strict):} 2,833 video scenes and 45,000 images with comprehensive criteria
        \item \textbf{v2 (relaxed):} 12,461 video scenes and 153,000 images with optimized thresholds
    \end{itemize}
    During fine-tuning, we sample from both datasets with calibrated probabilities to balance quality and diversity.
\end{itemize}

\begin{figure}[htbp]
    \centering
    \begin{subfigure}[b]{\textwidth}
        \centering
        \includegraphics[width=0.7\textwidth]{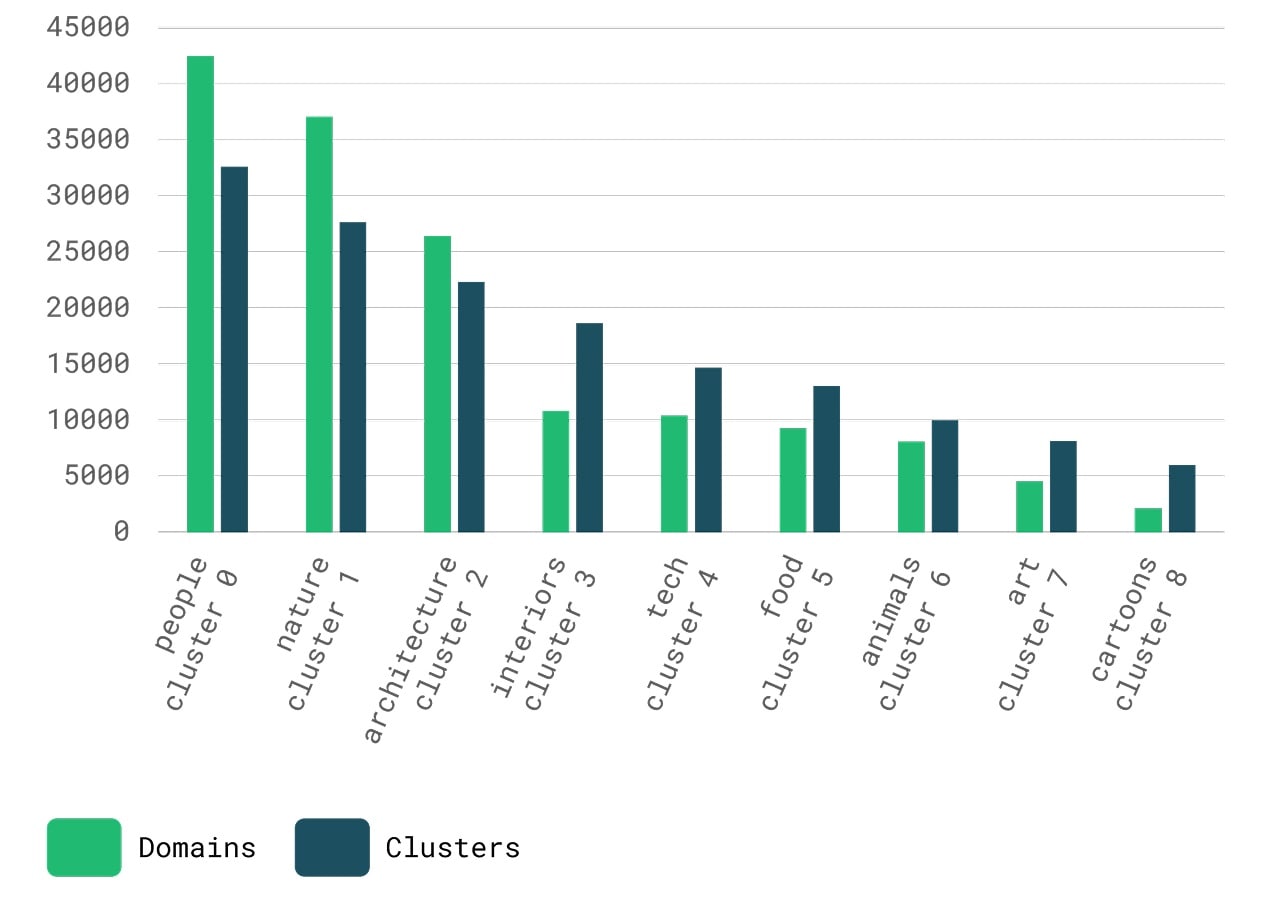}
        \caption{Difference in distributions between VLM domains and K-Means clusters.}
        \label{fig:cluster_domains}
    \end{subfigure}
    \vspace{0.5cm}
    
    \begin{subfigure}[b]{0.9\textwidth}
        \centering
        \includegraphics[width=\textwidth]{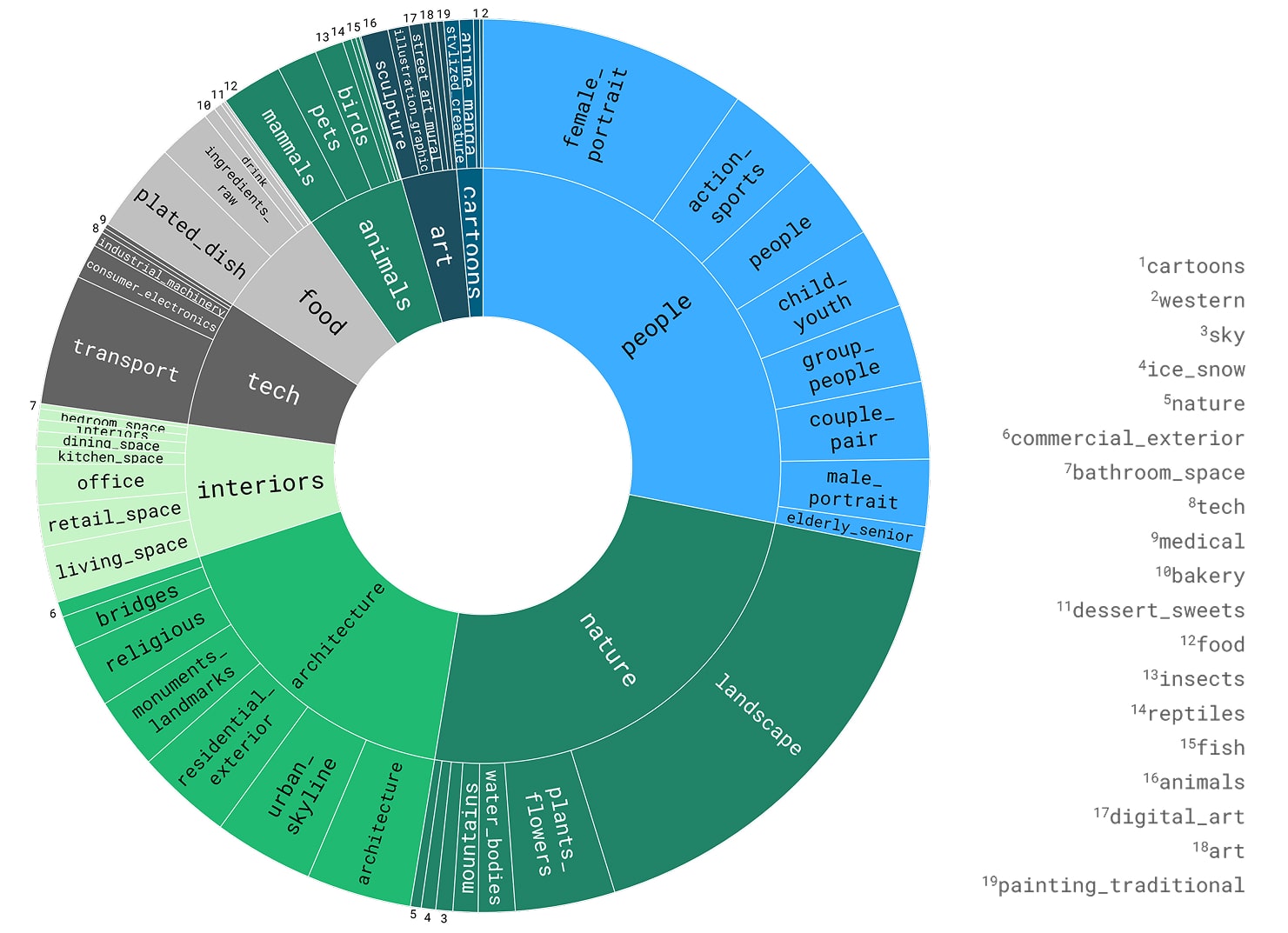}
        \caption{Distribution of data across VLM domains and subdomains.}
        \label{fig:hierarchy_domains}
    \end{subfigure}
    \caption{Data distributions for different domain classification and clustering strategies used during T2I SFT dataset organization.}
    \label{fig:domain_distributions}
\end{figure}

\subsubsection*{VLM Domain Organization}\label{sec:VLM_Domain_Organization}

Both video and image data were classified into 9 consistent domains using the Qwen2.5-VL-Instruct-32B model \cite{bai2025qwen25vltechnicalreport}. This ensures coherence between the video and image domains during mixed-dataset training of the Text-to-Video (T2V) model. The classification used a standardized prompt approach:

    \begin{verbatim}
CATEGORY_LIST = """
1) animals
2) architecture
3) art
4) cartoons
5) food
6) interiors
7) nature
8) people
9) tech
10) other
"""

PROMPT_TEMPLATE = """
You are a professional {media_type} classificator. You are given a {media_type}.
Your task is to return one of the following classes this {media_type} relates to the most.
Choose one class from the following classes:
{CATEGORY_LIST}
Return only the name of class.
""".strip()
    \end{verbatim}
The final SFT-soup model for T2V was composed by averaging weights of models fine-tuned on meaningful 9 domains, using a simplified approach with equal weighting.

\subsubsection*{Advanced Domain Organization and SFT-soup Composition Ablation Study}
For the Text-to-Image (T2I) model, we employed a more sophisticated hierarchical domain organization strategy. 

We experimented with multiple domain decomposition strategies for following optimal SFT-soup composition:

\begin{itemize}
    \item \textbf{VLM Domain Classification:} Full fine-tuning was performed on each VLM domain from the previous section with reduced batch size (64 vs. 4096 in pretrain) and learning rate (1e-5 vs. 1e-4).

    \item \textbf{CLIP Embedding Clustering:} We alternativeively used \textbf{CLIP-ViT-H-14-quickgelu} embeddings with k-means clustering into 9 clusters. This approach increased data diversity within individual components, reducing overfitting and enabling longer training for better realism.

    \item \textbf{Hierarchical VLM Domains:} Each VLM domain was further divided into 2-9 semantically coherent subdomains. Separate full fine-tuning on these smaller, more homogeneous subsets allowed for targeted improvement in specific visual characteristics while maintaining overall coherence.
\end{itemize}

For weight merging in SFT-soup composition, we compared three approaches: equal weights (1/N), weights proportional to dataset size, and weights proportional to the square root of dataset size. Our experiments showed that equal weights or root-proportional weights performed best. The hierarchical domain approach yielded the best results in side-by-side evaluations, achieving high realism, improved text rendering quality, and compositionally correct generated images.

The resulting SFT dataset represents a carefully curated collection of visual content that exemplifies high aesthetics and enables robust model fine-tuning across diverse visual domains.
\section[Kandinsky 5.0 Architecture]{Kandinsky 5.0 Architecture}\label{sec:base_model_architecture}

\subsection{Model Overview}

All models in the Kandinsky 5.0 family are built upon a unified architecture based on a latent diffusion pipeline~\cite{Rombach_2022_CVPR} and trained using the Flow Matching paradigm~\cite{lipman2023flow}. The core of this architecture is a specially designed Diffusion Transformer with cross-attention (\textbf{CrossDiT}) for multimodal fusion of visual and textual information. The number of CrossDiT blocks varies depending on the model size. The architecture also integrates text and visual encoders for efficient input processing:

\begin{itemize}
    \item \textbf{Text Encoding}: Text representations are extracted using the \textbf{Qwen2.5-VL model}~\cite{bai2025qwen25vltechnicalreport}, a transformer decoder architecture that generates rich text embeddings. These embeddings are further processed by a Linguistic Token Refiner module before being passed into the main CrossDiT backbone.

    \item \textbf{Visual Encoding}: We use \textbf{FLUX.1-dev VAE}\footnote{\url{https://huggingface.co/black-forest-labs/FLUX.1-dev}}~\cite{flux2024} to encode images. Video latents are obtained using an encoder from the \textbf{HunyuanVideo VAE}~\cite{kong2025hunyuanvideosystematicframeworklarge}, which produces compact latent representations suitable for the diffusion process while maintaining temporal consistency.
\end{itemize}

Below, we examine in more detail the structure of the CrossDiT backbone architecture (Section~\ref{sec:cross_dit}), the CrossDiT blocks (Section~\ref{sec:cross_dit_block}), and our proposed Neighborhood Adaptive Block-Level Attention (\textbf{NABLA}) mechanism~\cite{mikhailov2025nablanablaneighborhoodadaptiveblocklevel}, which enables significant acceleration and optimization for video generation tasks (Section~\ref{sec:nabla}). In our work, we place strong emphasis on computational efficiency and training stability, which are discussed further in Section~\ref{sec:optimization}.

\subsection{Diffusion Transformer (CrossDiT) Architecture}\label{sec:cross_dit}

\begin{figure}[htbp]
    \centering
    \includegraphics[width=\textwidth]{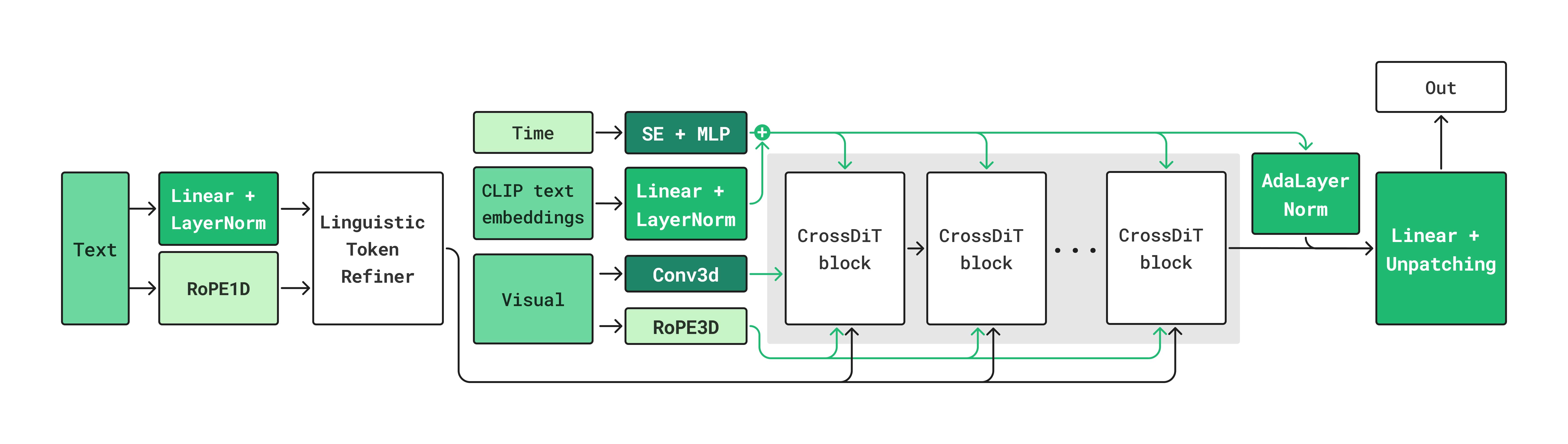}
    \caption{CrossDiT architecture.}
    \label{fig:CrossDiT}
\end{figure}

The CrossDiT architecture is illustrated in Figure~\ref{fig:CrossDiT}. Its core consists of a sequence of CrossDiT blocks. The model takes four distinct types of inputs:

\begin{itemize}
    \item \textbf{Text:} Text embeddings from Qwen2.5-VL are passed through a Linear layer and LayerNorm~\cite{ba2016layer}. One-dimensional positional embeddings are also generated for the text using Rotary Position Encoding (RoPE)~\cite{su2023roformerenhancedtransformerrotary}. These are combined and fed into the Linguistic Token Refiner (LTF) module. This module, which is a CrossDiT block without the cross-attention component (see Figure~\ref{fig:CrossDiTblock}), serves to enhance the text representation and eliminate the positional bias inherited from the pre-trained text encoder~\cite{10.5555/3737916.3741676}. The refined text queries are then fed into the main CrossDiT blocks via the cross-attention mechanism~\cite{lin2021catcrossattentionvision}.

    \item \textbf{Time:} The diffusion timestep value is processed by a block comprising Sinusoidal Encoding (SE)~\cite{vaswani2023attentionneed} and a Multi-Layer Perceptron (MLP).

    \item \textbf{CLIP Text Embedding:} A single text embedding of the full video description from CLIP ViT-L/14 model\footnote{\url{https://huggingface.co/sentence-transformers/clip-ViT-L-14}}~\cite{radford2021clip} is passed through a linear layer and LayerNorm, and then summed element-wise with the time embedding. This resulting sum, along with the output from the final CrossDiT block, is fed into the Adaptive Normalization Layer~\cite{peebles2023dit}.

    \item \textbf{Visual:} Image latents from FLUX.1-dev VAE or video latents from the HunyuanVideo VAE encoder are used to generate 3D Rotary Positional Embeddings, which are then fed into every CrossDiT block.
\end{itemize}

\paragraph{Architecture hyperparameters.} For all models, we use Qwen2.5-VL as the main text encoder with 7 billion parameters, an embedding size of 3584 and a maximum context length of 256. We also use CLIP ViT-L/14 with a text embedding size of 768 and a maximum context length of 77. Hyperparameters for the architecture of the main CrossDiT part of our models are shown in Table~\ref{table:model_hyperparams}.

\begin{table}[h!]
\centering
\caption{CrossDiT hyperparameters for the Kandinsky 5.0 family of models.}
\begin{tabular}{l c c c c c}
\hline
\textbf{Model} & \textbf{Number of} & \textbf{Number of} & \textbf{Linear} & \textbf{Model} & \textbf{Time} \\
 & \textbf{CrossDiT} & \textbf{LTF} & \textbf{layer} & \textbf{embedding} & \textbf{embedding} \\
& \textbf{blocks} & \textbf{blocks} & \textbf{dimension} & \textbf{dimension} & \textbf{dimension} \\
\hline
Image Lite &  50 & 2 & 10240 & 2560 & 512 \\
Video Lite &  32 & 2 & 7168 & 1792 & 512 \\
Video Pro &  60 & 4 & 16384 & 4096 & 1024 \\
\hline
\end{tabular}
\label{table:model_hyperparams}
\end{table}

\subsection{CrossDiT Block Architecture}\label{sec:cross_dit_block}

\begin{figure}[htbp]
    \centering
    \includegraphics[width=\textwidth]{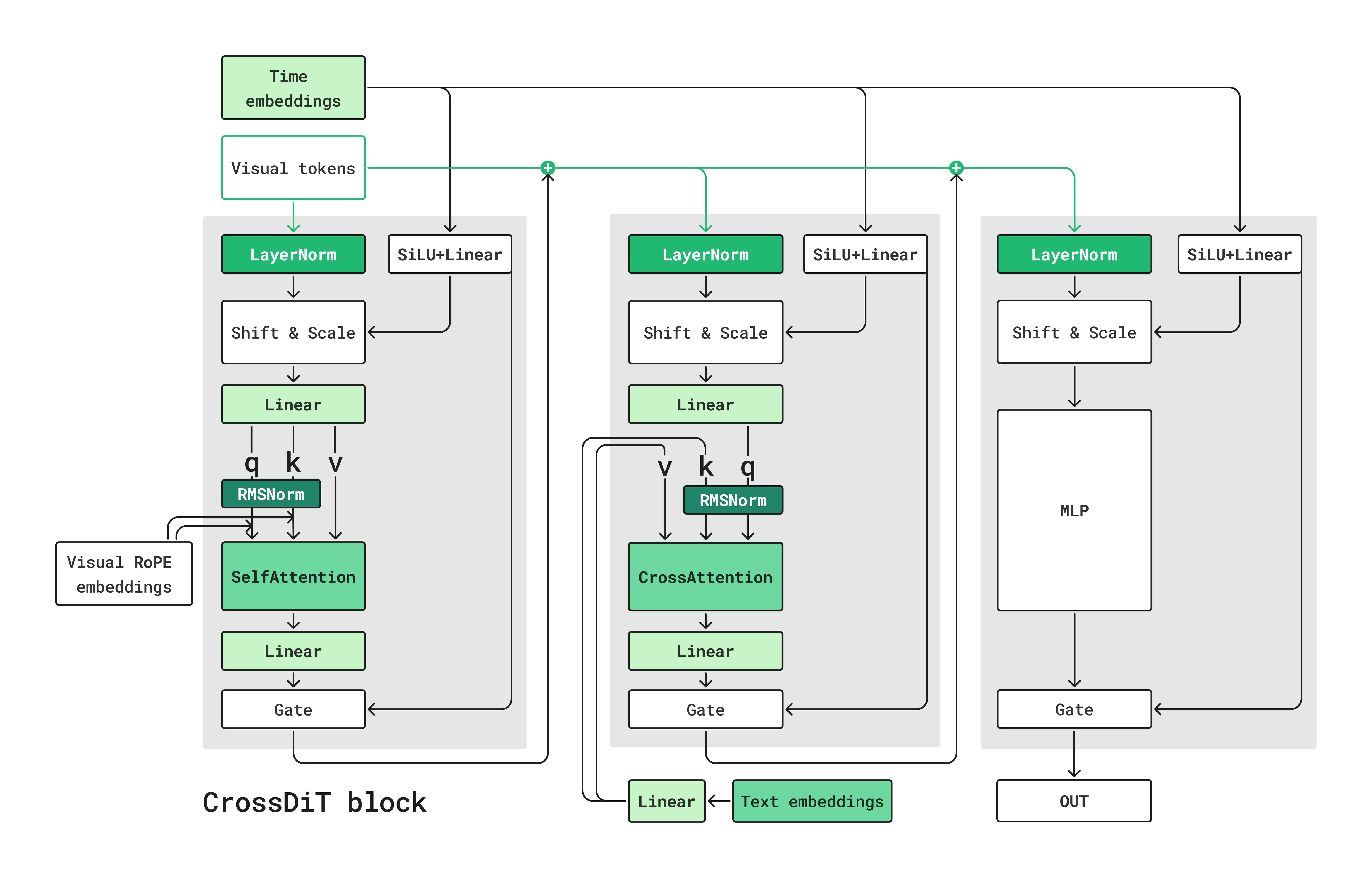}
    \caption{CrossDiT block architecture. From left to right: a Self-Attention Block, a Cross-Attention Block, and a MLP Block.}
    \label{fig:CrossDiTblock}
\end{figure}

At the heart of the \textbf{CrossDiT} block are classic residual connections: three sequential sub-blocks handling: \textbf{Self-attention}, \textbf{Cross-attention} and \textbf{Multi-Layer Perceptron (MLP)} (Figure ~\ref{fig:CrossDiTblock}). The sum of outputs from the Self-attention and Cross-attention attention with the input visual latents is shown on the diagram with a ``+'' symbol.

The advantages of this cross-attention architecture lie in its better compatibility with the \textbf{sparse attention mechanisms} required for processing videos of varying lengths within a single batch. In contrast, the MMDiT-like architecture~\cite{esser2024scalingrectifiedflowtransformers} used in Kandinsky 4.0 required a concatenation operation that significantly slowed down training. In Kandinsky 5.0, we have successfully eliminated this need.

\subsection{Neighborhood Adaptive Block-Level Attention}\label{sec:nabla}

To reduce the computational complexity of attention layers and accelerate the training process for high-resolution (up to 1024px) or long-duration (up to 10 seconds) video generation, we employ \textbf{NABLA} (Neighborhood-Adaptive Block-Level Attention) - a sparse attention mechanism that dynamically constructs content-aware masks for efficient video diffusion transformers.

\begin{figure}[ht]
    \centering
    \includegraphics[width=1.0\linewidth]{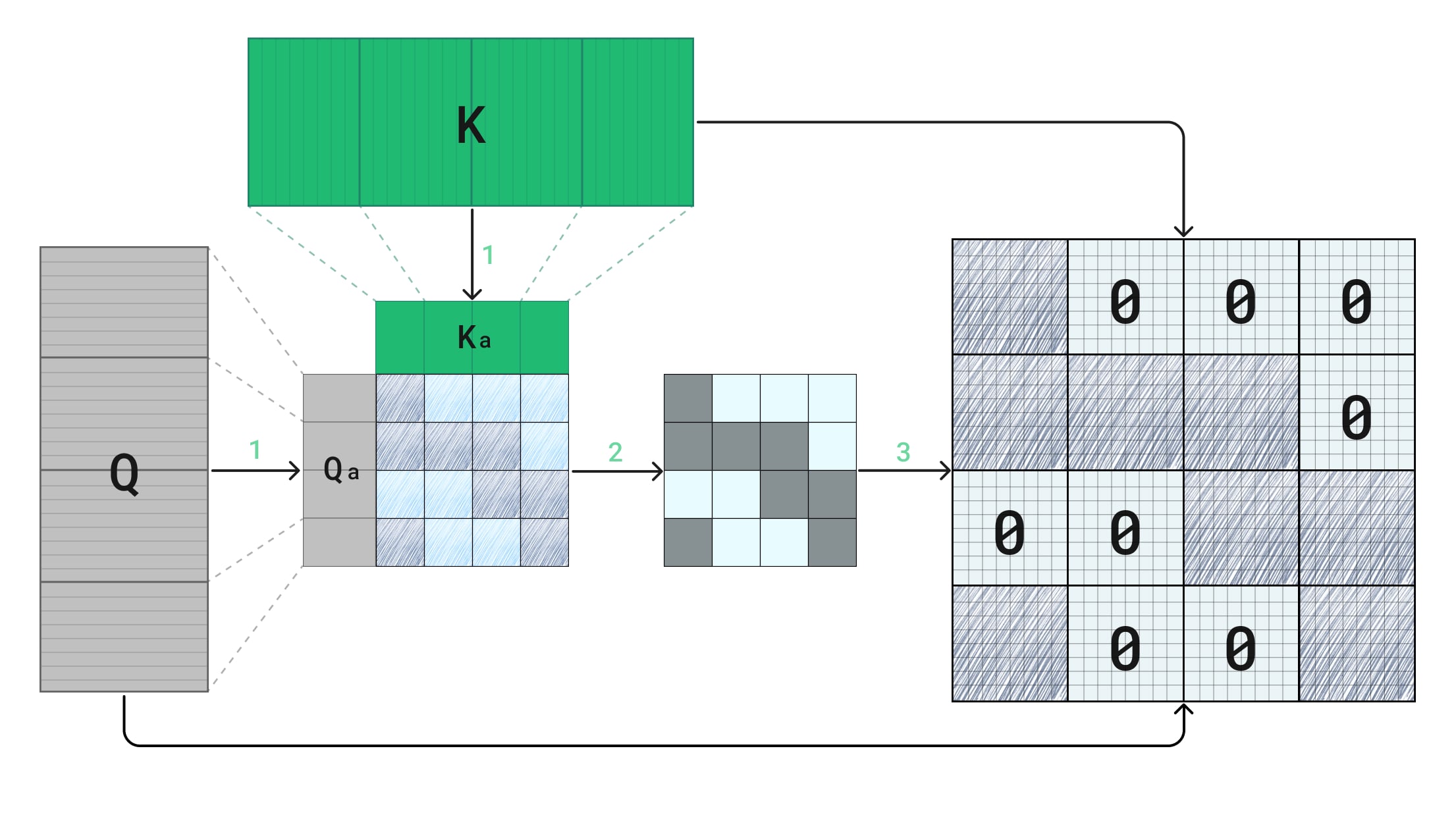}
    \caption{The block-sparse attention mask is computed by (1) reducing the dimensionality of queries (Q) and keys (K), (2) sparsifying the softmax distribution via a cumulative density function (CDF) threshold and binarizing the result, and (3) mapping the sparse mask back to the original input blocks.}
    \label{fig:nabla}
\end{figure}

As shown in Figure~\ref{fig:nabla}, NABLA constructs content-aware sparse attention masks through a three-stage process:

\begin{enumerate}
\item \textbf{Block-wise dimensionality reduction}: Queries and keys are average-pooled by groups of $N = 64$ elements, reducing the attention map computation complexity by a factor of $N^2 = 4096$ while maintaining the structural relationships essential for video coherence.

\item \textbf{Adaptive sparsification via CDF thresholding}: For each attention head, we compute the cumulative distribution function of the reduced attention map and dynamically select the most relevant blocks by thresholding at $1 - thr$, where $thr$ controls the sparsity level. This ensures that each head preserves its unique attention pattern tailored to the input content.

\item \textbf{Border artifact suppression}: The resulting adaptive mask is optionally combined with Sliding-Tile Attention patterns through union operation to maintain local continuity and suppress potential border artifacts that can occur in high-resolution generation.
\end{enumerate}

\begin{figure}[ht]
    \centering
    \begin{minipage}{0.47\textwidth}
    \begin{subfigure}{\linewidth}
        \includegraphics[width=\linewidth]{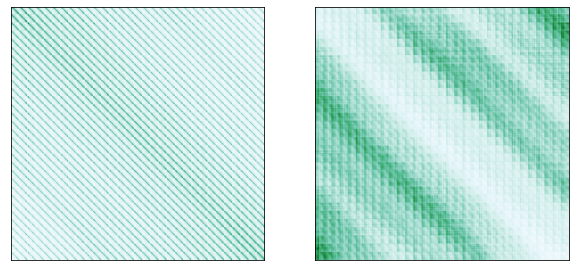}
    \end{subfigure}
    \caption{Real attention maps.}
    \label{fig:attn_maps}
    \vspace {1cm}
    \begin{subfigure}{\linewidth}
        \includegraphics[width=\linewidth]{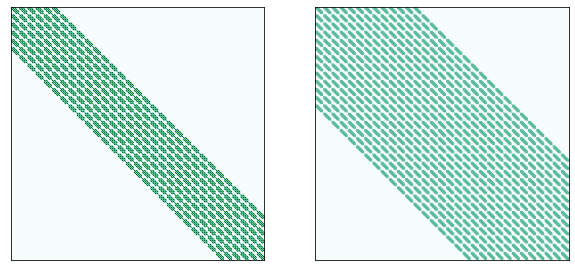}
    \end{subfigure}
    \caption{STA masks with different window sizes.}
    \label{fig:sta_masks}
    \end{minipage}
    \hfill
    \begin{minipage}{0.47\textwidth}
    \centering
    \includegraphics[width=\linewidth]{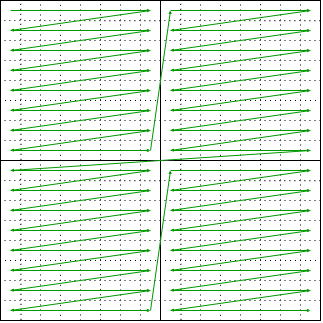}
    \caption{Token reordering illustration for a latent image with height 16, width 16, and patch size 8. The diagram shows how spatial tokens are reorganized into fractal-flattened sequences while preserving their semantic relationships.}
    \label{fig:token_reordering}
    \end{minipage}
 \end{figure}

NABLA employs token reordering through fractal flattening with spatial patches of size $P \times P$ ($P = 8$), grouping all tokens within each patch into contiguous sequences of $P^2 = N = 64$ tokens while preserving the original temporal ordering as illustrated in Figure~\ref{fig:token_reordering}. This reorganization optimizes memory access patterns by ensuring spatially adjacent tokens remain contiguous in memory, significantly improving computational efficiency during attention computation. The reordering operation is applied at the DiT network input with its inverse at the output, maintaining proper spatial relationships throughout the processing pipeline.

NABLA generates head-specific sparsity patterns that adapt to varying attention maps observed in real transformer layers (Figure~\ref{fig:attn_maps}), overcoming limitations of fixed sparse patterns like STA (Figure~\ref{fig:sta_masks}) that may not capture complex long-range dependencies required for coherent video generation. Extensive evaluation demonstrates that NABLA achieves up to \textbf{2.7× speedup} in training and inference while maintaining equivalent generation quality to full attention, as validated by both quantitative metrics (CLIP, VBench) and human evaluations.

NABLA integrates seamlessly with PyTorch's FlexAttention framework without requiring custom CUDA kernels or additional loss functions, making it practical for both training and inference of large-scale video generation models. For complete implementation details and algorithmic specifications, see \cite{mikhailov2025nablanablaneighborhoodadaptiveblocklevel}.

\section[Training Stages]{Training Stages}\label{sec:training}

\subsection{Training Infrastructure}

We pre-train our models on a standard NVIDIA multi-node cluster. Each node contains 8 GPUs connected by NVLink. InfiniBand was used for inter-node connection. We used S3 instead of NFS to store dataset because the weight of the dataset was O(10) Pb and NFS is too expensive for this case.

\subsubsection{Data Storage}

All data were stored in an S3-compatible object storage and streamed during training over a 100~Gbit/s link. The storage system allowed only $\mathcal{O}(10^3)$ concurrent connections, so the data pipeline was designed to keep the number of opened objects small and the throughput per connection high.

\paragraph{Latent pre-encoding and storage layout.}

To reduce I/O and avoid repeated encoder calls on the fly, all images and videos were pre-encoded with the VAE, and \emph{VAE latents} were stored instead of raw pixels. These latents were then packed into \texttt{.tar} archives so that each archive had approximately the same size. The number of samples per archive was chosen as a function of resolution and modality:

\begin{itemize}
    \item \textbf{Images}
    \begin{itemize}
        \item Low resolution: 1024 latents per tar
        \item Medium resolution: 256 latents per tar
        \item High resolution: 64 latents per tar
    \end{itemize}
    \item \textbf{Videos} (one ``latent'' here denotes the full VAE-latent sequence for a video)
    \begin{itemize}
        \item Low resolution: 16 video latents per tar
        \item Medium resolution: 4 video latents per tar
        \item High resolution: 1 video latent per tar
    \end{itemize}
\end{itemize}

This scheme keeps archive sizes roughly uniform across resolutions while minimizing the total number of S3 objects that need to be opened during training.

\paragraph{Text embeddings.}

Caption text embeddings were computed on the fly during training rather than stored in S3. A single text embedding is approximately 50$\times$ larger than a low resolution (e.g., 256$\times$256) image latent; precomputing and storing all text embeddings would therefore significantly increase both the storage footprint and the per-object size, putting additional pressure on the S3 bandwidth and connection limits. Computing them online keeps the S3 load dominated by compact image/video latents and simplifies the storage layout.

\subsubsection{DataLoader Design}

\begin{figure}[t]
    \centering
    \begin{subfigure}{\textwidth}
        \centering
        \includegraphics[width=\linewidth]{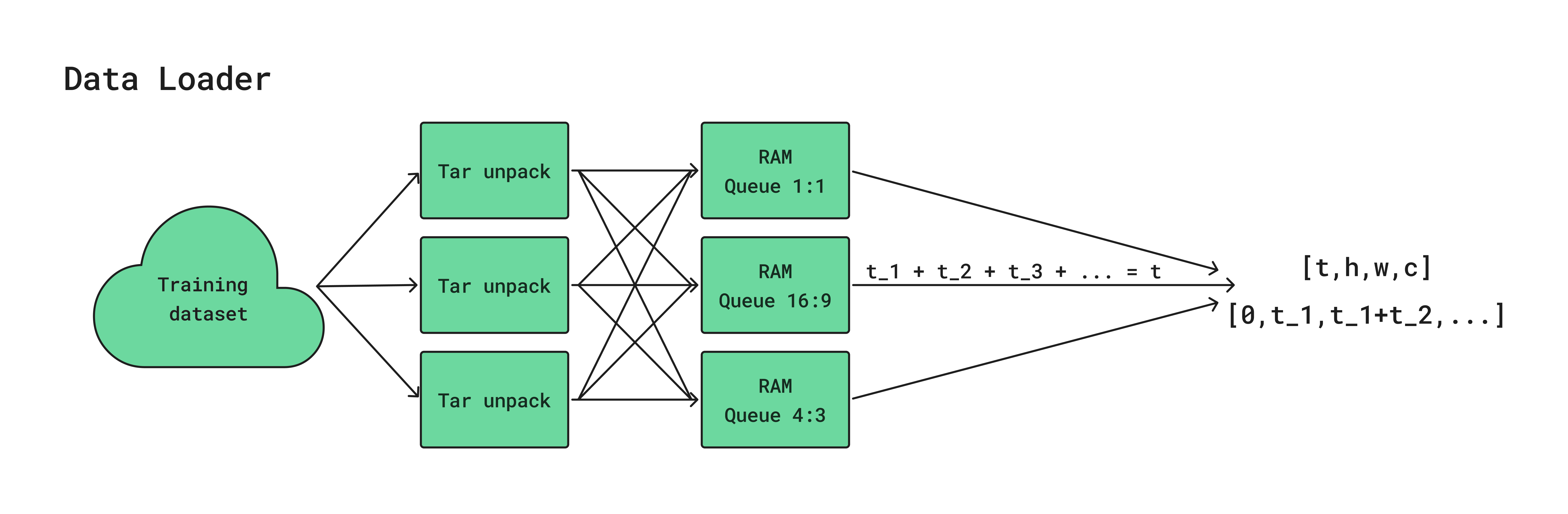}
        \caption{Data loader scheme with S3 streaming, tar unpacking, and aspect-ratio queues.}
        \label{fig:dataloader-scheme}
    \end{subfigure}
    \hfill
    \begin{subfigure}{\textwidth}
        \centering
        \includegraphics[width=0.5\linewidth]{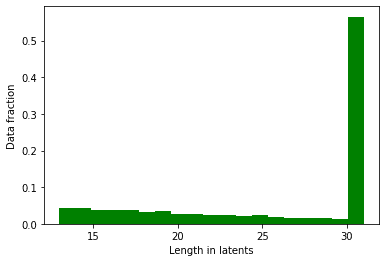}
        \caption{Histogram of video latent temporal lengths $t$ in the dataset.}
        \label{fig:tlen-hist}
    \end{subfigure}
    \caption{Data streaming pipeline and distribution of video temporal lengths.}
    \label{fig:dataloader}
\end{figure}

The data loader was implemented as a two--stage pipeline. A dedicated worker process
streamed \texttt{.tar} archives from S3, unpacked them on the fly, and pushed individual
VAE latents into in-memory queues. Each queue corresponded to a particular aspect ratio
(e.g., 1:1, 16:9, 4:3), so that the main training process could sample shape-compatible
clips without additional padding or resizing (see Figure~\ref{fig:dataloader-scheme}).

The main process constructed batches by repeatedly popping video latents from a selected
aspect-ratio queue until the \emph{sum of their temporal lengths} reached a predefined
maximum:
\[
\sum_i t_i \approx t_{\max}.
\]
Images were treated as videos of length 1, but they were always sampled in
\emph{image-only batches}. This allowed us to explicitly control the global fraction of
image steps during training: a high proportion of images in mixed batches was empirically
found to degrade convergence on video tasks.

Concatenating videos of different lengths into a single batch in this way significantly
reduced idle time and improved GPU utilization. For large temporal contexts we
additionally employed \emph{adaptive attention}, so the cost of processing one long video
was comparable to processing several shorter clips with the same total number of tokens.
The distribution of temporal lengths in the dataset (Figure~\ref{fig:tlen-hist}) is highly
skewed: most videos are close to the maximum allowed length, with a relatively small but
non-negligible portion of shorter clips, which makes this dynamic batching strategy
particularly beneficial.

\subsubsection{Distributed Training and Memory Optimization}

We used \emph{HSDP}~\cite{zhao2023pytorch} for distributed training at all stages, from low resolutions up to
the final high resolution setting. Starting from the medium resolution stage, we additionally
enabled \emph{Sequence Parallel}~\cite{korthikanti2205reducing}. The weights of the Diffusion Transformer were
partitioned across 64 GPUs, and the text encoder weights were partitioned across 32 GPUs.
This sharding scheme provided full overlap of computation and communication at all
training stages and resulted in very low per-GPU memory usage for both model parameters
and optimizer states.

We chose Sequence Parallel instead of Tensor Parallel because it requires only two
collective operations per transformer block (before and after the self-attention module).
Sharding the weights inside a single block was unnecessary, as the block itself was
relatively lightweight. For HD video training with 10-second sequences, we used the
maximum sequence-parallel sharding over 8 GPUs, which ensured that all collectives
stayed within a single NVLink island.

Checkpointing was performed in a non-blocking manner. Each of the 64 processes wrote
its parameter shard directly to storage, without reconstructing the full
\texttt{state\_dict} on any node.

\paragraph{Activation checkpointing and offloading.}

To further reduce memory usage we applied activation checkpointing~\cite{chen2016training} the granularity of
transformer blocks. During pre-training we used the classical scheme, storing only a
subset of activations and recomputing the rest during the backward pass. For the RL
fine-tuning stage, where we had to keep activations for the entire generation trajectory,
we used an extended variant with offloading: the inputs to each transformer block were
temporarily moved from GPU memory to host RAM between the forward and backward passes.

Offloading was implemented in a non-blocking fashion: device--host transfers were issued
asynchronously and overlapped with computation. As a result, the additional data movement
did not noticeably slow down training while still providing a substantial reduction in
peak activation memory.

\begin{figure}[t]
    \centering
    \includegraphics[width=\linewidth]{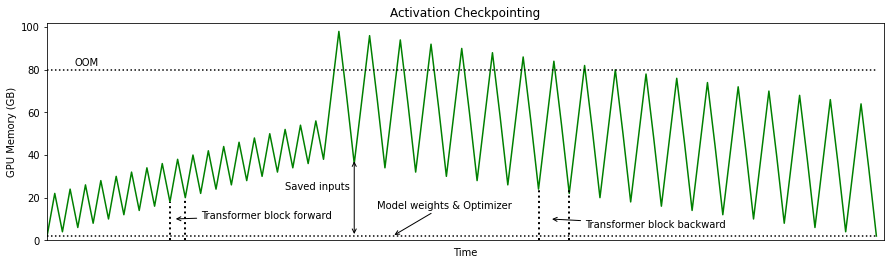}
    \includegraphics[width=\linewidth]{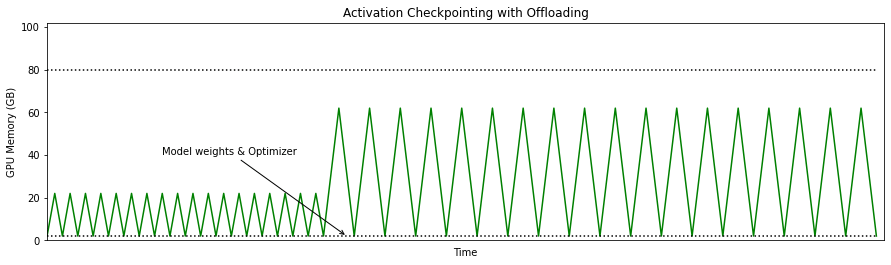}
    \caption{Activation memory traces for standard activation checkpointing (top) and
    activation checkpointing with host offloading (bottom).}
    \label{fig:act-checkpoint}
\end{figure}

The effect of this modification is illustrated in Figure~\ref{fig:act-checkpoint}.
Compared to standard activation checkpointing (top trace), activation checkpointing with
offloading (bottom trace) significantly lowers the peak activation footprint while
preserving a similar compute pattern.

\subsection{Training Procedure Overview}

\begin{figure}[h!]
    \centering
    \includegraphics[width=0.87\textwidth]{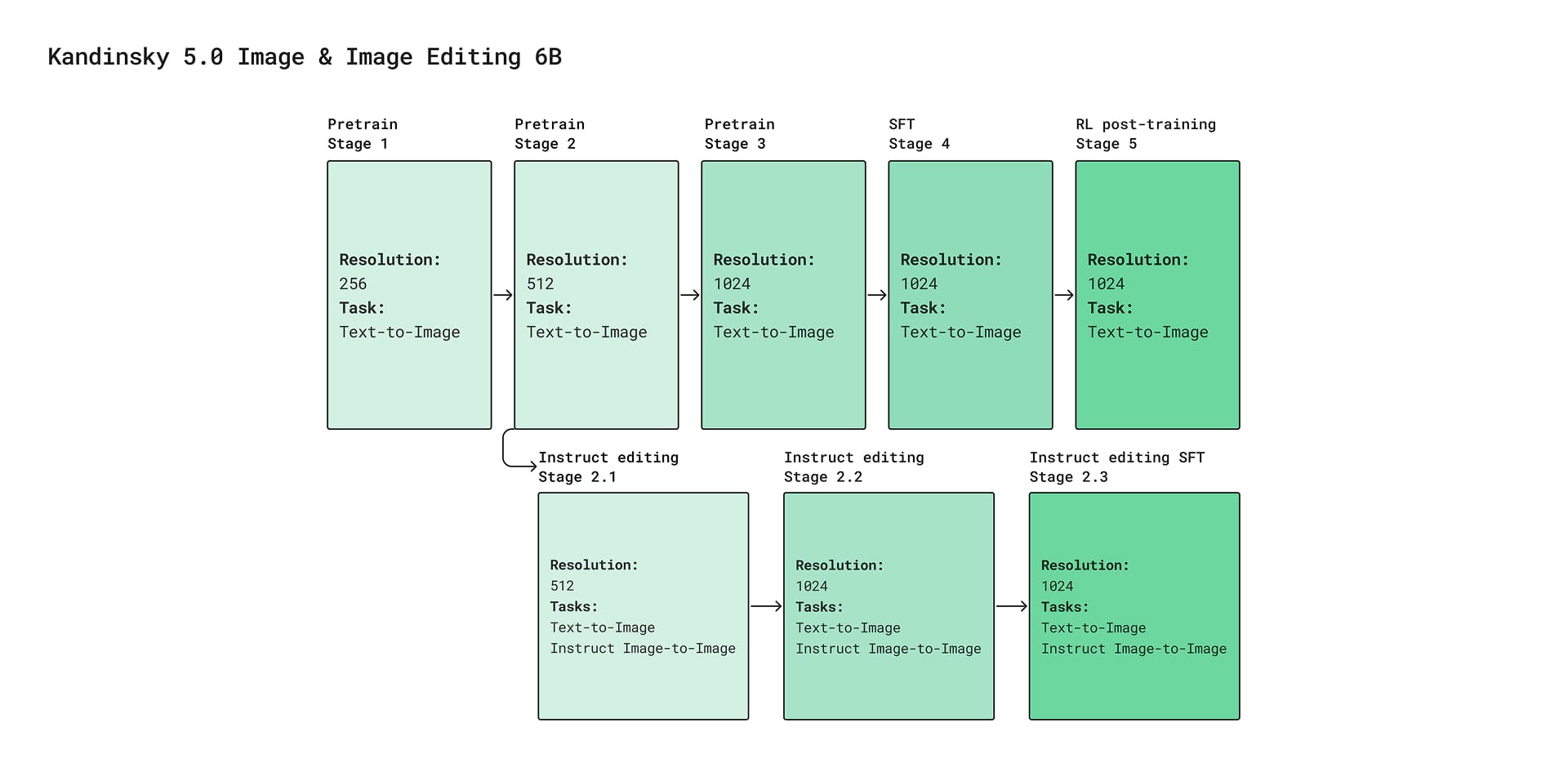}
    \includegraphics[width=0.87\textwidth]{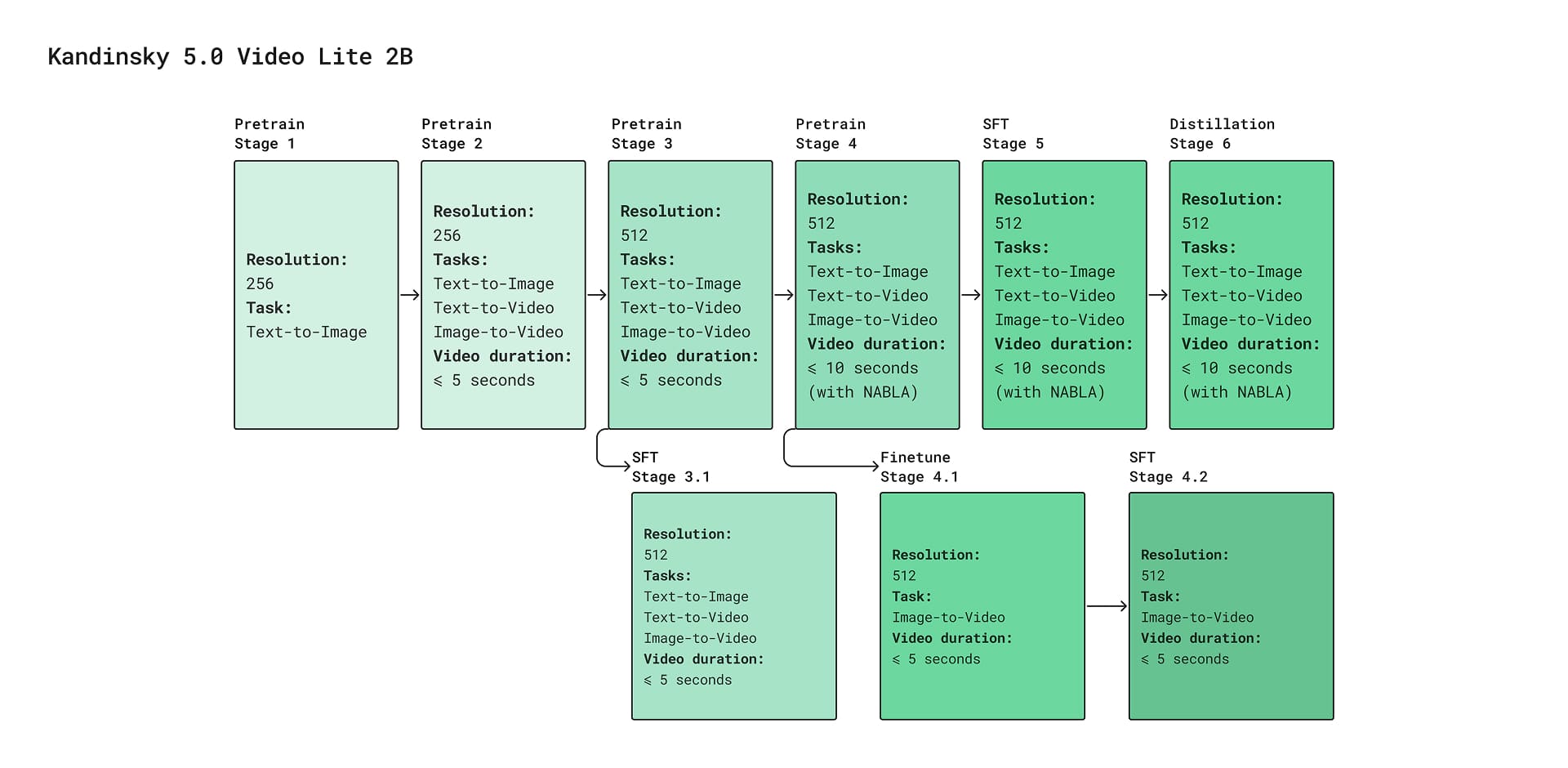}
    \includegraphics[width=0.87\textwidth]{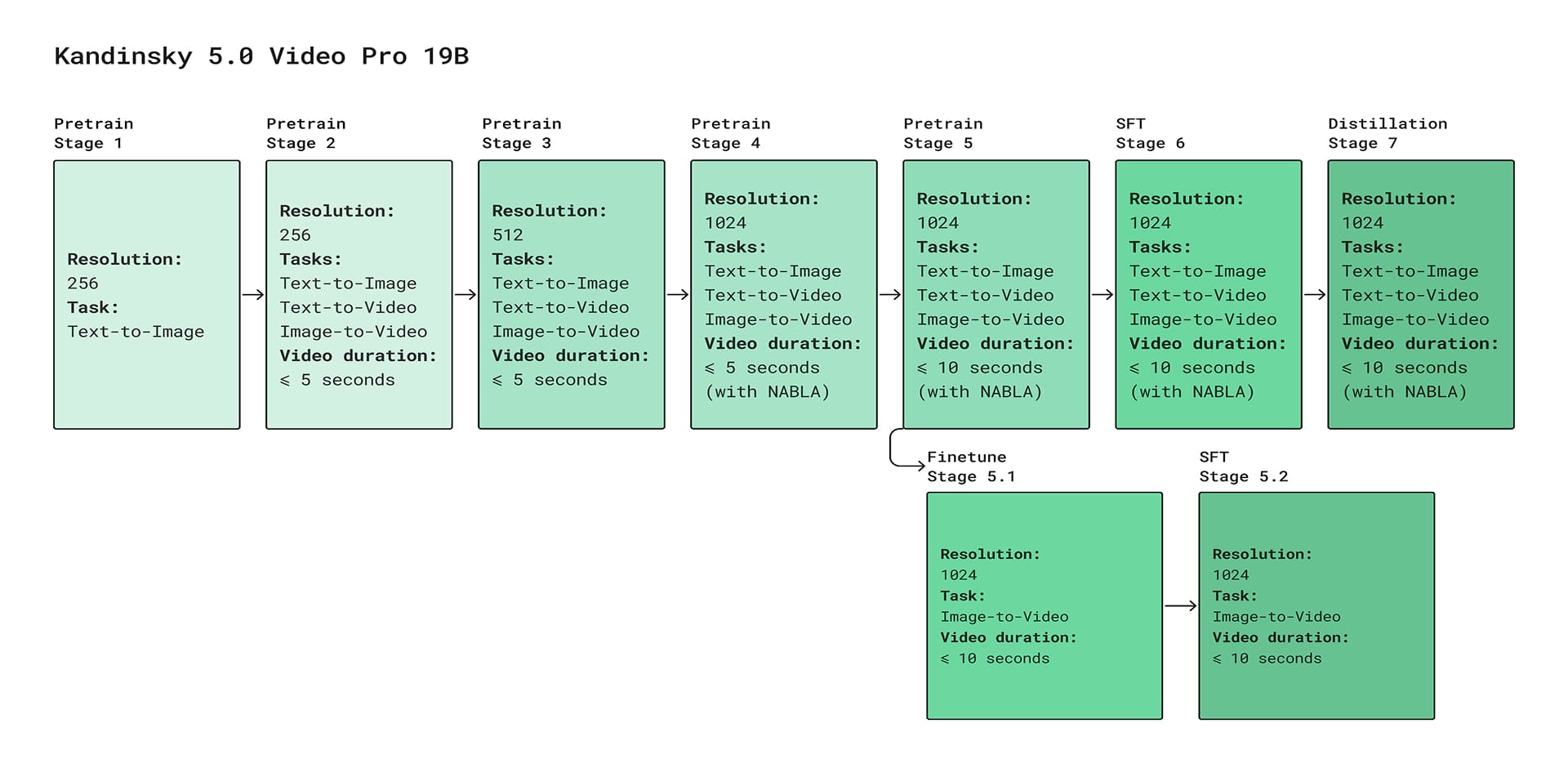}
    \caption{The training stages for models of the Kandinsky 5.0 family.}
    \label{fig:training_stages}
\end{figure}

The training process follows a multi-stage training pipeline, which includes \textbf{pre-training} (Section~\ref{sec:pretraining}), \textbf{supervised fine-tuning (SFT)} (Section~\ref{sec:sft}) and \textbf{distillation} (Section~\ref{sec:distillation}) for video models. The image generation model undergoes an additional \textbf{Reinforcement Learning (RL)-based Post-Training} stage (Section~\ref{sec:rl_posttrain}) for visual quality enhancement. Figure~\ref{fig:training_stages} illustrates the overall scheme of training stages. More specifically:

\paragraph{Kandinsky 5.0 Image Lite 6B models.} Both models share a common backbone from initial pre-training on text-to-image generation at \textbf{low} (192, 256, 320, 352, etc) and \textbf{medium} (384, 512, 640, etc) \textbf{resolutions}. Kandinsky 5.0 T2I Lite continues text-to-image pre-training at a \textbf{high resolution} (1024, 1280, 1408, etc) and then undergoes SFT and RL-based post-training for enhanced visual quality. Kandinsky 5.0 Image Editing inherits the checkpoint from the shared medium resolution pre-training stage and then undergoes two instructional editing fine-tuning stages at medium and high resolutions with subsequent SFT. During these three stages, the model performs instructional editing 80\% and text-to-image generation 20\% of cases.

\paragraph{Kandinsky 5.0 Video Lite 2B models.} The pre-training begins with independent text-to-image generation stage with a low resolution. Subsequently, during most stages, the model is trained to solve three tasks simultaneously -- text-to-image generation, text-to-video generation, and image-to-video generation with probabilities of 1\%, 79\%, and 20\%, respectively. This approach is applied during the pre-training stages at low and high resolutions with a maximum video length of 5 seconds, after which an SFT stage for this video length follows. During the pre-training stage with a maximum video length of 10 seconds, we employ the NABLA mechanism to enhance the efficiency of the attention operation (see Section~\ref{sec:nabla}). After this stage, we conduct fine-tuning for the image-to-video generation task and an SFT stage with a video length of 5 seconds to obtain a specialized 5-second version of the model for this task. In parallel, we continue to train the model for all three tasks simultaneously with a video length of 10 seconds during the subsequent SFT and distillation stages, ultimately yielding the text-to-video generation model \textbf{Kandinsky 5.0 T2V Lite}, image-to-video generation model \textbf{Kandinsky 5.0 I2V Lite} and accelerated version, \textbf{Kandinsky 5.0 Video Lite Flash}.

\paragraph{Kandinsky 5.0 Video Pro 19B models.} The pre-training for these models also begins with independent stage of text-to-image generation at a low resolution. We then conduct 4 stages of pre-training, as well as SFT, on the tasks of text-to-image, text-to-video, and image-to-video generation with probabilities of 2\%, 77\%, and 21\%, respectively. During pre-training stages, we increase the maximum resolution to 1024 pixels and the maximum video length from 5 to 10 seconds. For the high resolution, we use the NABLA mechanism (Section~\ref{sec:nabla}). We also conduct specialized fine-tuning on the image-to-video generation task for \textbf{Kandinsky 5.0 I2V Lite} model, followed by an SFT stage. As a result of distillation of the model, which solves the three aforementioned tasks, we obtain the accelerated version, \textbf{Kandinsky 5.0 Video Pro Flash}.

\subsection{Pre-training}\label{sec:pretraining}

\subsubsection{Regimes}

\begin{figure}[h!]
    \centering
    \begin{subfigure}[b]{0.495\textwidth}
         \centering
         \includegraphics[width=\textwidth]{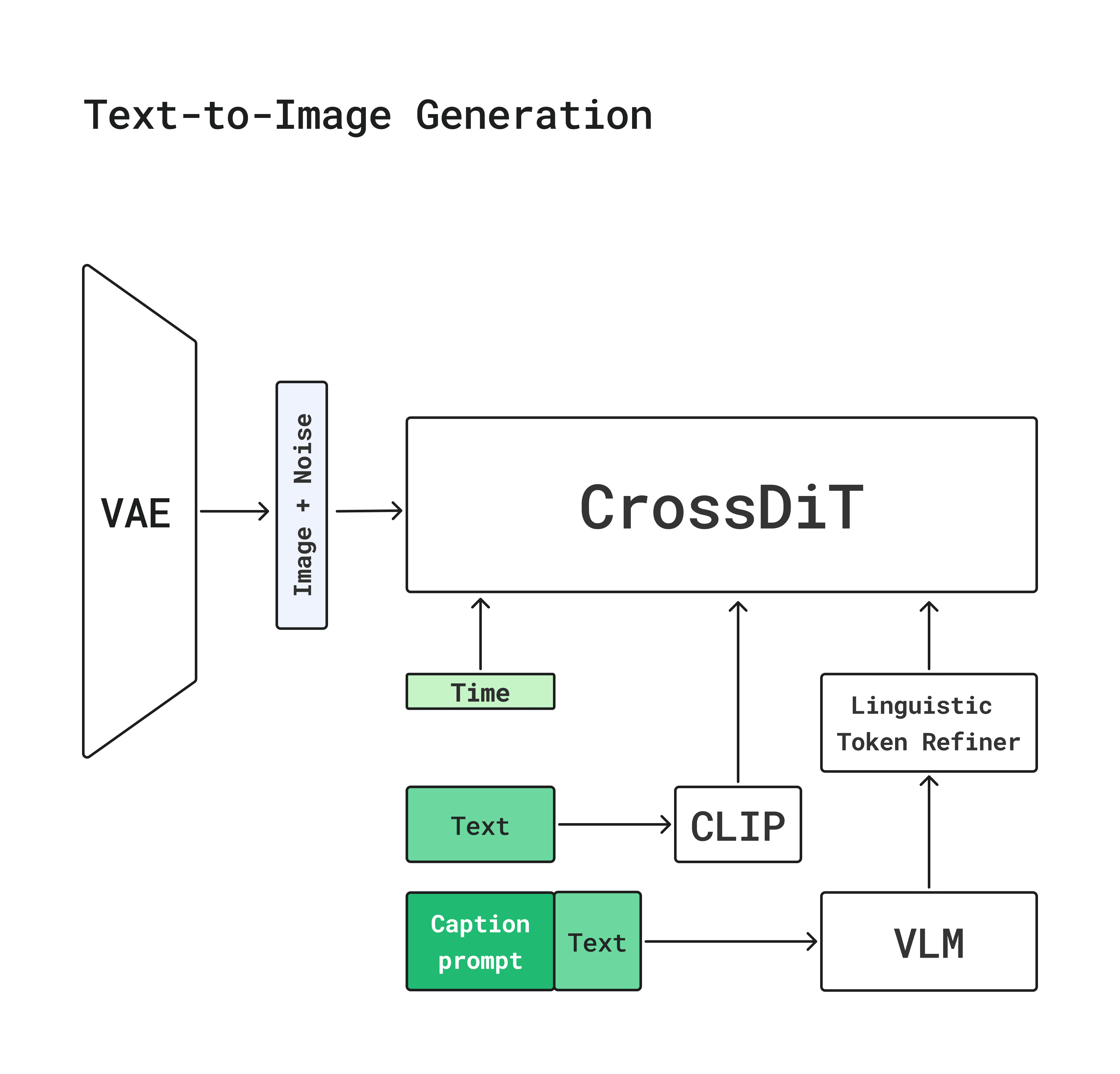}
         \caption{}
         \label{fig:pretraining_regimes_T2I}
     \end{subfigure}
     \hfill
     \begin{subfigure}[b]{0.495\textwidth}
         \centering
         \includegraphics[width=\textwidth]{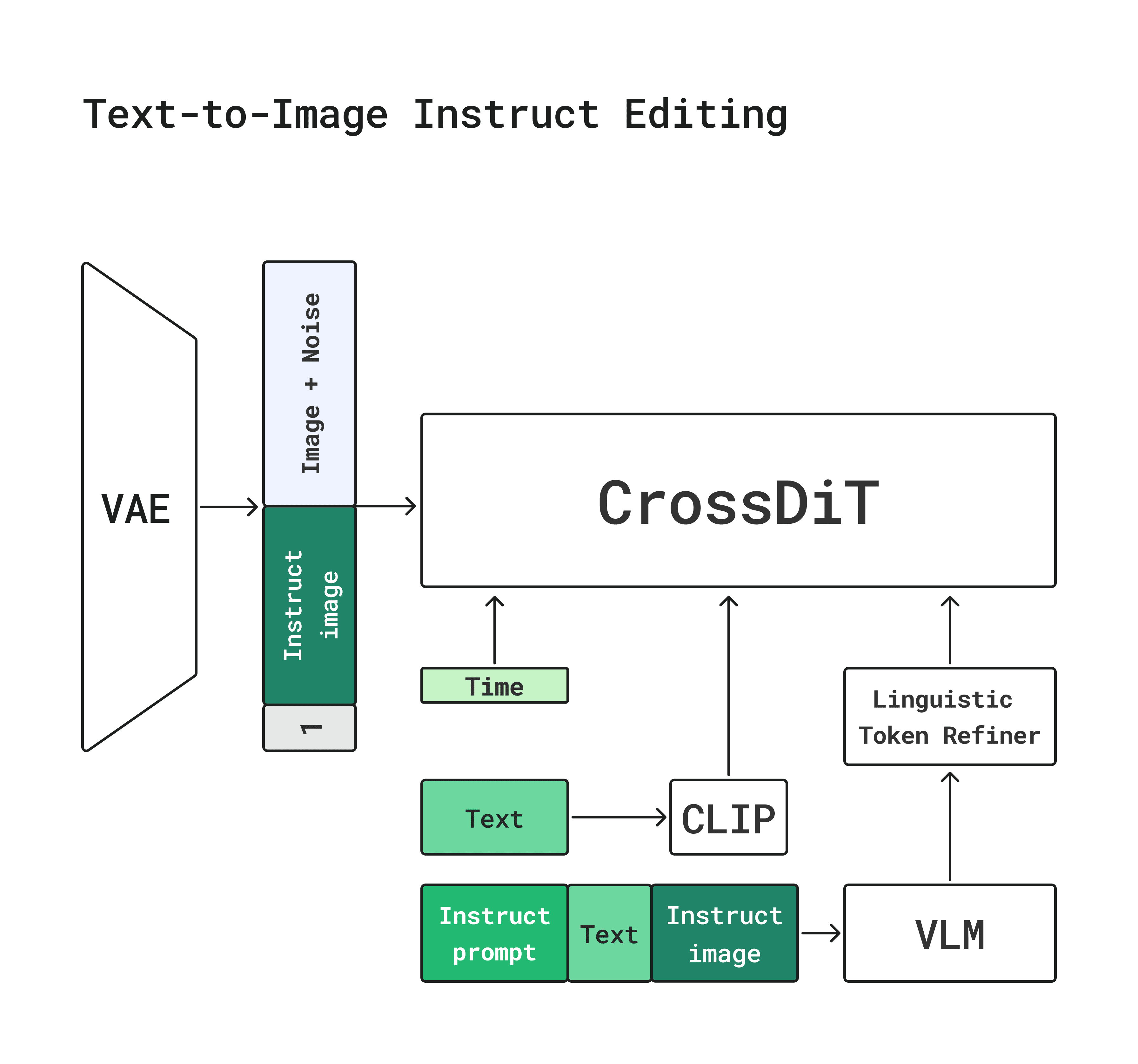}
         \caption{}
         \label{fig:pretraining_regimes_T2I_instruct}
     \end{subfigure}
     \hfill
     \begin{subfigure}[b]{0.495\textwidth}
         \centering
         \includegraphics[width=\textwidth]{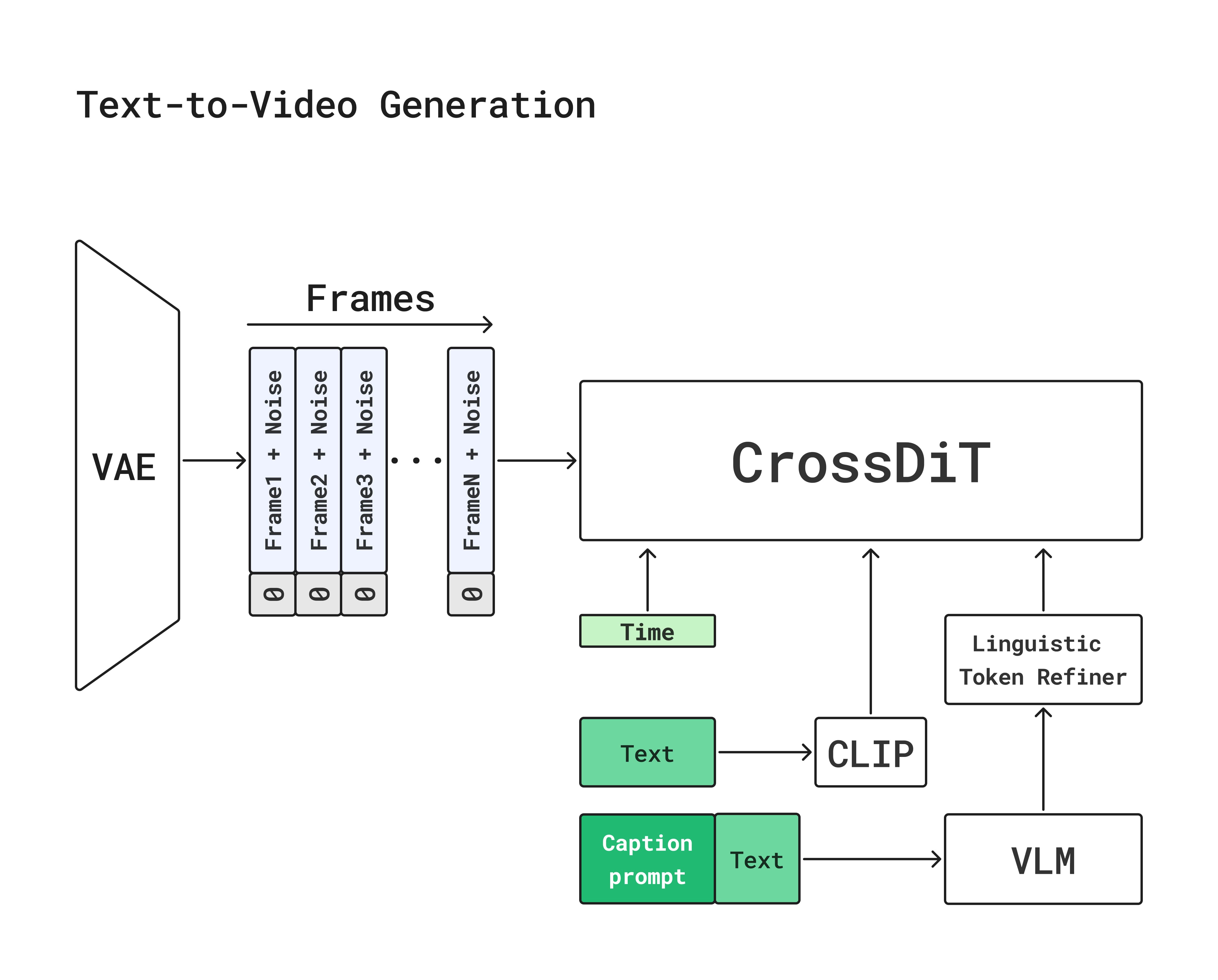}
         \caption{}
         \label{fig:pretraining_regimes_T2V}
     \end{subfigure}
     \hfill
     \begin{subfigure}[b]{0.495\textwidth}
         \centering
         \includegraphics[width=\textwidth]{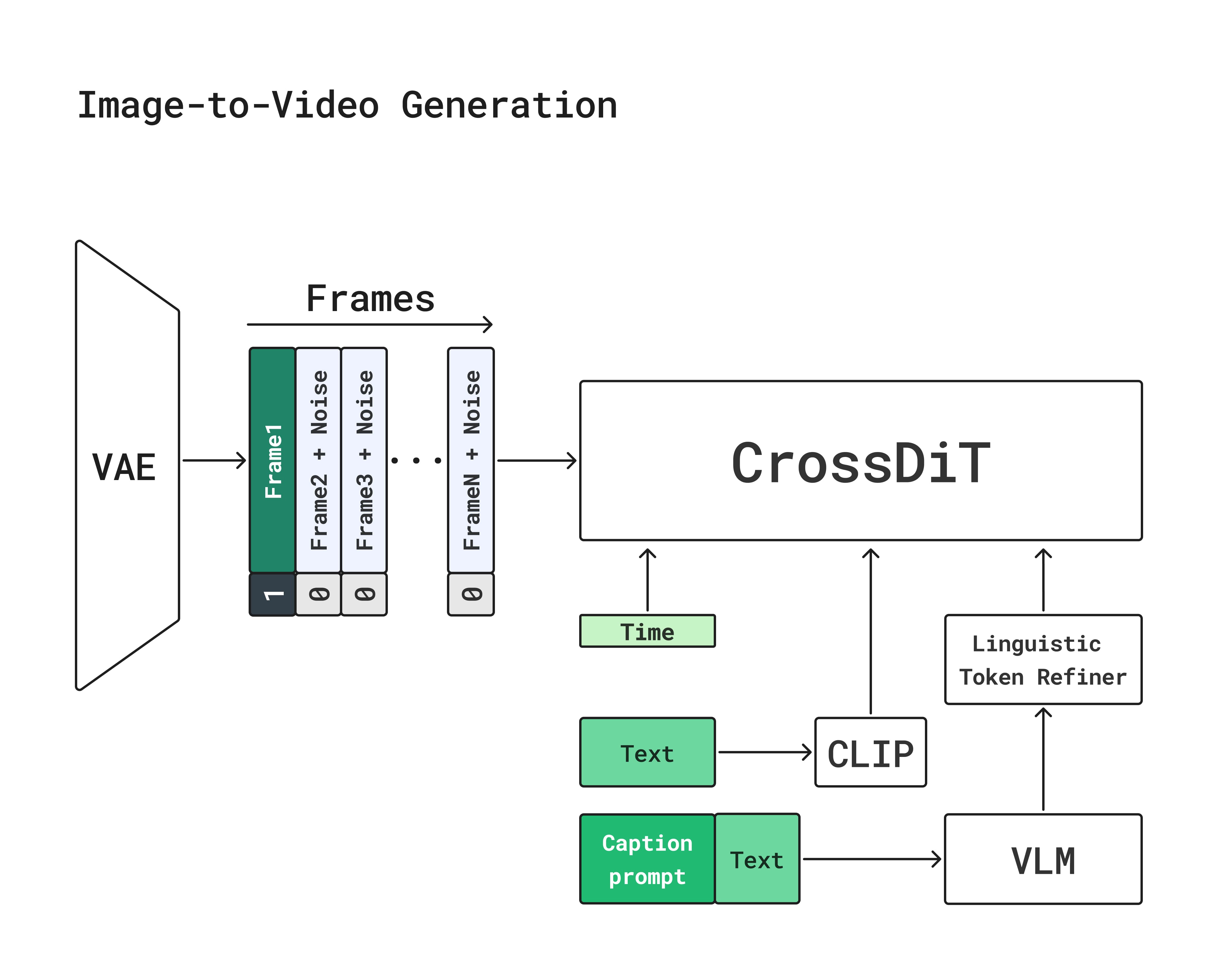}
         \caption{}
         \label{fig:pretraining_regimes_I2V}
     \end{subfigure}
    \caption{Pre-training setups for text-to-image generation (a), instruct image editing (b), text-to-video (c) and image-to-video regimes (d) for Kandinsky 5.0 models.}
    \label{fig:pretraining_regimes}
\end{figure}

We follow four core schemes for models pre-training in text-to-image, text-to-video, image-to-video, and instruct editing regimes (Figure~\ref{fig:pretraining_regimes}), each employing different weights of our CrossDiT architecture (Section~\ref{sec:cross_dit}). Below, we elaborate on each scheme in detail.

\paragraph{Text-to-Image Generation.} This scheme is applied for training all models, including the initial pre-training stages for video generation (see the training schemes in Figure~\ref{fig:training_stages}). The input to CrossDiT consists of a noisy image latent, the timestep in the diffusion process, and a text representation of the image description from the Qwen2.5-VL encoder~\cite{bai2025qwen25vltechnicalreport} using  the following template \textit{caption prompt}:

\begin{verbatim}
"<|im_start|>system\nYou are a promt engineer. Describe the image by detailing the
color, shape, size, texture, quantity, text, spatial relationships of the objects and
background:<|im_end|>",
"<|im_start|>user\n{}<|im_end|>".
\end{verbatim}

An image description is inserted into \texttt{user$\setminus$n\{\}}. This format of text input is more suitable for a VLM text encoder, which is trained on instructions~\cite{10.5555/3737916.3741676}. The model also receives a CLIP~\cite{radford2021clip} text embedding corresponding to the full image description. For pre-training the video model, a single-channel mask of zeros is added to the image.

\paragraph{Text-to-Image Instruct Editing.} This scheme is used for training Kandinsky 5.0 Image Editing. It differs from the previous one in that the input noisy image is channel-wise concatenated with an instruct image, which serves as a reference for the editing task, and a single-channel mask of ones. The instruct image is also fed into Qwen2.5-VL along with the textual image description, supplemented by the following \textit{instruct prompt}:

\begin{verbatim}
"<|im_start|>system\nYou are a promt engineer. Based on the provided source image
(first image) and target image (second image), create an interesting text prompt that 
can be used together with the source image to create the target image:<|im_end|>",
"<|im_start|>user\n{}".
\end{verbatim}

\paragraph{Text-to-Video Generation.} Here, the input to the video transformer is a sequence of noisy frames with a zero-valued single-channel mask. For 1024 resolution or for videos up to 10 seconds long at 512 resolution, CrossDiT incorporates the NABLA mechanism (Section~\ref{sec:nabla}); in other cases, full attention is applied to the video sequence. For this task, the corresponding caption prompt is also used to describe the video.

\paragraph{Image-to-Video Generation.} The first frame of the video sequence remains unnoised and, unlike the other frames, is accompanied by a single-channel mask of ones.

\subsubsection{Training details}

We train all our models at the pre-training stage with the AdamW optimizer~\cite{loshchilov2019decoupledweightdecayregularization} with \texttt{betas=(0.9, 0.95)} and \texttt{eps=1.0e-08}, changing the learning rate and \texttt{weight$\_$decay} depending on the training stage (see Table~\ref{tab:training_details}). We use scheduler warmup for all pre-training stages, limit the gradient norm to one, and also apply an exponential moving average with a parameter of 0.9999. During training, we use unconditional data examples with a probability of 0.1. Other parameters that depend on the models and training stages, such as the batch size, the number of optimizer steps, and the probabilities of spatial resolutions used in the trained data are presented in Table~\ref{tab:training_details}.

\begin{table}[h!]
\centering
\caption{Parameters for different stages of Kandinsky 5.0 models pre-training. Stages: LR -- low resolution, MR -- medium resolution, HR -- high resolution.}
\begin{tabular}{l c c c c c}
\hline
\textbf{Model \&} & \textbf{Training} & \textbf{Batch} & \textbf{Learning} & \textbf{Weight} & \textbf{Resolution} \\
\textbf{Training} & \textbf{steps} & \textbf{size} & \textbf{rate} & \textbf{decay} & \textbf{probabilities} \\
\textbf{stage} &  &  &  &  &  \\
\hline
Image Lite & 400k & 8000 & 1e-4 & 0.0 & 256$\times$256: 0.224, 192$\times$320: 0.11, \\
 (LR) &  &  &  &  & 320$\times$192: 0.332, 160$\times$352: 0.005, \\
&  &  &  &  & 352$\times$160: 0.012, 224$\times$288: 0.144, \\
 &  &  &  &  & 288$\times$224: 0.173 \\
Image Lite & 200k & 4000 & 4.0e-05 & 0.0 & 512$\times$512: 0.3152, 640$\times$384: 0.5301, \\
(MR) &  &  &  &  & 384$\times$640: 0.1547 \\
Image Lite & 200k & 2000 & 3.0e-05 & 0.0 & 1024$\times$1024: 0.183, 1408$\times$640: 0.017, \\
(HR) &  &  &  &  & 640$\times$1408: 0.010, 1280$\times$768: 0.315, \\
&  &  &  &  & 768$\times$1280: 0.115, 1152$\times$896: 0.175, \\
 &  &  &  &  &  896$\times$1152: 0.185 \\
 &  &  &  &  &  \\
Image Editing & 200k & 4000 & 4.0e-05 & 0.001 & 512$\times$512: 0.479, 384$\times$640: 0.4, \\
(MR) &  &  &  &  & 640$\times$384: 0.121 \\
Image Editing & 110k & 2000 & 2.0e-05 & 0.001 & 1024$\times$1024: 0.297, 640$\times$1408: 0.012, \\
(HR) &  &  &  &  & 1408$\times$640: 0.005, 768$\times$1280: 0.258, \\
&  &  &  &  & 1280$\times$768: 0.093, 896$\times$1152: 0.136, \\
&  &  &  &  & 1152$\times$896: 0.199 \\
&  &  &  &  &  \\
Video Lite & 220k & 8192 & 1e-4 & 0.0 & 256$\times$256: 0.25, 256$\times$384: 0.22, \\
(T2I, LR) &  &  &  &  & 384$\times$256: 0.53 \\
Video Lite & 10k & 2662 & 6.0e-05 & 0.001 & 256$\times$256: 0.25, 256$\times$384: 0.22, \\
(T2V + I2V, LR) &  &  &  &  & 384$\times$256: 0.53 \\
Video Lite & 50k & 1331 & 3.0e-05 & 0.001 & 512$\times$512: 0.28, 512$\times$768: 0.25, \\
(MR, 5 seconds) &  &  &  &  & 768$\times$512: 0.47 \\
Video Lite & 10k & 665 & 3.0e-05 & 0.001 & 512$\times$512: 0.28, 512$\times$768: 0.25, \\
(MR, 10 seconds) &  &  &  &  & 768$\times$512: 0.47 \\
 &  &  &  &  &  \\
Video Pro & 200k & 16384 & 1e-4 & 0.0 & 256$\times$256: 0.25, 256$\times$384: 0.22, \\
(T2I, LR) &  &  &  &  & 384$\times$256: 0.53 \\
Video Pro (LR) & 200k & 2662 & 1e-4 & 0.0 & 256$\times$256: 0.25, 256$\times$384: 0.22, \\
 &  &  &  &  & 384$\times$256: 0.53 \\
Video Pro (MR) & 45k & 1331 & 3.0e-05 & 0.001 & 512$\times$512: 0.28, 512$\times$768: 0.25, \\
 &  &  &  &  & 768$\times$512: 0.47 \\
Video Pro & 22.5k & 665 & 2.0e-05 & 0.001 & 1024$\times$1024: 0.183, 1408$\times$640: 0.017,\\
(HR, 5 seconds) &  &  &  &  & 640$\times$1408: 0.01, 1280$\times$768: 0.315, \\
&  &  &  &  & 768$\times$1280: 0.115, 1152$\times$896: 0.175,\\
&  &  &  &  & 896$\times$1152: 0.185 \\
Video Pro & 10k & 333 & 2.0e-05 & 0.001 & 1024$\times$1024: 0.183, 1408$\times$640: 0.017, \\
(HR, 10 seconds) &  &  &  &  & 640$\times$1408: 0.01, 1280$\times$768: 0.315, \\
&  &  &  &  & 768$\times$1280: 0.115, 1152$\times$896: 0.175, \\
&  &  &  &  & 896$\times$1152: 0.185 \\
\hline
\end{tabular}
\label{tab:training_details}
\end{table}

\subsection{Supervised Fine-tuning}\label{sec:sft}

This section outlines the general concept of supervised fine-tuning (SFT) for our models, following the established training protocol (see Figure~\ref{fig:training_stages}). The approaches described below are applicable for all SFT stages in the training procedure.

\subsubsection{Image Generation}

Following the pretraining stage, we perform supervised fine-tuning on a dataset of 153 thousand high-quality, realistic images with detailed text descriptions. Data selection was conducted by annotators based on a range of criteria, including technical quality, composition, perspective, lighting, colour palette, and overall aesthetics. For more details, please refer to Section~\ref{sec:sft_dataset_processing}.

We found that direct full fine-tuning of the pretrained model on the entire SFT dataset yielded unsatisfactory results; namely, it maintained a non-photorealistic style, degraded text alignment, and reduced the quality of text rendering within images. Consequently, we employ a \textbf{model souping} technique~\cite{wortsman2022modelsoup, biggs2024diffusionsoupmodelmerging} with \textbf{hierarchical clustering}. The source dataset is clustered into 9 thematic domains (e.g., ``people'', ``nature'') using a VLM. To further enhance composition, realism, and text rendering, each domain is subsequently segmented into 2-9 semantically homogeneous subdomains (Figure~\ref{fig:domain_distributions}).

We perform independent full fine-tuning on each subdomain with a batch size of 64 and a peak learning rate of 1e-5. Training on a subdomain is halted upon the emergence of generation artifacts on the validation set, such as geometrical distortions or colour aberrations. The resulting checkpoints within a single main domain are then aggregated via a weighted summation of their model parameters, with weights proportional to the square root of the size of each subdomain.

\textbf{The final model is obtained by performing a weighted averaging of the nine domain-specific models.} This technique leads to significant improvements across all target metrics in side-by-side human evaluation, enabling the model to achieve a high level of realism, text alignment, and compositional quality.

\subsubsection{Video Generation}

For video model, we investigated two approaches: standard fine-tuning and model souping. The video SFT dataset consists of approximately 2.8 thousand videos and 45 thousand images that underwent a strict manual selection process based on multiple criteria. For more details on the video data collection for SFT refer to Section~\ref{sec:sft_dataset_processing}.

During standard fine-tuning of the Kandinsky 5.0 Video Lite model, we observed overfitting after approximately 10 thousand steps. The best results for the 5-second model were achieved using an Exponential Moving Average (EMA) checkpoint after 10 thousand iterations.

\textbf{We implement the souping method for video similarly to the approach for images:} the data is partitioned into nine thematic domains using a VLM. We train a separate model on each domain and then perform uniform averaging of their weights. Although the individual domain models are prone to overfitting and generating artifacts, their averaged version demonstrates high visual quality and generation stability. In our internal side-by-side evaluations, the model obtained through souping outperformed the standard SFT model in visual quality, motion consistency, and absence of artifacts, while maintaining equivalent text alignment. Examples of comparing the generations before and after the SFT stage are presented in Figure~\ref{fig:progress}.

\subsection{Distillation}\label{sec:distillation}

\begin{figure}[h!]
    \centering
    \includegraphics[width=\textwidth]{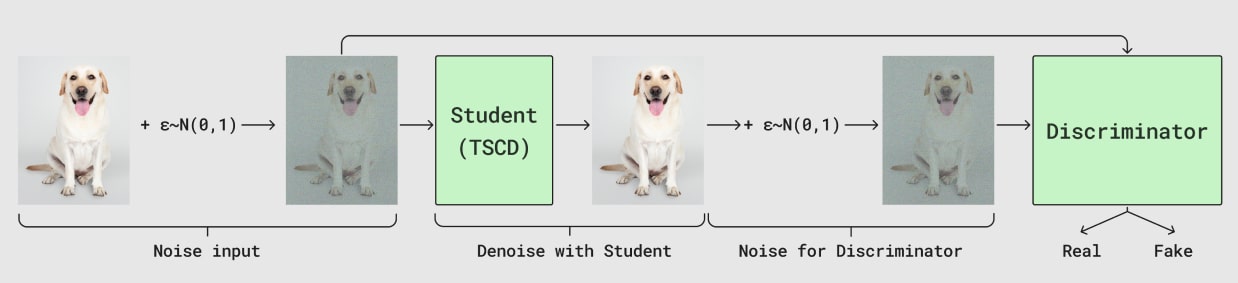}
    \caption{Adversarial post-training for diffusion distilled model.}
    \label{fig:distillation}
\end{figure}

We present two distinct classes of distilled models for video generation, each optimized for a different balance of sampling speed and visual fidelity:

\begin{enumerate}
    \item \textbf{Guidance Distilled Model.} The first model variant is produced via Classifier-Free Guidance (CFG) Distillation \cite{meng2023distillationguideddiffusionmodels}. This process directly distills the sampling trajectory of a base model, conditioned on a CFG scale, into a model that requires fewer number of function evaluations (NFEs). Starting from a base model requiring 100 NFEs, we applied CFG distillation with a null prompt and an optimal CFG scale of 5.0. This yielded a distilled model capable of generating samples of comparable quality in just 50 NFEs -- a 2x speedup.
    
    According to our internal side-by-side human evaluation, the CFG Distilled model preserves the semantic composition, texture detail, motion coherence, and prompt alignment of the original model without perceptible degradation.

    \item \textbf{Diffusion Distilled Model.} The second variant is a more aggressively distilled model, produced through a two-stage pipeline designed to maximize inference speed. This version is the basis for the \textbf{Kandinsky 5.0 Video Lite Flash} and \textbf{Kandinsky 5.0 Video Pro Flash} models.

    \begin{enumerate}
        \item \textbf{Initial CFG Distillation:} We first applied Trajectory Segmented Consistency Distillation (TSCD) \cite{ren2024hypersdtrajectorysegmentedconsistency} to the CFG-checkpoint, distilling it into a model requiring only 16 NFEs. This stage prioritizes a significant reduction in inference cost.
        \item \textbf{Adversarial Post-Training:} The model resulting from the first stage achieves high speed but exhibits a deficit in visual fidelity. To address this, we performed a second-stage refinement using an adversarial training framework inspired by LADD \cite{sauer2024fasthighresolutionimagesynthesis} and (Figure~\ref{fig:distillation}). 
        
        The adversarial training was conducted using a Hinge loss objective. We used the RMSprop optimizer~\cite{rmsprop} for both the generator (student model) and the discriminator, with learning rates of 1e-6 and 1e-4, respectively. Gradient clipping with a maximum norm of 1.0 was applied to stabilize training.

        A critical component for stabilizing our adversarial training was the application of a stochastic perturbation to images before feeding them to the discriminator. Specifically, both real and generated images are perturbed with Gaussian noise, with the noise level sampled from a Logit-Normal(-4, 1) timestep distribution. We found that this ``re-noising'' strategy, as also explored in \cite{lin2025diffusionadversarialposttrainingonestep}, significantly improves training stability compared to feeding clean images to the discriminator. Furthermore, this approach eliminates the need for an R1 gradient penalty, simplifying the training objective and reducing computational overhead.
    \end{enumerate}
\end{enumerate}

\subsection{RL-based Post-training for Image Generation}\label{sec:rl_posttrain}

\begin{figure}[h!]
    \centering
    \includegraphics[width=0.8\textwidth]{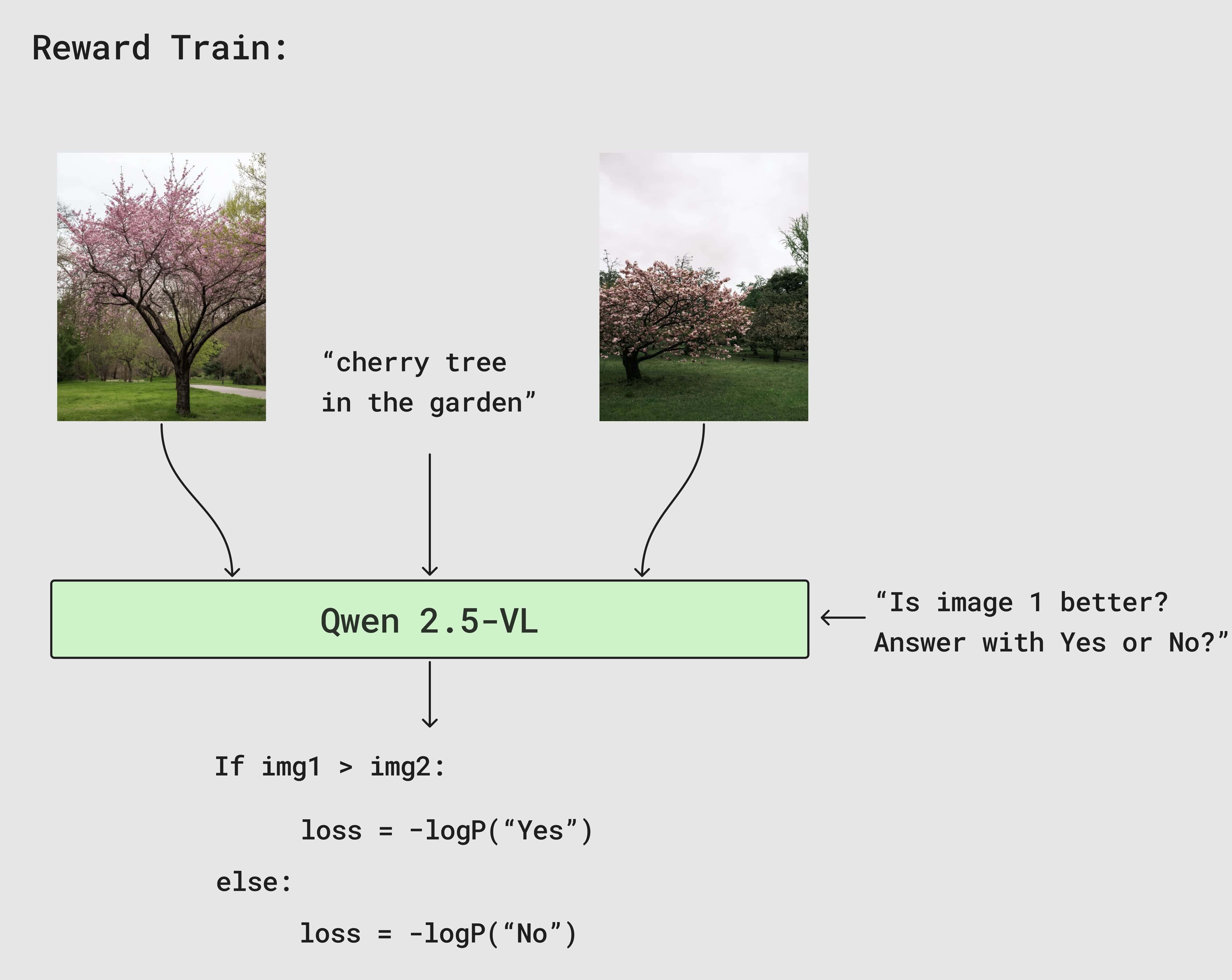}
    \caption{Reward model training.}
    \label{fig:reward}
\end{figure}

\begin{figure}[h!]
    \centering
    \includegraphics[width=0.32\textwidth]{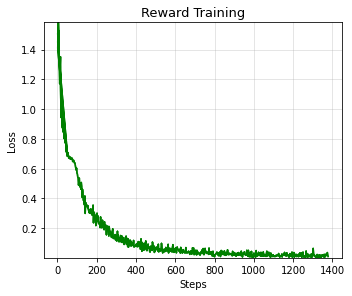}
    \includegraphics[width=0.32\textwidth]{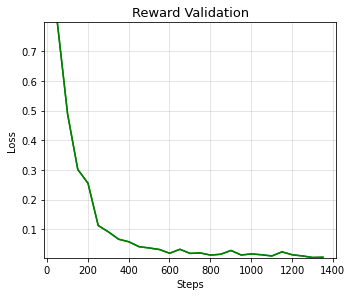}
    \includegraphics[width=0.32\textwidth]{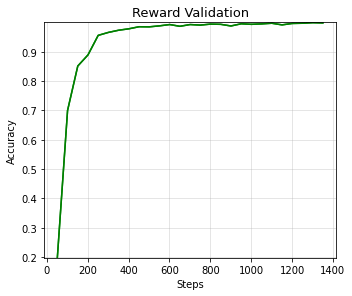}
    \caption{Training loss, validation loss and validation accuracy for reward model.}
    \label{fig:reward_loss}
\end{figure}

\begin{figure}[h!]
    \centering
    \includegraphics[width=0.7\textwidth]{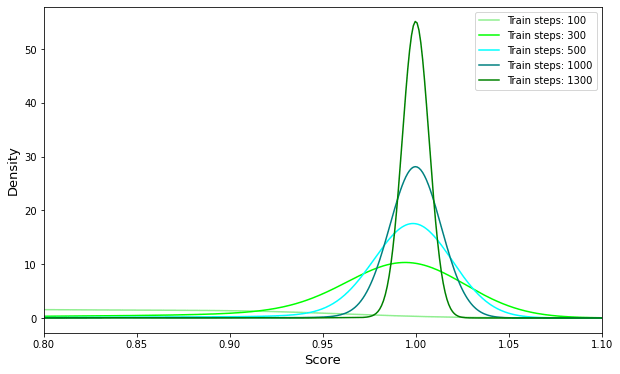}
    \caption{The scores distribution of the reward model on the validation set. As the number of steps increases, the distribution looks more like a delta function.}
    \label{fig:reward_scores}
\end{figure}

\begin{figure}[h!]
    \centering
    \includegraphics[width=0.9\textwidth]{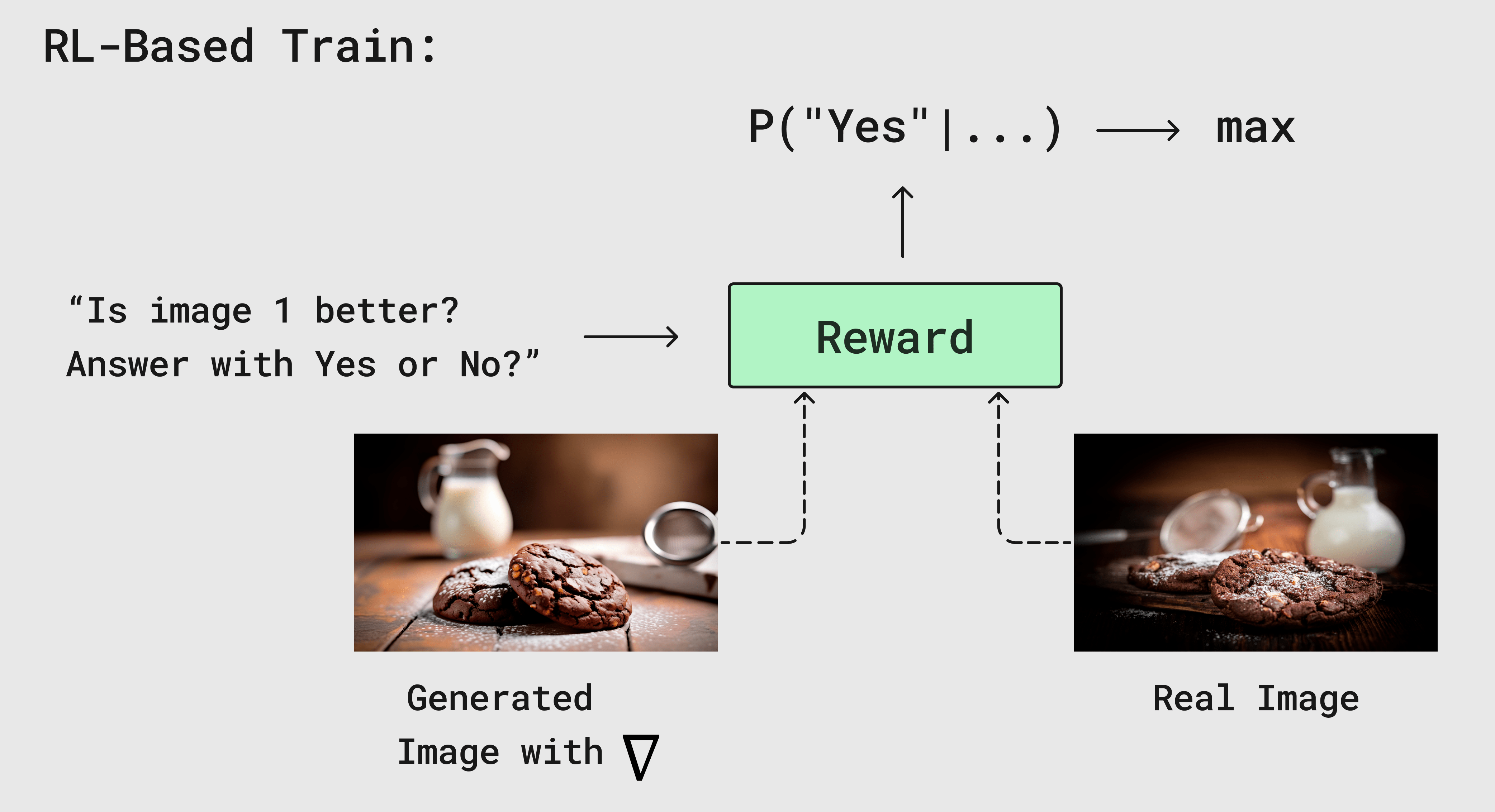}
    \caption{Fine-tuning of the image generation model with feedback from Reward Model.}
    \label{fig:RLHF}
\end{figure}

To further enhance the visual quality and realism of generated images, we apply RL-based post-training techniques, which consist of two main parts -- training a \textbf{reward model} (Figure~\ref{fig:reward}) and \textbf{RL-based} fine-tuning stage of the image generation model with feedback from trained reward model (Figure~\ref{fig:RLHF}).

\subsubsection*{Reward Model Training}

\textbf{Data.} We conduct experiments to train the reward model using real data collected for the supervised fine-tuning (SFT) stage (see Section~\ref{sec:sft_dataset_processing}), as well as generations from the Kandinsky 5.0 Image model after the pre-training and SFT stages. To form the dataset for reward model training, we used heuristic, that allowed us to skip the usual for RLHF~\cite{christiano2023deepreinforcementlearninghuman} (Reinforcement Learning on Human Feedback) data annotation process and still get very good results from RL-based post-training stage. We supposed that by design image generated from pre-train checkpoint is worse, than image generated from SFT checkpoint, which is worse, than the real, not generated, image from SFT set. This way, we collected our $(x_w,x_l,y)$ samples for reward model training.

\paragraph{Relative Reward Training.} Our approach is based on the Reward Dance method~\cite{wu2025rewarddancerewardscalingvisual}. For the reward, we are training a visual-language model, which we initialize from Qwen 2.5VL-7B~\cite{bai2025qwen25vltechnicalreport}. The reward model takes two annotated images with a text description as input, and then returns whether the first image is better or not. The model is trained to return ``Yes'', when the first image is better and ``No'', when the first image is worse. The value, that reward returns, that will be used as feedback for the generative model is the probability of the token ``Yes'' as the answer for whether the image on position 1 is better, i.e.: $R(x_1,x_2,y)=P(\text{``Yes''}|x_1,x_2,y)$.

Figure~\ref{fig:reward_loss} shows the training loss for the reward model, as well as the loss and accuracy values on the validation set for the reward outputs. Figure~\ref{fig:reward_scores} shows a plot of the reward model scores for different numbers of training steps on the validation set. It can be seen that as the number of steps increases, overfitting occurs and the plot increasingly resembles a delta function. During training, we monitor this plot and the standard deviation value $\sigma$ for Gaussian kernel density estimation to avoid overfitting. As a result, \textbf{we selected the reward model trained for 1300 steps}.

\subsubsection*{RL-based post-training}

\begin{figure}[h!]
    \centering
    \includegraphics[width=0.37\textwidth]{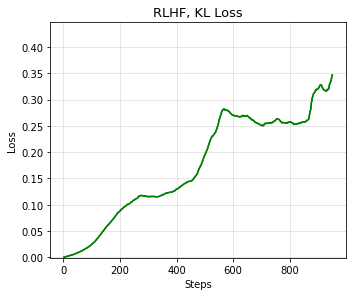}
    \includegraphics[width=0.37\textwidth]{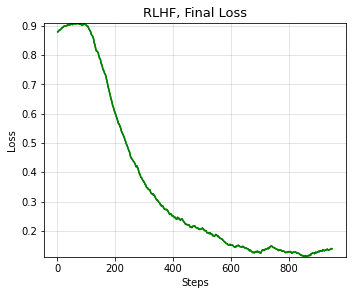}
    \caption{KL loss and final loss for RL-based post-training of image model. }
    \label{fig:rlhf_loss}
\end{figure}

For RL-based tuning we utilize Direct Reward Fine-Tuning (DRaFT) method~\cite{clark2024directlyfinetuningdiffusionmodels}, specifically we use the DRaFT-K version (Algorithm 1 from the original paper). Specifically, we generate an image using our model that has undergone the SFT stage and only backpropagate gradients through the last few generation steps. For the second image, we take a real image from the SFT dataset and maximize the probability that the generated image will be chosen as better by the reward model. We also use the Kullback–Leibler divergence between the distributions of images generated by the SFT-stage model $p_{SFT}$ and the model being trained $p_{RL}$. In the case of flow-matching models~\cite{lipman2023flow}, the KL-divergence takes the following form:

\[\text{KL}(p_{RL} \parallel p_{SFT}) = \sum_t \|v_{RL}(x_t, t) - v_{SFT}(x_t, t)\|^2,\]
where $v_{RL}$, $v_{SFT}$ are the outputs of the image generation model in the trainable ($p_{RL}$) and frozen post-SFT ($p_{SFT}$) versions respectively, $x_t$ is the denoising result after the first $T - t$ steps of the $p_{RL}$ model, and the sum is taken only over those final generation steps $t$ for which gradients are backpropagated through the $p_{RL}$ model. The final loss looks as follows:

\[L = L_{RL} + \beta_{KL} \cdot \text{KL}(p_{RL} \parallel p_{SFT}),\]
where $L_{RL}=1-R(x_\nabla,x_\text{real},y)$ is RL loss term, $x_\nabla$ is the image generated with gradients for the last K steps, as proposed by DRaFT-K algorithm from paper~\cite{clark2024directlyfinetuningdiffusionmodels}, $x_\text{real}$ is the real image from SFT set, that corresponds to the text prompt $y$. For text-to-image model, optimal $\beta_{KL}$ equals to 2e-2 and $K=10$ for DRaFT-K. The training plots for the generation model during the RL-based training are shown in Figure~\ref{fig:rlhf_loss}.

\paragraph{Other Experiments.} We also investigated other approaches, including the use of a differentiable absolute reward~\cite{clark2024directlyfinetuningdiffusionmodels, xu2023imagerewardlearningevaluatinghuman, liu2025improvingvideogenerationhuman}, as well as using two generated images for relative reward -- with and without gradient backpropagation on the final steps, selecting the best image from N generated ones as the reference image as proposed in ~\cite{wu2025rewarddancerewardscalingvisual}. We tried using two generations as input to the reward model, but eventually in our experiments we found that using a relative reward with a real image from the SFT dataset as a reference sample results in the best quality.

\section[Optimizations]{\color{Green}Optimizations}\label{sec:optimization}

\subsection{VAE Encoder Acceleration}

We have significantly accelerated the Hunyuan Video VAE encoder~\cite{kong2025hunyuanvideosystematicframeworklarge} through code optimization and by replacing several operations with more efficient equivalents. This resulted in an average 2.5$\times$ speed-up in the encoding process without requiring any additional training. Alongside the performance gains, we also improved video reconstruction quality: we found that temporal blending of tiles reduces the quality, fewer tiles means less quality loss.

Key enhancements include:
\begin{itemize}
    \item \textbf{Optimized Tiling.} Since Kandinsky 5 supports a limited set of input spatial resolutions, we determined optimal tile sizes. This approach efficiently utilizes GPU memory, improves GPU utilization, and minimizes artifacts at tile boundaries blending;
    \item \textbf{Integration of \texttt{torch.compile}.} We added support for \texttt{torch.compile} and identified a configuration that provides the optimal balance between compilation time and inference speed.
\end{itemize}

\subsection{CrossDiT Optimization}\label{sec:dit_optimization}

The primary computational bottleneck in image and video generation is the diffusion transformer, making its optimization especially critical. In addition to diffusion step distillation, described in Section~\ref{sec:distillation}, we apply the following optimization techniques:

\begin{enumerate}
    \item Detailed performance profiling and careful refactoring of the inference code to eliminate GPU idle time and achieve maximum efficiency with \texttt{torch.compile}.
    \item Caching diffusion steps using the MagCache~\cite{ma2025magcachefastvideogeneration} method. In our experiments, this technique delivered a 46\% speedup with no visible quality drop.
    \item For SD resolution and generation durations up to 5 seconds, we employ either Flash Attention~\cite{dao2023flash2, shah2024flashattention3fastaccurateattention} or Sage Attention~\cite{zhang2024sageattention2}, depending on the GPU hardware.
    \item For longer generation times (over 5 seconds) or HD resolution, we use our custom NABLA method (see Section~\ref{sec:nabla}), which accelerates inference by a factor of 2.7 compared to full attention.
\end{enumerate}

Table~\ref{table:performance_evaluation} summarizes the computational performance measurements for all models in the Kandinsky~5.0 family. Memory requirements can be decreased by using quantized text encoder.

\begin{table}[h!]
\centering
\begin{tabular}{l c c c c c}
\hline
\textbf{Model} & \textbf{Frame} & \textbf{Resolution} & \textbf{NFE} & \textbf{Generation} & \textbf{GPU Memory, GB} \\
 & \textbf{Number} &  &  &  \textbf{Time, s} & \textbf{(with offloading)} \\
\hline
Video Lite 5s & 121 & 512$\times$768 & 100 & 139 & 21 \\
Video Lite 10s & 241 & 512$\times$768 & 100 & 224 & 21 \\
Video Lite 5s Flash & 121 & 512$\times$768 & 16 & 35 & 21 \\
Video Lite 10s Flash & 241 & 512$\times$768 & 16 & 61 & 21 \\
Video Pro 5s & 121 & 512$\times$768 & 100 & 560 & 47 \\
Video Pro 10s & 241 & 512$\times$768 & 100 & 1158 & 51 \\
Video Pro 5s Flash & 121 & 512$\times$768 & 16 & 123 & 47 \\
Video Pro 10s Flash & 241 & 512$\times$768 & 16 & 242 & 51 \\
Video Pro 5s & 121 & 768$\times$1280 & 100 & 1241 & 53 \\
Video Pro 10s & 241 & 768$\times$1280 & 100 & 3218${}^{\ast}$ & 68\\
Video Pro 5s Flash & 121 & 768$\times$1280 & 16 & 235 & 53 \\
Video Pro 10s Flash & 241 & 768$\times$1280 & 16 & 576${}^{\ast}$ & 68 \\
Image Lite & 1 & 1024$\times$1024 & 100 & 13 & 17 \\
\hline
\end{tabular}
\captionsetup{justification=centering}
\caption{Kandinsky 5.0 computational performance evaluation. All measurements are performed using single 80 GB H100 GPU. \\
${}^{\ast}$Evaluation is performed with offloading enabled.}
\label{table:performance_evaluation}
\end{table}

\subsection{Training}

To estimate the training step duration and GPU memory consumption during the model training, we developed a mathematical model that relates these quantities to the main parameters of the training and model configuration.

\subsubsection{Training Step Estimation.}

The training step time is estimated using the following expression:
\begin{equation}
Step = \frac{d}{d_{0}} \cdot \frac{S}{S_{0}} \cdot 
\left( 9 + 14 \cdot \frac{S}{S_{0}} + 6 \cdot \frac{d}{d_{0}} \right)
\cdot L \cdot B,
\end{equation}
where:
\begin{itemize}
    \item $d_{0}$ and $S_{0}$ are constants: $d_{0} = 1792$ (hidden dimension for the 2B model) and $S_{0} = 256 \times 384 \times 31$ (reference video resolution);
    \item $d$ is the hidden dimension of the model being analyzed;
    \item $S$ is the video resolution;
    \item $L$ is the number of transformer blocks;
    \item $B$ is the batch size.
\end{itemize}

\subsubsection{GPU Memory Consumption.}

The GPU memory consumption is estimated as:
\begin{equation}
Memory = 12L \frac{(9d_{t}d + 8d^{2} + 2d_{f}d)}{N} 
+ \max \left( 4L \frac{(9d_{t}d + 8d^{2} + 2d_{f}d)}{N}, \,
2S(Ldo + 18d + 2d_{f}) \right),
\end{equation}
where:
\begin{itemize}
    \item $N$ is the number of GPUs;
    \item $S$ is the sequence length;
    \item $L$ is the number of blocks;
    \item $d$ is the hidden dimension ($hidden\_dim$);
    \item $d_{t}$ is the time dimension ($time\_dim$);
    \item $d_{f}$ is the feed-forward dimension ($ff\_dim$);
    \item $o = 0$ if activations are offloaded, otherwise $o = 1$.
\end{itemize}

This model enables fast analytical estimation of the scaling behaviour of the training pipeline with respect to model size, resolution, and hardware configuration and according to our experiments well fit real training step time and memory consumption. We used theoretical estimation for experiments planning, correct batch and parallel setup selection.

\section[{Results}]{\color{Green}Results}\label{sec:evaluation}

\subsection{Quality Progress}

\begin{figure}[htbp]
    \centering
    \begin{minipage}{\textwidth}
    	\centering
        \subfloat[Pre-training]{\includegraphics[width=0.28\linewidth]{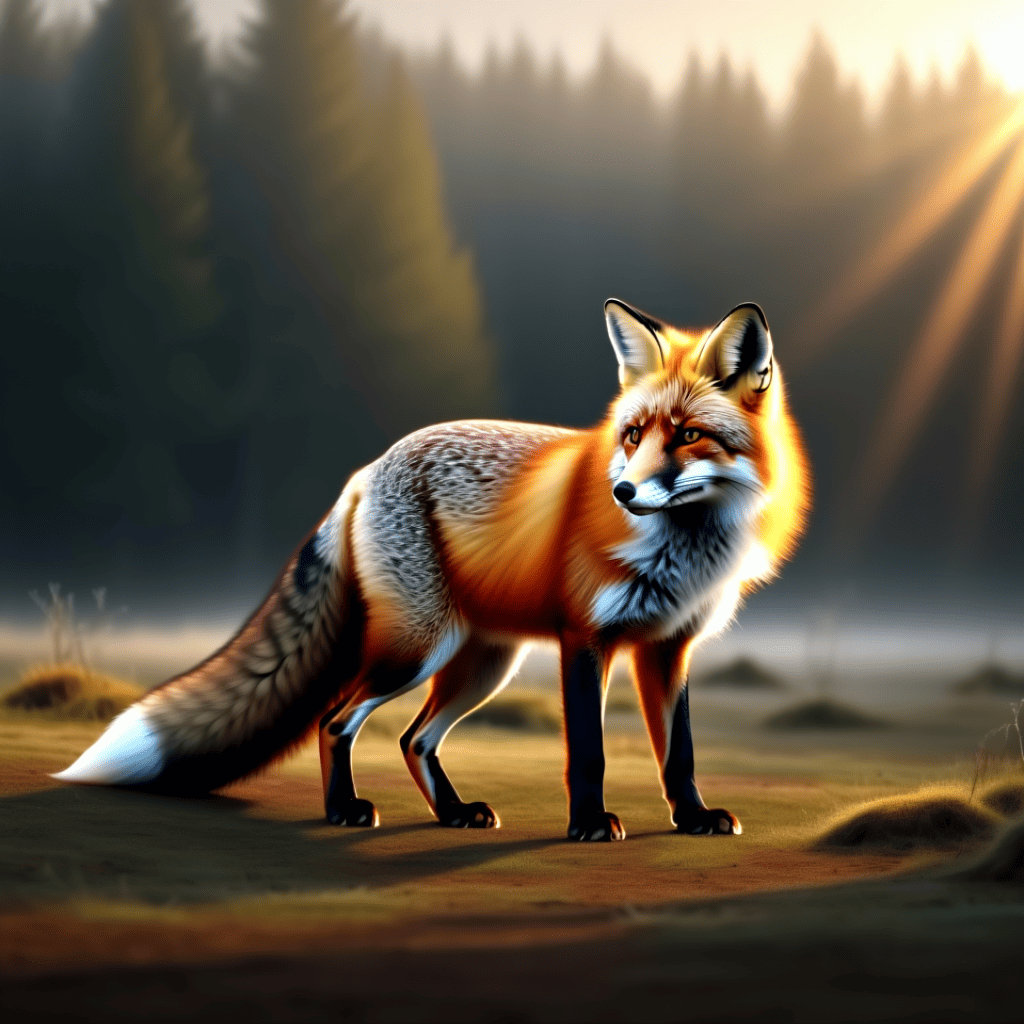}}
        \hfill
        \subfloat[SFT]{\includegraphics[width=0.28\linewidth]{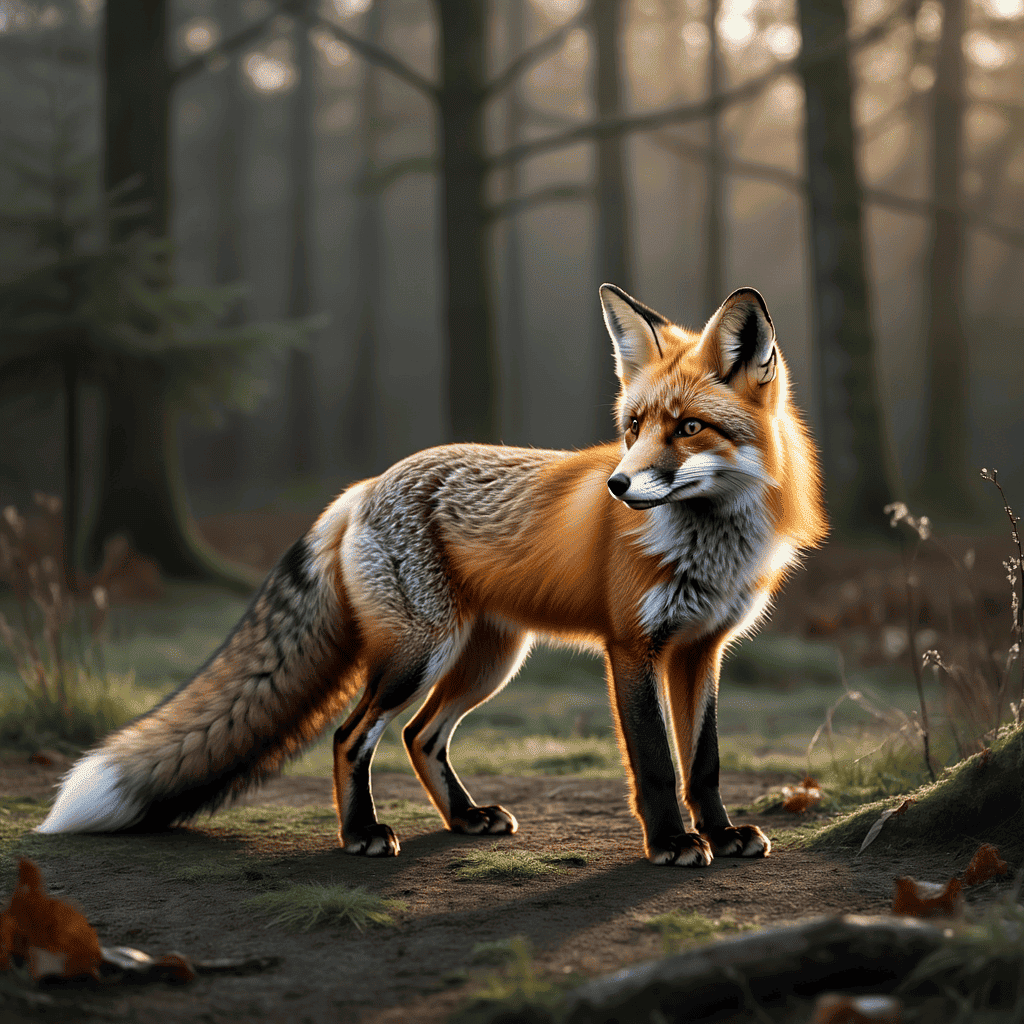}}
        \hfill
        \subfloat[RL post-training]{\includegraphics[width=0.28\linewidth]{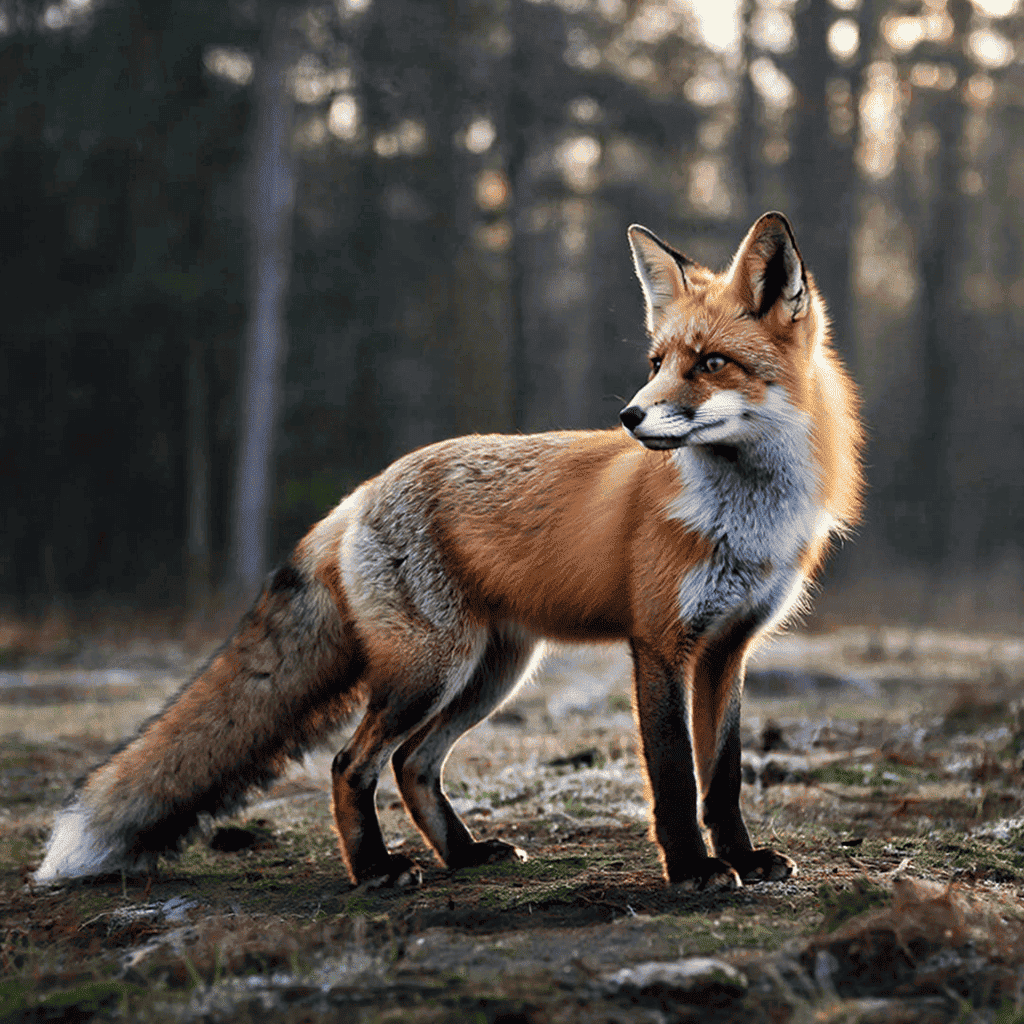}}
        \caption*{\tiny A graceful red fox with a fluffy tail stands in a forest clearing early in the morning. The rays of the rising sun gently illuminate its silky fur with golden hues. The fox looks attentively into the distance, ears alertly raised up. In the background, dense trees covered with morning dew create an atmosphere of mystery and tranquility. Detailed rendering of fur, eyes, and paws emphasizes the natural beauty of the animal. Realistic image with high resolution, focus on light and shadow transmission. Photographic accuracy, natural surroundings, no extraneous objects.}
    \end{minipage} \\
    \begin{minipage}{\textwidth}
    	\centering
        \subfloat[Pre-training]{\includegraphics[width=0.28\linewidth]{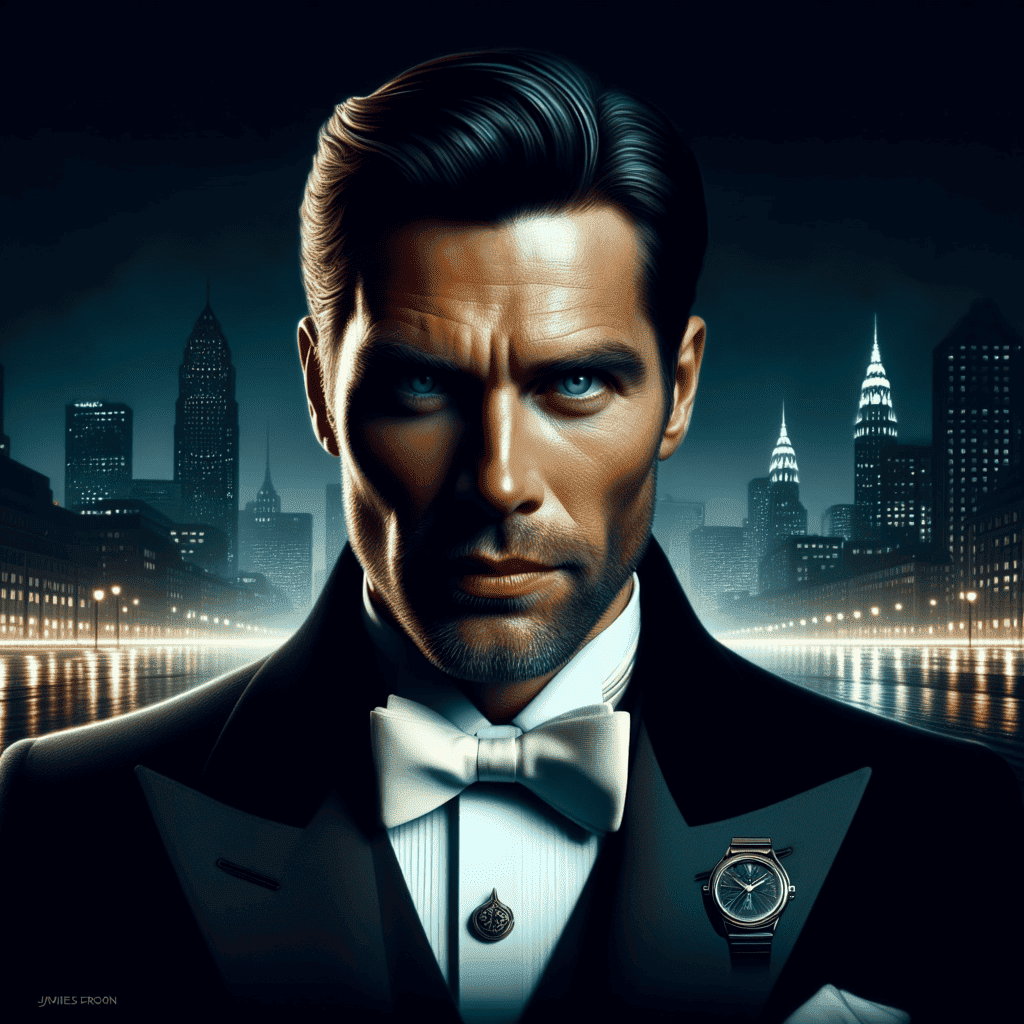}}
        \hfill
        \subfloat[SFT]{\includegraphics[width=0.28\linewidth]{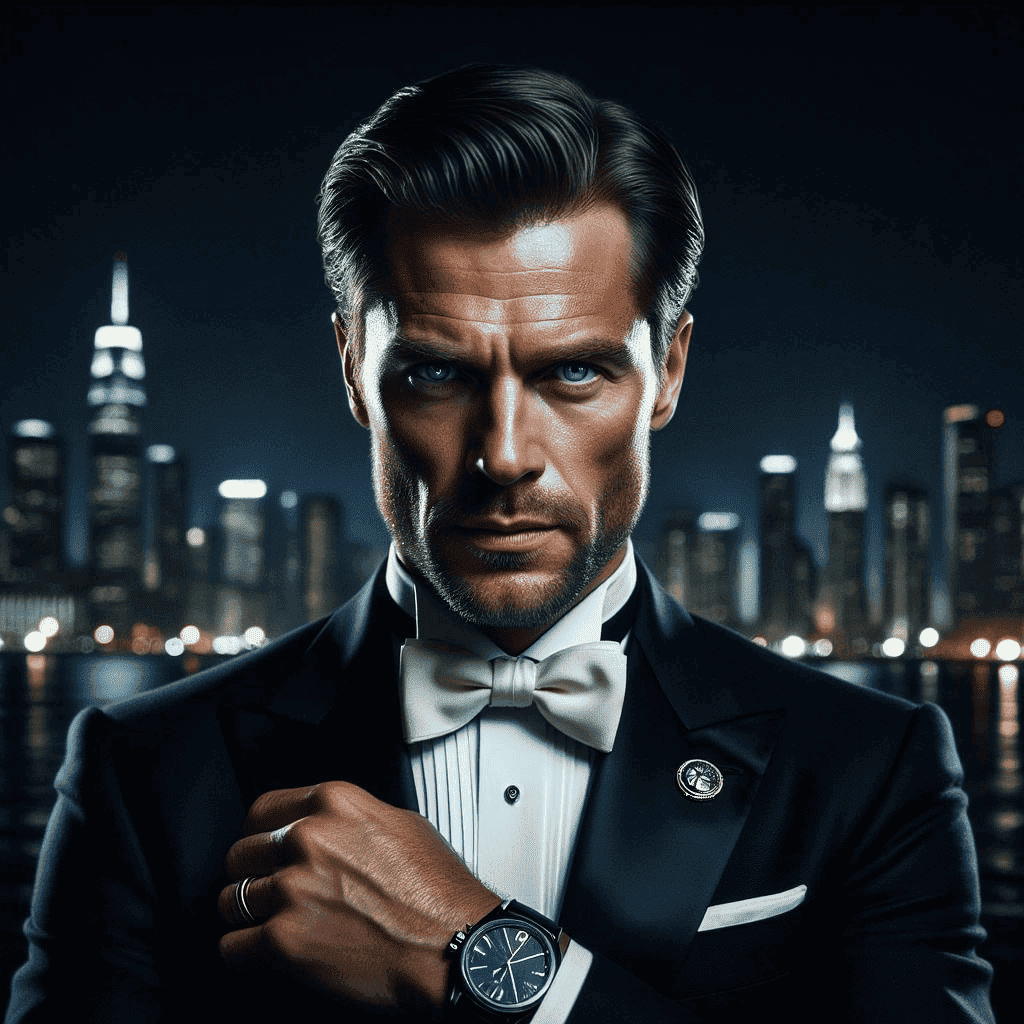}}
        \hfill
        \subfloat[RL post-training]{\includegraphics[width=0.28\linewidth]{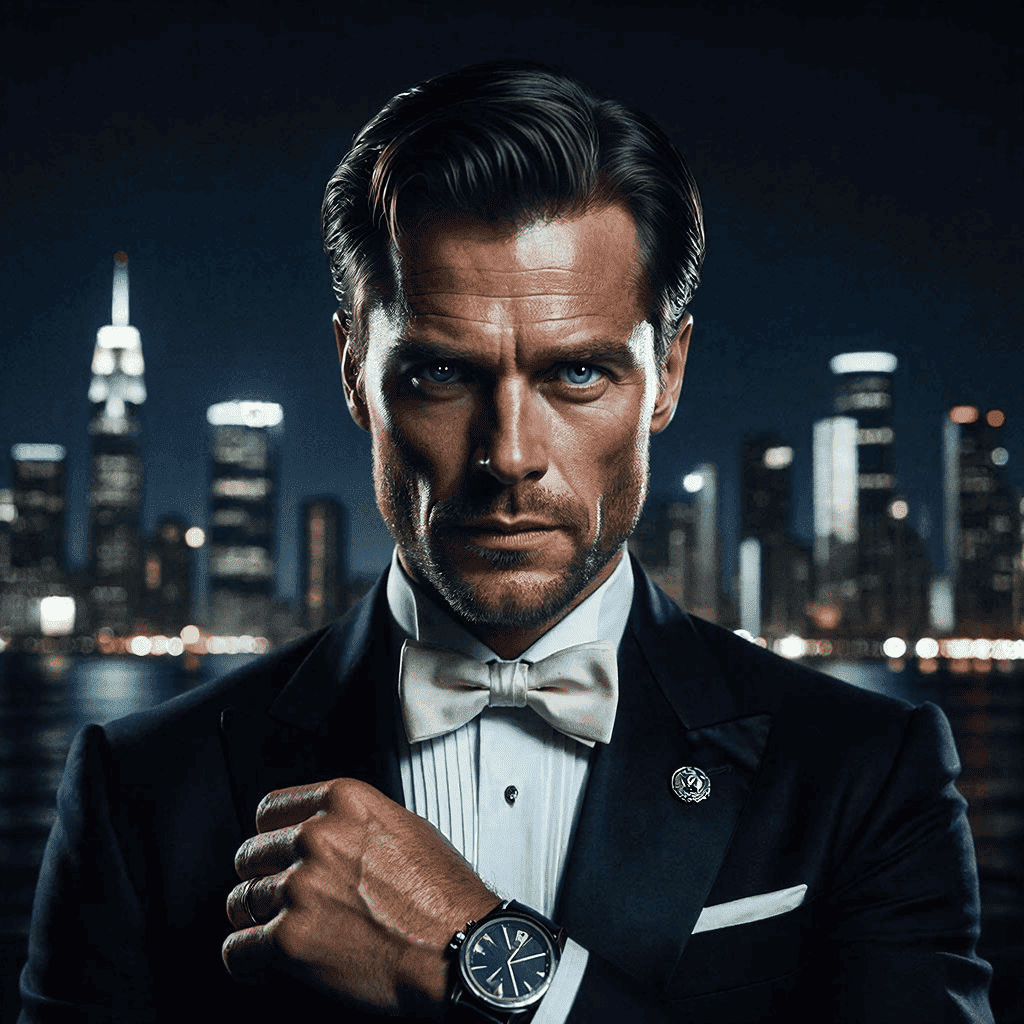}}
        \caption*{\tiny Close-up portrait of James Bond: a rugged face, confident gaze from cold blue eyes, perfectly styled dark hair, three-day stubble adds brutality to his image. He wears a classic black tuxedo with a white shirt and bow tie, an elegant tie clip, and premium wristwatch. Behind him is a night cityscape with skyscraper silhouettes and reflections of lights on wet pavement. The composition emphasizes an atmosphere of mystery and danger. Realistic depiction with deep shadows and sharp light accents, cinematic post-processing, high-quality photography.}
    \end{minipage}
    \begin{minipage}{\textwidth}
    	\centering
        \subfloat[Pre-training]{\includegraphics[width=0.28\linewidth]{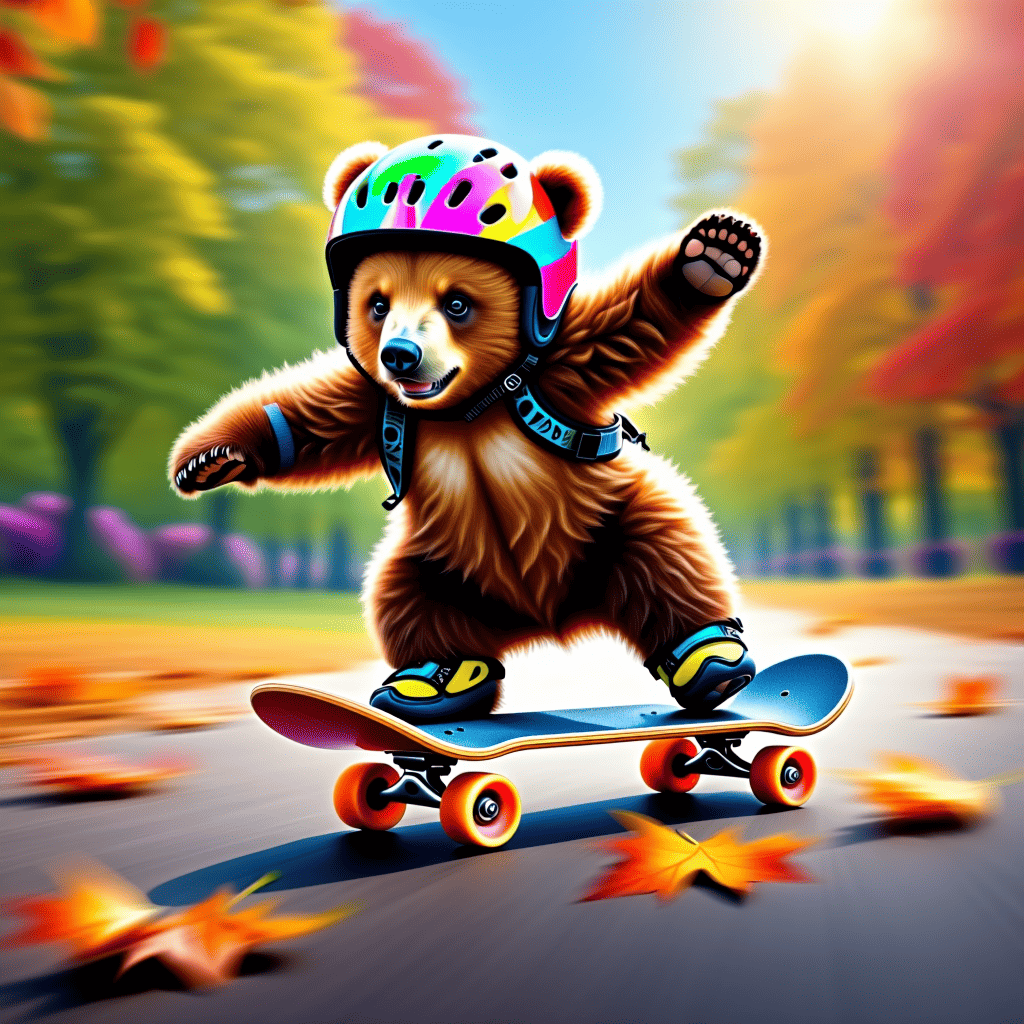}}
        \hfill
        \subfloat[SFT]{\includegraphics[width=0.28\linewidth]{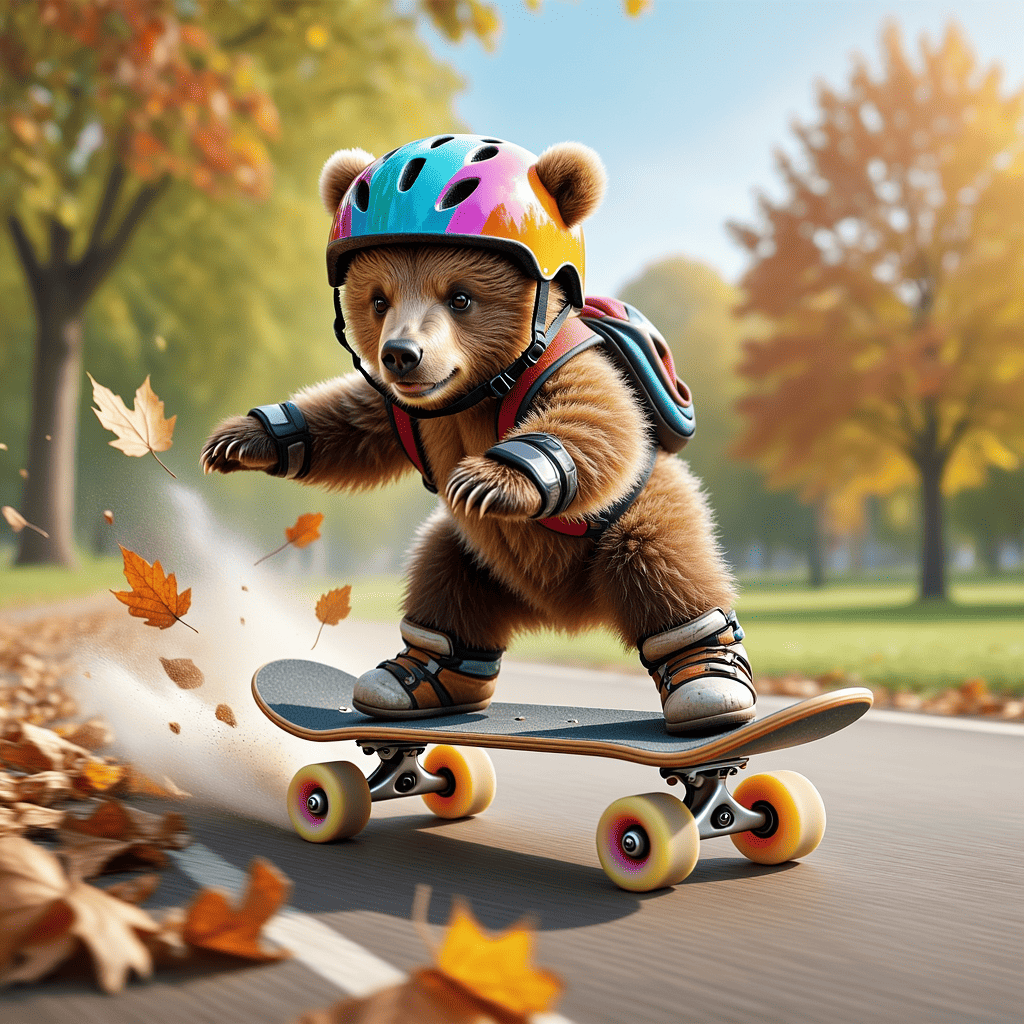}}
        \hfill
        \subfloat[RL post-training]{\includegraphics[width=0.28\linewidth]{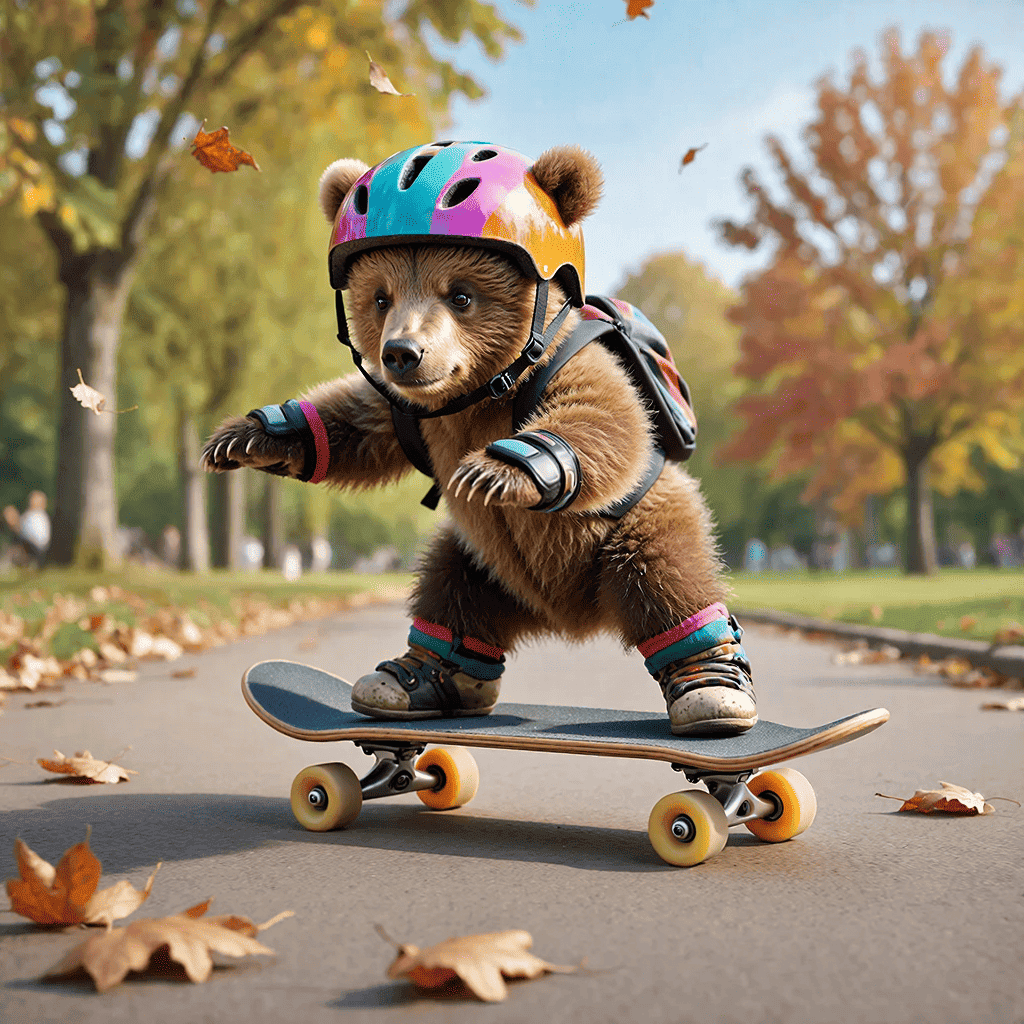}}
        \caption*{\tiny Frame in dynamic motion: a small bear cub confidently stands on a brightly colored skateboard, skillfully maneuvering among autumn leaves. The bear wears a colorful helmet and protective accessories that reflect safety and fun. Behind him is a picturesque park landscape with multicolored trees illuminated by soft sunlight. The animal's fur, board texture, and leaves are detailed to create a sense of realism. The artistic style combines elements of hyperrealistic photography and cute cartoon accents, emphasizing both naturalness and playfulness. The frame is filled with positive mood and carefree childhood atmosphere.}
    \end{minipage}
    \caption{Visual quality progress for text-to-image generation after different training stages. From left to right: pretraining, SFT, RL-based fine-tuning.}
    \label{fig:progress_image}
\end{figure}

\begin{figure}[h!]
    \centering
    \captionsetup[subfloat]{labelfont=scriptsize,textfont=scriptsize}
    {\label{subfig:155-pretrain}
        \includegraphics[width=0.195\textwidth]{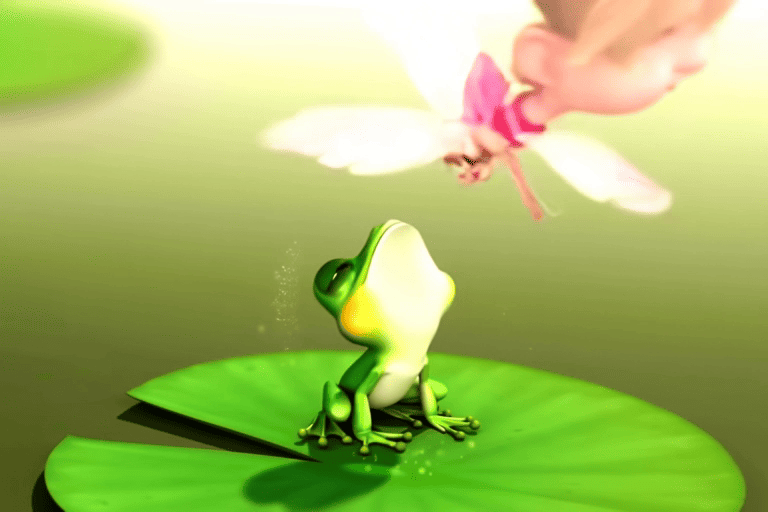} \hfill 
        \includegraphics[width=0.195\textwidth]{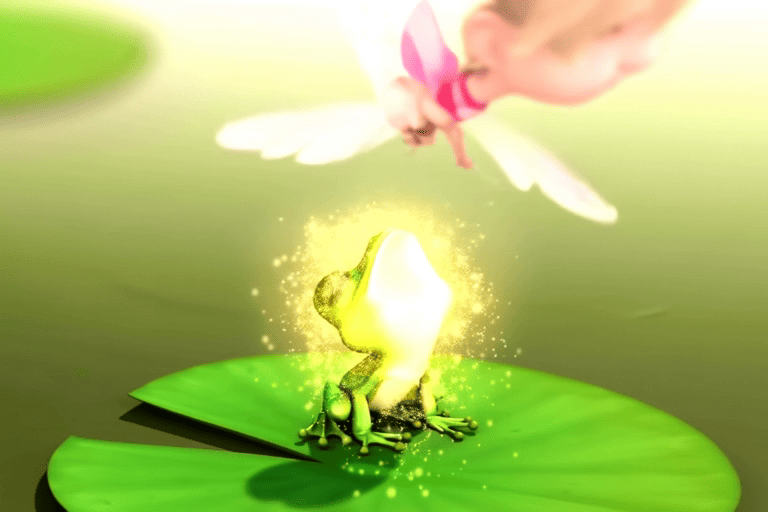} \hfill 
        \includegraphics[width=0.195\textwidth]{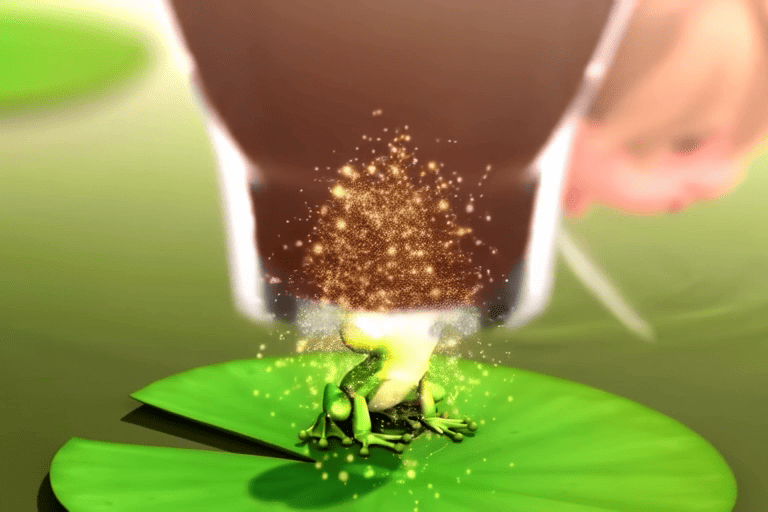} \hfill 
        \includegraphics[width=0.195\textwidth]{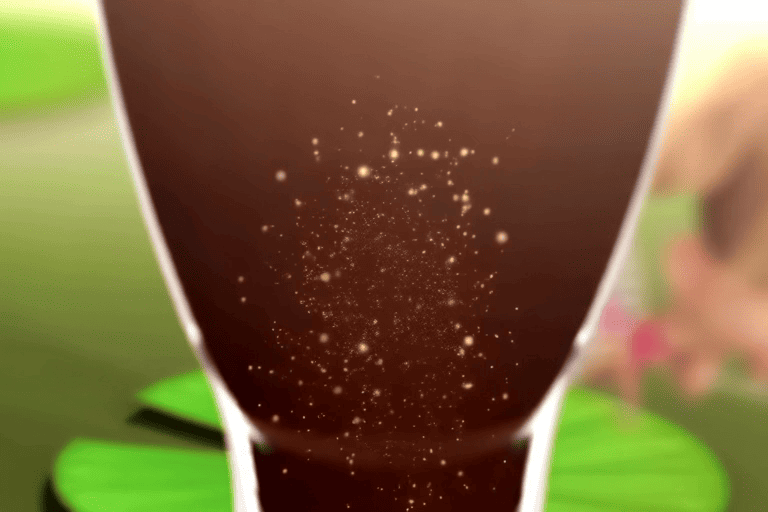} \hfill 
        \includegraphics[width=0.195\textwidth]{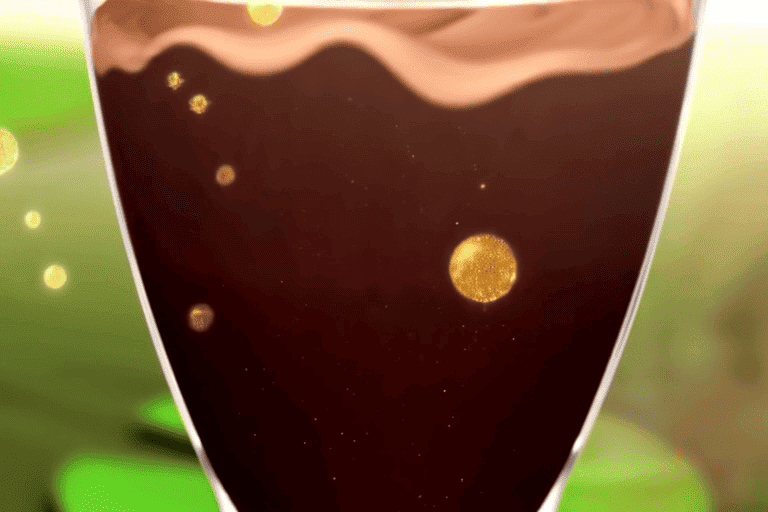}
    }
    \\
    \vspace{0.1cm}

    \subfloat[A small green frog sits on a lily pad in a calm pond, its eyes wide and curious. A gentle kiss lands on its forehead from a passing fairy, sending a shimmering glow across its body. The frog begins to shimmer and dissolve, its form transforming into a rich swirl of chocolate and milk. Bubbles rise as the once-living creature fully becomes a frothy, delicious chocolate milkshake in a glass. The transformation is magical and fluid, with golden sparkles floating around as the final product glistens under soft sunlight.]
    {\label{subfig:155-sft}
        \includegraphics[width=0.193\textwidth]{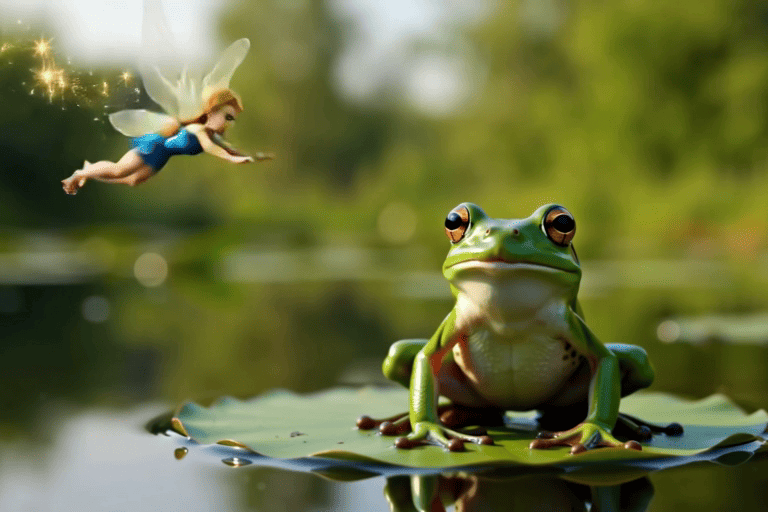} \hfill 
        \includegraphics[width=0.193\textwidth]{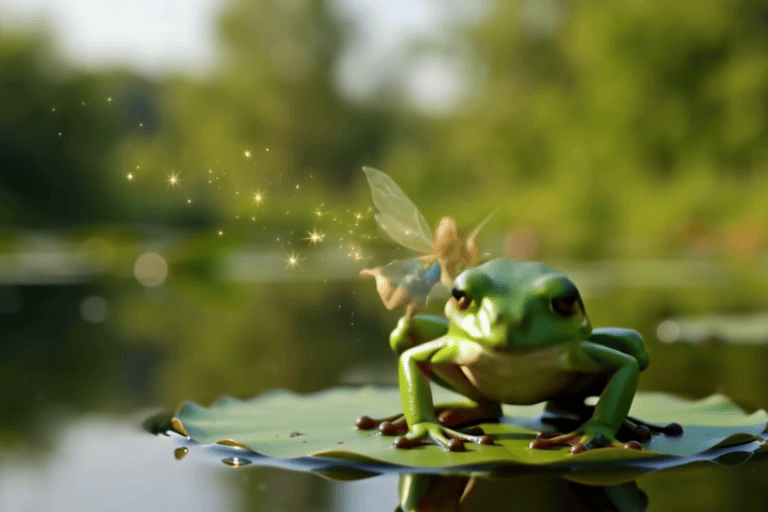} \hfill 
        \includegraphics[width=0.193\textwidth]{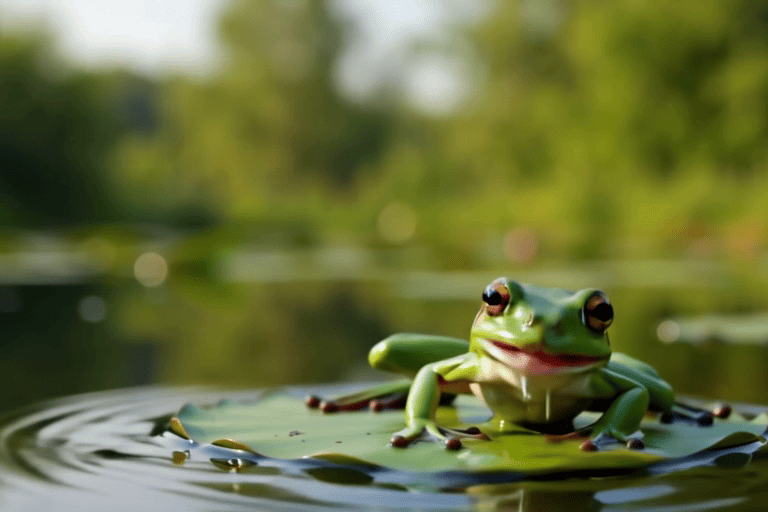} \hfill 
        \includegraphics[width=0.193\textwidth]{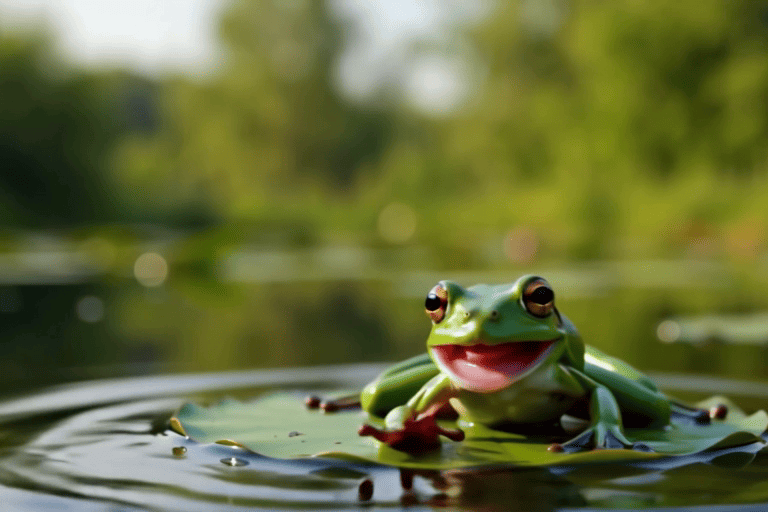} \hfill 
        \includegraphics[width=0.193\textwidth]{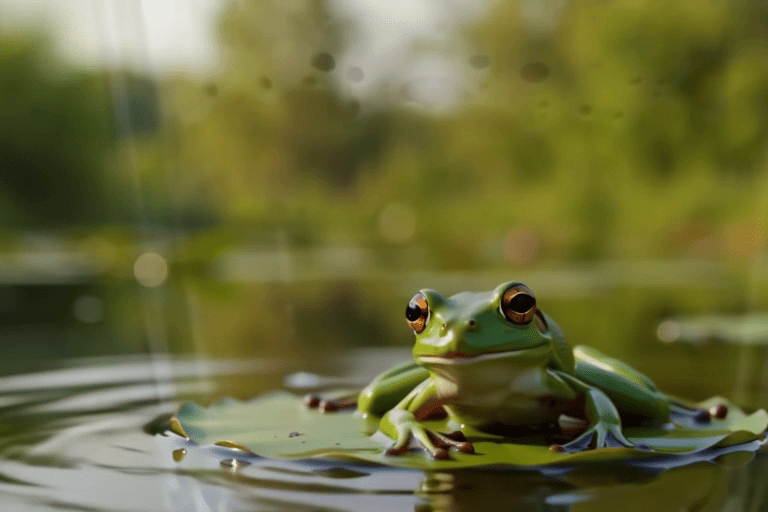}
    }
    \\
    \vspace{0.5cm}
    
    {\label{subfig:256-pretrain}
        \includegraphics[width=0.195\textwidth]{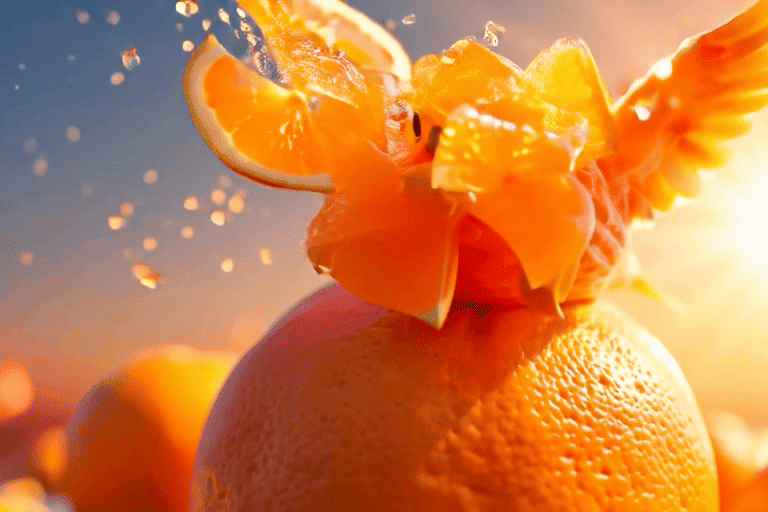} \hfill 
        \includegraphics[width=0.195\textwidth]{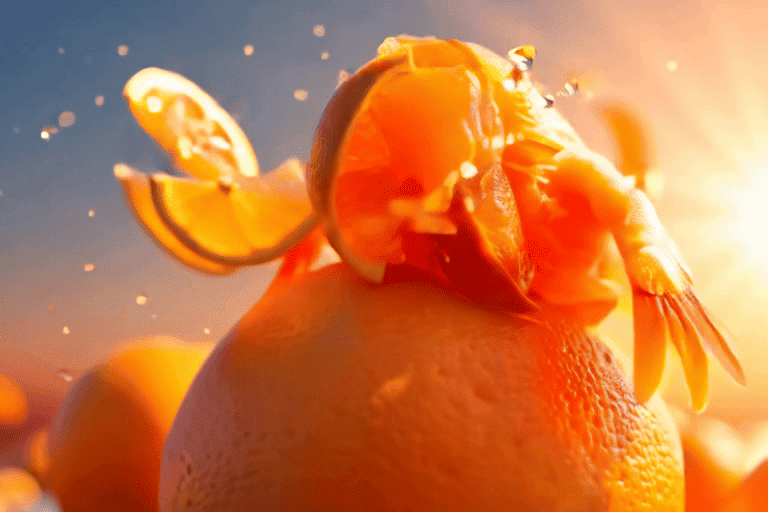} \hfill 
        \includegraphics[width=0.195\textwidth]{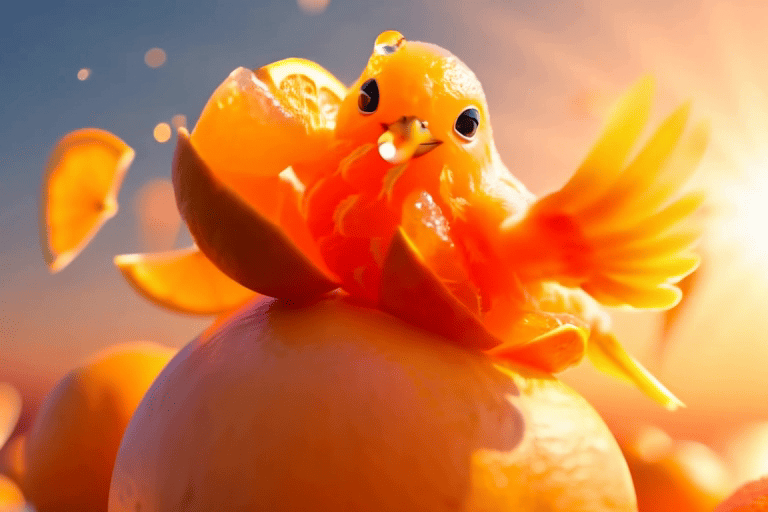} \hfill 
        \includegraphics[width=0.195\textwidth]{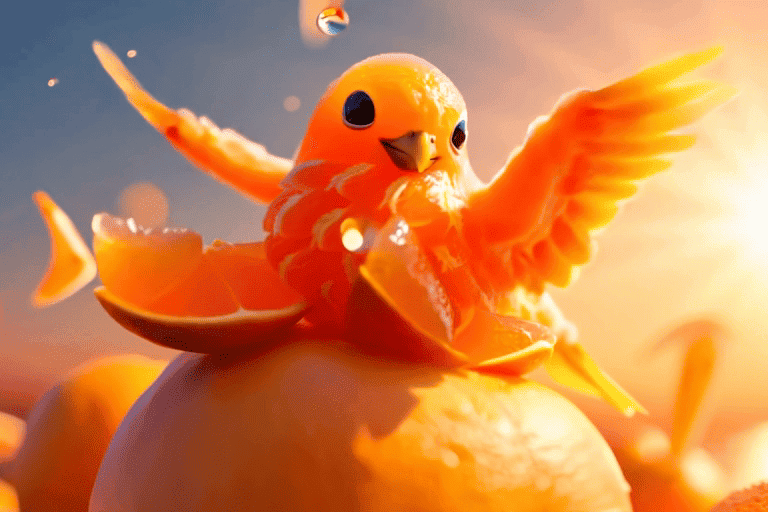} \hfill 
        \includegraphics[width=0.195\textwidth]{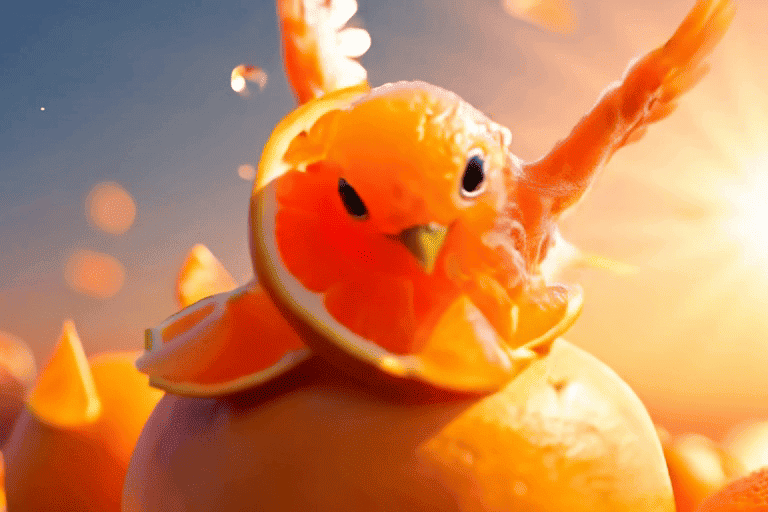}
    }
    \\
    \vspace{0.1cm}

    \subfloat[A vibrant bird crafted entirely from fresh, juicy oranges suddenly bursts forth from a large, ripe orange. Its wings, made of plump citrus slices, flap rapidly as it soars into the air, leaving behind a trail of sparkling orange juice droplets. Sunlight glimmers off the bird’s glossy orange feathers, casting a warm, golden glow around it. The bird twists and turns gracefully, its beak pecking playfully at floating citrus peels. As it flies higher, the background shifts to a bright orange sunset, completing the surreal and lively scene.]
    {\label{subfig:256-sft}
        \includegraphics[width=0.193\textwidth]{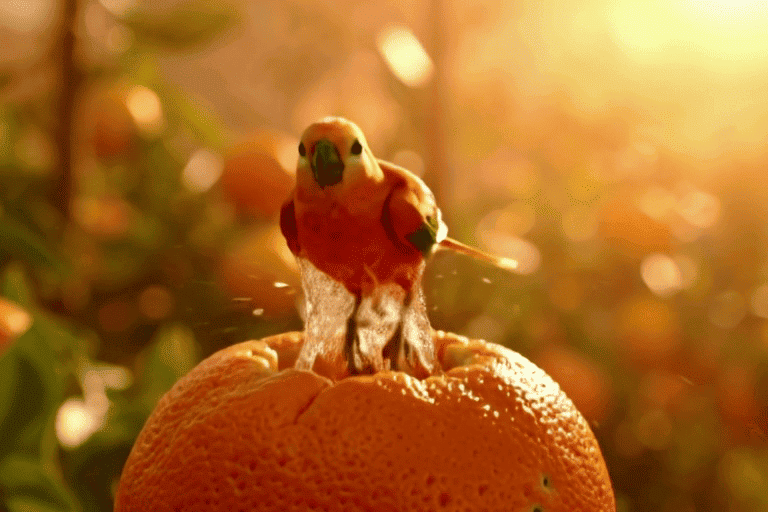} \hfill 
        \includegraphics[width=0.193\textwidth]{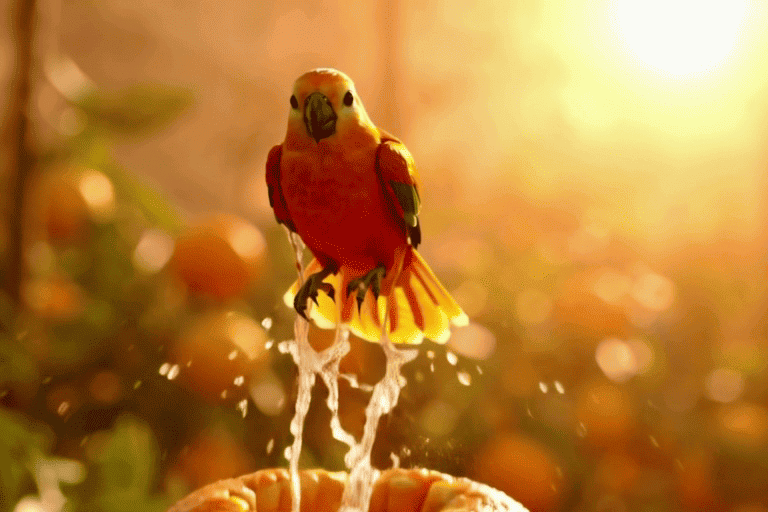} \hfill 
        \includegraphics[width=0.193\textwidth]{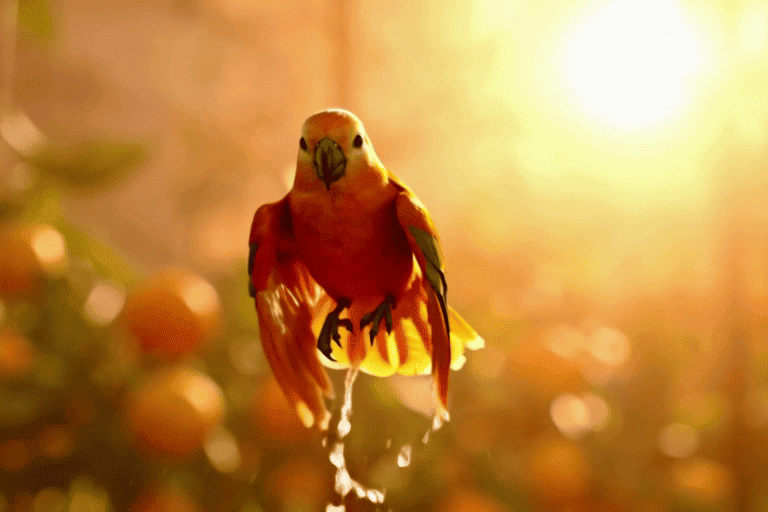} \hfill 
        \includegraphics[width=0.193\textwidth]{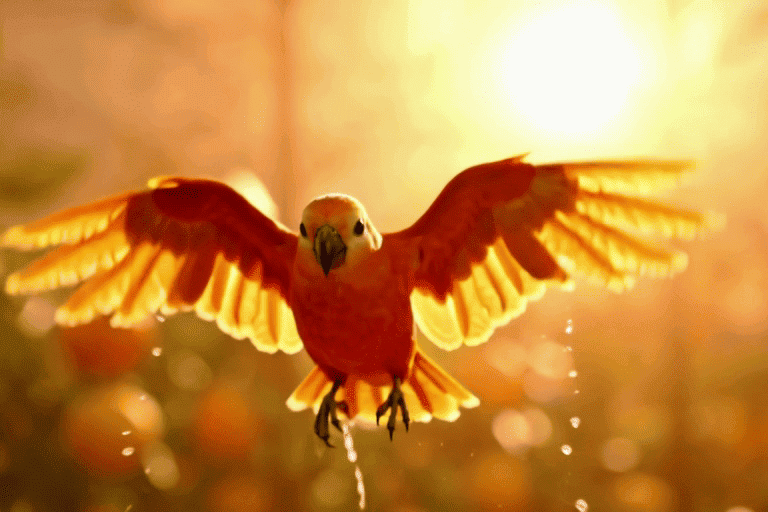} \hfill 
        \includegraphics[width=0.193\textwidth]{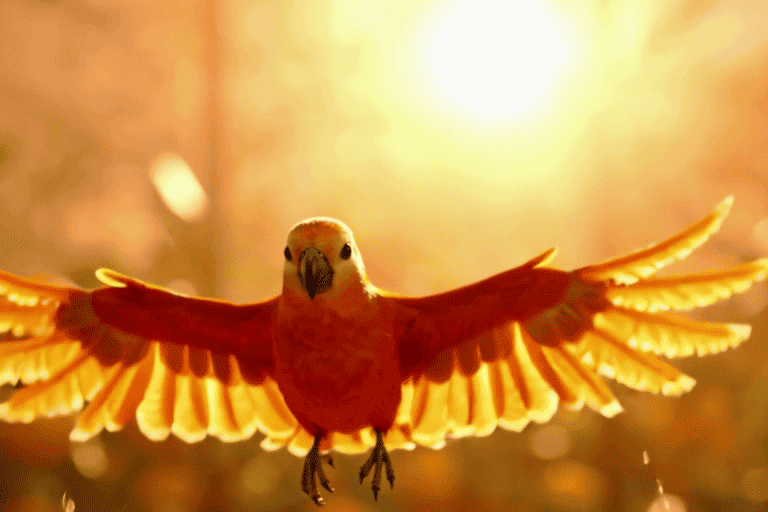}
    }
    \\
    \vspace{0.5cm}

    {\label{subfig:864-pretrain}
        \includegraphics[width=0.195\textwidth]{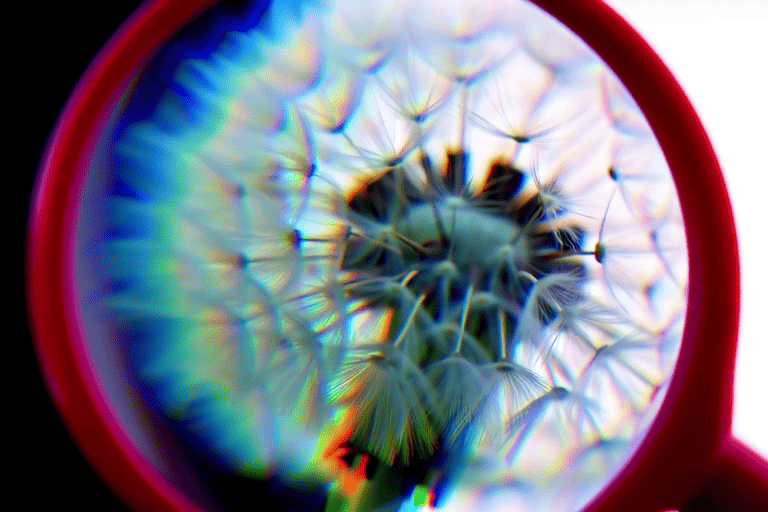} \hfill 
        \includegraphics[width=0.195\textwidth]{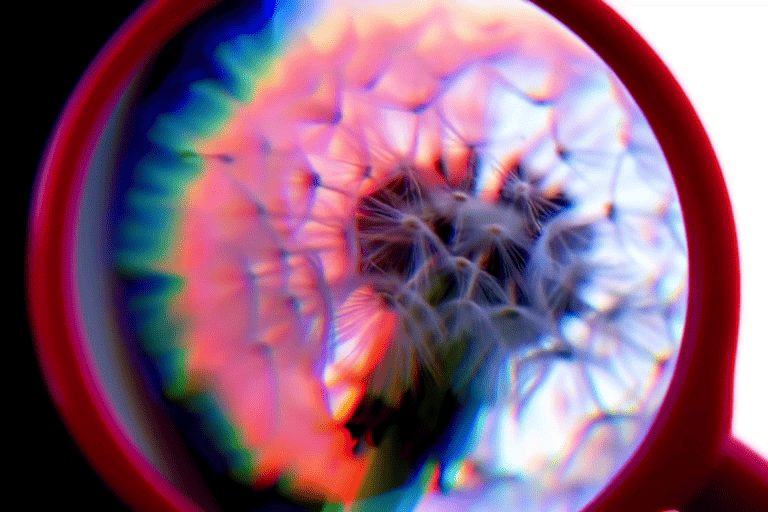} \hfill 
        \includegraphics[width=0.195\textwidth]{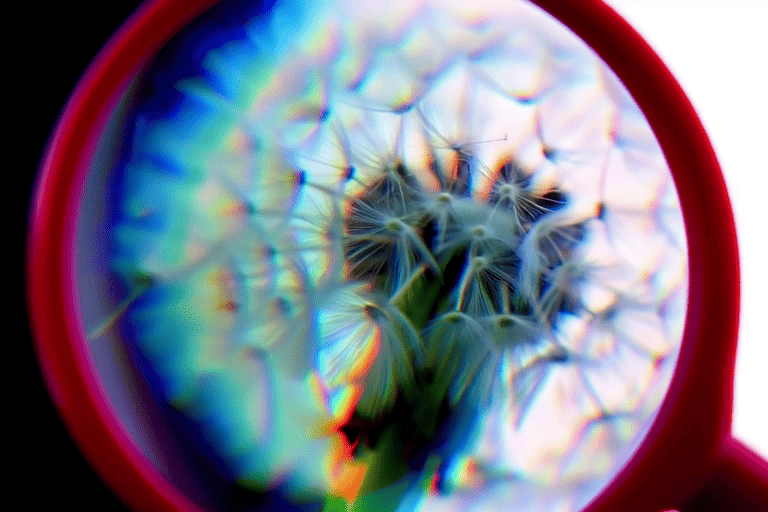} \hfill 
        \includegraphics[width=0.195\textwidth]{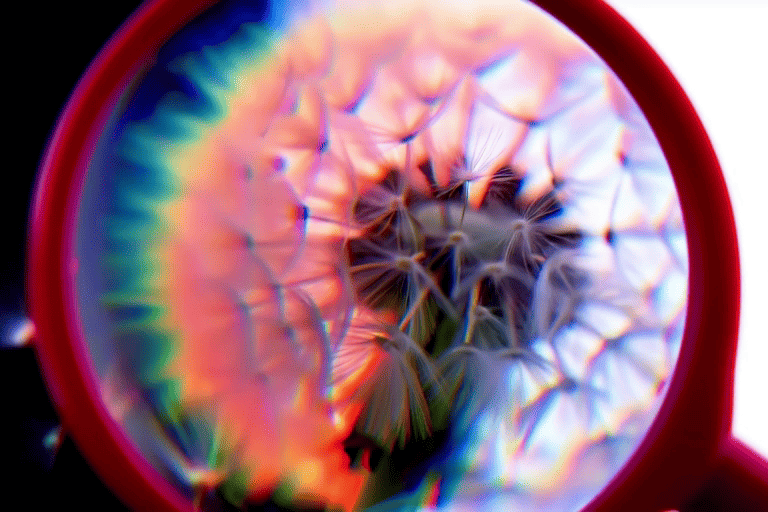} \hfill 
        \includegraphics[width=0.195\textwidth]{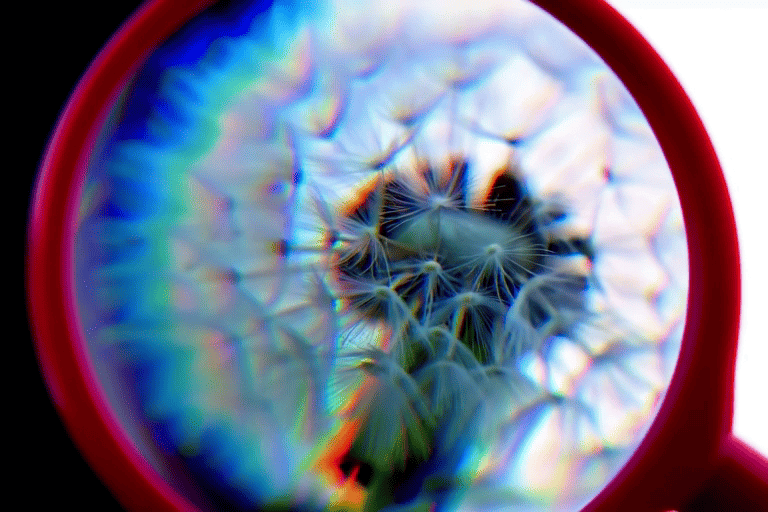}
    }
    \\
    \vspace{0.1cm}

    \subfloat[A dynamic, extremely macro closeup view of a delicate white dandelion, its feathery seeds gently swaying in a soft breeze. The scene is seen through the curved, highly reflective surface of a large red magnifying glass, distorting and refracting the light in colorful patterns. The dandelion’s tiny details—each seed, petal, and dewdrop—are magnified and illuminated dramatically. A slight movement of the magnifying glass causes the image to shift and shimmer, creating a mesmerizing visual effect. As the wind picks up, the dandelion’s seeds begin to lift and drift, captured in slow motion through the magnifying lens.]
    {\label{subfig:864-sft}
        \includegraphics[width=0.193\textwidth]{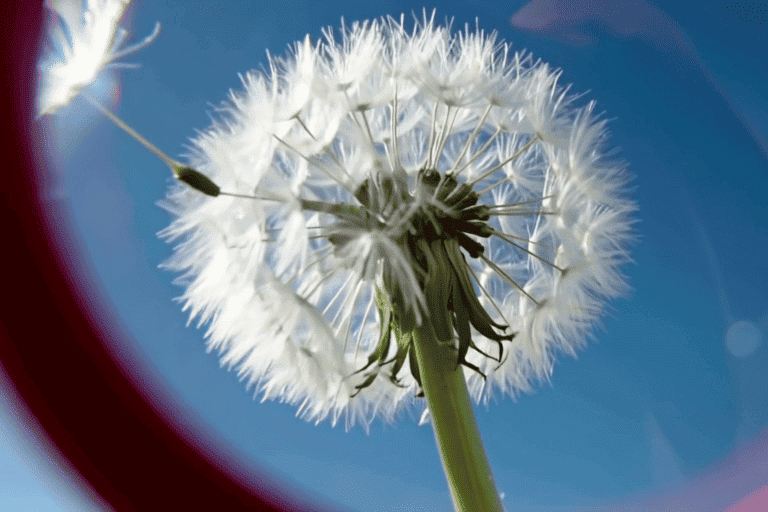} \hfill 
        \includegraphics[width=0.193\textwidth]{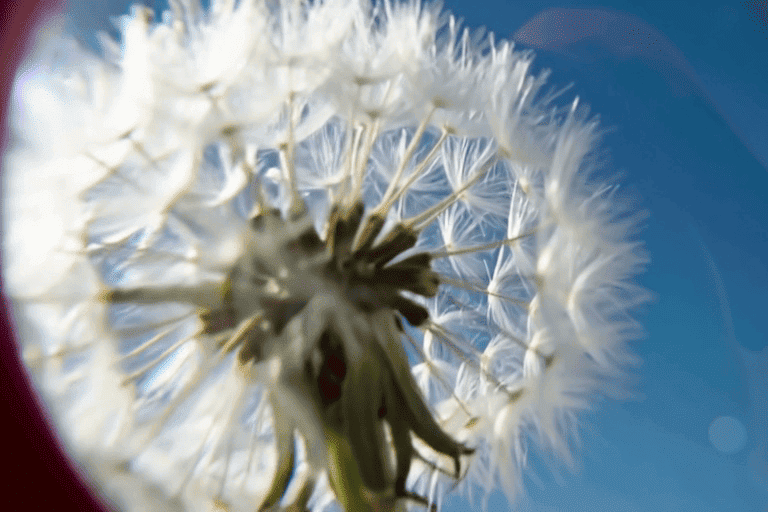} \hfill 
        \includegraphics[width=0.193\textwidth]{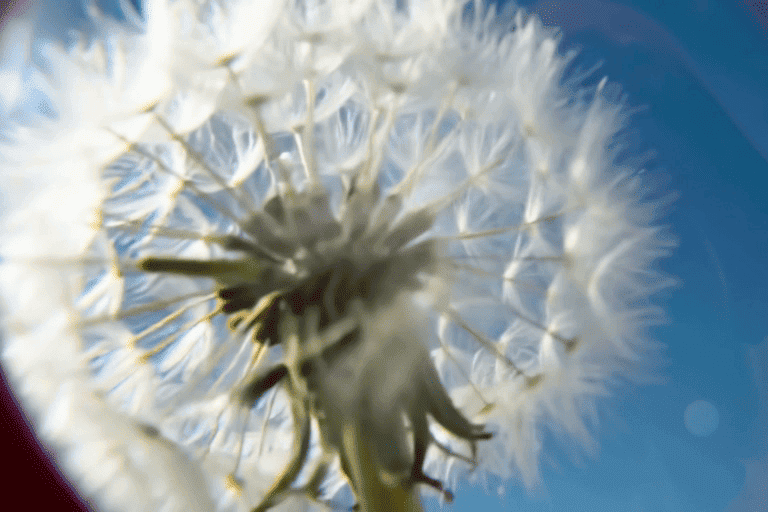} \hfill 
        \includegraphics[width=0.193\textwidth]{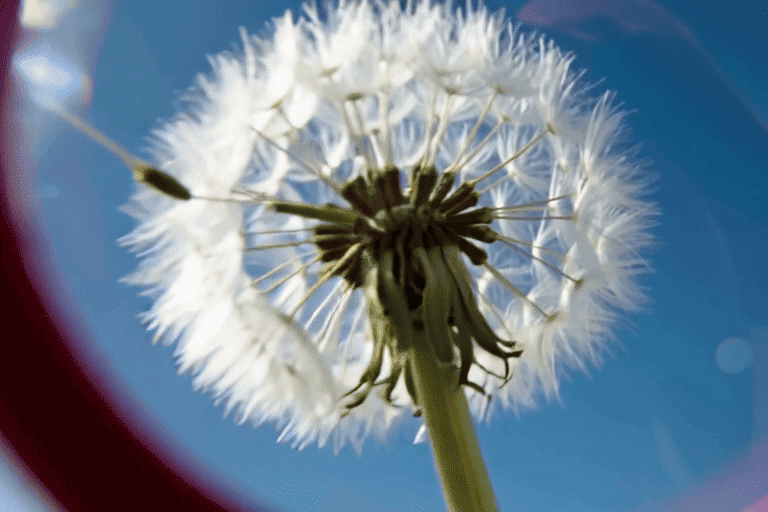} \hfill 
        \includegraphics[width=0.193\textwidth]{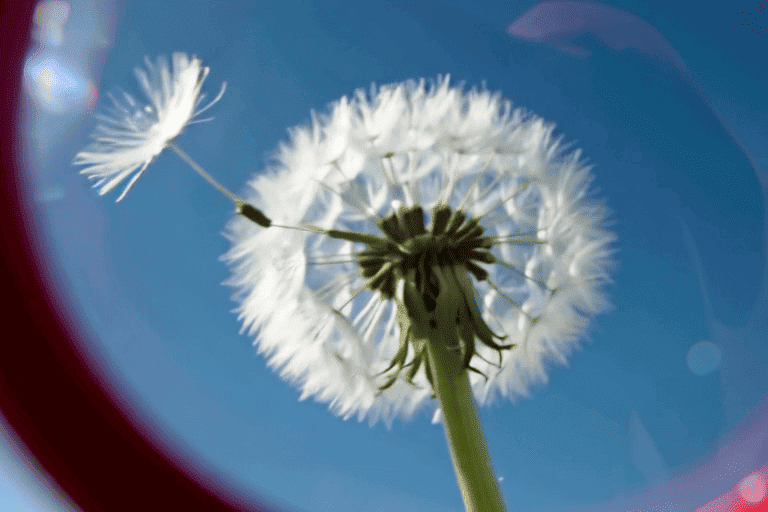}
    }
    \caption{Visual quality progress for text-to-video generation. \textbf{The results after pre-training stage (the top example from the pair) and after SFT (the bottom example).}}
    \label{fig:progress}
\end{figure}

We have seen significant qualitative improvements in terms of visual quality, realism, detail, and video dynamics following our multi-step training procedure. Figures~\ref{fig:progress_image} and~\ref{fig:progress} show a side-by-side comparison of the generation results for images (after pre-training, SFT and RL stages) and videos (after pre-training and SFT stages), respectively.



\subsection{Human Evaluation}

The Side-by-Side (SBS) evaluation is conducted on the \textbf{Elementary platform}\footnote{\url{https://elementary.center}}. Typically, a project is annotated by approximately 20 annotators who are not novices. The evaluation is split into two main phases:
\begin{enumerate}
    \item \textbf{Visual Quality Assessment}: Annotators evaluate generated visuals (images or videos) without access to the prompt to prevent bias.
    \item \textbf{Prompt Following Assessment}: After completing the visual assessment, annotators evaluate how well the generations adhere to the given prompt.
\end{enumerate}

Generations from different models are mixed and displayed in random order (left/right). An overlap of 5 is used for annotation consistency.

For video generation, most SBS evaluations are performed on the \textbf{Moviegen}\cite{polyak2024movie} (1003 prompts).





\bigskip
\bigskip

\subsubsection{Prompt Following}

Evaluators assessed prompt adherence using the criteria below. For each criterion, annotators selected one of the following judgments:

\textbf{Judgment Options:}
\begin{center}
\begin{tabular}{lp{11cm}}
\hline
\textbf{Choice} & \textbf{Description} \\
\hline
Model 1 Better & Model 1 better satisfies the prompt requirement \\
Model 2 Better & Model 2 better satisfies the prompt requirement \\
Both Fully Correct & Both models fully satisfy the prompt requirement \\
Both Fully Incorrect & Neither model satisfies the prompt requirement \\
Equally & Both models partially satisfy the requirement, with no clear advantage \\
\hline
\end{tabular}
\end{center}

\medskip

\textbf{Evaluation Criteria:}
\begin{center}
\begin{tabular}{lp{10cm}}
\hline
\textbf{Criterion} & \textbf{Description} \\
\hline
Object Presence (Count) & Number of unique objects from the prompt correctly generated \\
Object Quantity & Whether the correct number of each object is present \\
Object Properties & Accuracy of object attributes (e.g., color, size, shape) \\
Object Placement & Correct spatial relationships and relative positioning of objects \\
Action Presence (Count) & Number of unique actions from the prompt successfully realized \\
Action Properties & Accuracy in execution, timing, and dynamics of described actions \\
\hline
\end{tabular}
\end{center}

\textbf{Aggregation Formula:}
\begin{itemize}
    \item Model 1 Score = Model 1 Better + Both Fully Correct + Equally / 2
    \item Model 2 Score = Model 2 Better + Both Fully Correct + Equally / 2
\end{itemize}

Scores are normalized and averaged across prompts to compute an overall \textbf{Prompt Following} score, with detailed results visualized per criterion.

\subsubsection{Visual Quality}
Annotators evaluate the following aspects:

\begin{center}
\begin{tabular}{l p{14cm}}
\hline
\textbf{No.} & \textbf{Description} \\
\hline
1 & Composition: Framing, balance, and visual structure  \\
2 & Lighting: Realism and consistency of illumination  \\
3 & Color and Contrast: Accuracy and harmony of color palette and contrast levels \\
4 & Object and Background Distinctness: Clarity of foreground/background separation \\
5 & Frame Transition Smoothness: Temporal coherence between consecutive frames \\
6 & Dynamism: Energy, motion intensity, and scene activity \\
7 & Realism of Object Motion: Physical plausibility and naturalness of movement \\
8 & Face Generation Consistency: Temporal stability of facial features (if applicable) \\
9 & Overall Impression: Holistic aesthetic quality \\
10 & Number of Artifacts: Visual defects (e.g., distortions, glitches, blur) \\
11 & Number of Semantic Breaks: Unintended transformations or content shifts (per KandiVideoPrompts) \\
\hline
\end{tabular}
\end{center}
\begin{minipage}[t]{0.48\textwidth}
\begin{center}
\begin{tabular}{l}
\textbf{Judgment Options (Criteria 1–9):} \\
\hline
Model 1 Better \\
Model 2 Better \\
Equally \\
Model 1 Better (Unconfident) \\
Model 2 Better (Unconfident) \\
\hline
\end{tabular}
\end{center}
\end{minipage}
\hfill
\begin{minipage}[t]{0.48\textwidth}
\begin{center}
\begin{tabular}{l}
\textbf{Judgment Options (Criteria 10–11):}\\
\hline
Model 1 Better \\
Model 2 Better \\
Both Fully Correct \\
Equally \\
\hline\\
\end{tabular}
\end{center}
\vspace{0.5cm}
\end{minipage}

\textbf{Aggregation Formula (Criteria 1-9):}
\begin{itemize}
    \item Model 1 Score = Model 1 Better + Model 1 Better (Unconfident) / 2 + Equally / 2
    \item Model 2 Score = Model 2 Better + Model 2 Better (Unconfident) / 2 + Equally / 2
\end{itemize}

\textbf{Aggregation Formula (Criteria 10-11):}
\begin{itemize}
    \item Model 1 Score = Model 1 Better + Both Fully Correct + Equally / 2
    \item Model 2 Score = Model 2 Better + Both Fully Correct + Equally / 2
\end{itemize}

\textbf{Results are averaged into:}
\begin{itemize}
    \item Overall \textbf{Visual Quality} score (Criteria 1-5, 9-10).
    \item Overall \textbf{Dynamism and Motion Quality} score (Criteria 6-8, 11).
\end{itemize}

Averaging is done by percentage due to uneven distribution of ratings across criteria.

\subsubsection{Kandinsky 5.0 Video Lite vs. Sora}

\begin{figure}[htbp]
\centering
\begin{subfigure}[b]{0.7246376811594204\textwidth}
    \centering
    \includegraphics[width=\textwidth]{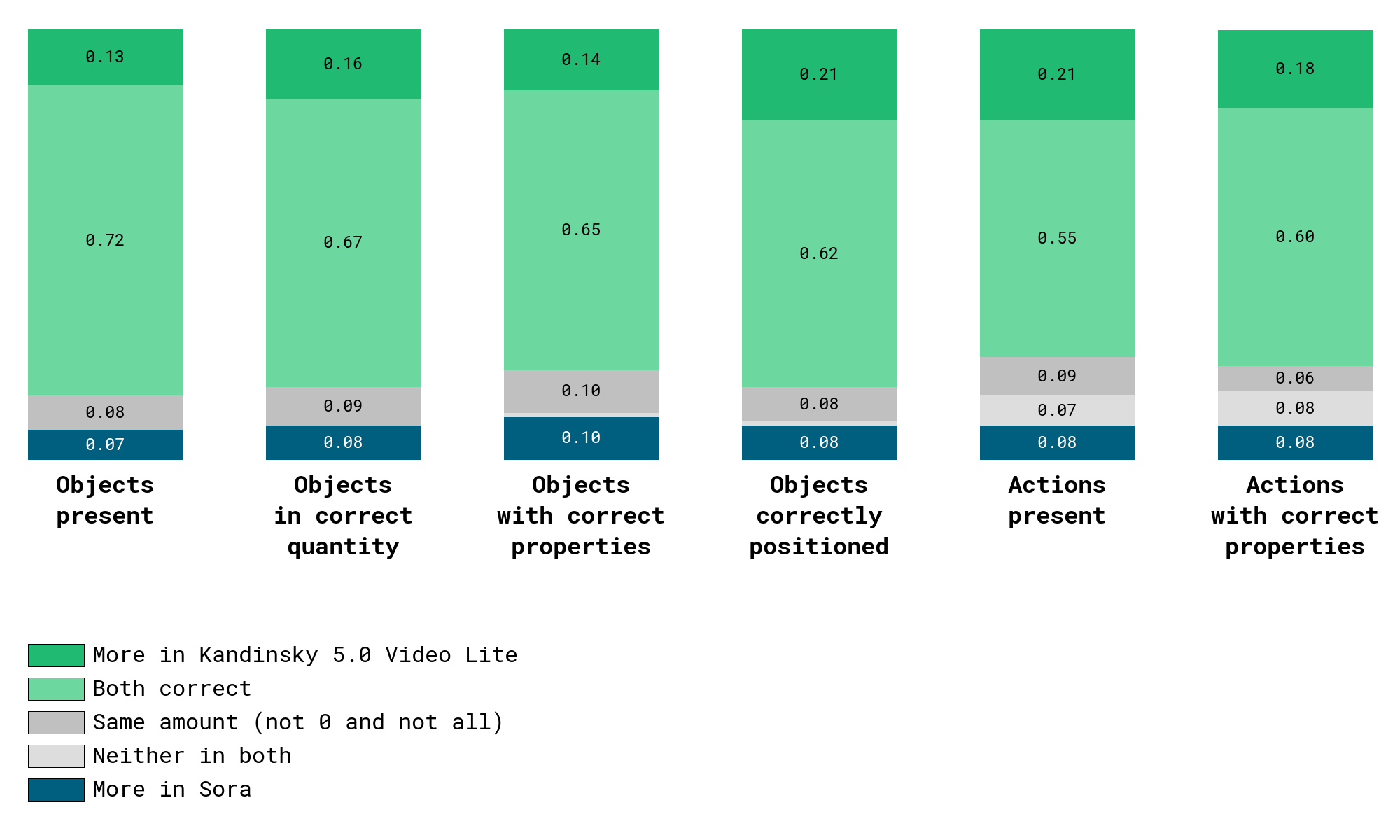}
    \caption{Object and Action Fidelity}
    \label{fig:object_action}
\end{subfigure}
\hfill
\begin{subfigure}[b]{0.26376811594202904\textwidth}
    \centering
    \includegraphics[width=\textwidth]{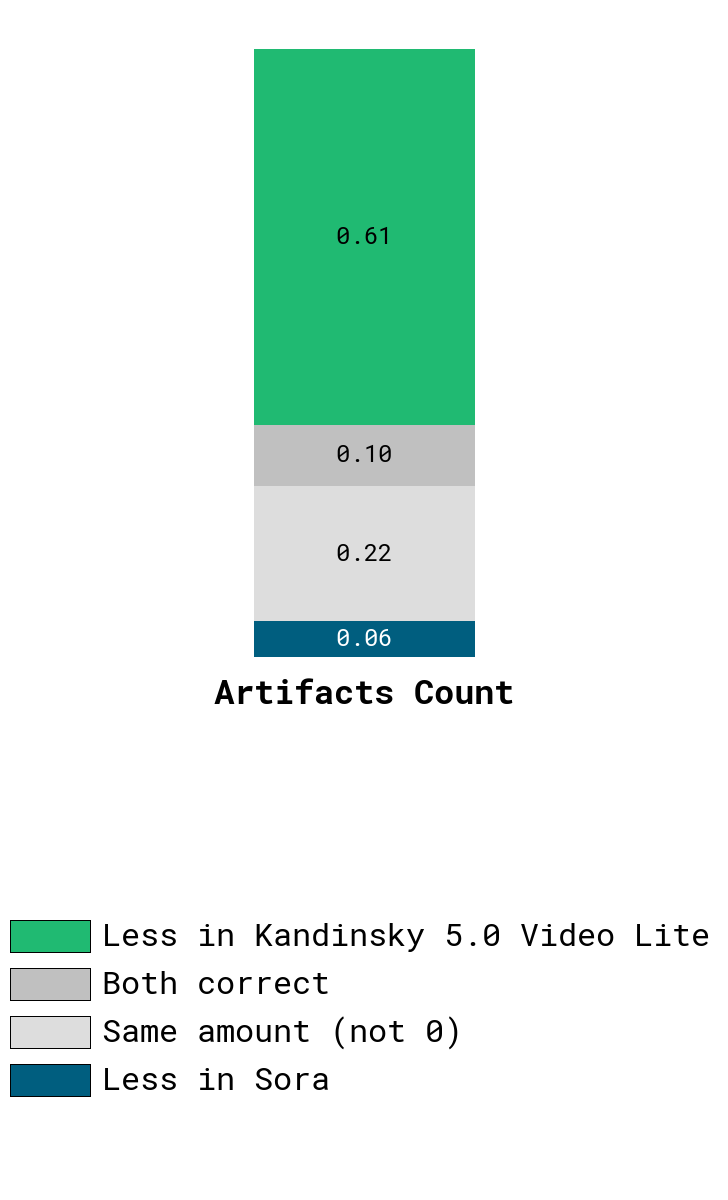}
    \caption{Artifacts Comparison}
    \label{fig:artifacts}
\end{subfigure}

\begin{subfigure}[b]{0.4803921568627451\textwidth}
    \centering
    \includegraphics[width=\textwidth]{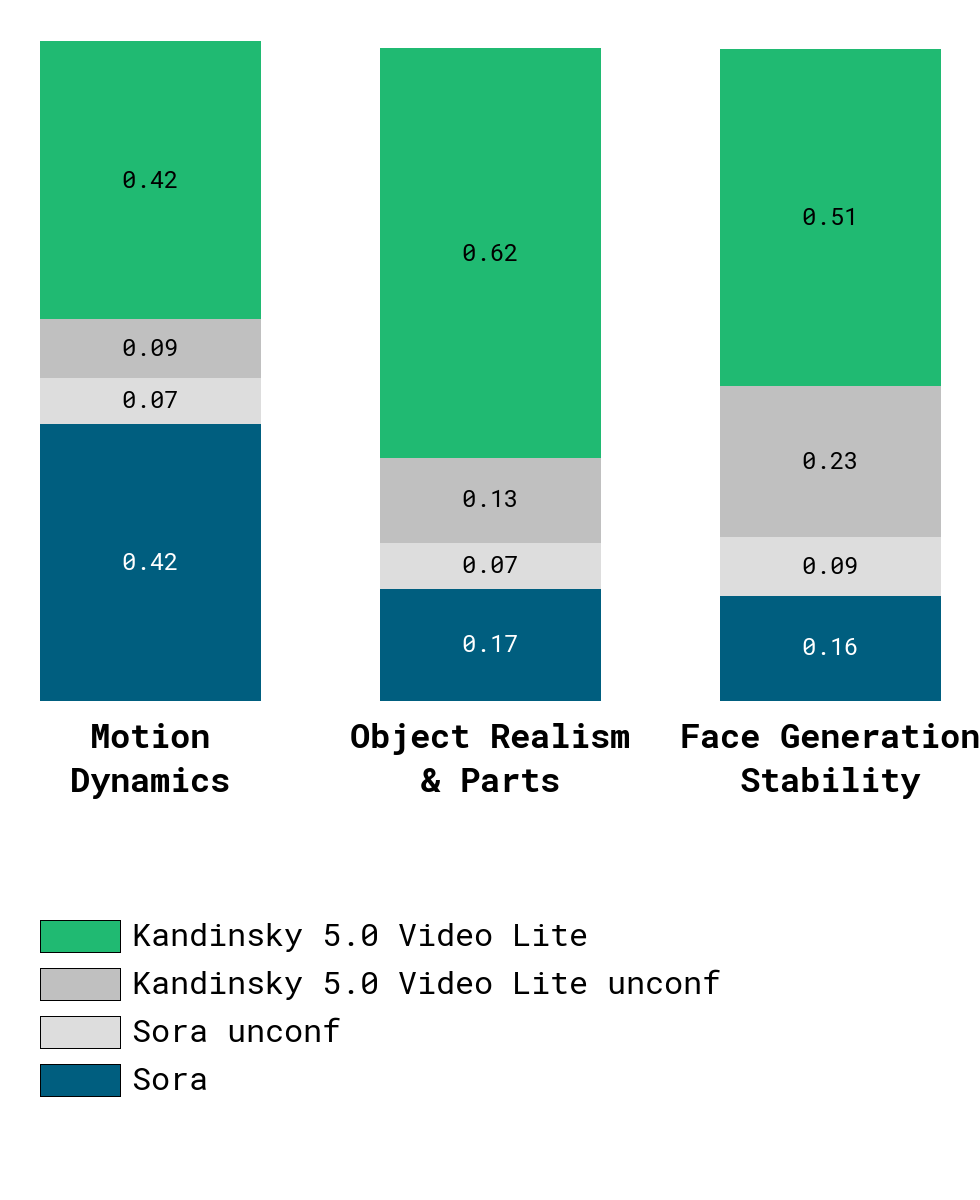}
    \caption{Component-wise Performance}
    \label{fig:components}
\end{subfigure}
\hfill
\begin{subfigure}[b]{0.5\textwidth}
    \centering
    \includegraphics[width=\textwidth]{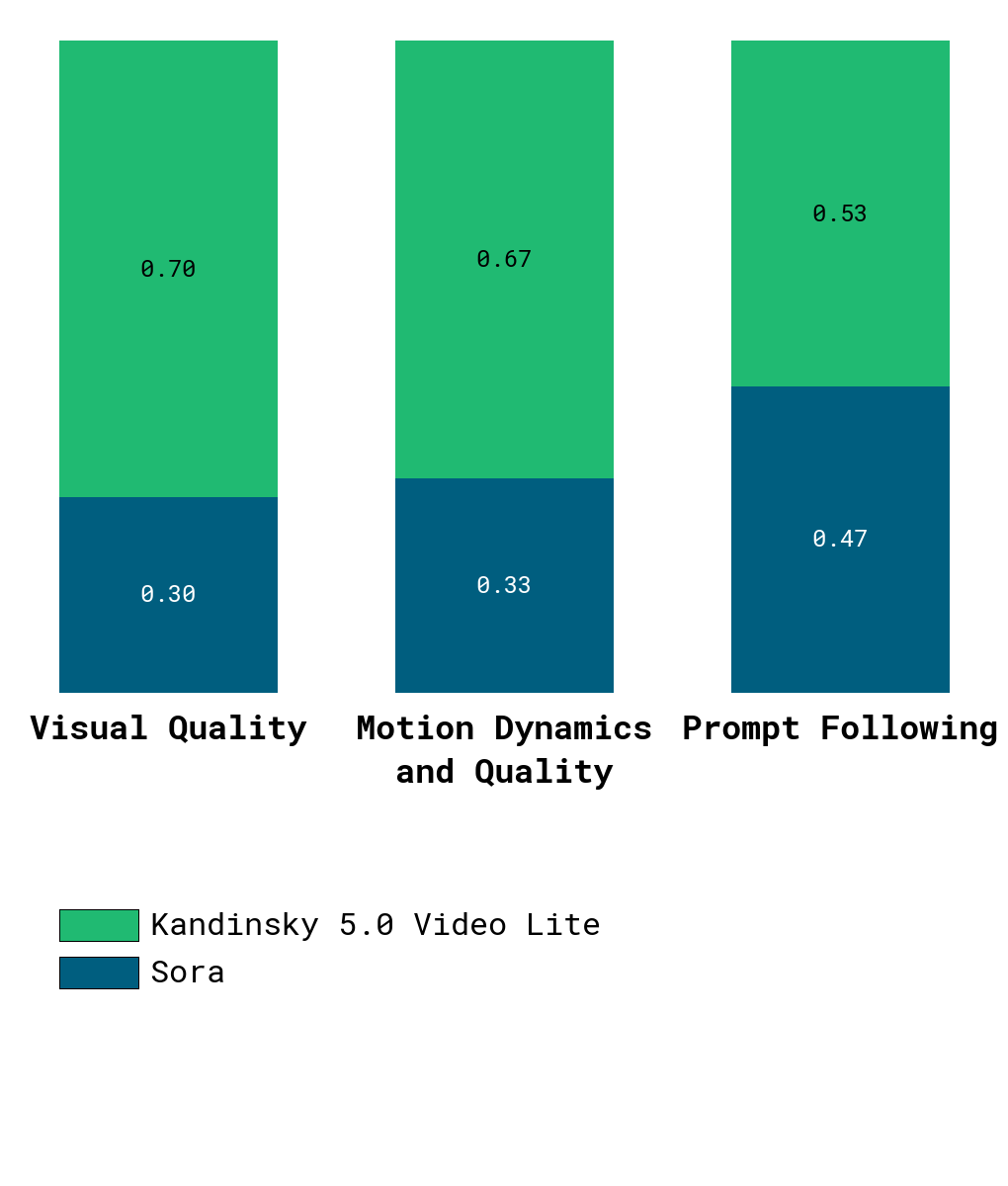}
    \caption{Key Video Quality Dimensions}
    \label{fig:other}
\end{subfigure}
\caption{
Side-by-side (SBS) human evaluation of \textbf{Kandinsky 5.0 Video Lite} versus \textbf{Sora} on the full \textsc{MovieGen} benchmark.
We collected \textasciitilde65K pairwise judgments from 44 trained raters (239 person-hours) across 1,002 prompt-video pairs (with 5-way overlap per item).
Each subplot shows the distribution of preferences: green segments indicate cases where \textbf{Kandinsky 5.0 Video Lite} was rated higher, blue — where Sora was preferred, and intermediate shades represent ties or neutral outcomes. Inter-rater agreement is approximately 71\%.}
\label{fig:sbs-sora}
\end{figure}

We conducted a side-by-side (SBS) human evaluation study comparing \textbf{Kandinsky 5.0 Video Lite} and \textbf{Sora} across six key dimensions of video generation quality on the full MovieGen benchmark. For each criterion, raters were presented with paired outputs and asked to select the better-performing model or indicate a tie. The results are aggregated over a representative sample of prompts and visualized as stacked bar charts, where each segment reflects the proportion of judgments favoring one model, both, or neither. See Figure \ref{fig:sbs-sora} for details.

\begin{figure}[htbp]
    \centering
    \begin{subfigure}[t]{0.32\textwidth}
    \centering
    \includegraphics[width=\linewidth]{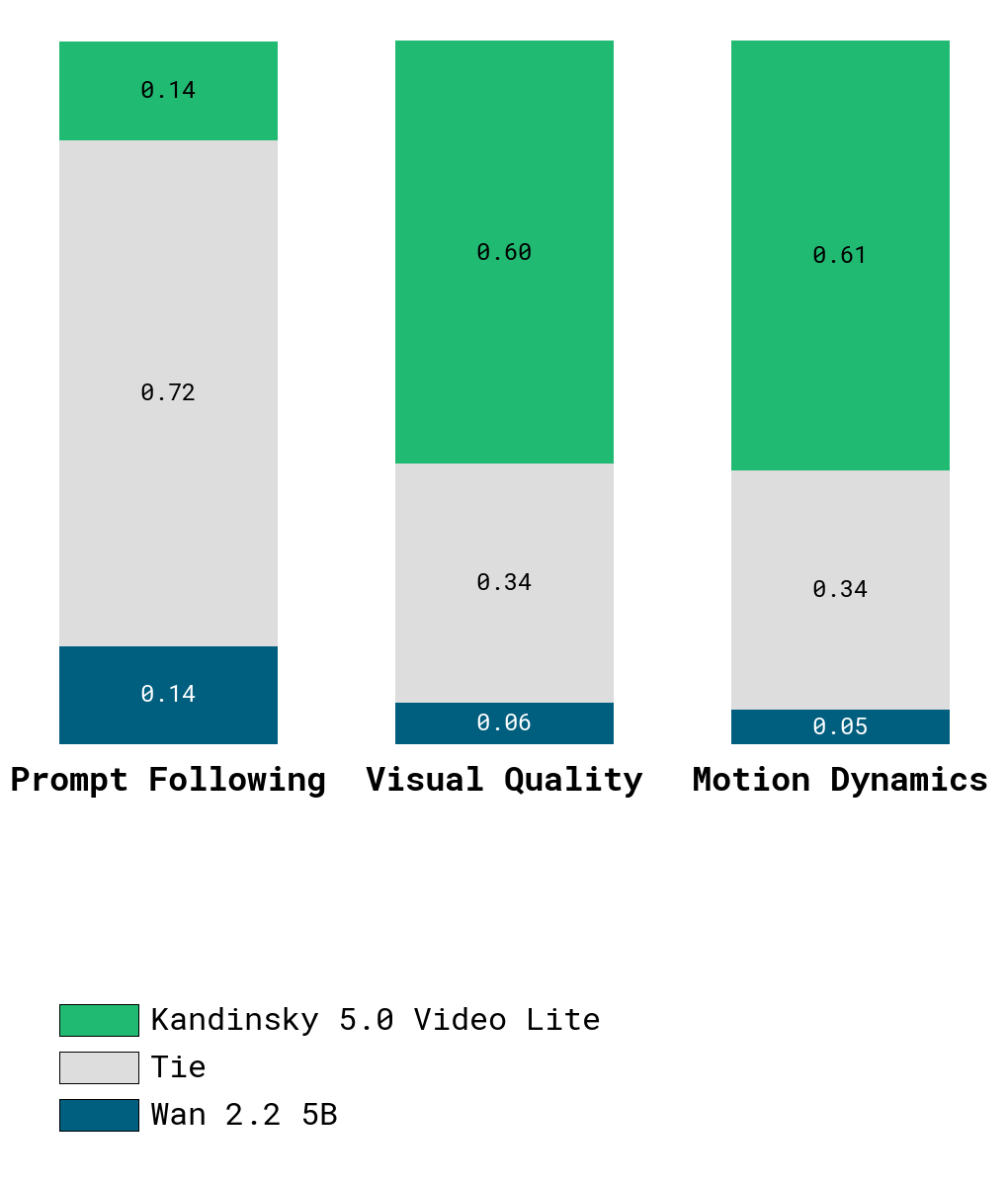}
    \caption{Comparison with Wan 2.2 5B} 
    \end{subfigure}
    \hfill
    \begin{subfigure}[t]{0.32\textwidth}
    \centering
    \includegraphics[width=\linewidth]{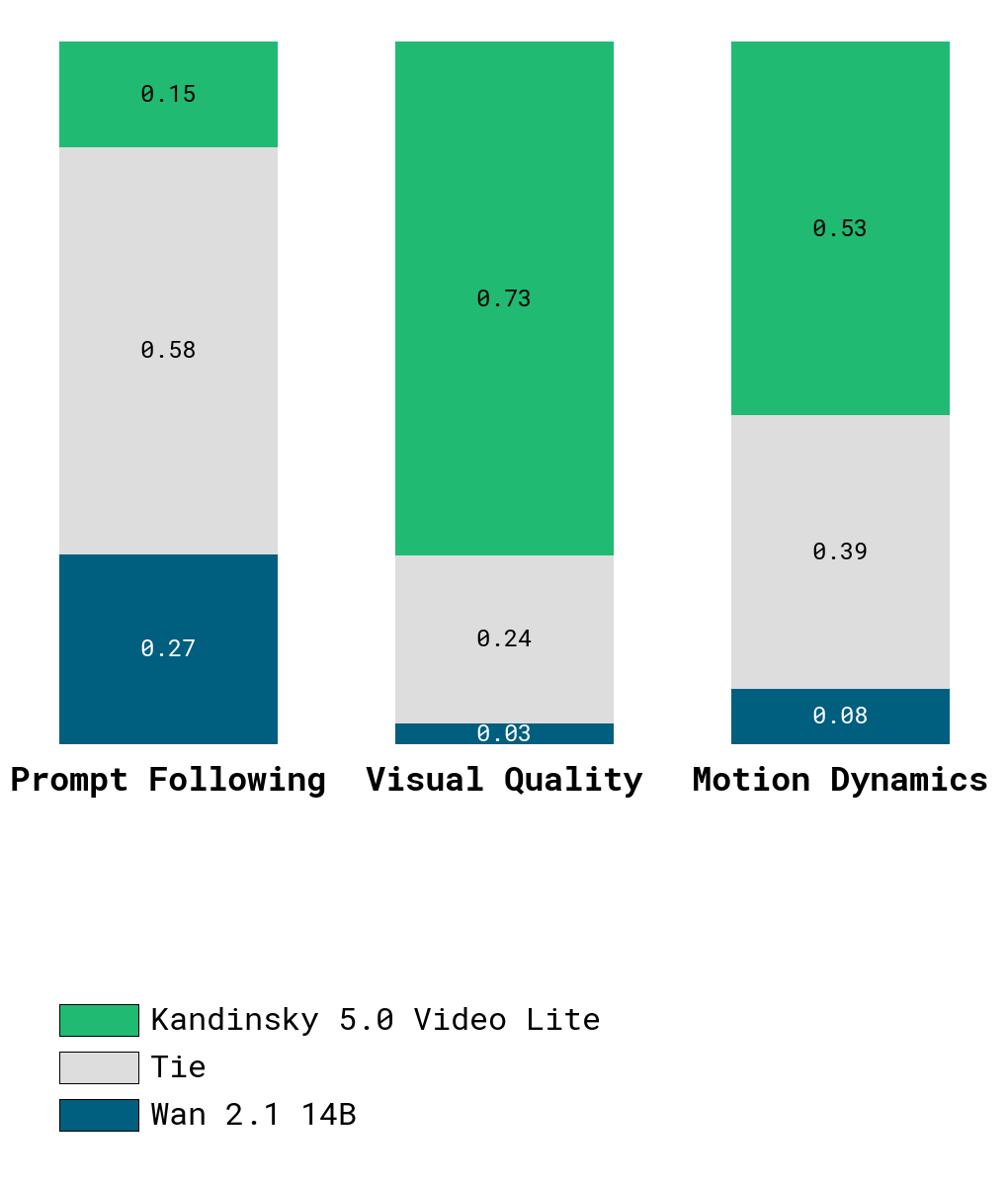}
    \caption{Comparison with Wan 2.1 14B} \end{subfigure}
    \hfill
    \begin{subfigure}[t]{0.32\textwidth}
    \centering
    \includegraphics[width=\linewidth]{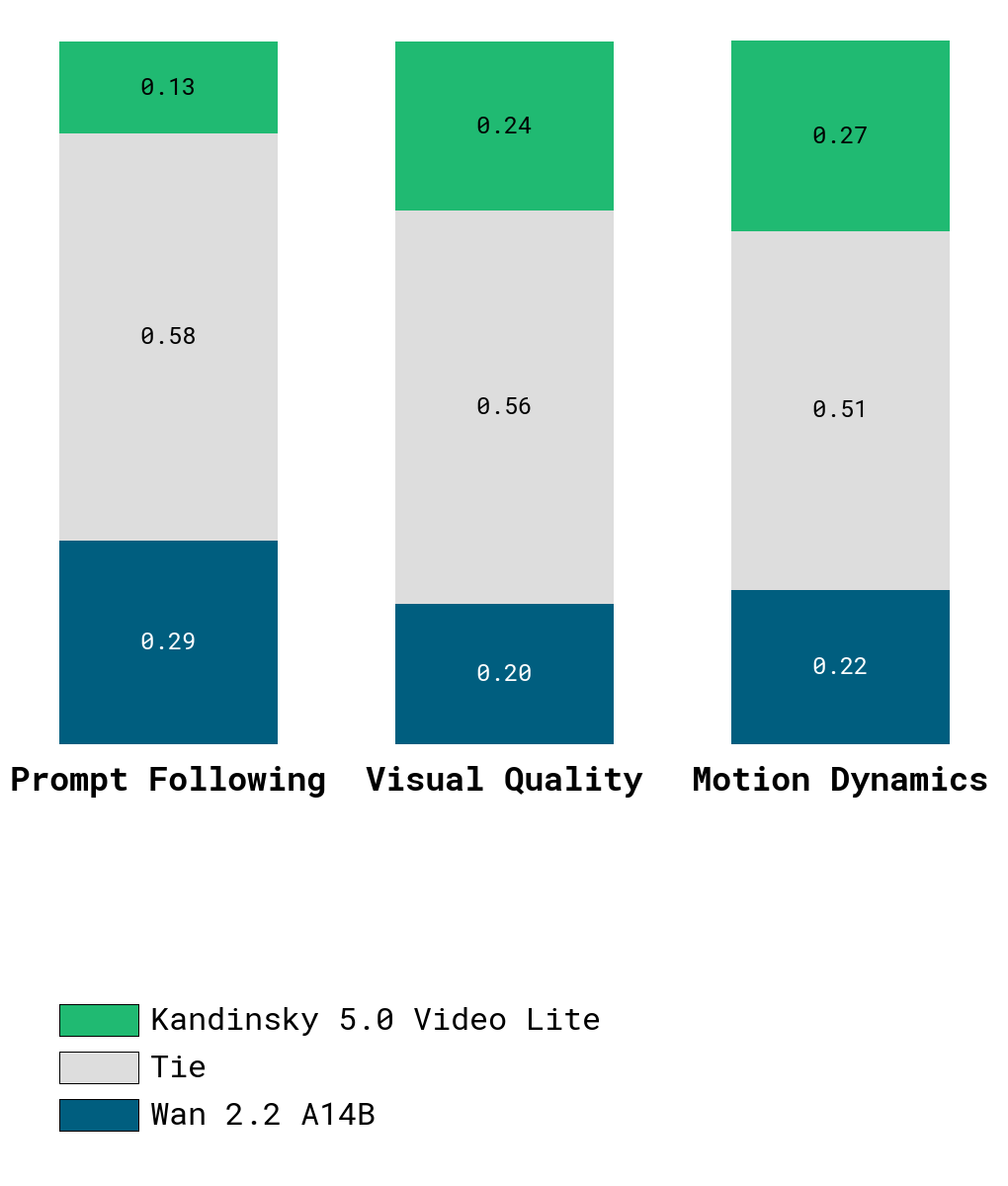}
    \caption{Comparison with Wan 2.2 A14B} \end{subfigure}
    \caption{Kandinsky 5.0 Video Lite outperforms Wan models in Visual Quality and Motion Dynamics.}
    \label{fig:sbs-lite-wan}
\end{figure} 

\begin{figure}[htbp]
    \centering
        \includegraphics[width=0.8\linewidth]{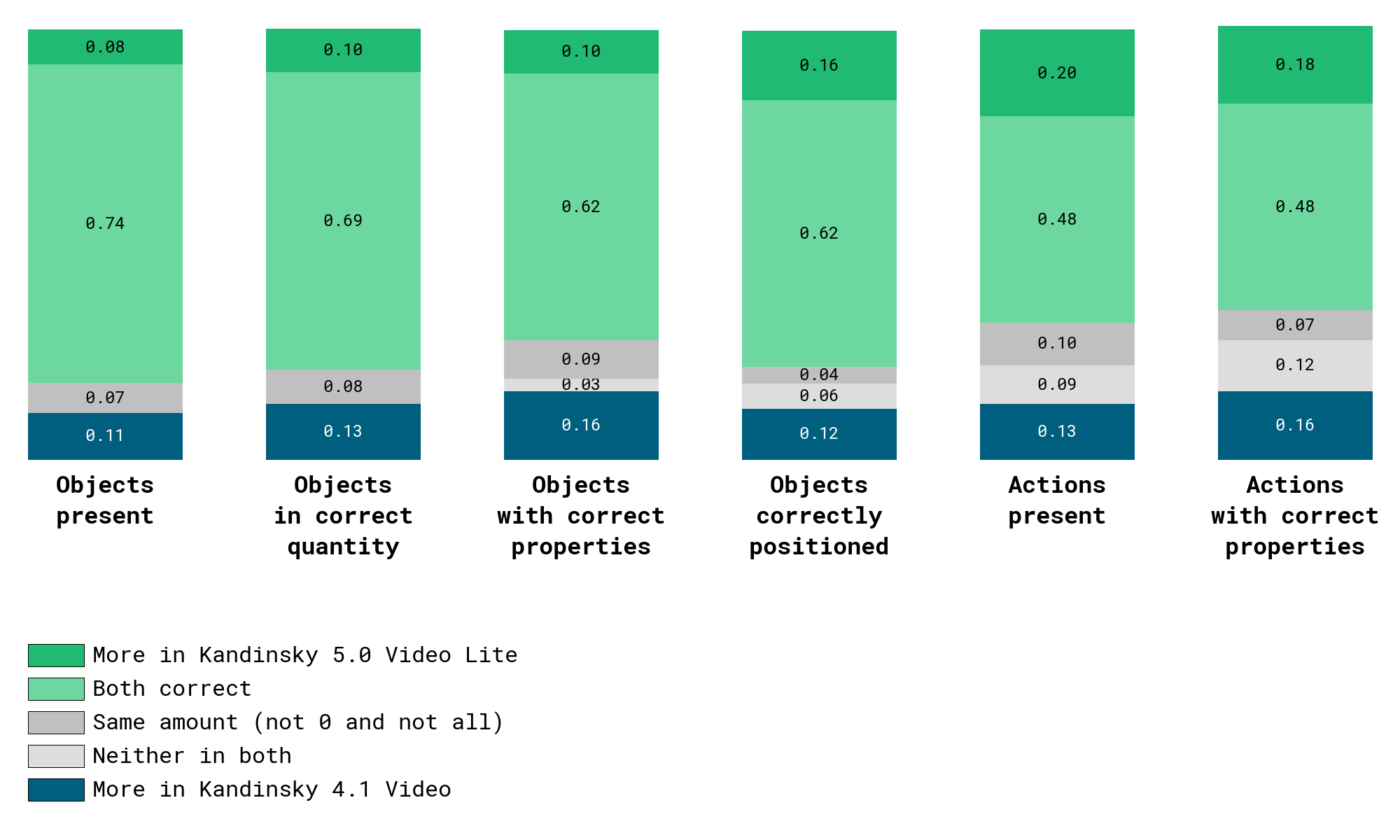}
        \caption{
        Object and Action Presence, Quantity, Properties, and Positioning. 
        \textbf{Kandinsky 5.0 Video Lite} outperforms \textbf{Kandinsky 4.1 Video} in action-related metrics — both in the presence of actions and their alignment with prompt semantics — and shows better object positioning. 
        Conversely, \textbf{Kandinsky 4.1 Video} is slightly preferred in basic object presence, quantity, and attribute fidelity, though both models in most cases produce correct outputs in these categories.}
        \label{fig:sbs-k41-fig1}
\end{figure}

\begin{figure}[htb]
    \begin{subfigure}[t]{0.3402777777777778\textwidth}
    \centering
    \includegraphics[width=\linewidth]{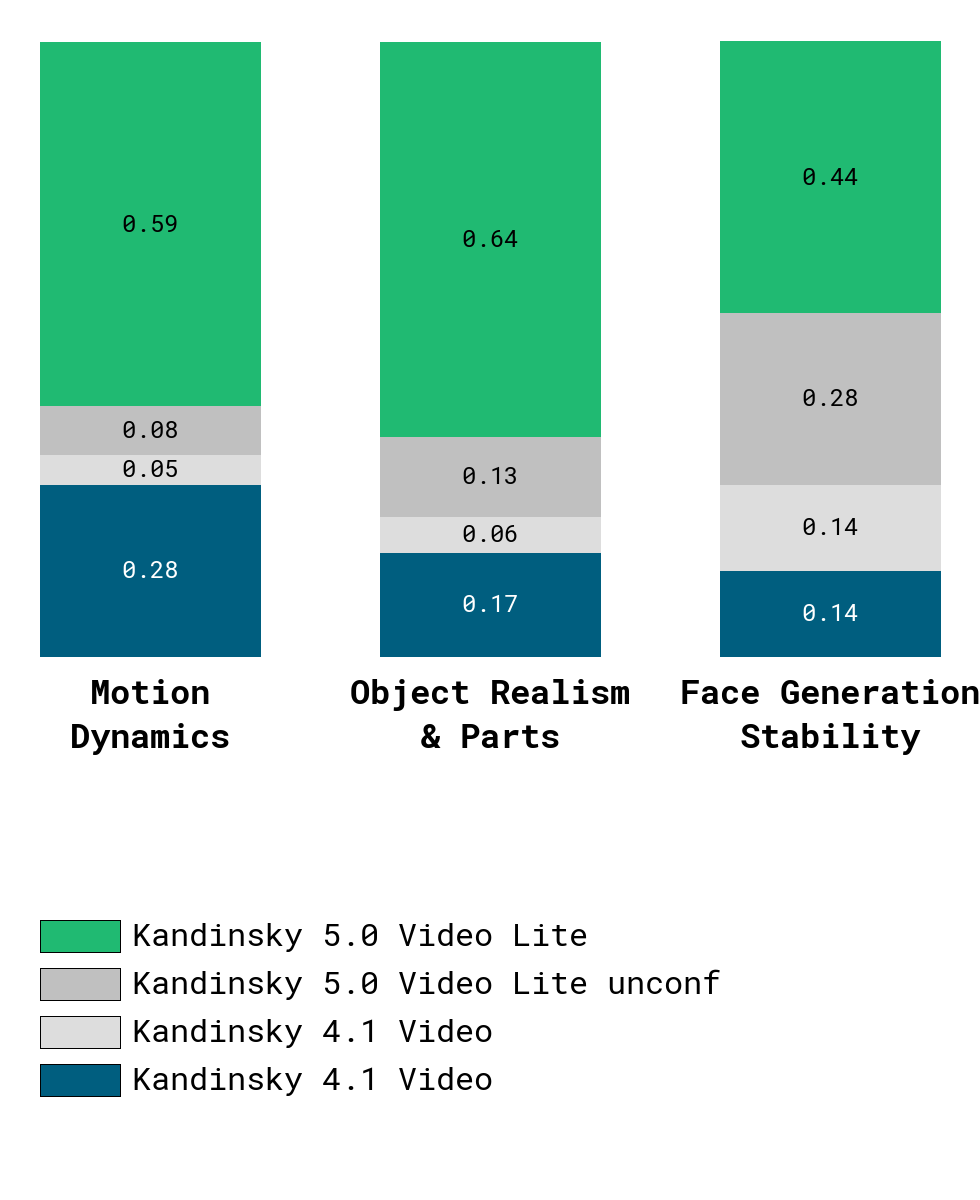}
    \caption{Dynamics and object motion realism.} \label{fig:sbs-k41-3}
    \end{subfigure}
    \hfill
    \begin{subfigure}[t]{0.2527777777777778\textwidth}
    \centering
    \includegraphics[width=\linewidth]{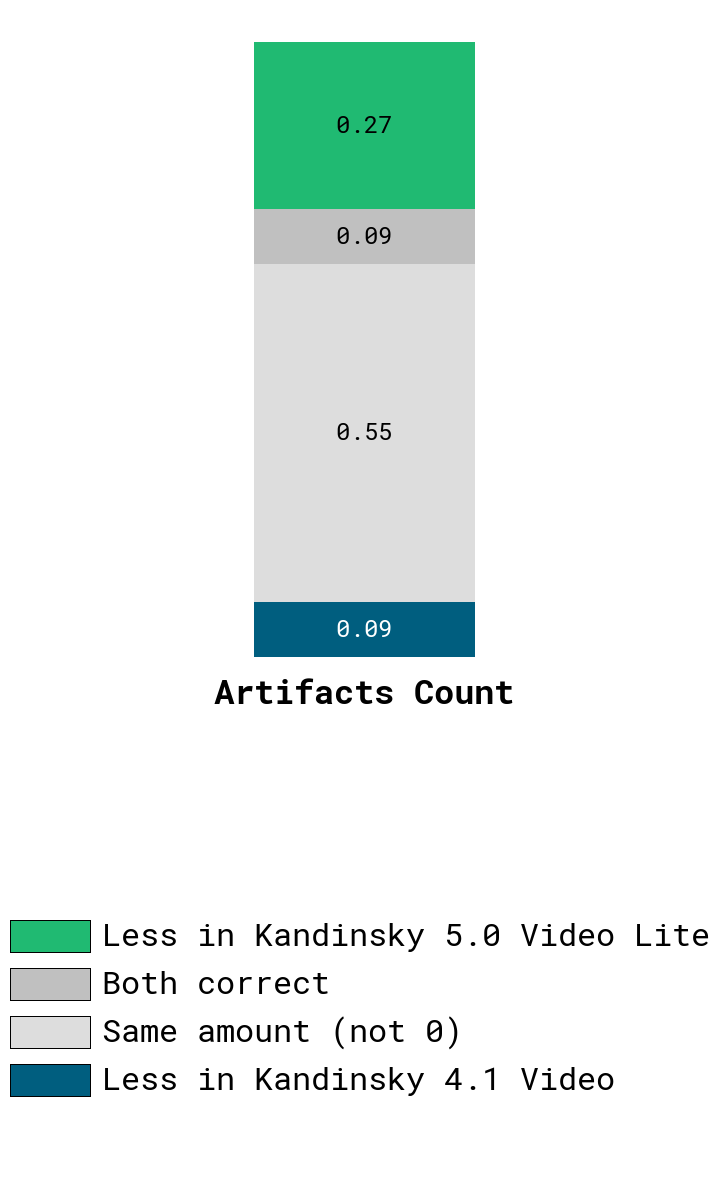}
    \caption{Artifact comparison} \label{fig:sbs-k41-1}
    \end{subfigure}
    \hfill
    \begin{subfigure}[t]{0.3888888888888889\textwidth}
    \centering
    \includegraphics[width=\linewidth]{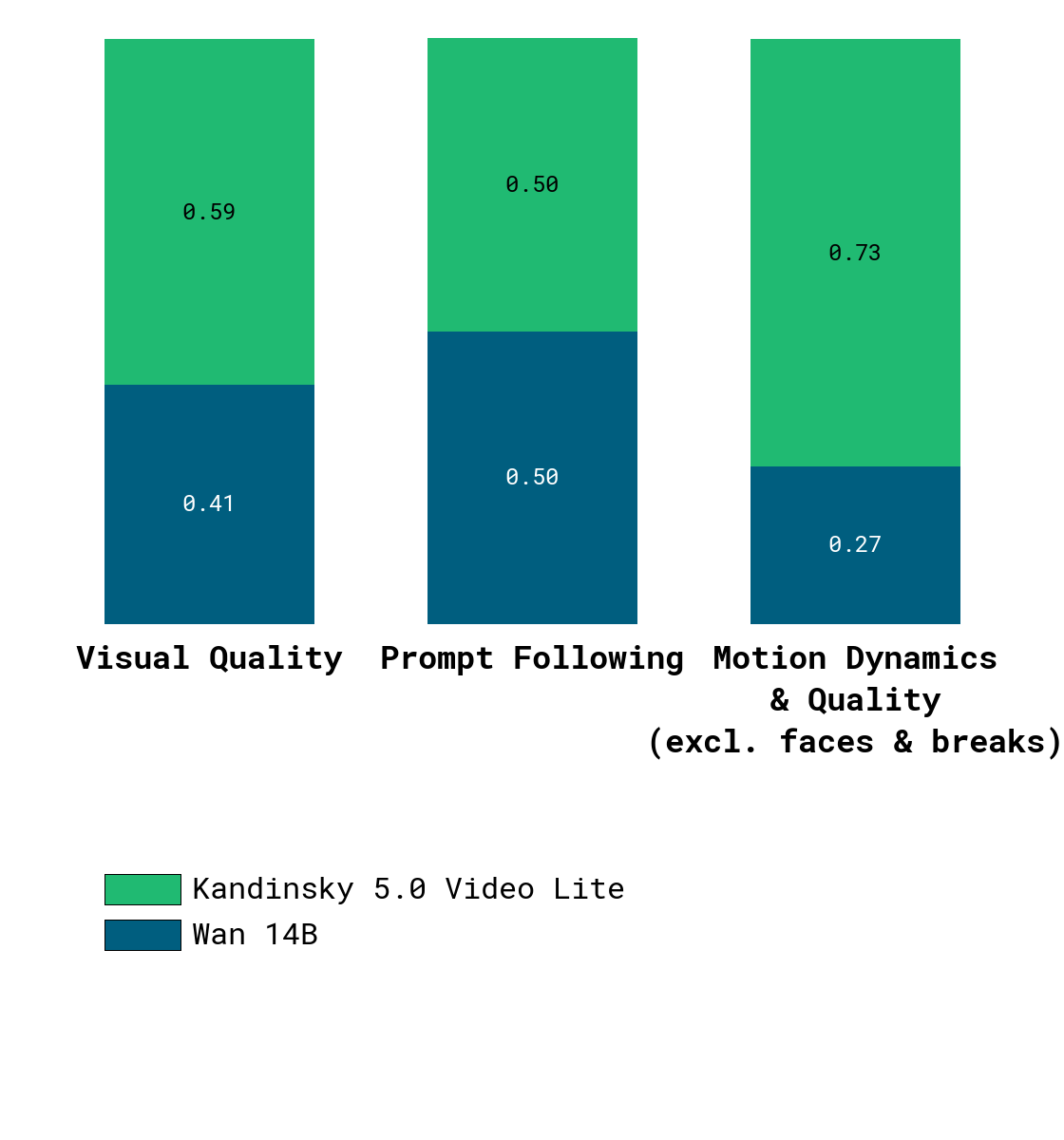}
    \caption{Key Video Quality Dimensions}
    \label{fig:sbs-k41-4}
    \end{subfigure}
    \caption{
    (a) Motion Dynamics: \textbf{Kandinsky 5.0 Video Lite} is preferred in 59\% of cases, \textbf{Kandinsky 4.1 Video} in 28\%, with 13\% undecided — indicating substantial improvement in temporal coherence and fluidity. \
    (b) Artifacts: In 55\% of comparisons, both models exhibit similar artifact levels; \textbf{Kandinsky 5.0 Video Lite} has fewer artifacts in 27\% of cases, while Kandinsky 4.1 Video does so in only 9\%. This confirms significant artifact reduction in the newer version. \
    (c) Overall Quality Dimensions: \textbf{Kandinsky 5.0 Video Lite} leads decisively in Visual Quality (0.59) and Motion Dynamics (0.73), while matching \textbf{Kandinsky 4.1 Video} in Prompt Following (0.50). The upgrade delivers consistent gains across all core metrics, especially in motion and aesthetics.}
    \label{fig:sbs-k41-fig2}
\end{figure}

\subsubsection{Kandinsky 5.0 Video Lite vs. Wan Models}

We conducted a simplified side-by-side (SBS) evaluation of our \textbf{Kandinsky 5.0 Video Lite} in Text-to-Video mode against three models of Wan series — \textbf{Wan 2.1 14B}, \textbf{Wan 2.2 5B}, and \textbf{Wan 2.2 A14B} — on the \textsc{MovieGen} benchmark, comparing performance across three key dimensions: Prompt Following, Visual Quality, and Motion Dynamics. Expert raters assessed paired video outputs per prompt, with results aggregated into preference scores per criterion.

Results, visualized in Figure~\ref{fig:sbs-lite-wan} reveal a consistent pattern:
\begin{itemize}
\item \textbf{Visual Quality} and \textbf{Motion Dynamics} consistently favor \textbf{Kandinsky 5.0 Video Lite} across all comparisons, with clear advantages in aesthetic coherence, object realism, and temporal fluidity.
\item \textbf{Prompt Following} shows stronger performance from Wan models, particularly \textbf{Wan 2.2 A14B} and \textbf{Wan 2.1 14B}, which better capture fine-grained semantic details and action specifications.
\end{itemize}

While Wan variants demonstrate superior alignment with textual prompts, Kandinsky 5.0 Video Lite maintains a decisive edge in perceptual quality and motion naturalness — suggesting a clear trade-off between semantic fidelity and visual fluency. The gap in prompt adherence is moderate and varies by scenario, indicating that Kandinsky 5.0 Video Lite remains a compelling choice for applications prioritizing visual output over precise instruction following.

\subsubsection{Kandinsky 5.0 Video Lite vs. Kandinsky 4.1 Video}

We also conducted a comprehensive side-by-side (SBS) human evaluation comparing \textbf{Kandinsky 5.0 Video Lite} and our previous model \textbf{Kandinsky 4.1 Video} across key dimensions of video generation quality on the full \textsc{MovieGen} benchmark. Evaluations were performed by trained raters using paired outputs per prompt, with judgments aggregated over a representative sample of video generations. Results are visualized in stacked bar charts, where green segments indicate preference for \textbf{Kandinsky 5.0 Video Lite}, blue for \textbf{Kandinsky 4.1 Video}, and intermediate shades denote ties or neutral outcomes. See Figures~\ref{fig:sbs-k41-fig1}--\ref{fig:sbs-k41-fig2} for detailed breakdowns.

Across all metrics, \textbf{Kandinsky 5.0 Video Lite} demonstrates marked improvements over \textbf{Kandinsky 4.1 Video}, particularly in motion dynamics, object realism, artifact suppression, and semantic accuracy. The new version excels in generating coherent, visually rich sequences with higher fidelity to prompts, while maintaining strong performance in face stability and component-level realism. These results confirm that \textbf{Kandinsky 5.0 Video Lite} model represents a substantial leap forward in video generation capability within the Kandinsky family.

\subsubsection{Kandinsky 5.0 Video Pro vs Veo 3 and Veo 3 fast}

We conducted a side-by-side (SBS) comparison of our Kandinsky 5.0 Video Pro text-to-video model with leading video generation models Veo 3 and Veo 3 Fast, using the MovieGen benchmark dataset. The evaluation focused on three key aspects: Prompt Following (how accurately the generated video aligns with the input text description), Video Quality (including aesthetic appeal, visual coherence, and technical execution), and Motion Dynamics (the naturalness, smoothness, and realism of motion over time). Expert evaluators assessed the outputs based on these criteria. Results show that Veo 3 and Veo 3 Fast significantly outperform Kandinsky 5.0 Video Pro in Prompt Following, demonstrating superior understanding and fidelity to complex textual instructions. However, Kandinsky 5.0 Video Pro achieves higher scores in Video Quality and Motion Dynamics, delivering more visually compelling and dynamically coherent sequences.

\begin{figure}[htbp]
    \centering
    \begin{subfigure}[t]{0.48\textwidth}
    \centering
    \includegraphics[width=\linewidth]{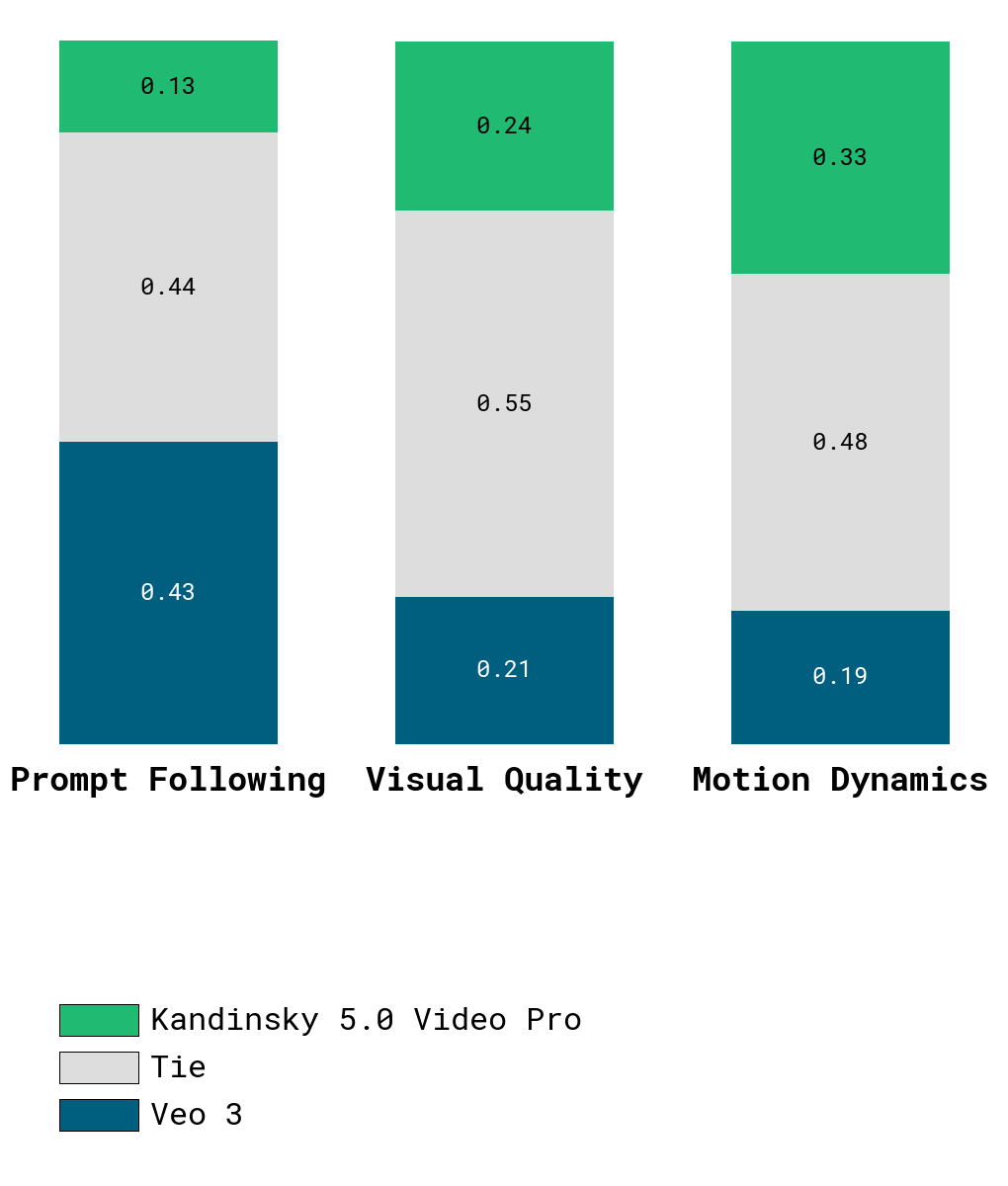}
    \caption{Comparison with Veo 3} 
    \end{subfigure}
    \hfill
    \begin{subfigure}[t]{0.48\textwidth}
    \centering
    \includegraphics[width=\linewidth]{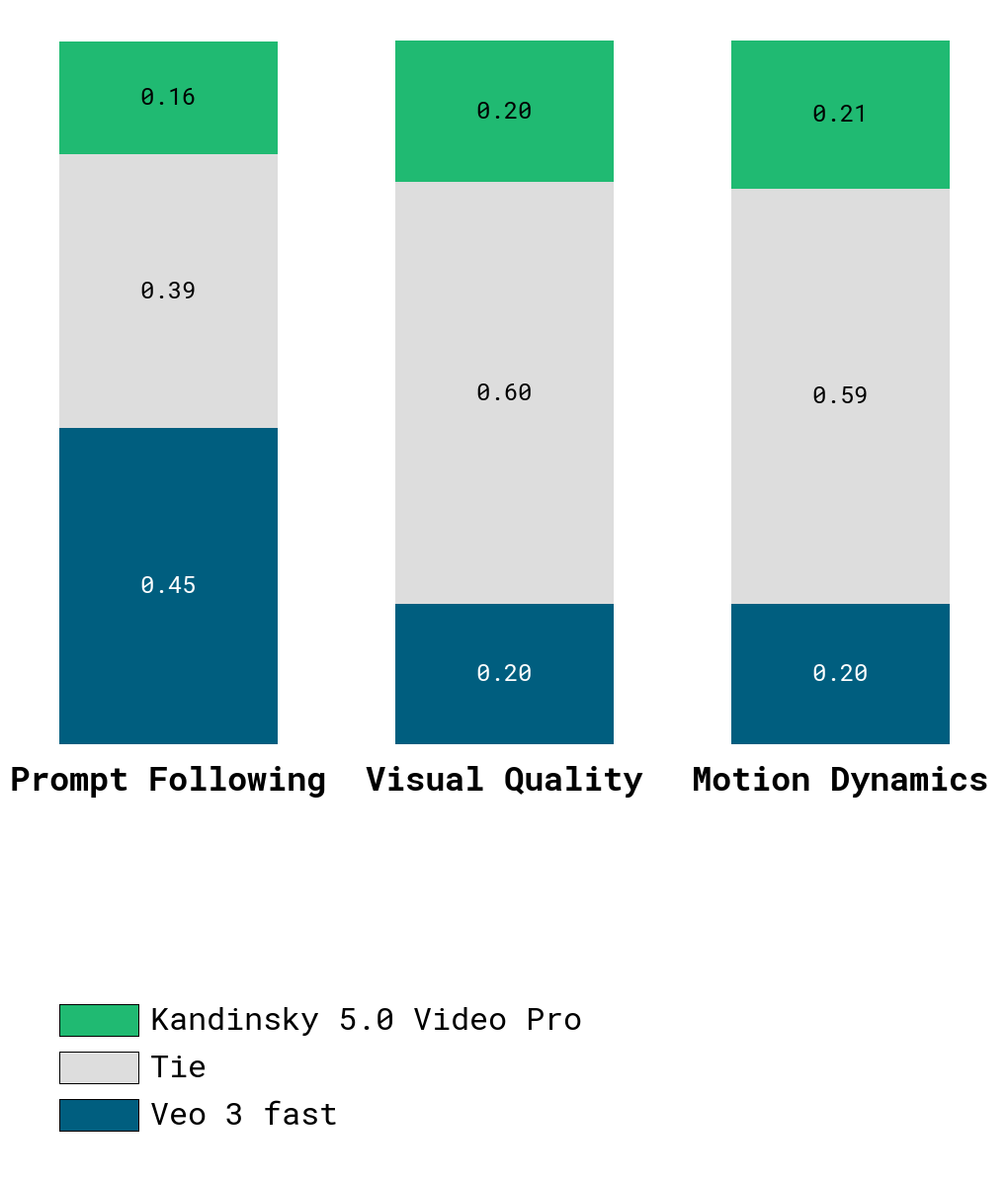}
    \caption{Comparison with Veo 3 fast} \end{subfigure}
    \caption{Kandinsky 5.0 Video Pro excelled in Visual Quality and Motion Dynamics, whereas Prompt Following remained a relative weakness compared to Veo 3 variants.}
    \label{fig:sbs-veo}
\end{figure}

We recognize the importance of precise prompt adherence and will prioritize further improvements in this area to close the gap, while continuing to leverage our strengths in visual and temporal realism. The results are presented in Figure~\ref{fig:sbs-veo}.

\subsubsection{Kandinsky 5.0 Video Pro vs Wan 2.2 A14B}

\begin{figure}[htbp]
    \centering
    \begin{subfigure}[t]{0.48\textwidth}
    \centering
    \includegraphics[width=\linewidth]{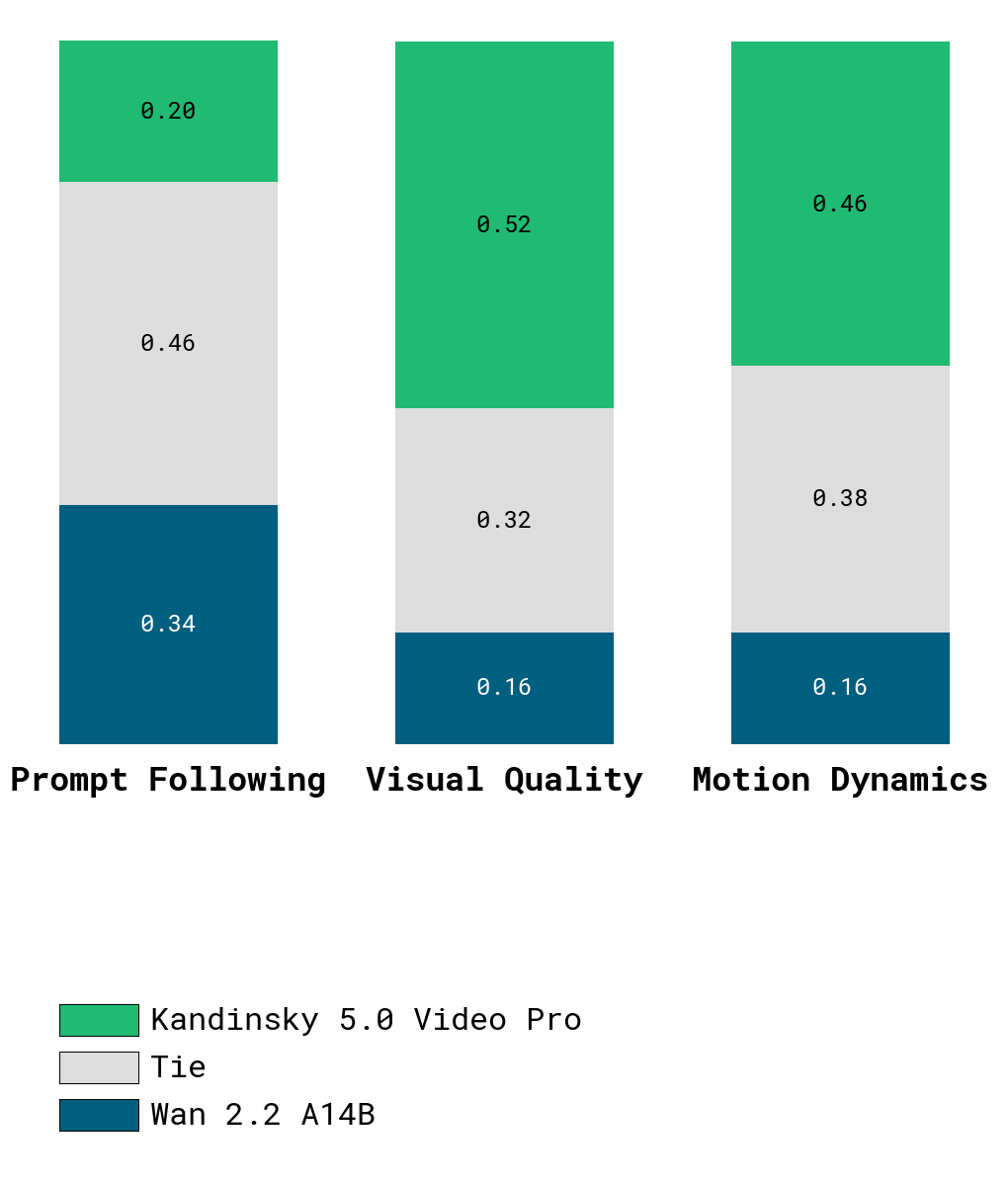}
    \caption{Text-to-Video mode} 
    \end{subfigure}
    \hfill
    \begin{subfigure}[t]{0.48\textwidth}
    \centering
    \includegraphics[width=\linewidth]{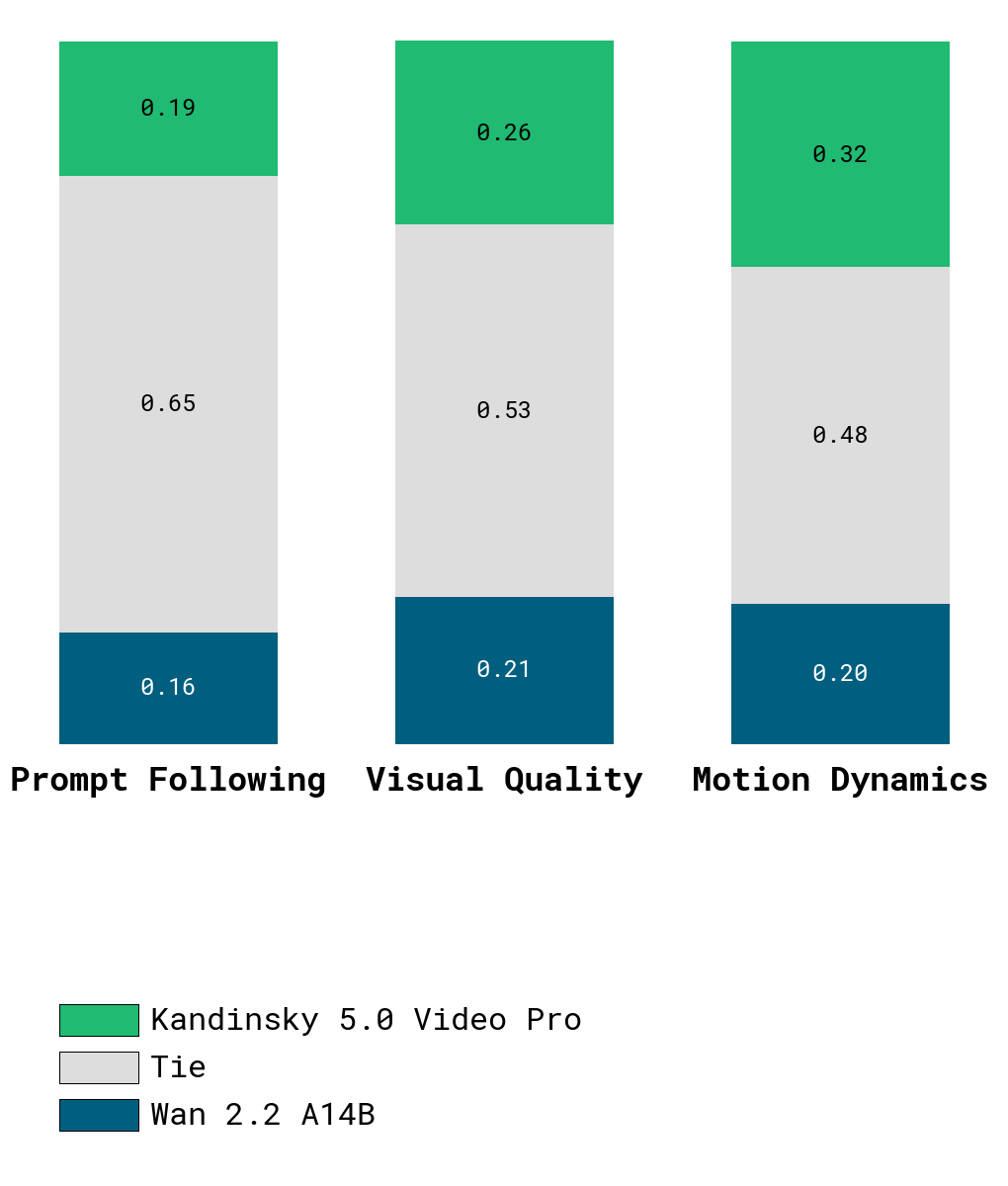}
    \caption{Image-to-Video mode} \end{subfigure}
    \caption{Kandinsky 5.0 Video Pro outperforms Wan 2.2 A14B models in Visual Quality and Motion Dynamics.}
    \label{fig:sbs-wan22-a14b}
\end{figure}

We also conducted a simplified side-by-side (SBS) comparison of Kandinsky 5.0 Video Pro with Wan 2.2 A14B in both text-to-video and image-to-video modes, evaluating performance on the same three criteria: Prompt Following, Visual Quality, and Motion Dynamics. The results are presented in Figure~\ref{fig:sbs-wan22-a14b}.

\begin{figure}[htbp]
    \centering
    \begin{minipage}{\textwidth}
    \begin{subfigure}[t]{0.45\textwidth}
    \centering
    \includegraphics[width=0.8\linewidth]{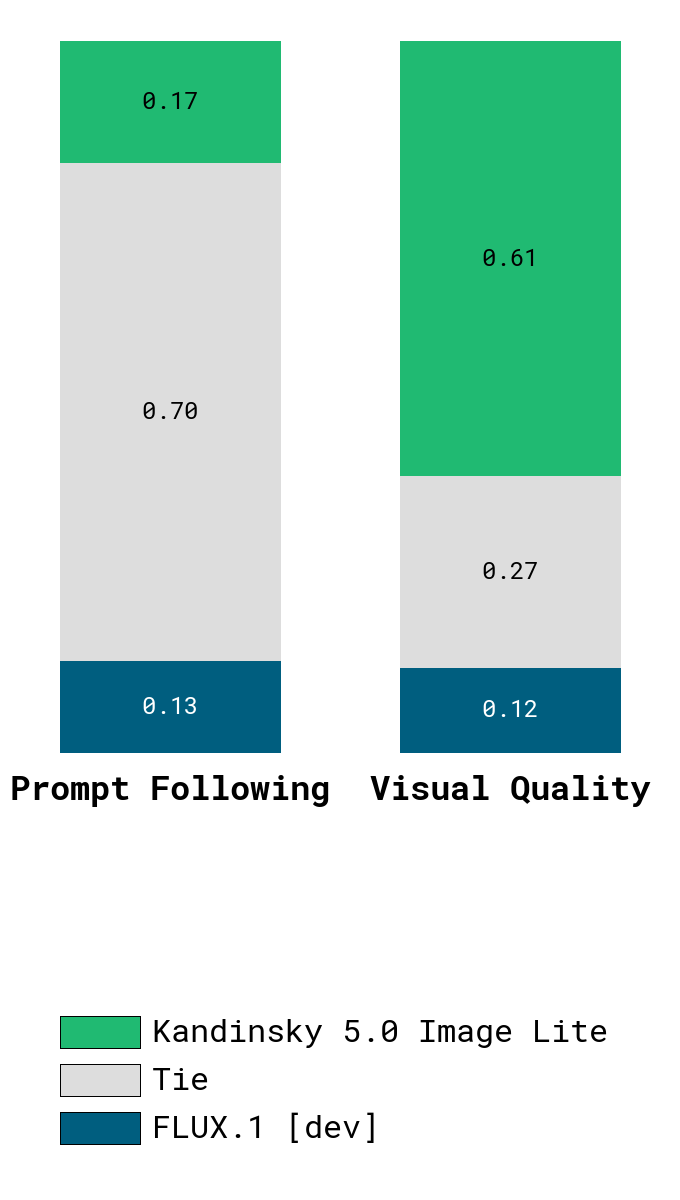}
    \caption{Comparison with FLUX.1 dev} 
    \end{subfigure}
    \hfill
    \begin{subfigure}[t]{0.45\textwidth}
    \centering
    \includegraphics[width=0.8\linewidth]{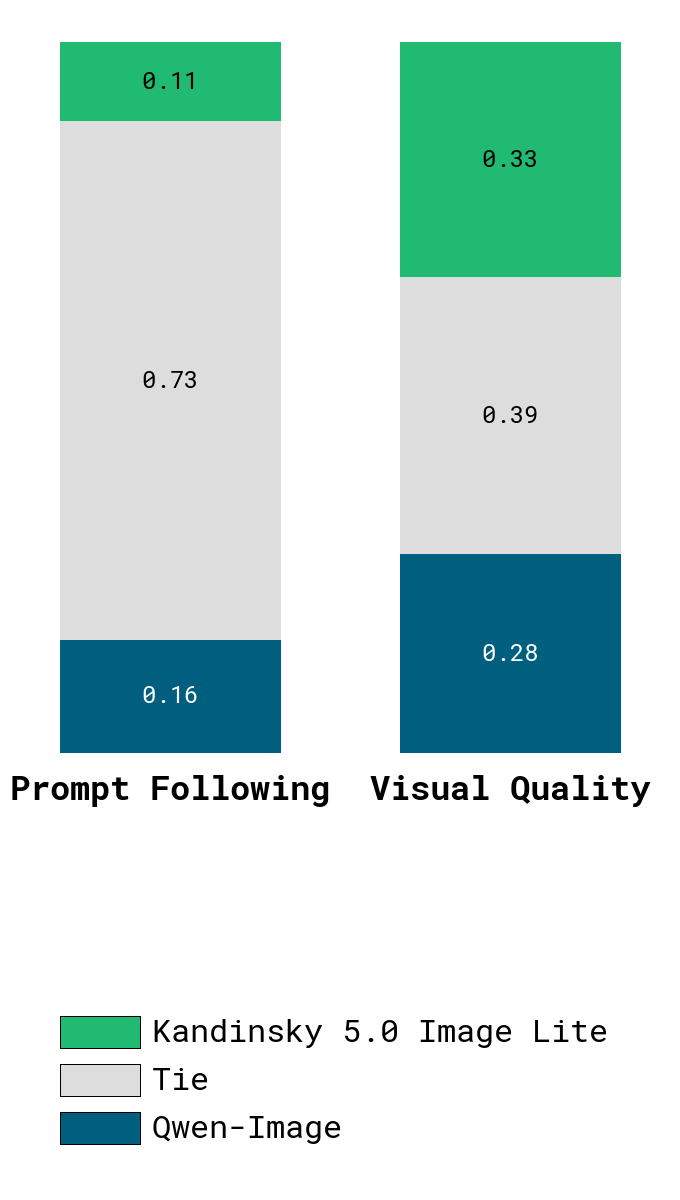}
    \caption{Comparison with Qwen-Image} \end{subfigure}
    \caption{\textbf{Kandinsky 5.0 Image Lite} demonstrated stronger performance in Visual Quality while remaining competitive in Prompt Following.}
    \label{fig:sbs-T2I}
    \end{minipage} \\
    \begin{minipage}{\textwidth}
    	\begin{subfigure}[t]{0.45\textwidth}
            \centering
            \includegraphics[width=0.8\linewidth]{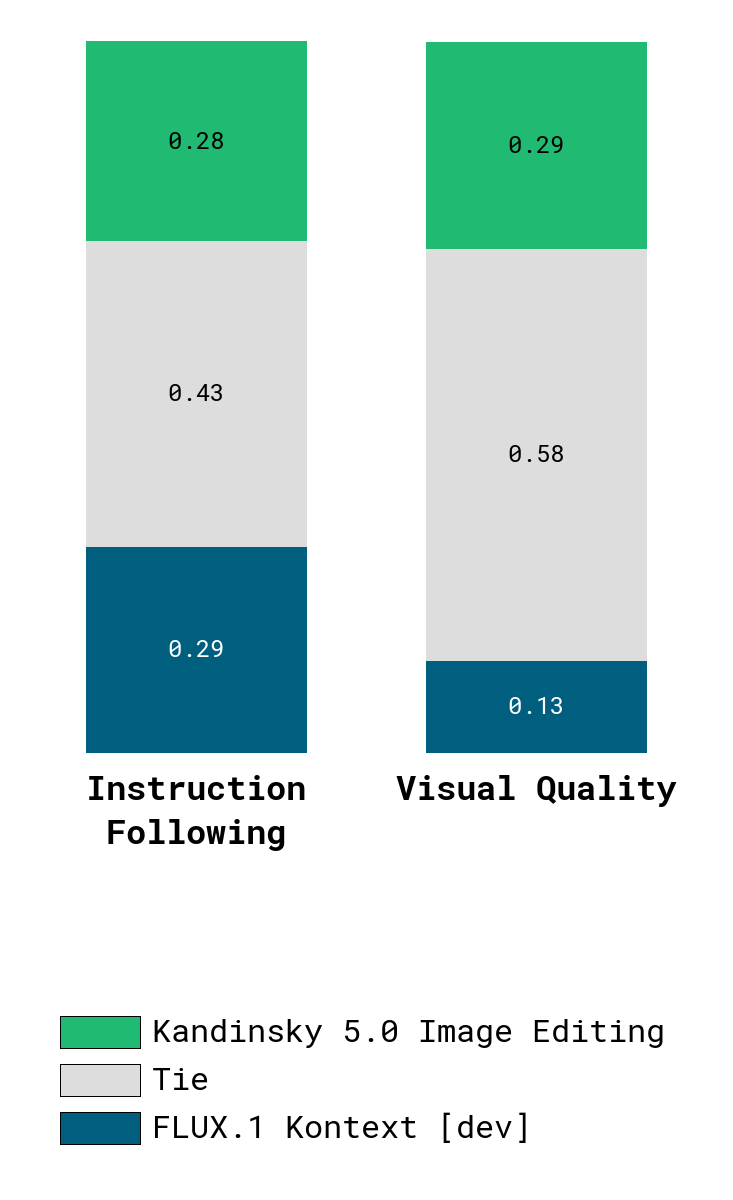}
            \caption{Comparison with FLUX.1~Kontext~[dev]}
        \end{subfigure}
        \hfill
        \begin{subfigure}[t]{0.45\textwidth}
            \centering
            \includegraphics[width=0.8\linewidth]{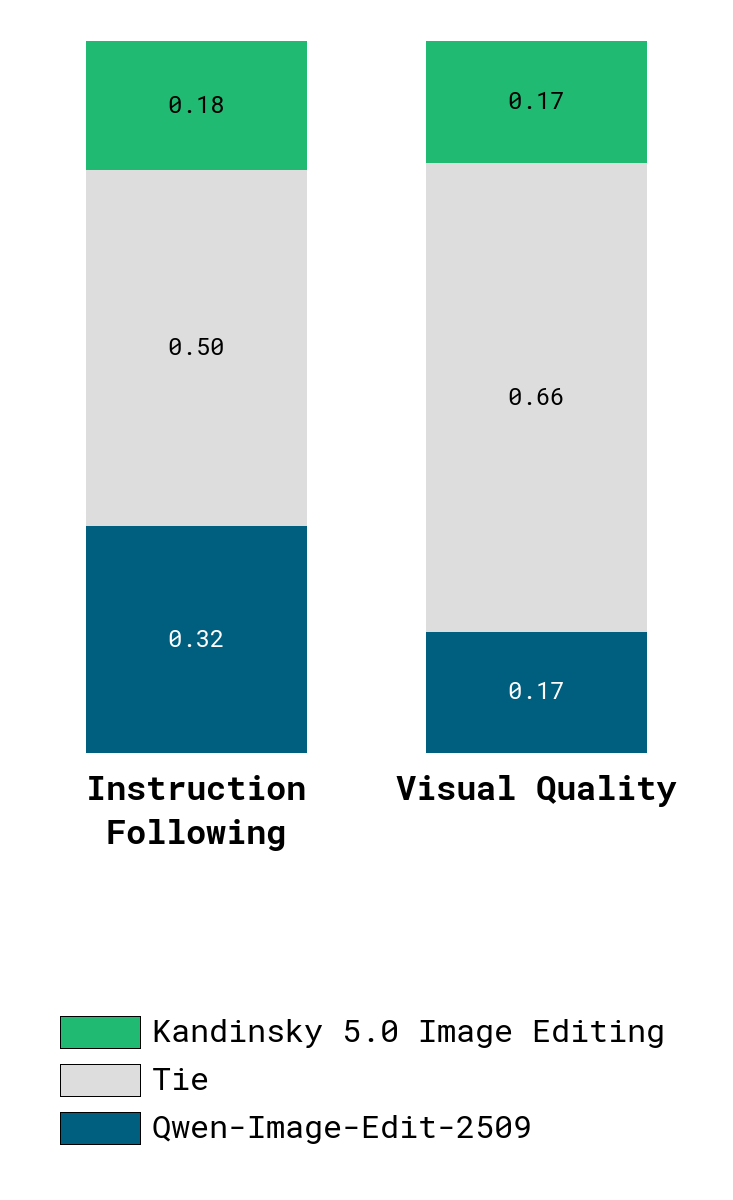}
            \caption{Comparison with Qwen-Image-Edit-2509}
        \end{subfigure}
        \caption{\textbf{Kandinsky 5.0 Image Editing} demonstrated competitive performance in against the evaluated models.}
        \label{fig:sbs-I2I}
    \end{minipage}
\end{figure}

\subsubsection{Kandinsky 5.0 Image Lite and Image Editing}

We conducted an internal simplified side-by-side (SBS) comparison of our \textbf{Kandinsky 5.0 Image Lite} text-to-image model with popular models, specifically evaluating \textbf{FLUX.1~[dev]} and \textbf{Qwen-Image}.

The test was performed using a custom prompt dataset based on the PartiPrompts (P2) dataset~\footnote{Original dataset available at: \url{https://huggingface.co/datasets/nateraw/parti-prompts}}, which was further expanded using the Giga Max large language model. Expert evaluators then compared the generated images from the models based on only two key parameters: \textbf{prompt following} (how accurately the image reflects the given text description) and \textbf{visual quality} (encompassing aesthetics, coherence, and technical execution).
The results of this comparison are presented in Figure~\ref{fig:sbs-T2I}.

We also conducted an internal simplified side-by-side (SBS) comparison of our \textbf{Kandinsky 5.0 Image Editing} capabilities with popular image editing models, specifically evaluating \textbf{FLUX.1~Kontext~[dev]} and \textbf{Qwen-Image-Edit-2509}.

The test utilized a Kontext Bench dataset~\footnote{\url{https://huggingface.co/datasets/black-forest-labs/kontext-bench}}  of image-instruction pairs. Expert evaluators compared the edited images from all models based on two key parameters: \textbf{instruction following} (how accurately the edit reflects the given instruction) and \textbf{visual quality} (assessing the coherence, realism, and aesthetic quality of the edited regions within the final image).

The results of this comparison are presented in Figure~\ref{fig:sbs-I2I}.

\subsubsection{Kandinsky 5.0 Video Lite Flash}
\begin{figure}[tbp]
    \centering
    \begin{subfigure}[t]{0.48\textwidth}
    \centering
    \includegraphics[width=\linewidth]{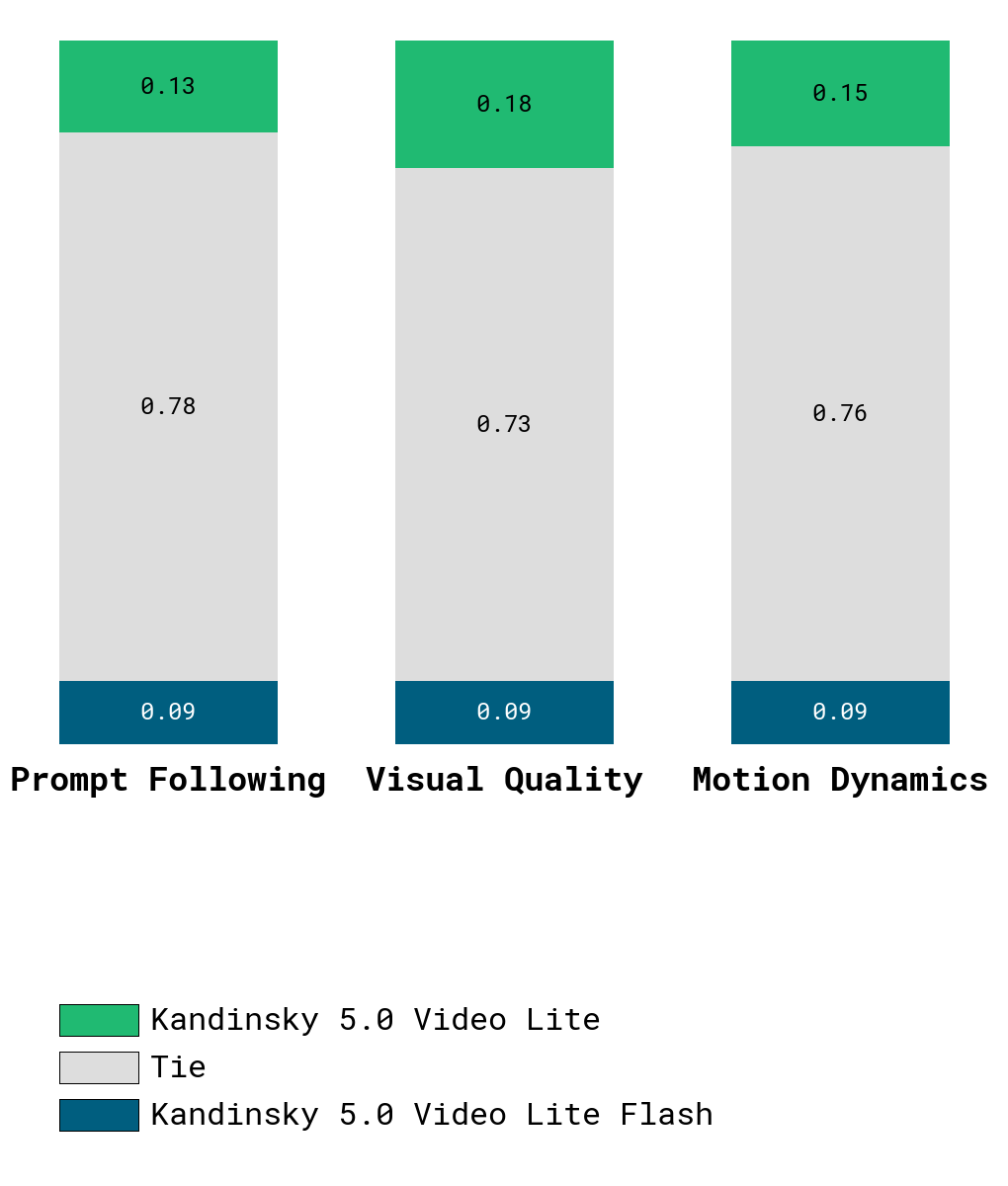}
    \caption{5-second model} 
    \end{subfigure}
    \hfill
    \begin{subfigure}[t]{0.48\textwidth}
    \centering
    \includegraphics[width=\linewidth]{images/sbs/sbs-T2V-Flash-10s.png}
    \caption{10-second model} \end{subfigure}
    \caption{Evaluation Kandinsky 5.0 Video Lite Flash model against not distilled Kandinsky 5.0 Video Lite.}
    \label{fig:sbs-Flash}
\end{figure} 
We conducted a simplified side-by-side (SBS) comparison between Kandinsky 5.0 Video Lite and Kandinsky 5.0 Video Lite Flash to assess the quality trade-offs associated with model distillation and optimization for speed. The evaluation covered both 5-second and 10-second generation lengths, using the same criteria as in prior comparisons: Prompt Following, Visual Quality, and Motion Dynamics.

Results (Figure~\ref{fig:sbs-Flash})indicate a measurable but generally moderate drop in performance for the Flash variant, particularly in fine detail rendering, temporal coherence, and handling of complex prompt semantics. This reflects the inherent trade-off between reduced computational cost, faster inference, and output fidelity.

However, the degradation is not critical and varies depending on the use case — lighter prompts and shorter durations show minimal perceptible difference. These findings suggest that Kandinsky 5.0 Video Lite Flash remains a viable option for applications where speed and efficiency are prioritized over maximum visual precision. The results support flexible deployment across the model family based on scenario-specific requirements.

\section[Use cases]{\color{Green} Use cases}\label{sec:use_cases}

\subsection{Text-to-Image}
\begin{figure}[htbp]
    \centering
    \begin{minipage}{\textwidth}
    	\centering
        \begin{subfigure}{0.32\textwidth}
            \centering
            \includegraphics[width=0.9\linewidth]{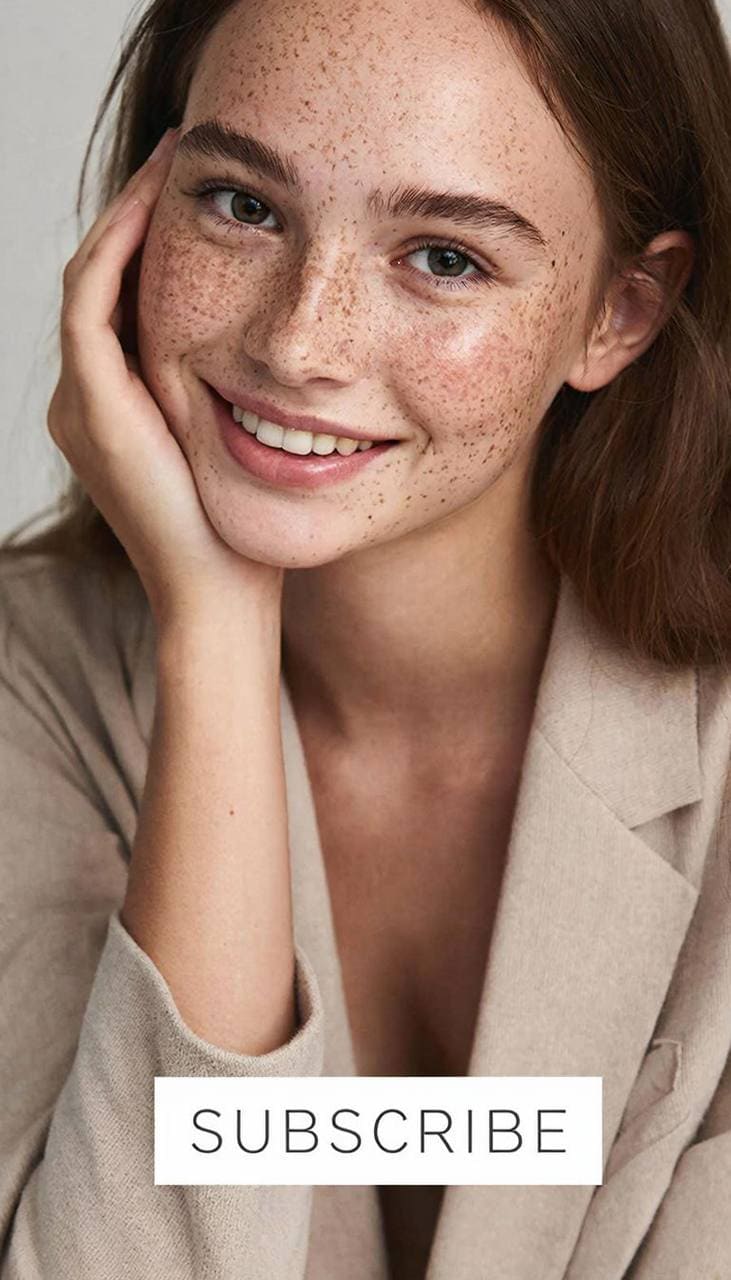}
            \caption{Original prompt: A beautiful, stylish girl with freckles looks at the camera and smiles sweetly. The caption reads: "Subscribe".}
        \end{subfigure}    
        \hfill
        \begin{subfigure}{0.32\textwidth}
            \centering
            \includegraphics[width=0.9\linewidth]{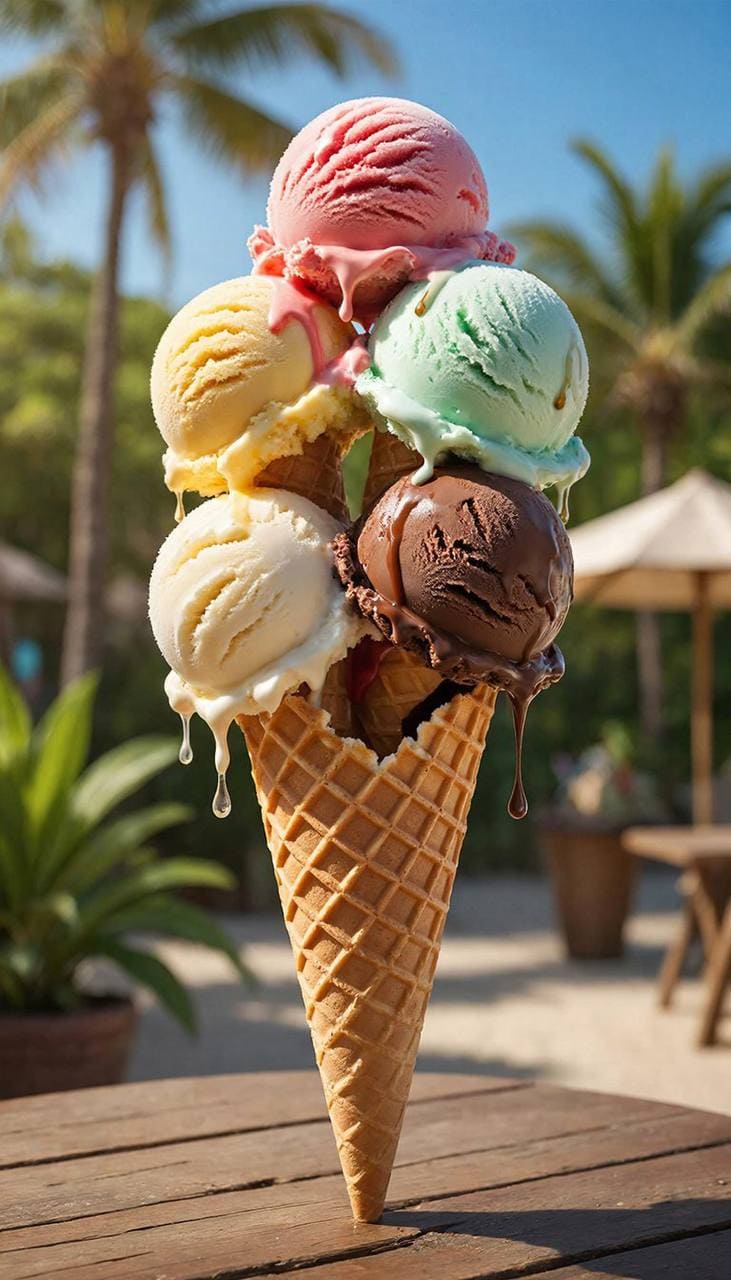}
            \caption{Original prompt: 5 scoops of ice cream, hot summer day.\\ \\}
        \end{subfigure}    
        \hfill
        \begin{subfigure}{0.32\textwidth}
            \centering
            \includegraphics[width=0.9\linewidth]{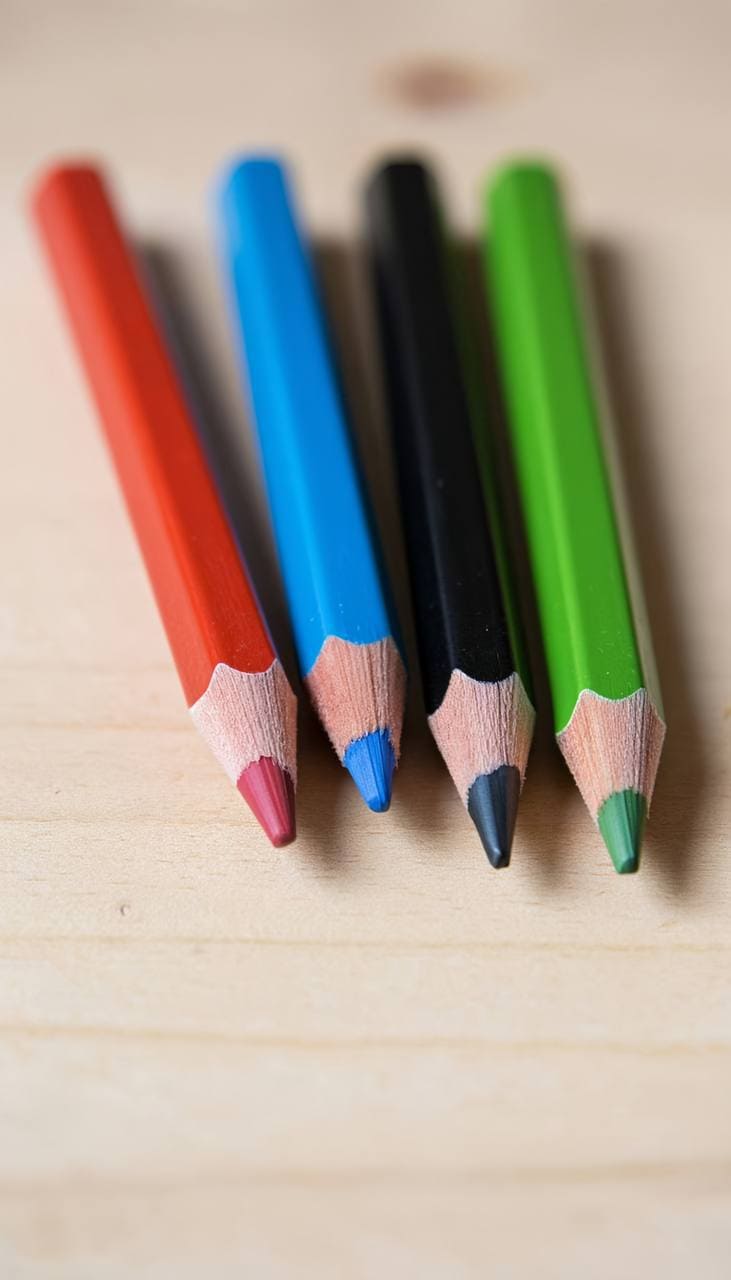}
            \caption{Original prompt: The pencils are laid out in order: red, blue, black, green.\\}
        \end{subfigure}
    \end{minipage} \\
    \vspace{0.5cm}
    \begin{minipage}{\textwidth}
    	\centering
        \begin{subfigure}{0.45\textwidth}
            \centering
            \includegraphics[width=\linewidth]{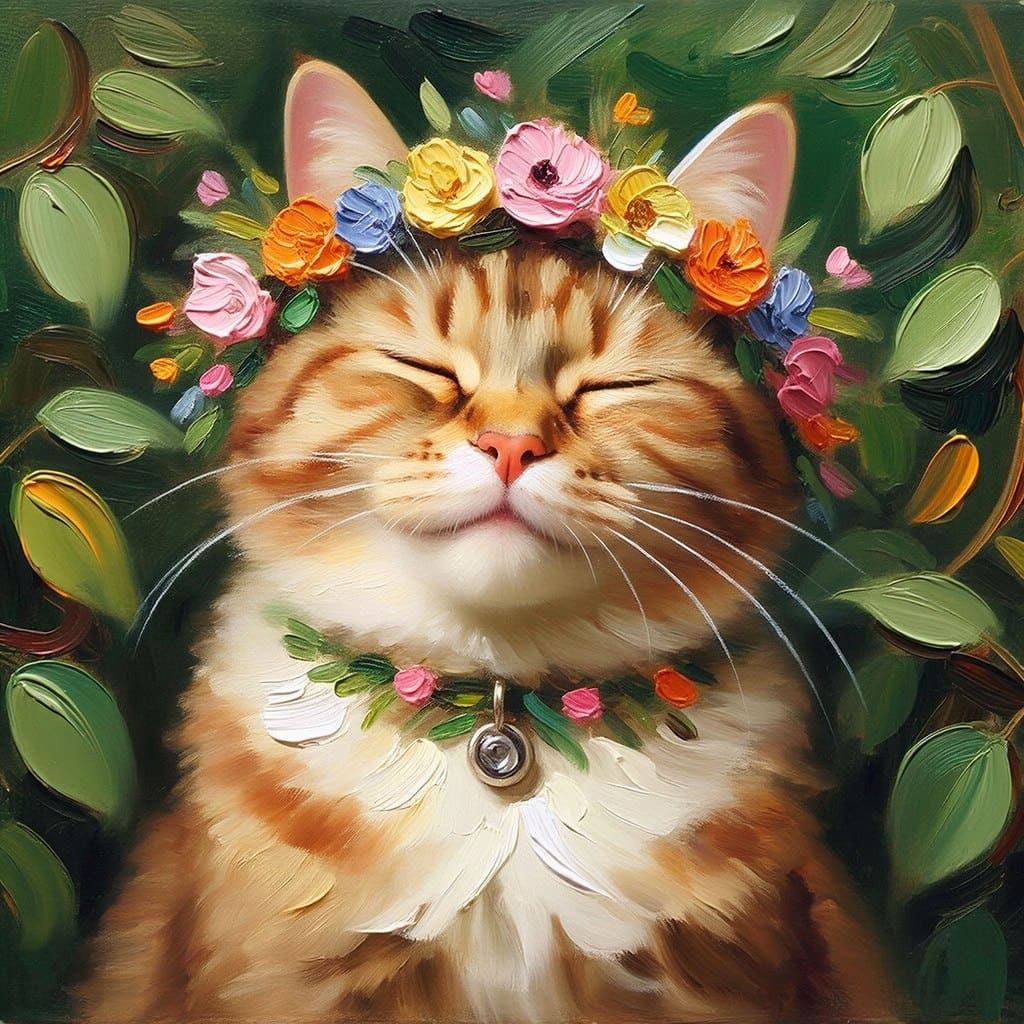}
            \caption{Original prompt: an oil painting depicting a contented cat in a bright floral crown and matching collar. The cat's eyes are closed, and there is a gentle smile on its face. The background is a rich green with texture. Artistic style: Impressionism. Color palette: warm and bright.\\ \\ \\ \\}
        \end{subfigure}    
        \hfill
        \begin{subfigure}{0.45\textwidth}
            \centering
            \includegraphics[width=\linewidth]{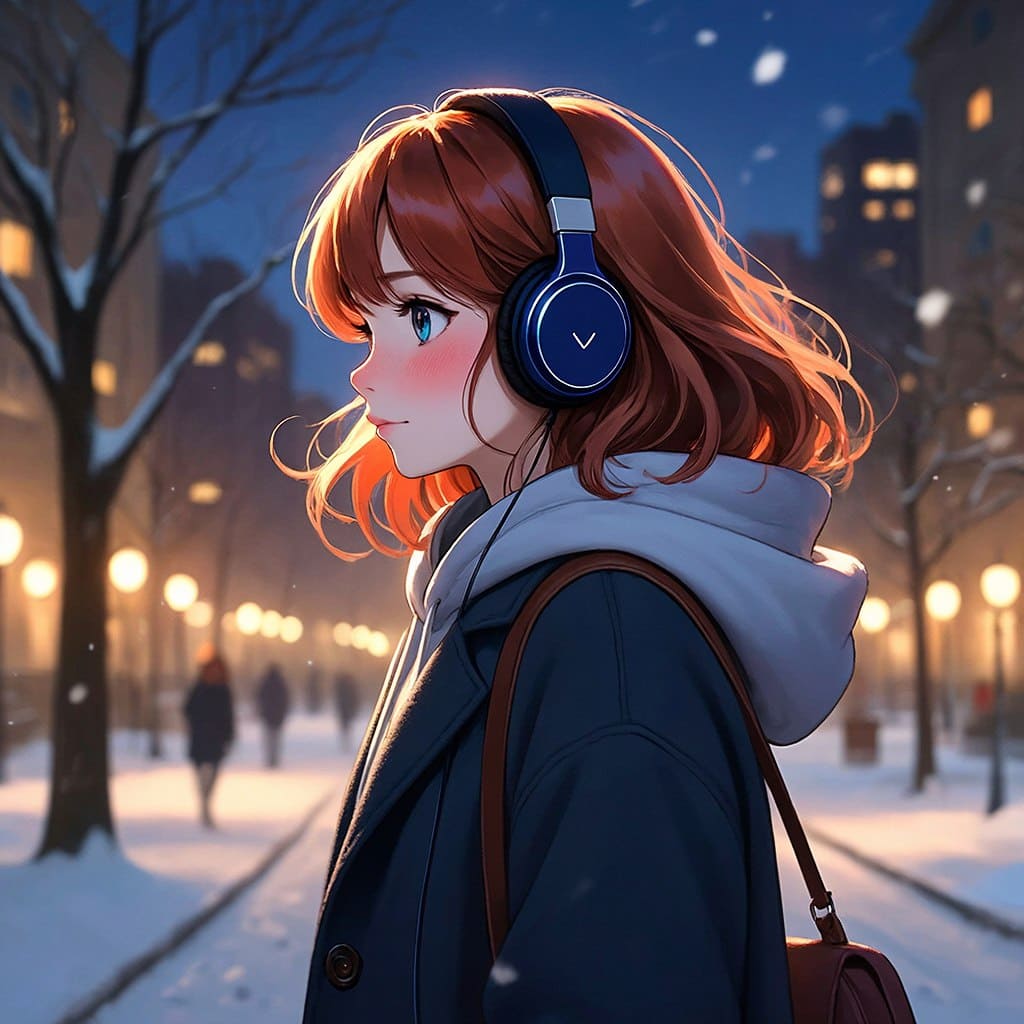}
            \caption{Original prompt: Anime-style winter evening scene: a girl in profile, facing left. Her copper-red wavy hair flutters in the air as she walks through the park. She is wearing a light hoodie and a dark coat. Large in-ear headphones with hearts complete the look. In the background is an unfocused night city with warm glowing windows, creating a pleasant contrast. The delicate glow and fine texture of the brush give the illustration a special depth and comfort.}
        \end{subfigure}
        \caption{Text-to-Image generation examples by Kandinsky 5.0 Image Lite}
        \label{fig:T2I_fig1}
    \end{minipage}
\end{figure}

\begin{figure}[htbp]
    \centering

    \begin{subfigure}{0.45\textwidth}
        \centering
        \includegraphics[width=\linewidth]{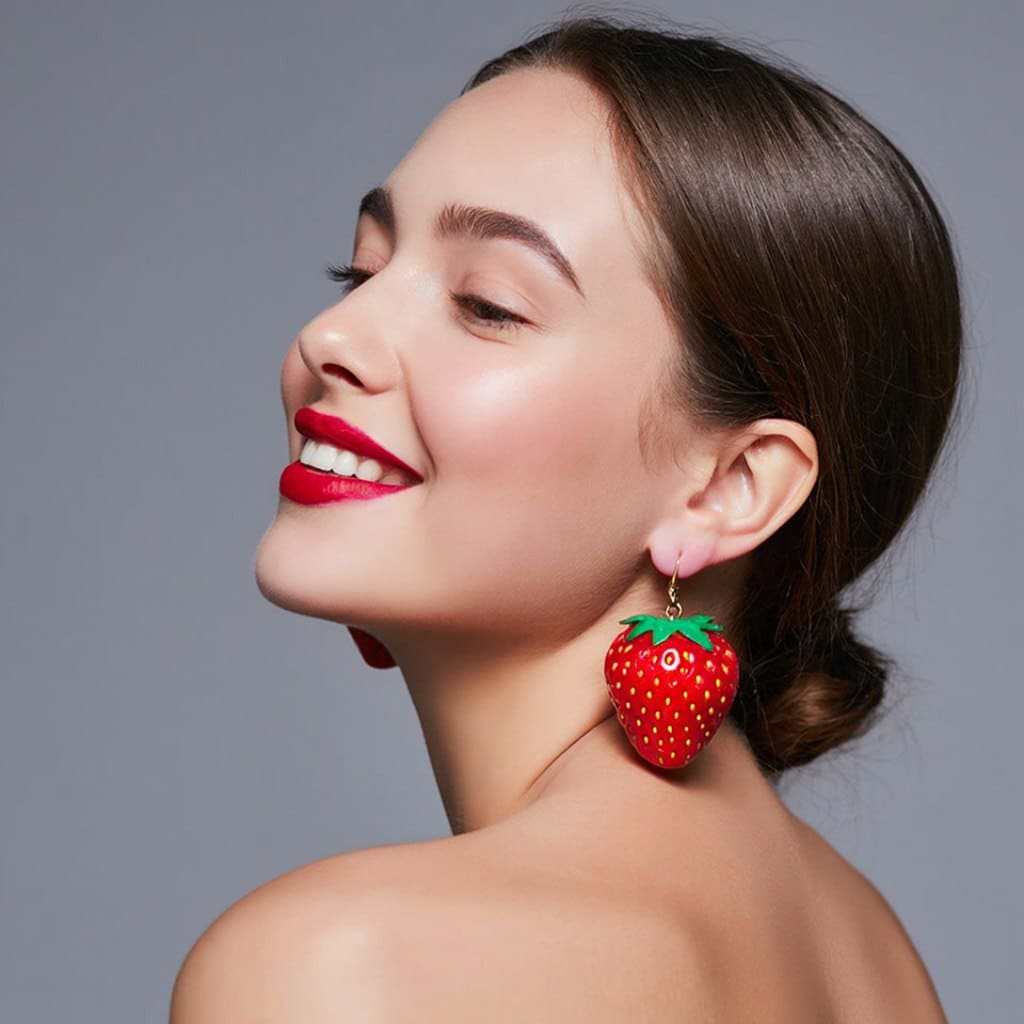}
        \caption{Original prompt: With a slight smile, the model smoothly turns her upper body, and it seems that the realistic strawberry-shaped earrings begin to shine. In a static photo, she is captured in full growth on a monochrome background, demonstrating deliberate minimalism, which draws attention to a bold fashionable image with red lips and fruit decorations in even studio lighting.}
        \label{fig:sample_t2i_4}
    \end{subfigure}    
    \hfill
    \begin{subfigure}{0.45\textwidth}
        \centering
        \includegraphics[width=\linewidth]{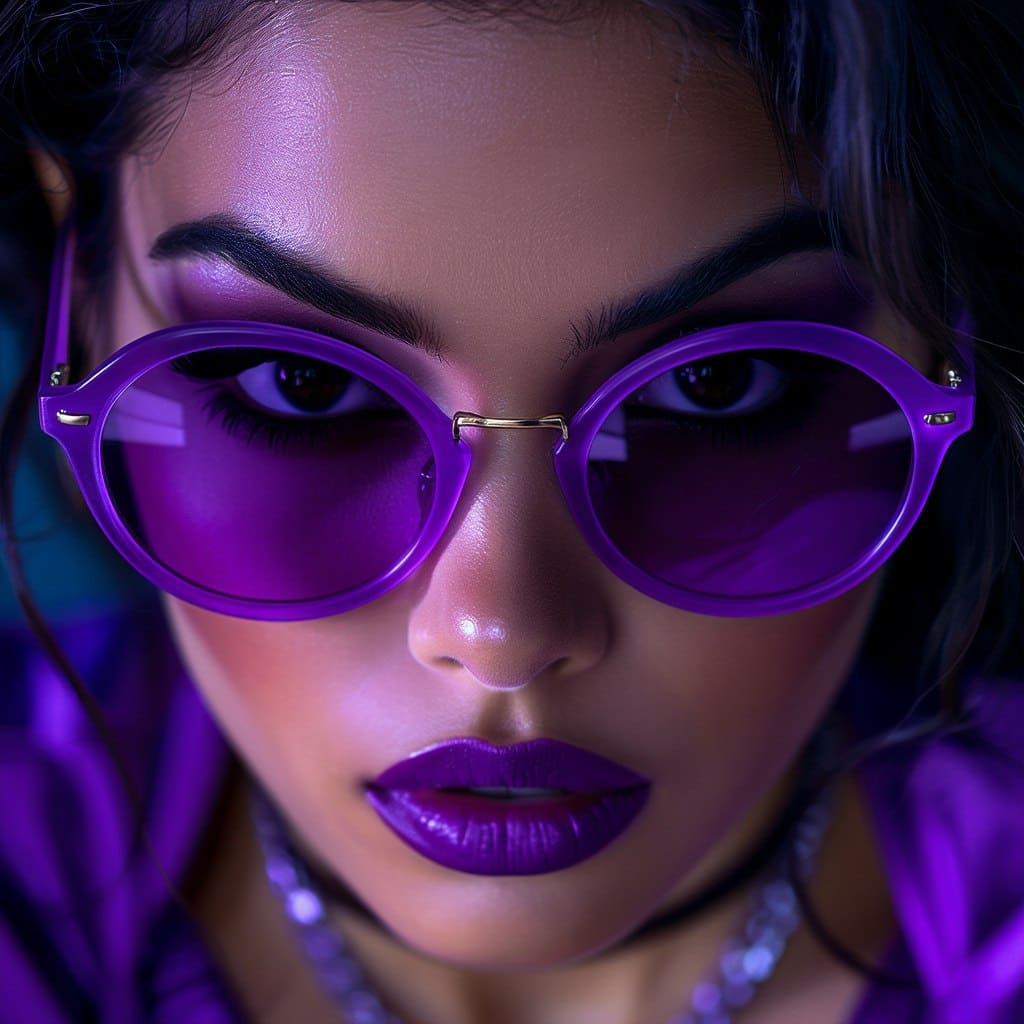}
        \caption{Original prompt: Digital image of a woman's face in close-up, low-angle view, slightly downcast gaze. She's wearing round purple sunglasses, shiny purple lipstick, and a necklace. The spectacular, high-contrast lighting highlights her facial features with bright highlights and deep shadows. Anime. Bright purple and dark tones. Mysterious, hyper-realistic, highly detailed, 4k, cinematic.}
        \label{fig:sample_t2i_1}
    \end{subfigure}    
    \vspace{0.5cm}
    
    \begin{subfigure}{0.45\textwidth}
        \centering
        \includegraphics[width=\linewidth]{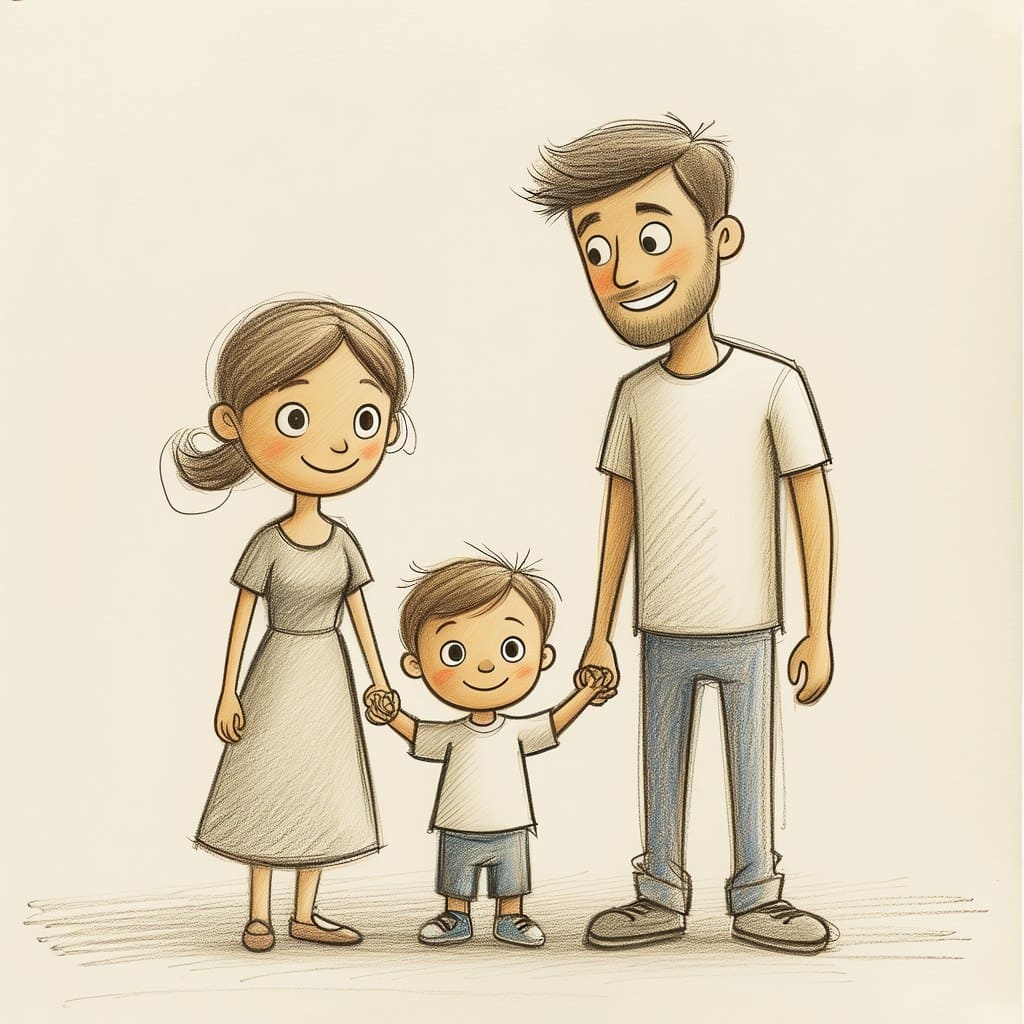}
        \caption{Original prompt: Children's pencil illustration: family — mom, dad, child. A naive, touching drawing made by a child.\\ \\}
        \label{fig:sample_t2i_3}
    \end{subfigure}
    \hfill
    \begin{subfigure}{0.45\textwidth}
        \centering
        \includegraphics[width=\linewidth]{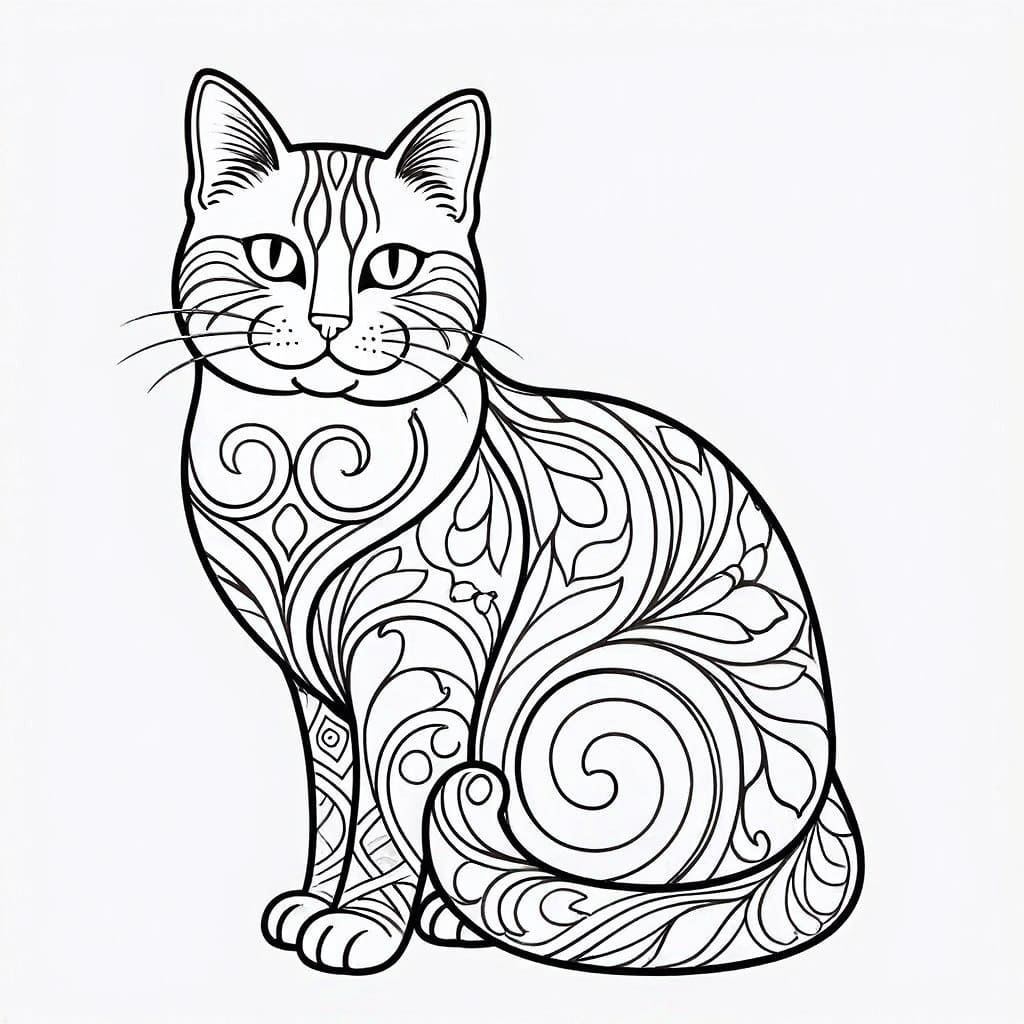}
        \caption{Original prompt: Linear illustration on a white background, a Meinkun cat whose coat smoothly merges into swirling ornamental floral patterns and arabesques. Suitable for coloring pages.}
    \end{subfigure}    
    \caption{Text-to-Image generation examples by Kandinsky 5.0 Image Lite}
    \label{fig:T2I_fig2}
\end{figure}

\textbf{Kandinsky 5.0 Image Lite} model line-up contains a powerful text-to-image generative model capable of producing highly diverse visual content with strong semantic alignment to input prompts. The model excels in photorealistic image synthesis, accurately rendering lighting, textures, and fine details that closely match real-world appearances. Beyond realism, it supports a wide range of artistic styles and media simulations — including oil and acrylic paintings, watercolor washes, pencil and charcoal sketches, and wax crayon drawings — allowing users to generate images that mimic specific traditional or digital art techniques. Additionally, the model can generate custom logos, typographic compositions, and even render legible text within images when explicitly prompted. Representative examples of these capabilities are illustrated in Figures~\ref{fig:T2I_fig1} and \ref{fig:T2I_fig2}.

\begin{figure}[htbp]
    \centering
    \begin{subfigure}{0.49\textwidth}
        \centering
        \includegraphics[width=\linewidth]{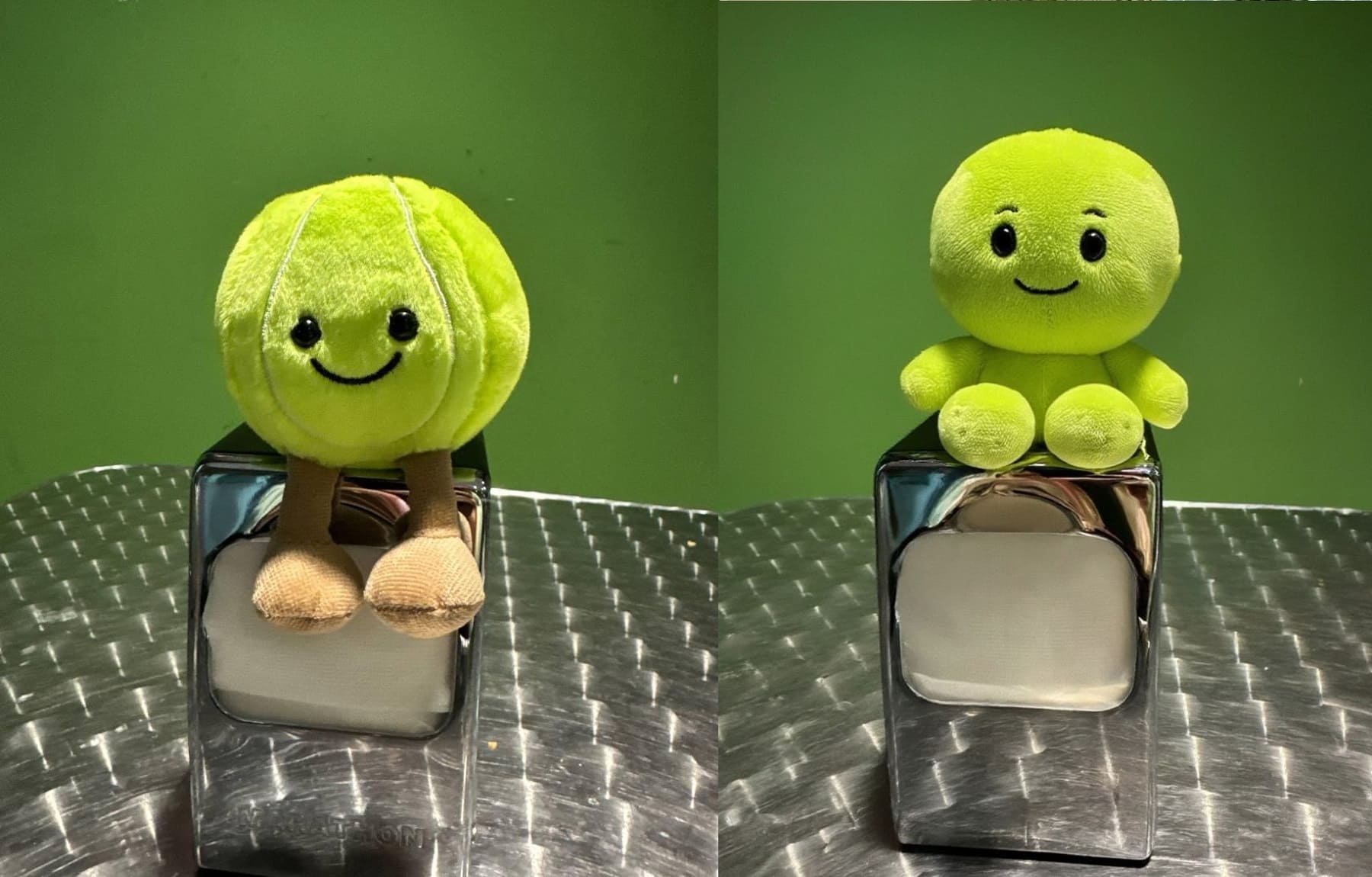}
        \caption{Instruction: Transform into a human.}
    \end{subfigure}    
    \hfill
    \begin{subfigure}{0.49\textwidth}
        \centering
        \includegraphics[width=\linewidth]{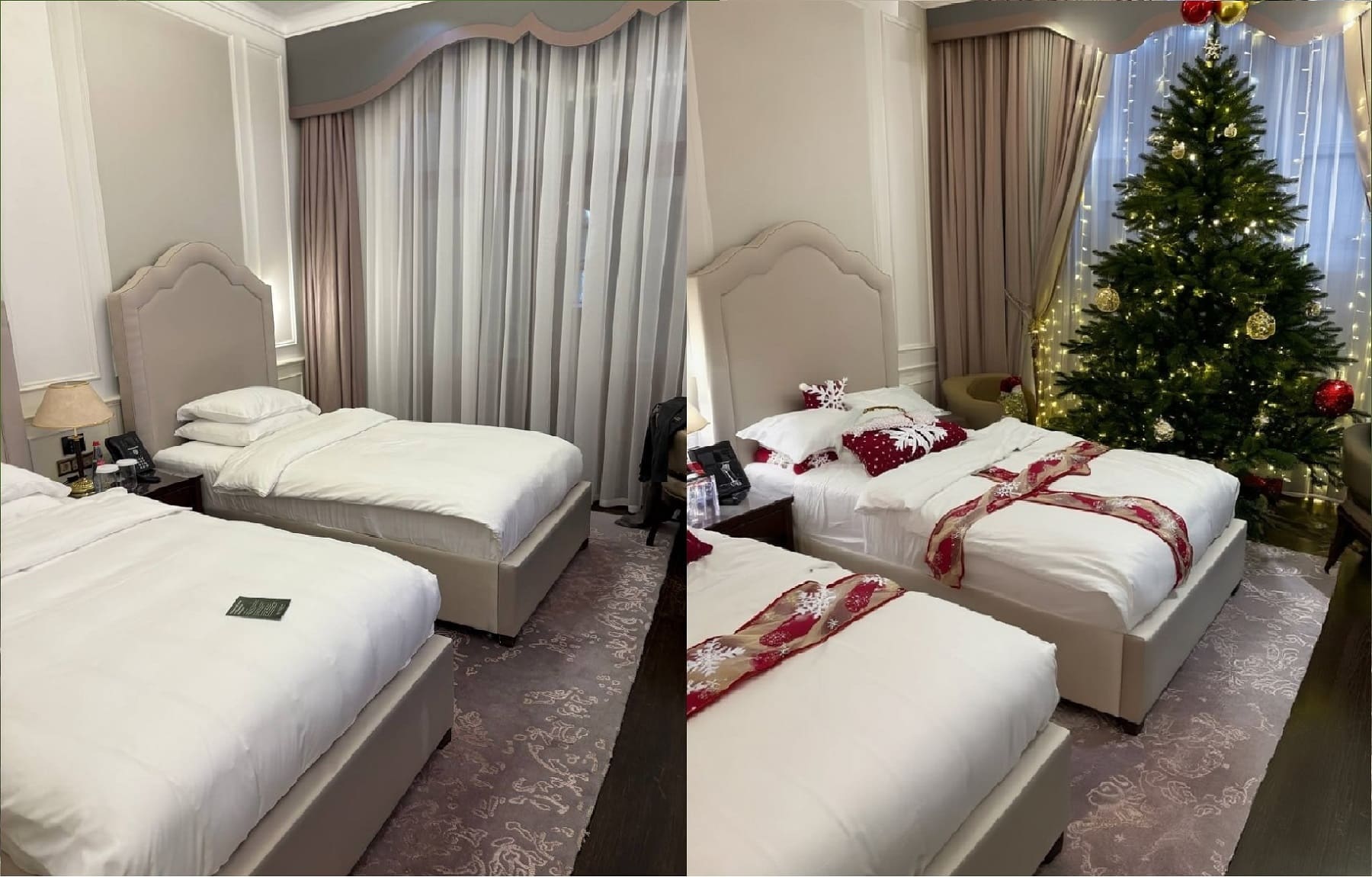}
        \caption{Instruction: Decorate the room for New Year.}
    \end{subfigure}    

    \vspace{0.5cm}
    \begin{subfigure}{0.49\textwidth}
        \centering
        \includegraphics[width=\linewidth]{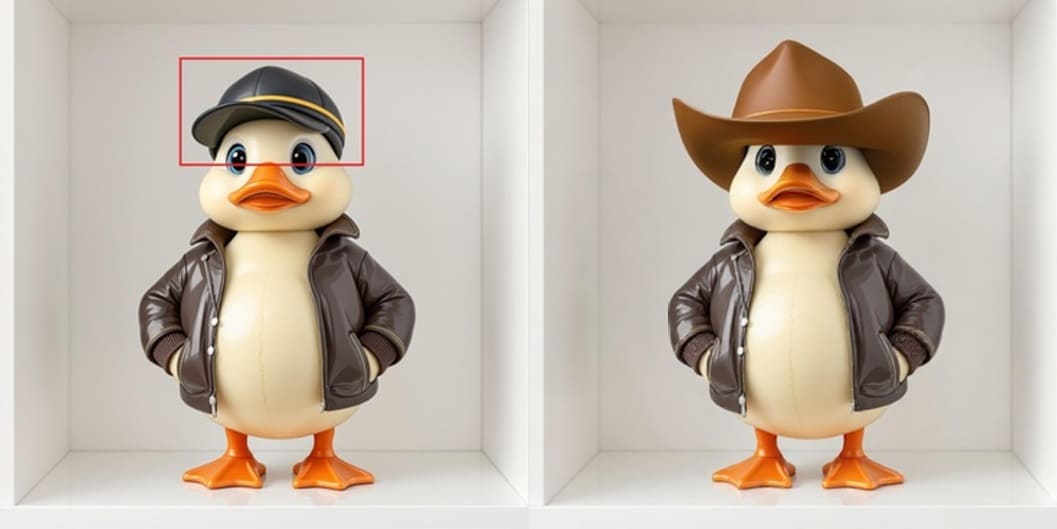}
        \caption{Instruction: Change this to a cowboy hat.}
    \end{subfigure}
    \hfill
    \begin{subfigure}{0.49\textwidth}
        \centering
        \includegraphics[width=\linewidth]{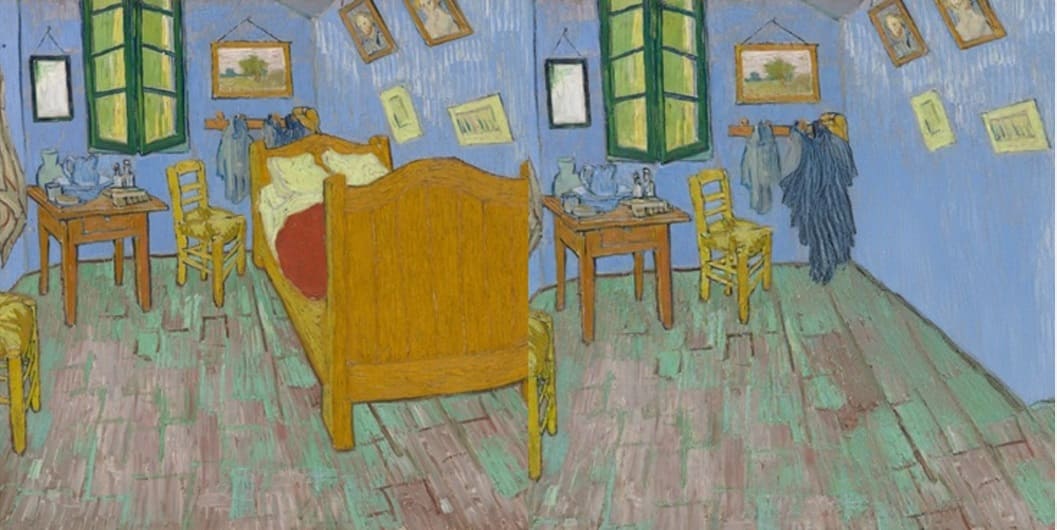}
        \caption{Instruction: Remove the bed.}
    \end{subfigure}    

    \vspace{0.5cm}
    \begin{subfigure}{0.49\textwidth}
        \centering
        \includegraphics[width=\linewidth]{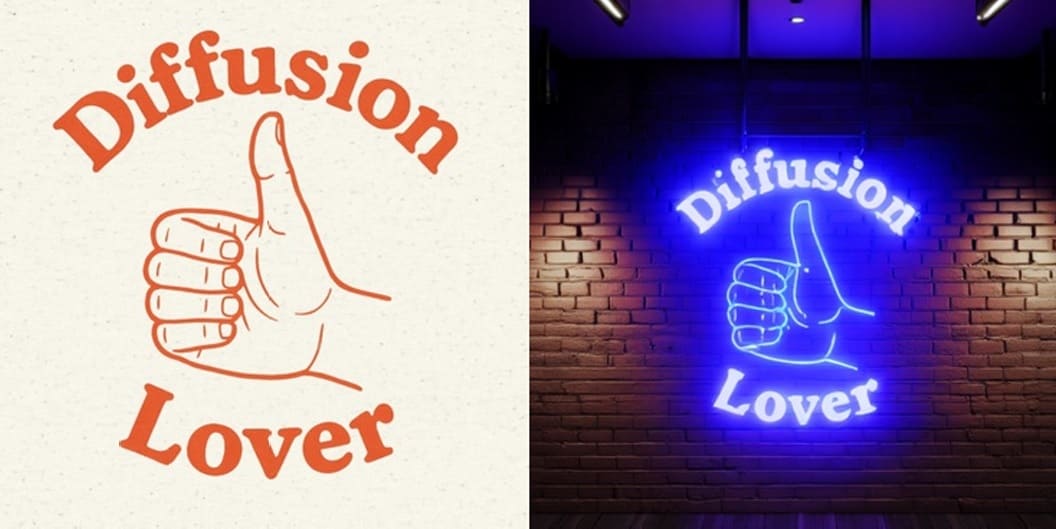}
        \caption{Instruction: Turn this into a neon sign hanging on a brick wall in a cool modern office.\\S}
    \end{subfigure}
    \hfill
    \begin{subfigure}{0.49\textwidth}
        \centering
        \includegraphics[width=\linewidth]{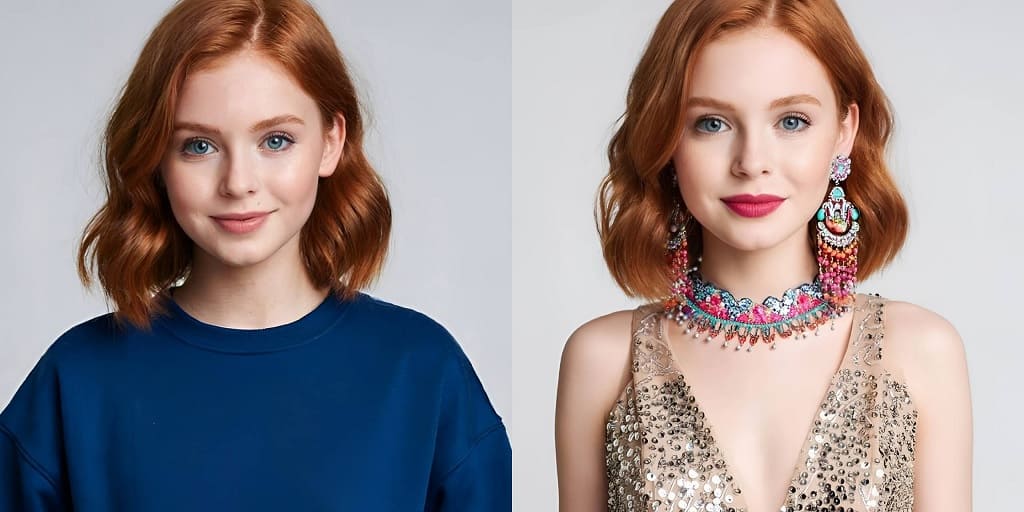}
        \caption{Instruction: Swap your sweatshirt for a sequined evening dress, add some bright jewelry, and brighten your lips and eyes. Keep the angle.}
    \end{subfigure}    

    \vspace{0.5cm}
    \begin{subfigure}{0.49\textwidth}
        \centering
        \includegraphics[width=\linewidth]{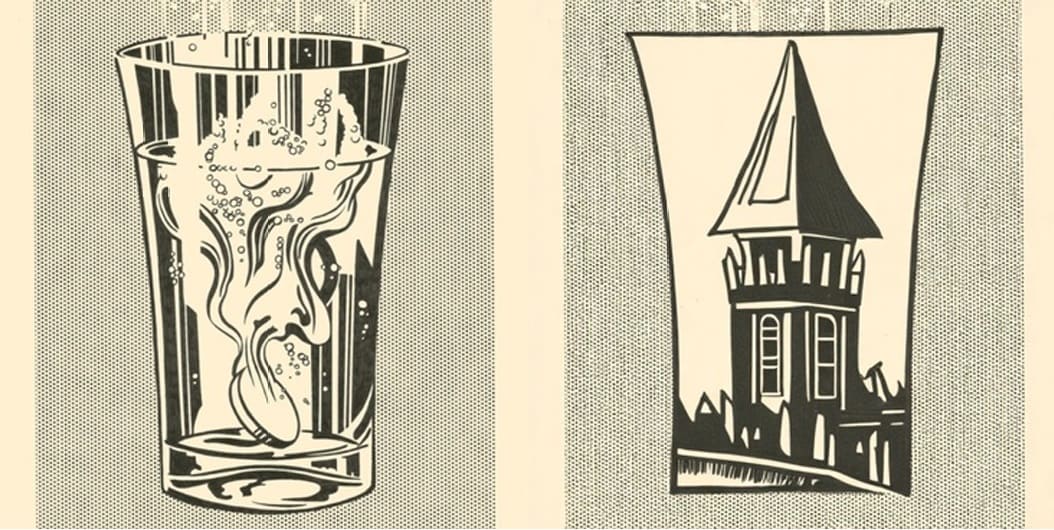}
        \caption{Instruction: Using this style create art of the wizards tower.}
    \end{subfigure}
    \hfill
    \begin{subfigure}{0.49\textwidth}
        \centering
        \includegraphics[width=\linewidth]{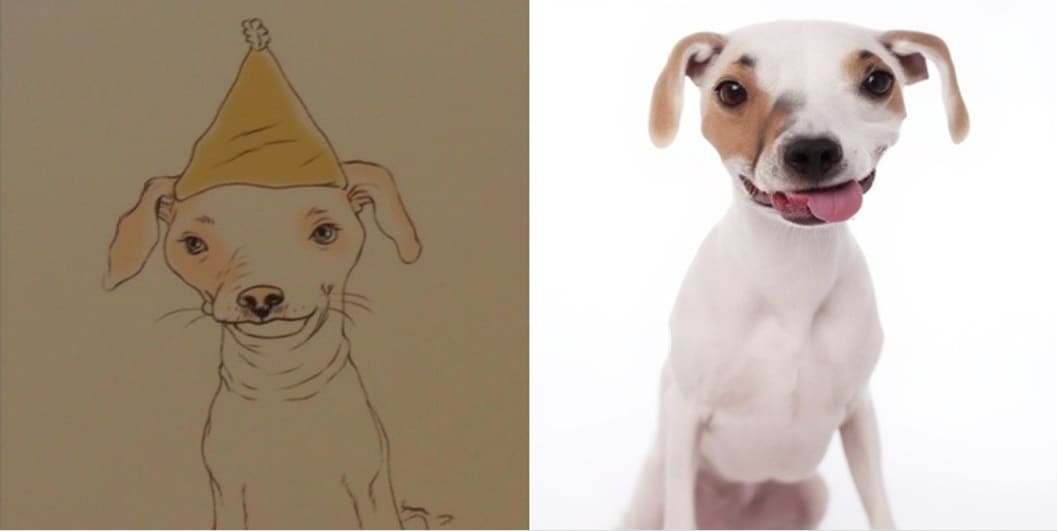}
        \caption{Instruction: Turn this into a real photograph of the same dog.}
    \end{subfigure}    
    
    \caption{Kandinsky 5.0 image-to-image examples}
    \label{fig:image_editing}
\end{figure}

\begin{figure}[htbp]
    \centering
    \captionsetup[subfloat]{labelfont=scriptsize,textfont=scriptsize}
    \subfloat[A small, animated rooster with fluffy white and brown feathers, a bright red comb and wattle, and a yellow beak stands on a person's open palm. The rooster has large, expressive eyes and is initially looking to the side. It then turns its head forward, spreads its wings wide in a welcoming or excited gesture, and opens its beak as if crowing or speaking. The background is a simple, out-of-focus indoor setting with a wooden door frame visible on the left.]
    {\label{subfig:chicken}
        \includegraphics[width=0.200\textwidth]{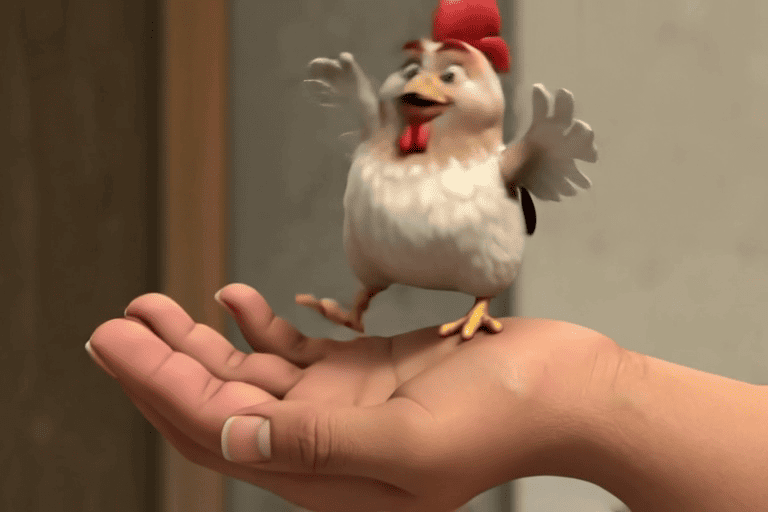} \hfill 
        \includegraphics[width=0.200\textwidth]{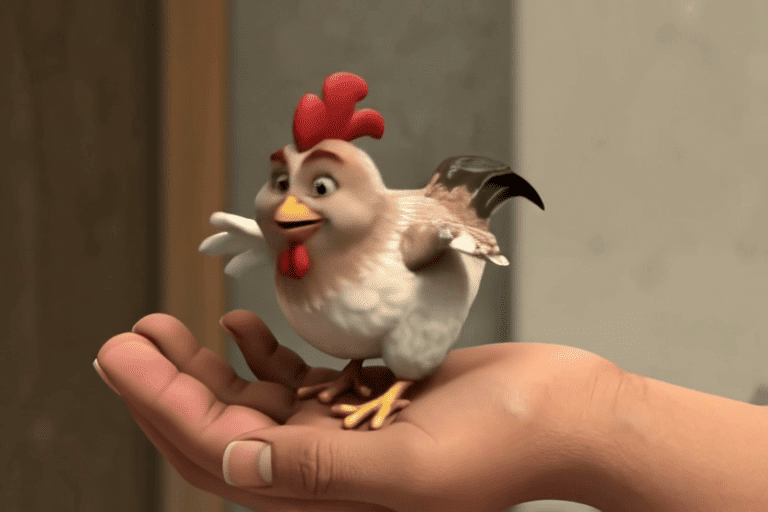} \hfill 
        \includegraphics[width=0.200\textwidth]{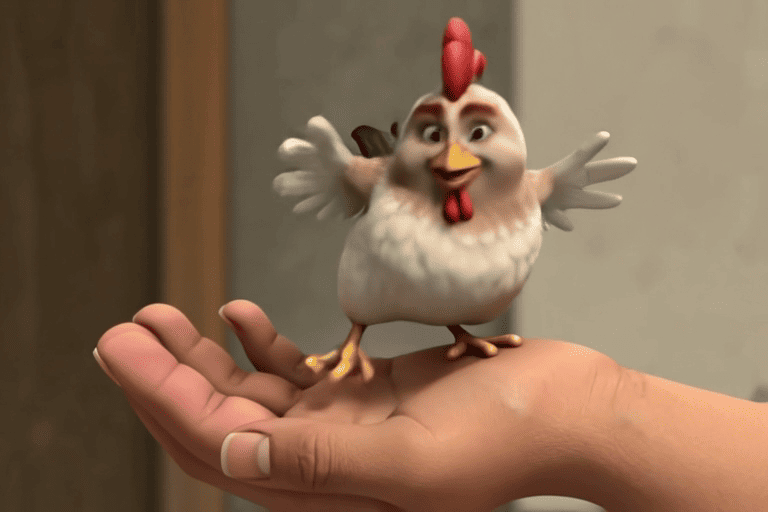} \hfill 
        \includegraphics[width=0.200\textwidth]{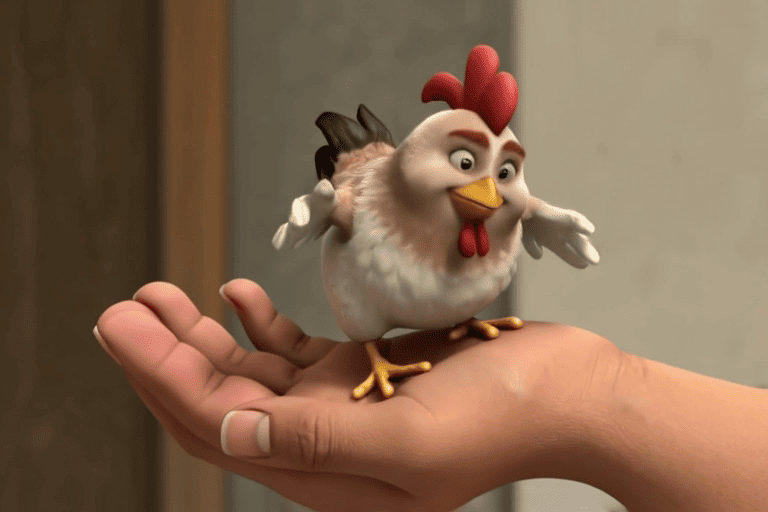} \hfill 
        \includegraphics[width=0.200\textwidth]{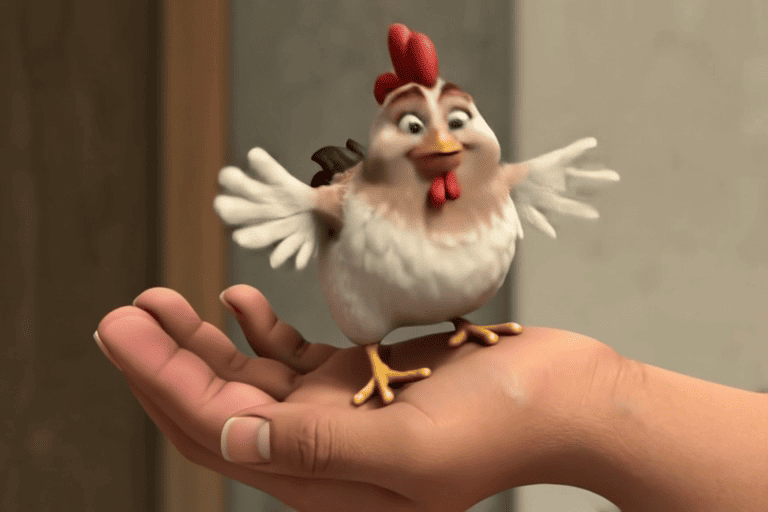}
    }
    \\
    \vspace{0.5cm}

    \subfloat[In a vibrant, futuristic city at night, a sleek, red sports car with glowing blue headlights speeds down a wide, empty highway. The cityscape is dominated by towering skyscrapers adorned with dazzling neon lights in shades of pink, purple, and blue, creating a cyberpunk atmosphere. The car's aerodynamic design and sharp angles reflect the advanced technology of this world. As the car moves forward, its headlights illuminate the road ahead, casting dynamic reflections on the glossy surface of the highway. The scene captures the essence of speed and innovation, with the car's motion suggesting a journey through this high-tech metropolis.]
    {\label{subfig:transformer}
        \includegraphics[width=0.200\textwidth]{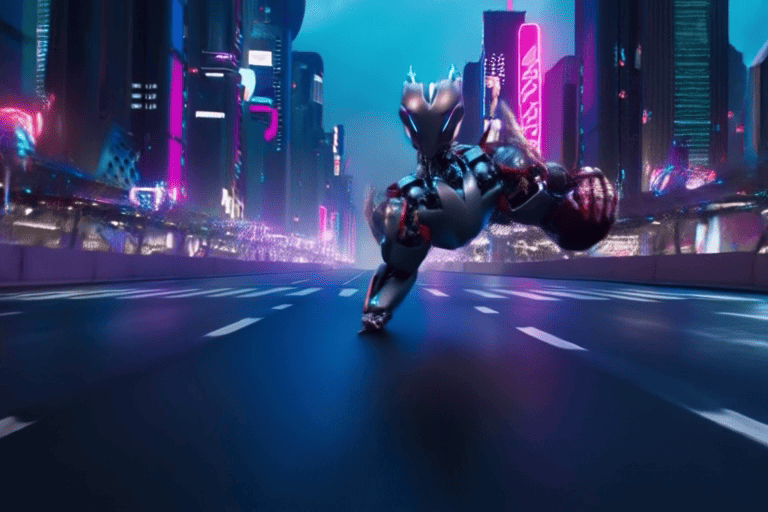} \hfill 
        \includegraphics[width=0.200\textwidth]{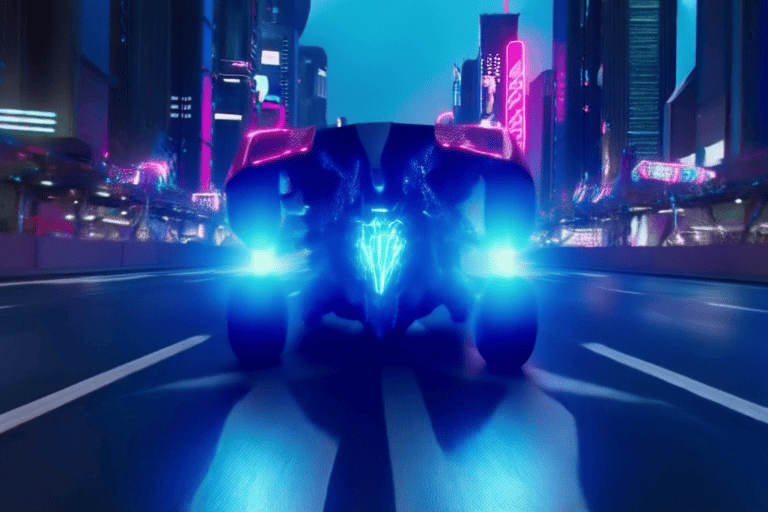} \hfill 
        \includegraphics[width=0.200\textwidth]{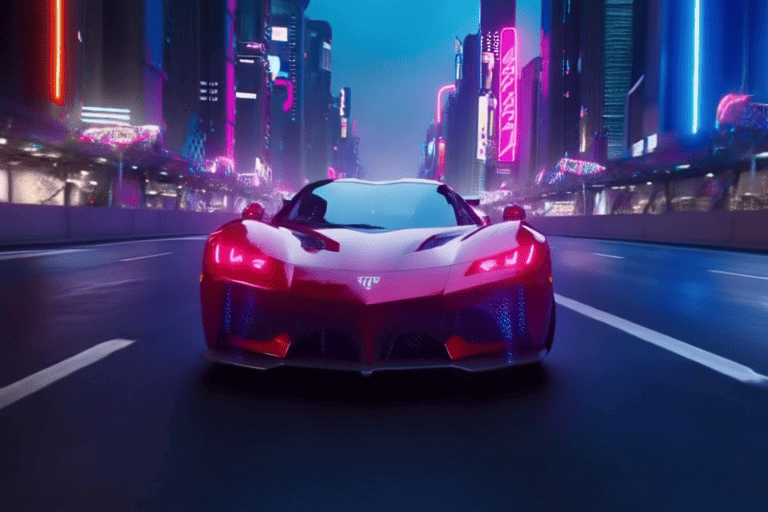} \hfill 
        \includegraphics[width=0.200\textwidth]{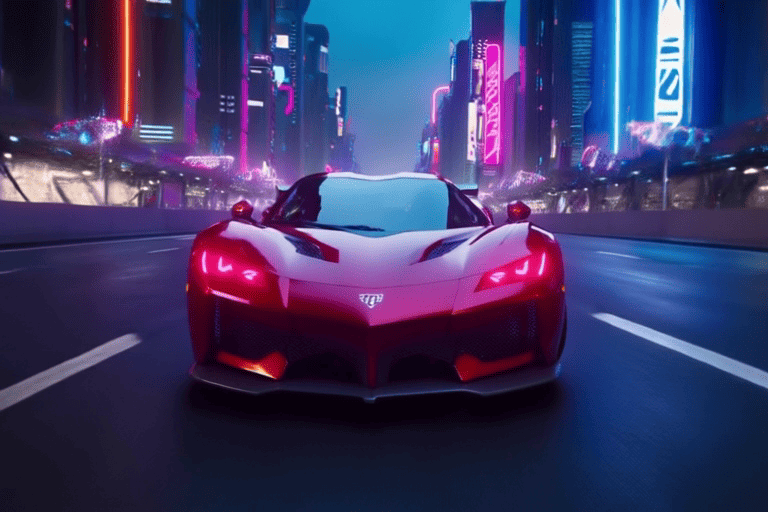} \hfill 
        \includegraphics[width=0.200\textwidth]{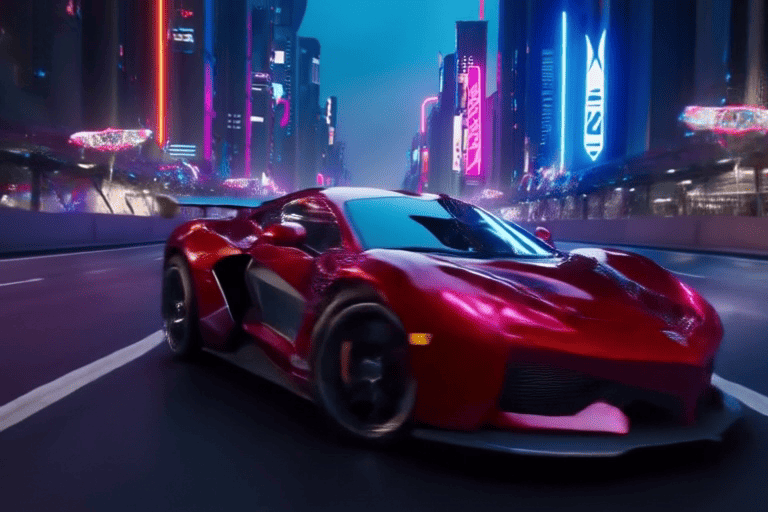}
    }
    \\
    \vspace{0.5cm}

    \subfloat[In the depths of the ocean, a colossal octopus with a deep red hue dominates the scene. Its massive body is adorned with intricate patterns, and its eight tentacles, each lined with rows of suction cups, extend outward in a menacing display. The octopus's eyes glow with an intense, fiery orange light, piercing through the murky blue-green water. The surrounding environment is rocky and rugged, with large boulders and coral formations scattered throughout. The water is thick with sediment, creating a hazy atmosphere that adds to the sense of mystery and danger. The octopus appears to be in a state of alertness, its tentacles twitching slightly as it surveys its surroundings. The overall scene is one of awe and intimidation, capturing the raw power and beauty of this deep-sea creature in its natural habitat.]
    {\label{subfig:octopus}
        \includegraphics[width=0.200\textwidth]{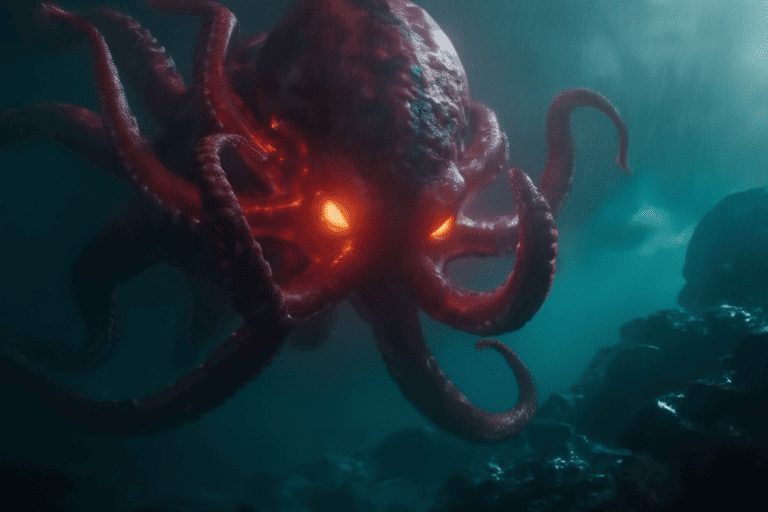} \hfill 
        \includegraphics[width=0.200\textwidth]{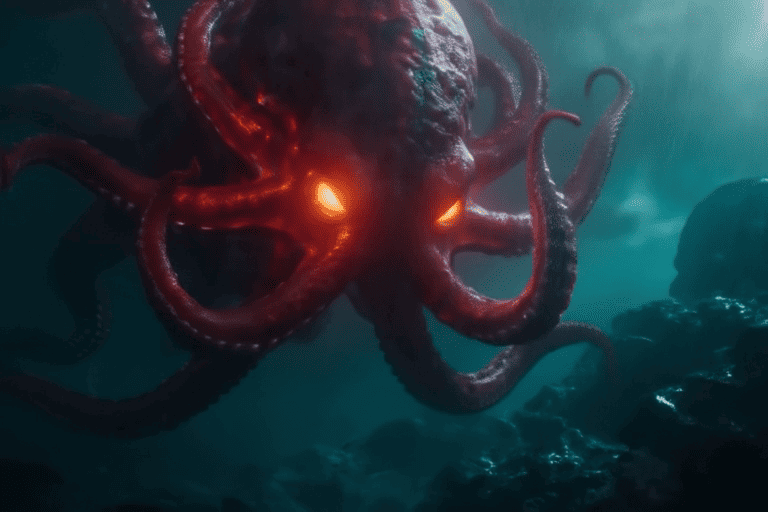} \hfill 
        \includegraphics[width=0.200\textwidth]{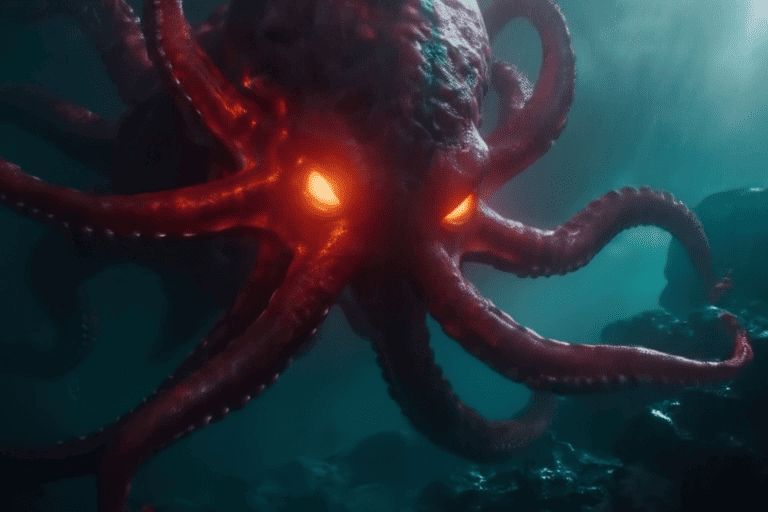} \hfill 
        \includegraphics[width=0.200\textwidth]{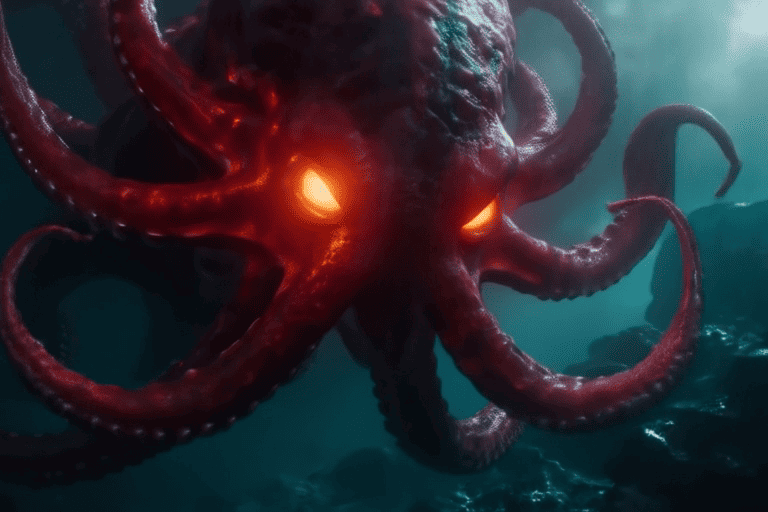} \hfill 
        \includegraphics[width=0.200\textwidth]{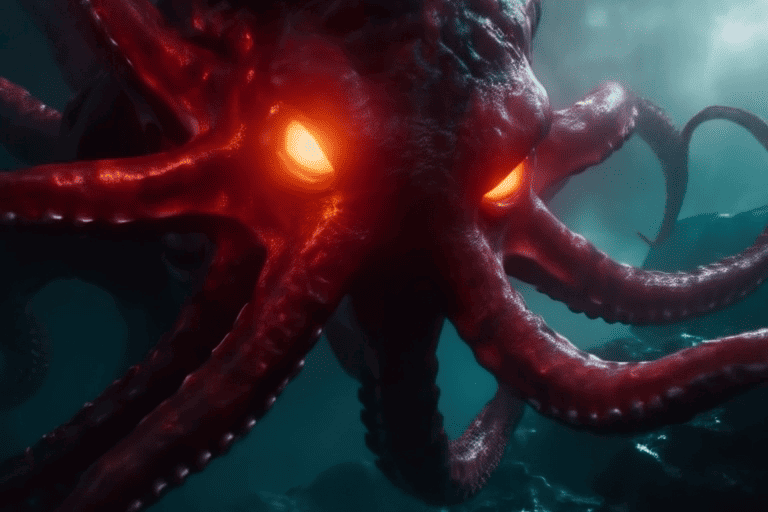}
    }
    \\
    \vspace{0.5cm}

    \subfloat[A young woman with shoulder-length brown hair is standing on a city sidewalk, leaning against a textured stone wall. She is wearing a dark, silky, three-quarter-sleeved top and has a black shoulder bag slung over her right shoulder. In her hands, she holds a white disposable coffee cup with a white lid. She is looking off to her left, her mouth slightly open as if she is in the middle of speaking or reacting to something. Her expression is engaged and animated. The background shows a modern building with large glass windows and a paved sidewalk with a few other pedestrians in the distance.]
    {\label{subfig:coffee}
        \includegraphics[width=0.200\textwidth]{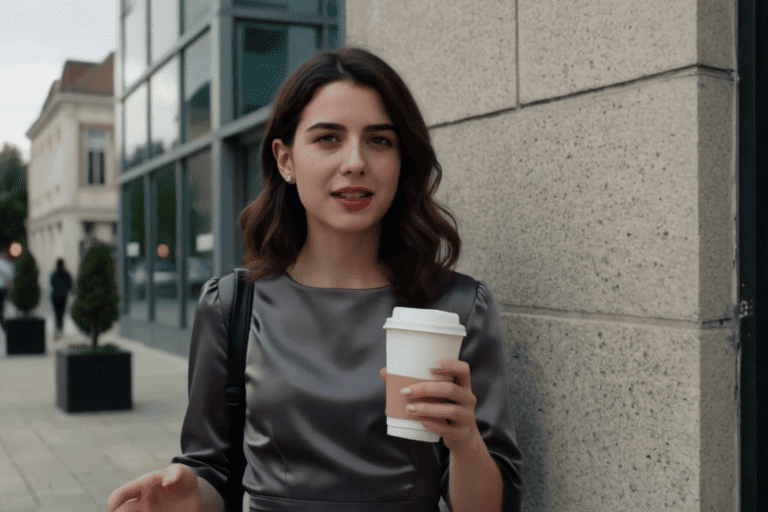} \hfill 
        \includegraphics[width=0.200\textwidth]{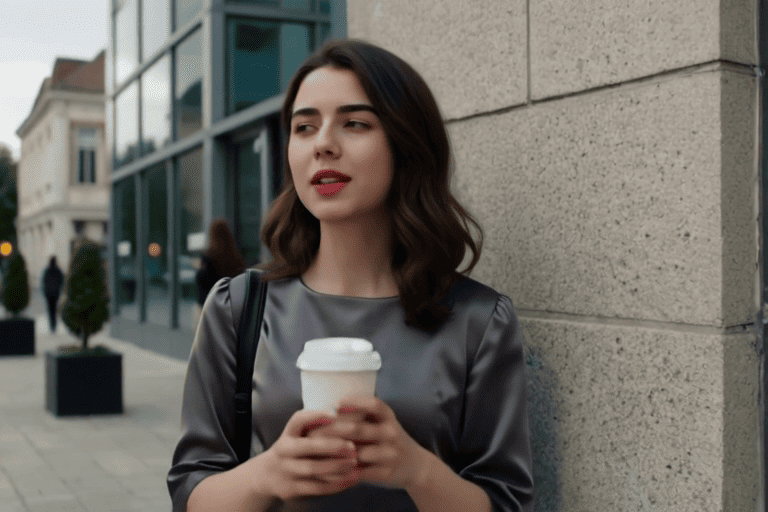} \hfill 
        \includegraphics[width=0.200\textwidth]{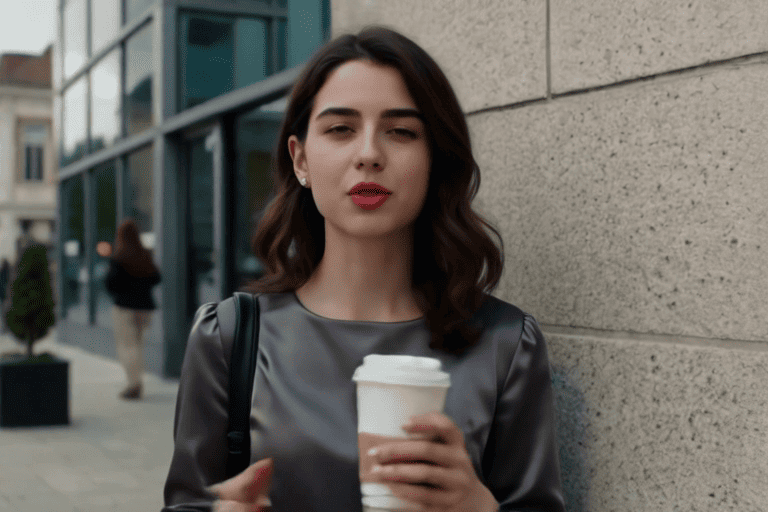} \hfill 
        \includegraphics[width=0.200\textwidth]{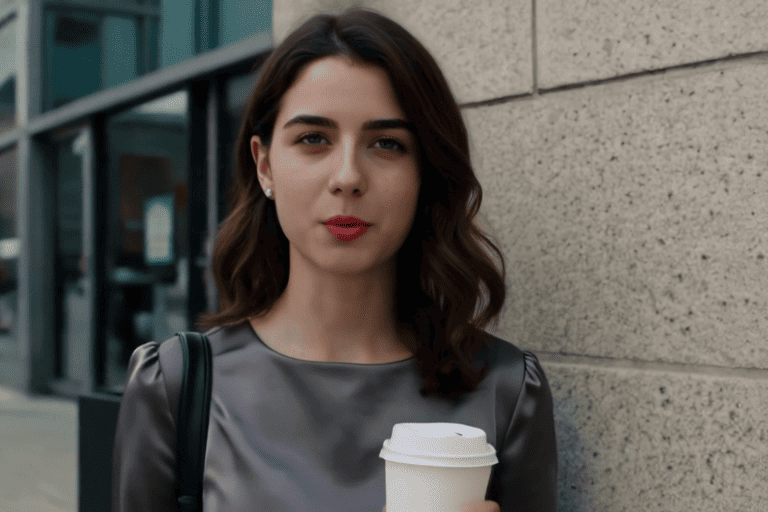} \hfill 
        \includegraphics[width=0.200\textwidth]{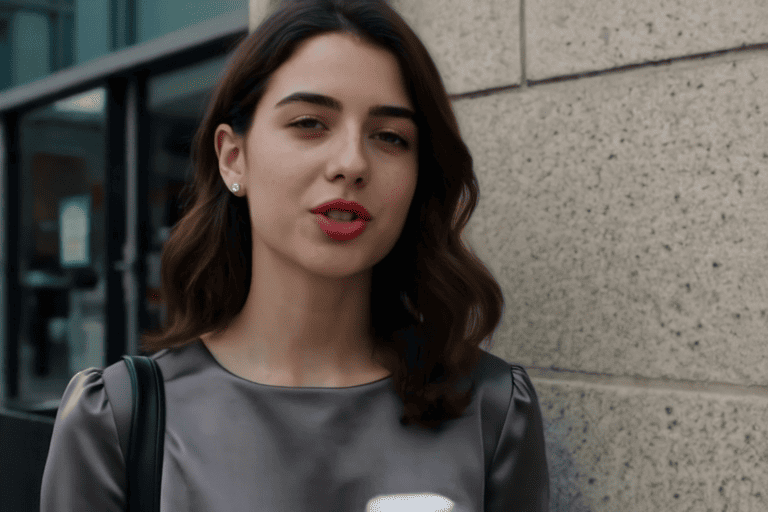}
    }
    \\
    \vspace{0.5cm}

    \subfloat[A beautiful woman with long, dark brown hair is sitting on a couch in a cozy, well-lit room. She is wearing a dark blazer over a light-colored top and has a necklace and earrings. She is holding a slice of pepperoni pizza in her hands and is in the process of taking a bite. The room has a warm, inviting atmosphere with a round table and chairs in the background, and a window letting in natural light. The woman appears to be enjoying her meal in a relaxed setting.]
    {\label{subfig:pizza}
        \includegraphics[width=0.200\textwidth]{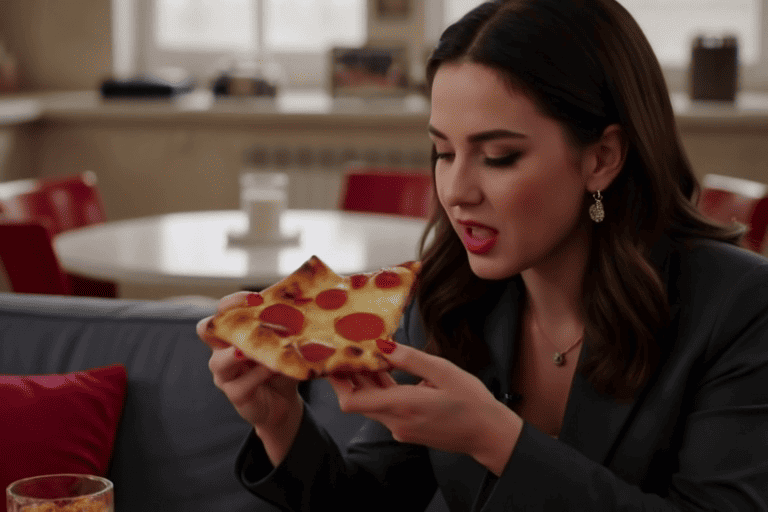} \hfill 
        \includegraphics[width=0.200\textwidth]{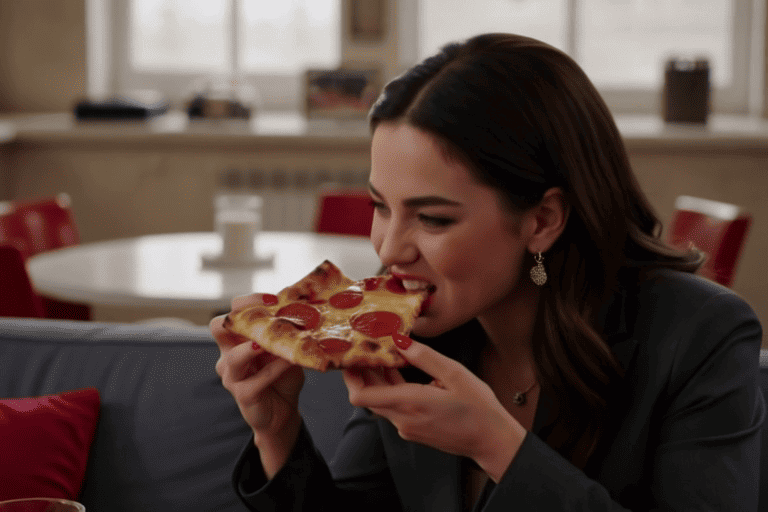} \hfill 
        \includegraphics[width=0.200\textwidth]{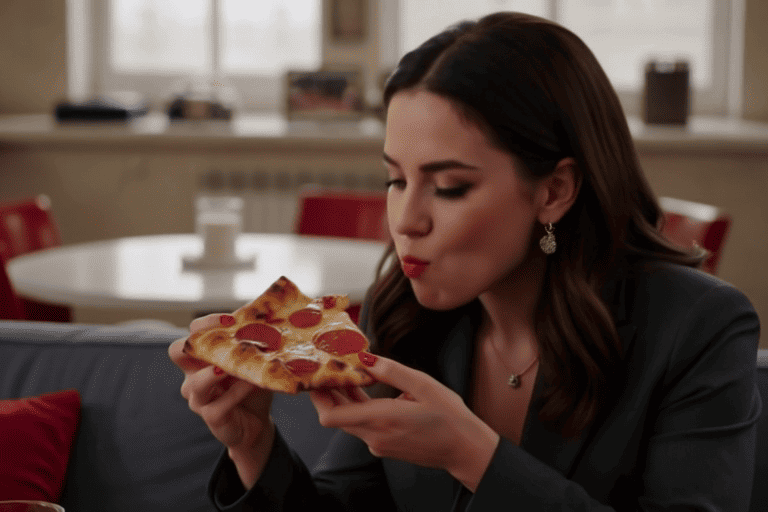} \hfill 
        \includegraphics[width=0.200\textwidth]{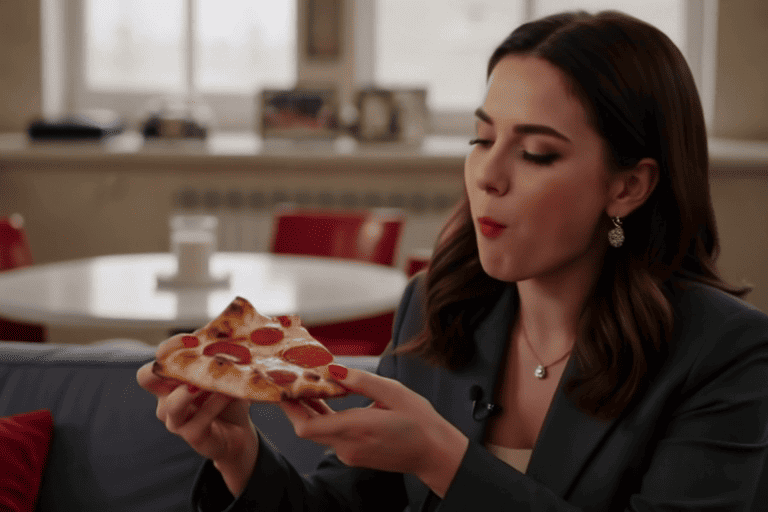} \hfill 
        \includegraphics[width=0.200\textwidth]{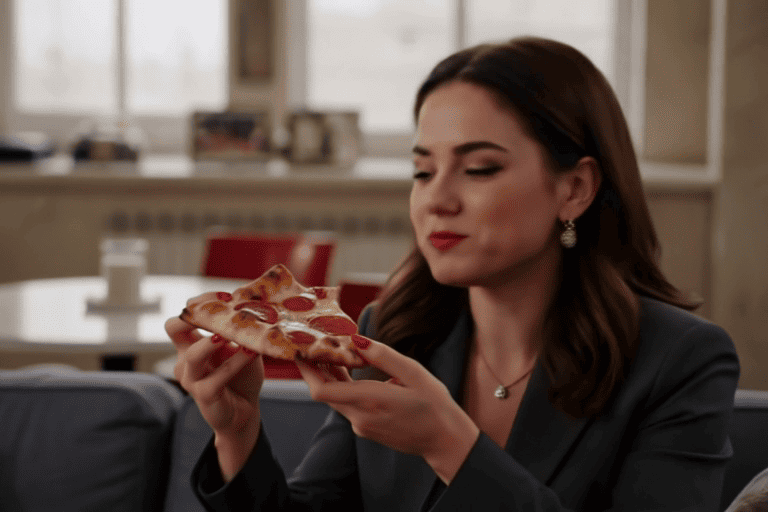}
    }
    \caption{Kandinsky 5.0 Video Lite text-to-video generation examples}
    \label{fig:lite_T2V}
\end{figure}
\begin{figure}[hbtp]
    \centering
    \captionsetup[subfloat]{labelfont=scriptsize,textfont=scriptsize}
    \subfloat[The stylish doors of an office elevator open, revealing a vast array of rubber yellow ducks that leap out directly at the camera and fill the entire space. The scene is captured from a dynamic low-angle perspective, with dramatic studio lighting creating sharp contrasts and deep shadows.]
    {\label{subfig:ducks}
        \includegraphics[width=0.200\textwidth]{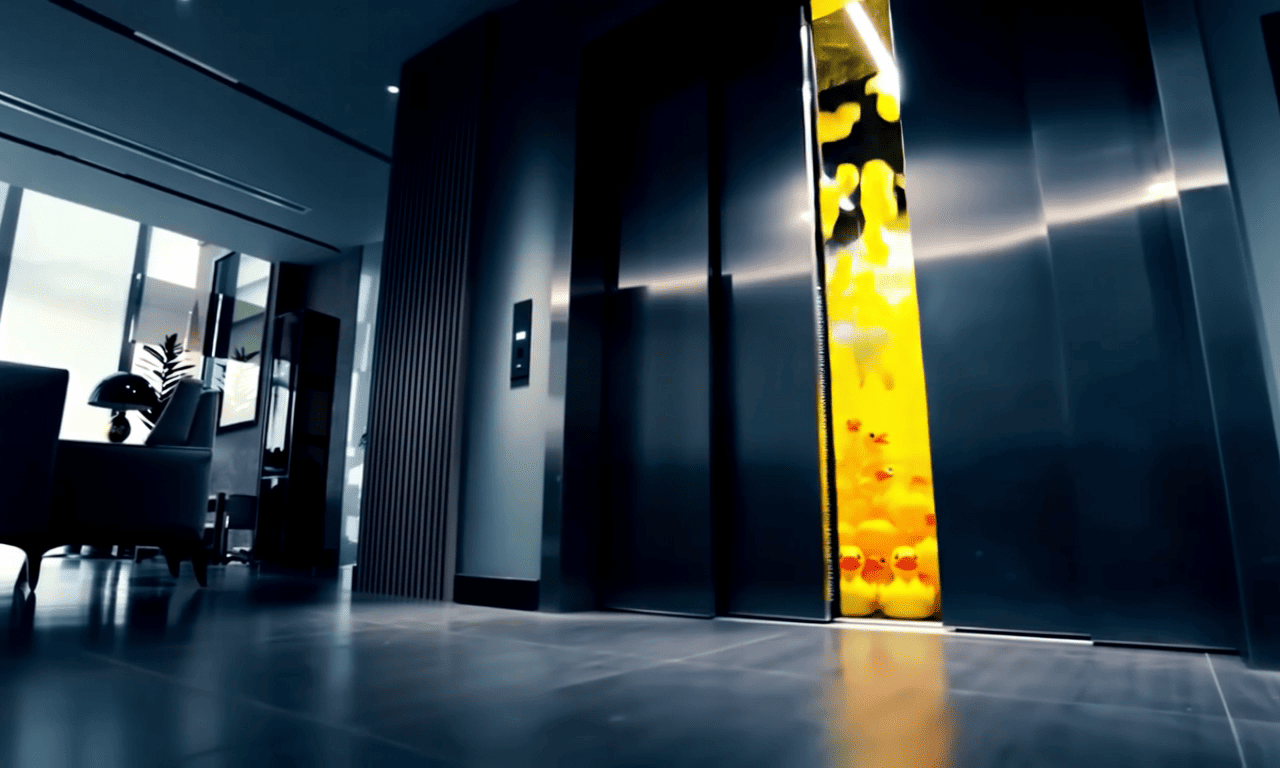} \hfill 
        \includegraphics[width=0.200\textwidth]{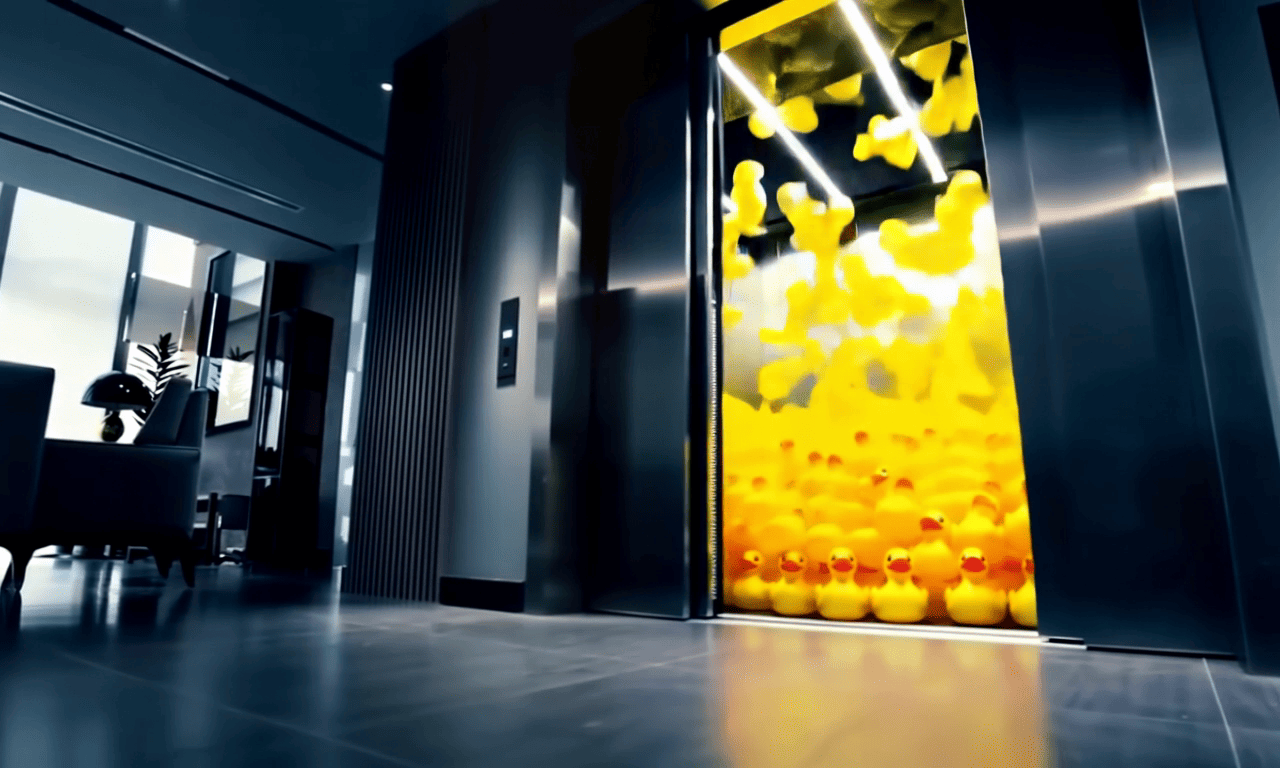} \hfill 
        \includegraphics[width=0.200\textwidth]{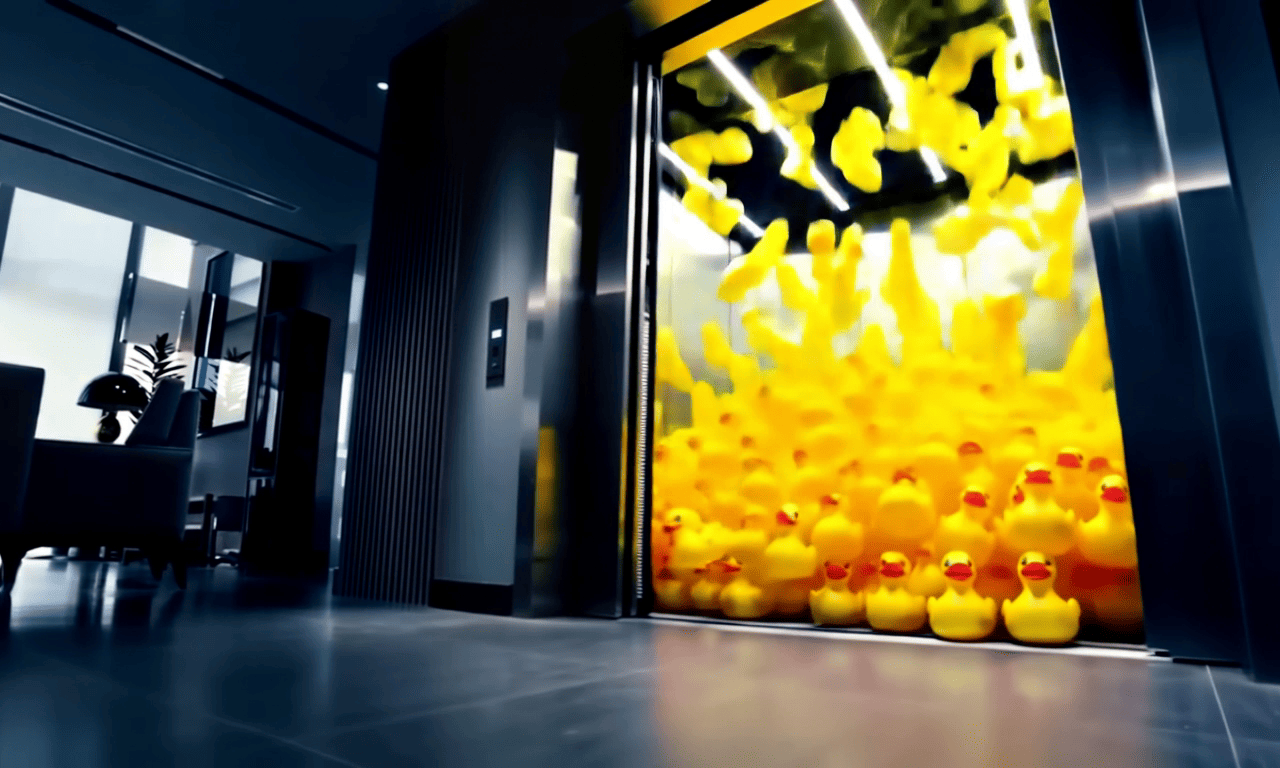} \hfill 
        \includegraphics[width=0.200\textwidth]{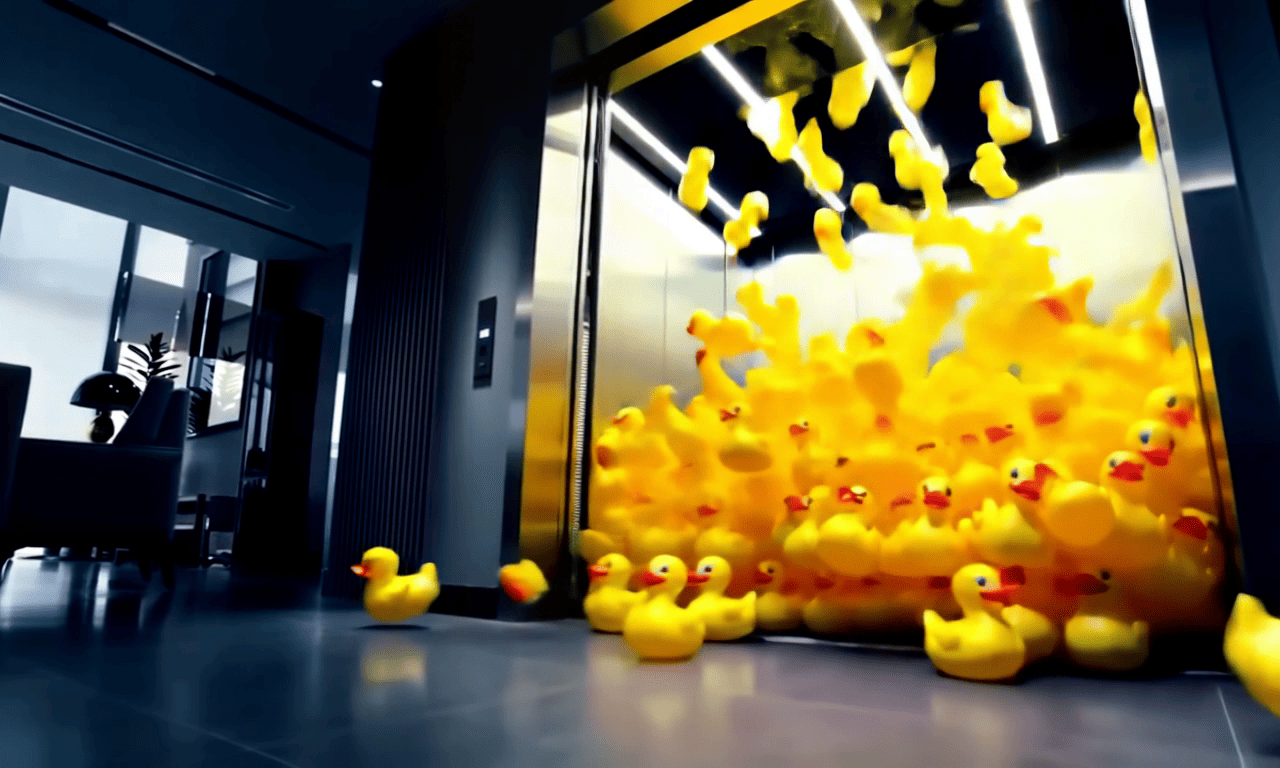} \hfill 
        \includegraphics[width=0.200\textwidth]{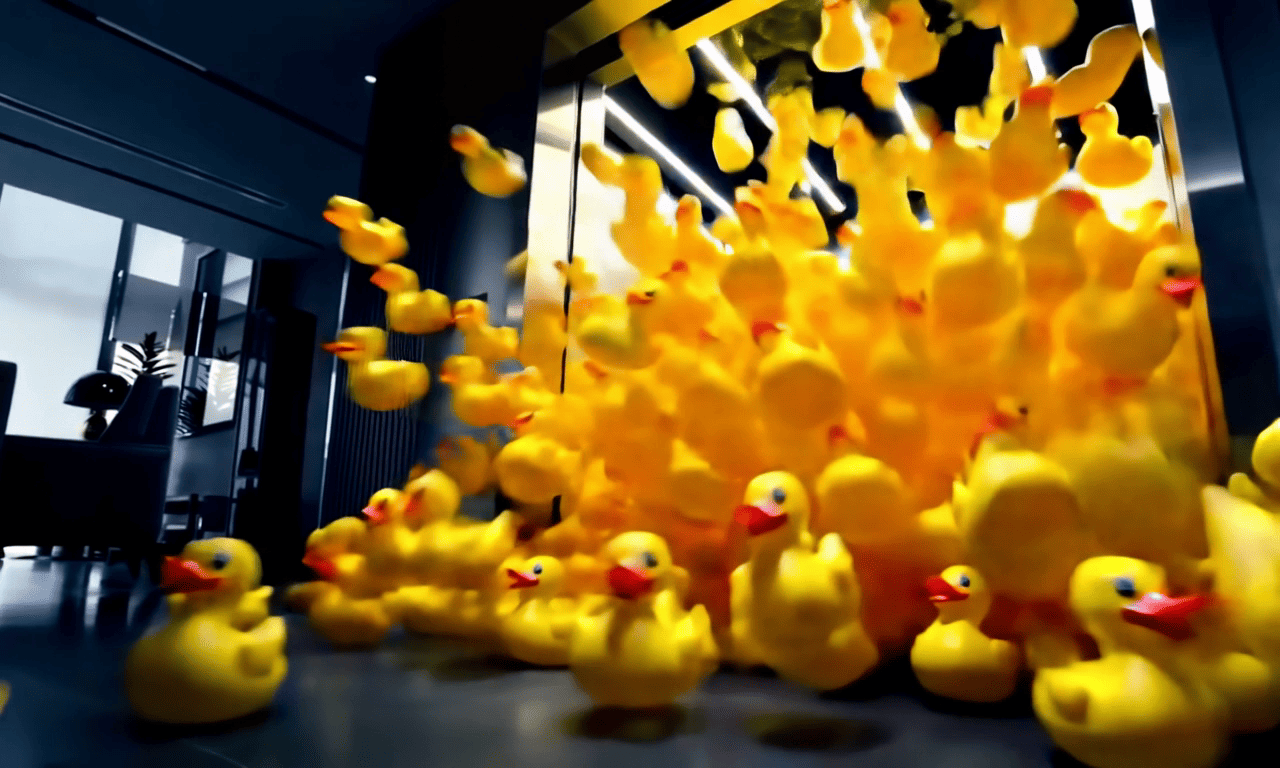}
    }
    \\
    \vspace{0.5cm}

    \subfloat[A single golden kernel of popcorn lies on a dark, smooth surface. In slow motion, the glowing cloud changes shape—rounding into a plump body, the fluff gradually transforming into fluffy yellow feathers. A tiny beak slowly emerges from the haze, followed by two round, dark eyes. Small wings slowly unfurl at the sides, and slender legs emerge. Puffs of steam rise upward like gentle breath, completing the transformation. Ultra-realistic 8K, a magical yet calming cinematic atmosphere.]
    {\label{subfig:popcorn}
        \includegraphics[width=0.200\textwidth]{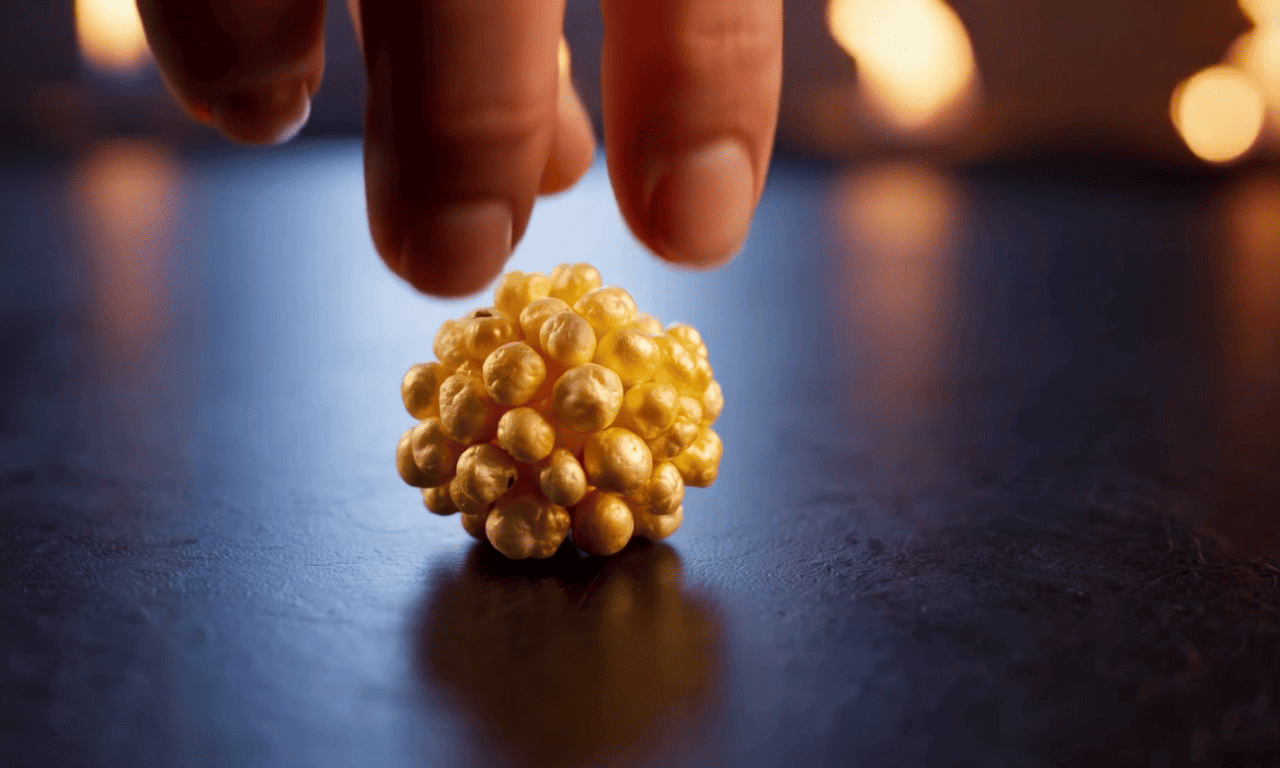} \hfill 
        \includegraphics[width=0.200\textwidth]{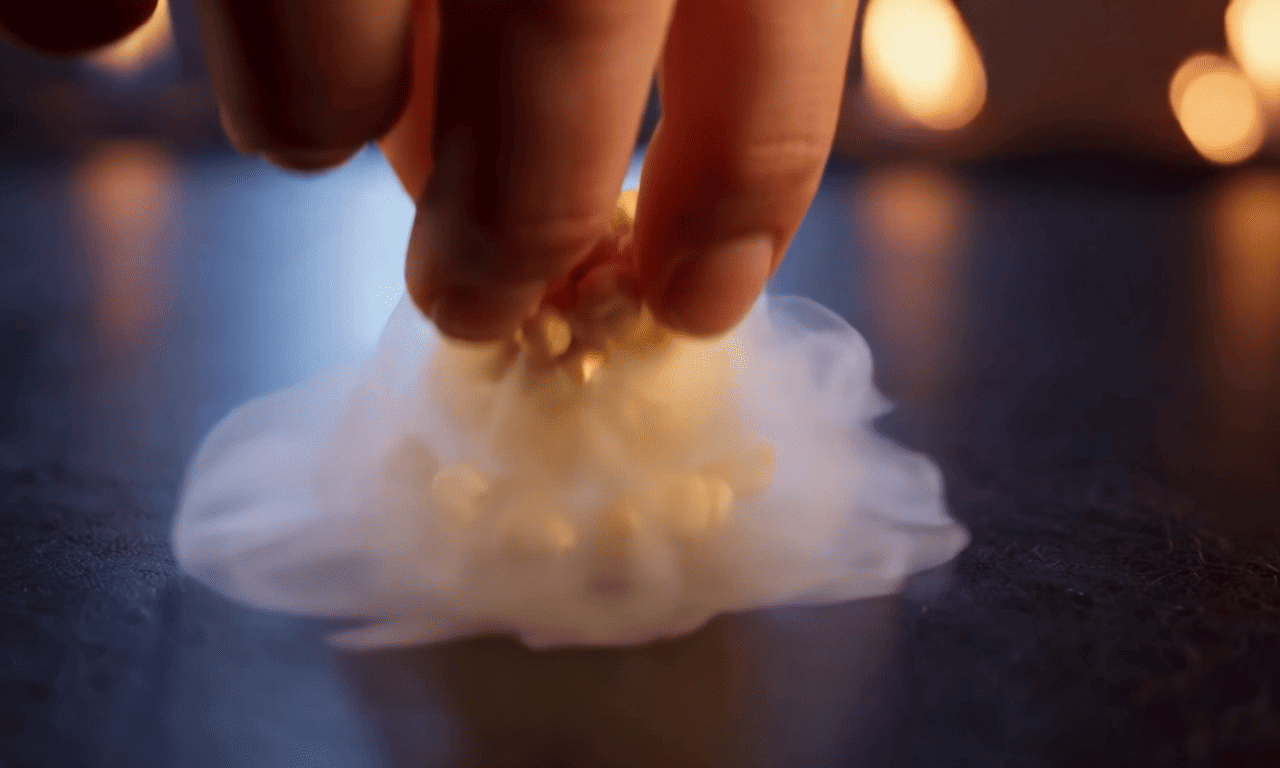} \hfill 
        \includegraphics[width=0.200\textwidth]{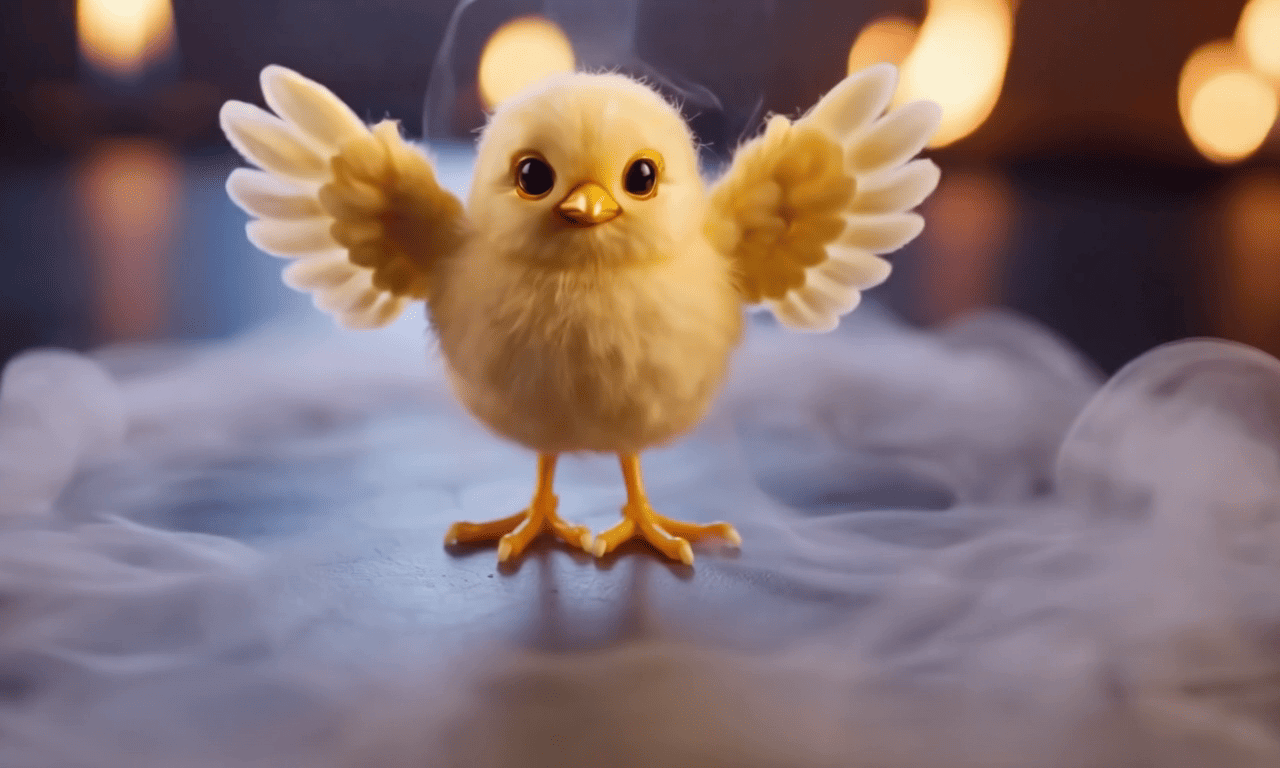} \hfill 
        \includegraphics[width=0.200\textwidth]{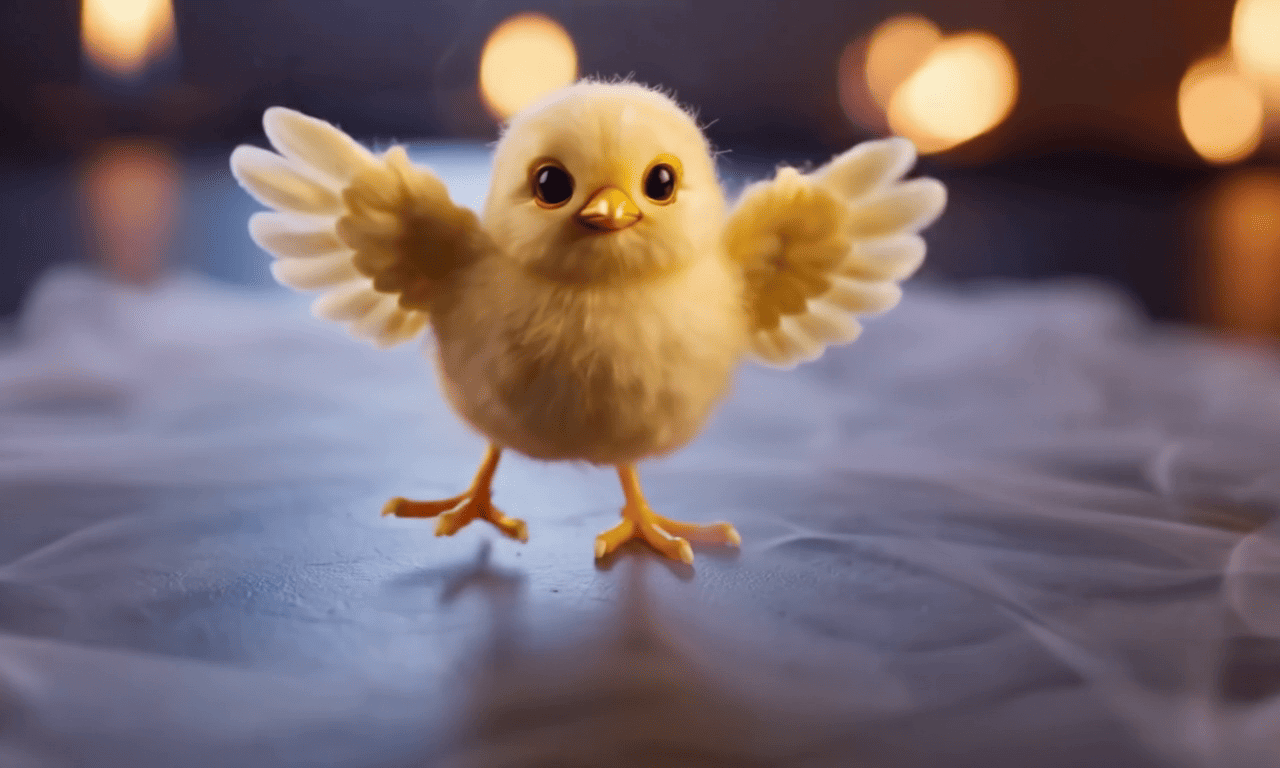} \hfill 
        \includegraphics[width=0.200\textwidth]{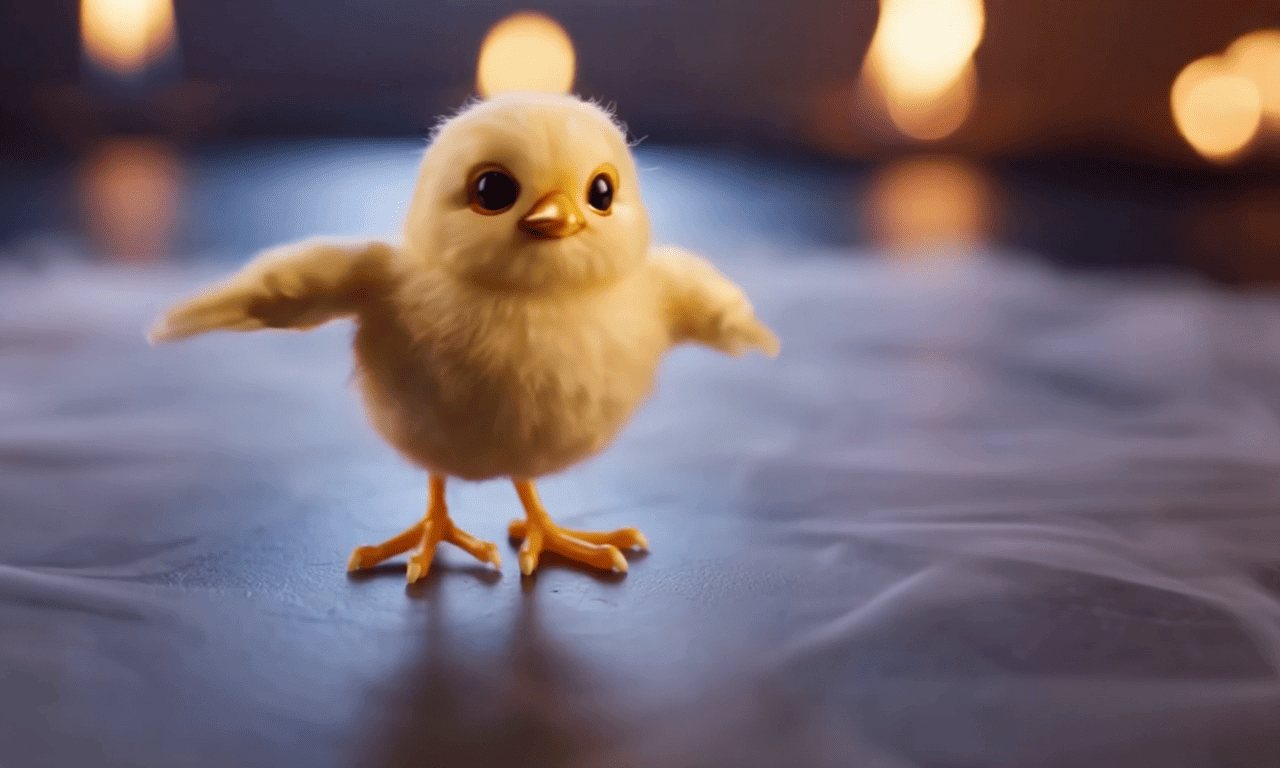}
    }
    \\
    \vspace{0.5cm}

    \subfloat[Astronaut snowboarding on planet surface, deep space, dynamic pose, snowboard jump, Saturn background, 8k]
    {\label{subfig:snowboard}
        \includegraphics[width=0.200\textwidth]{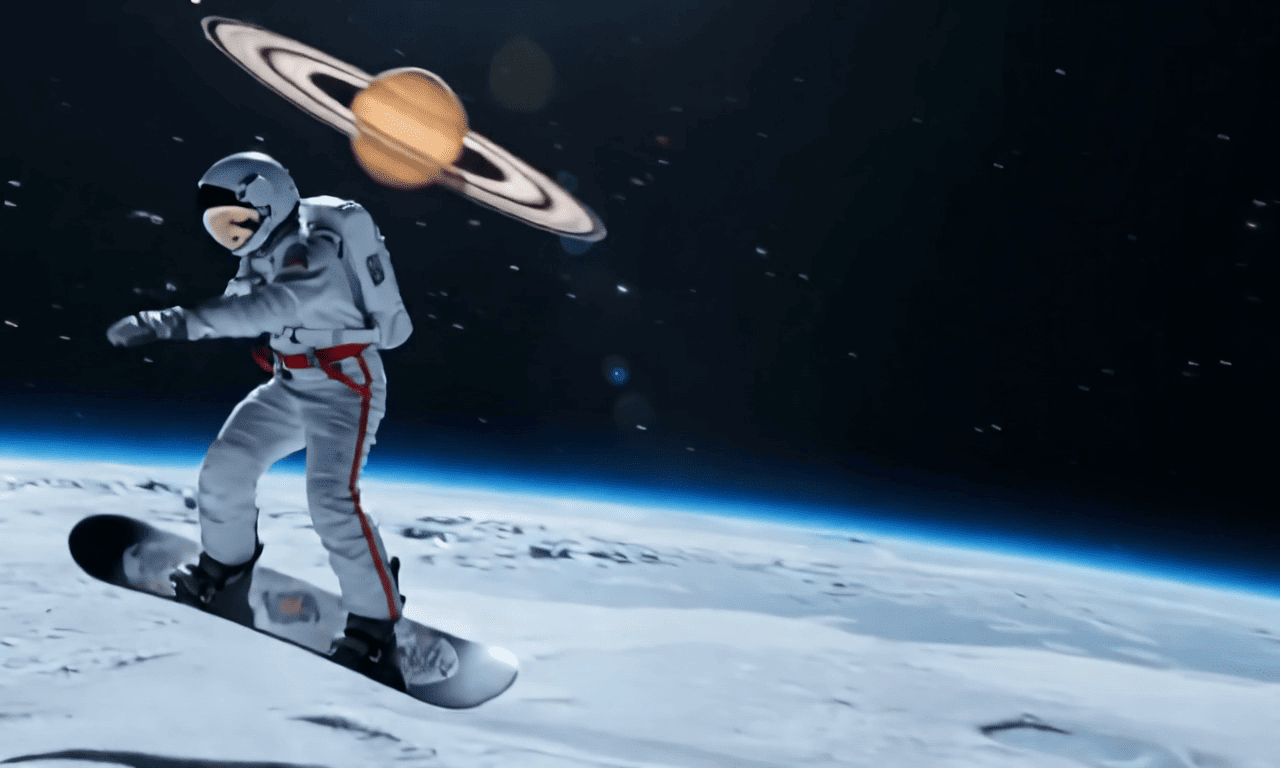} \hfill 
        \includegraphics[width=0.200\textwidth]{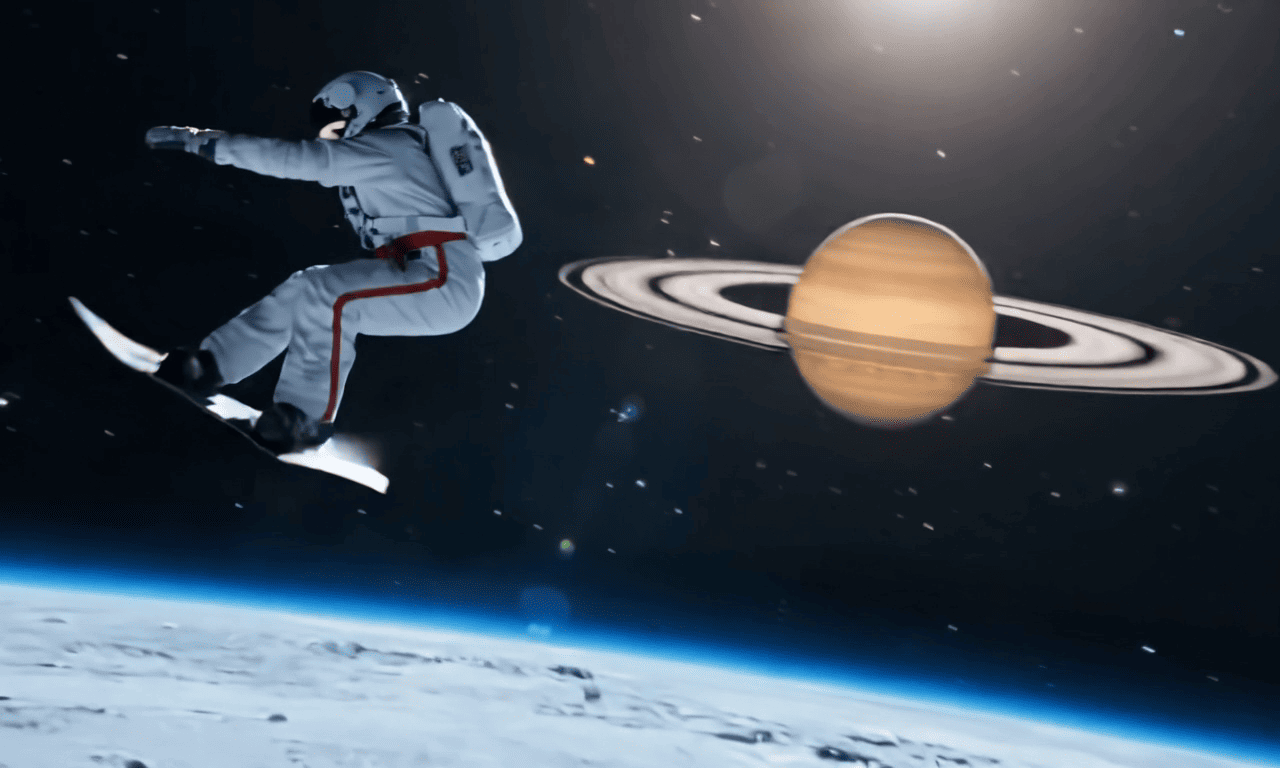} \hfill 
        \includegraphics[width=0.200\textwidth]{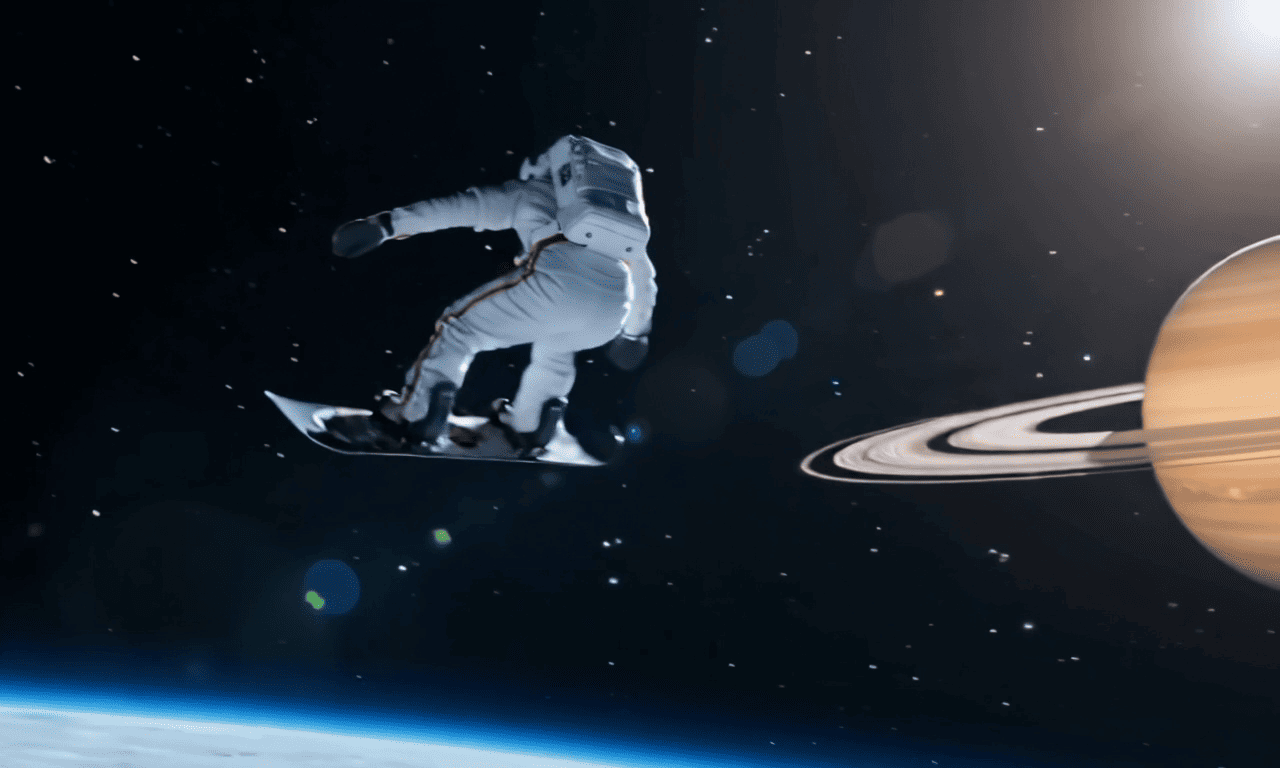} \hfill 
        \includegraphics[width=0.200\textwidth]{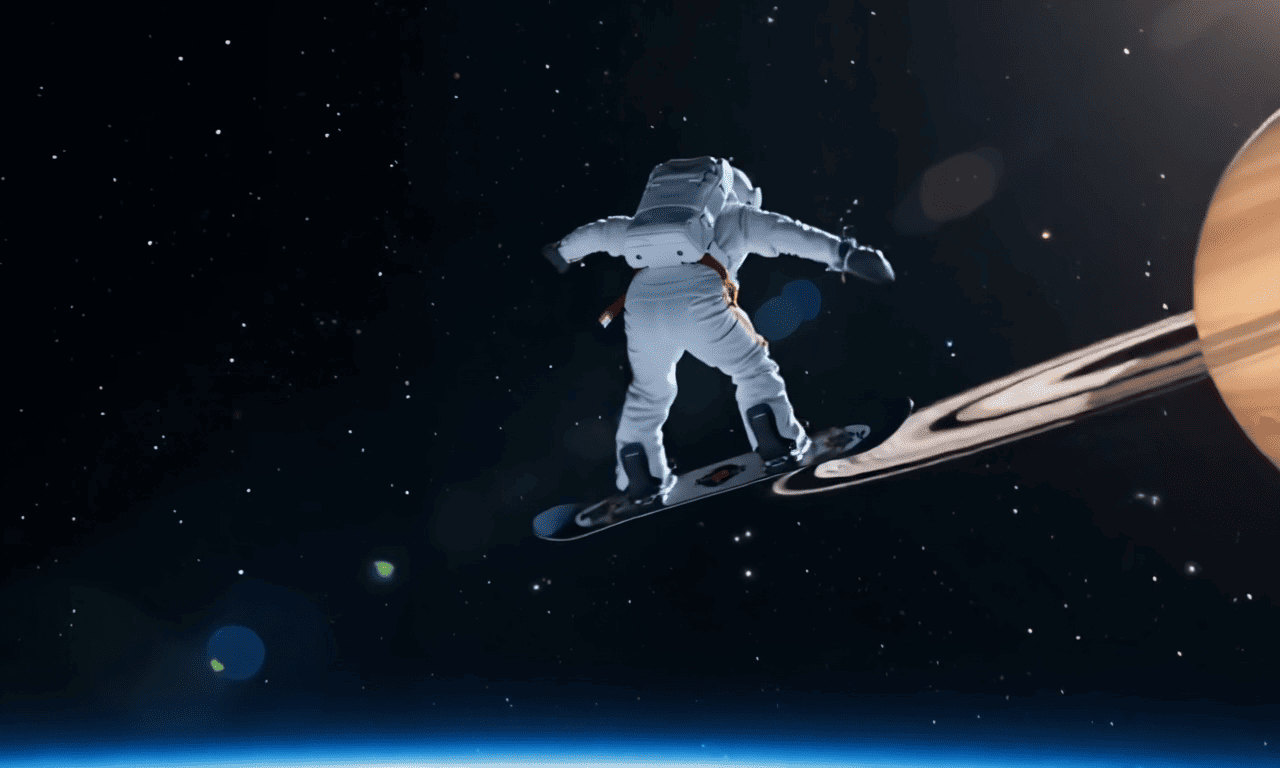} \hfill 
        \includegraphics[width=0.200\textwidth]{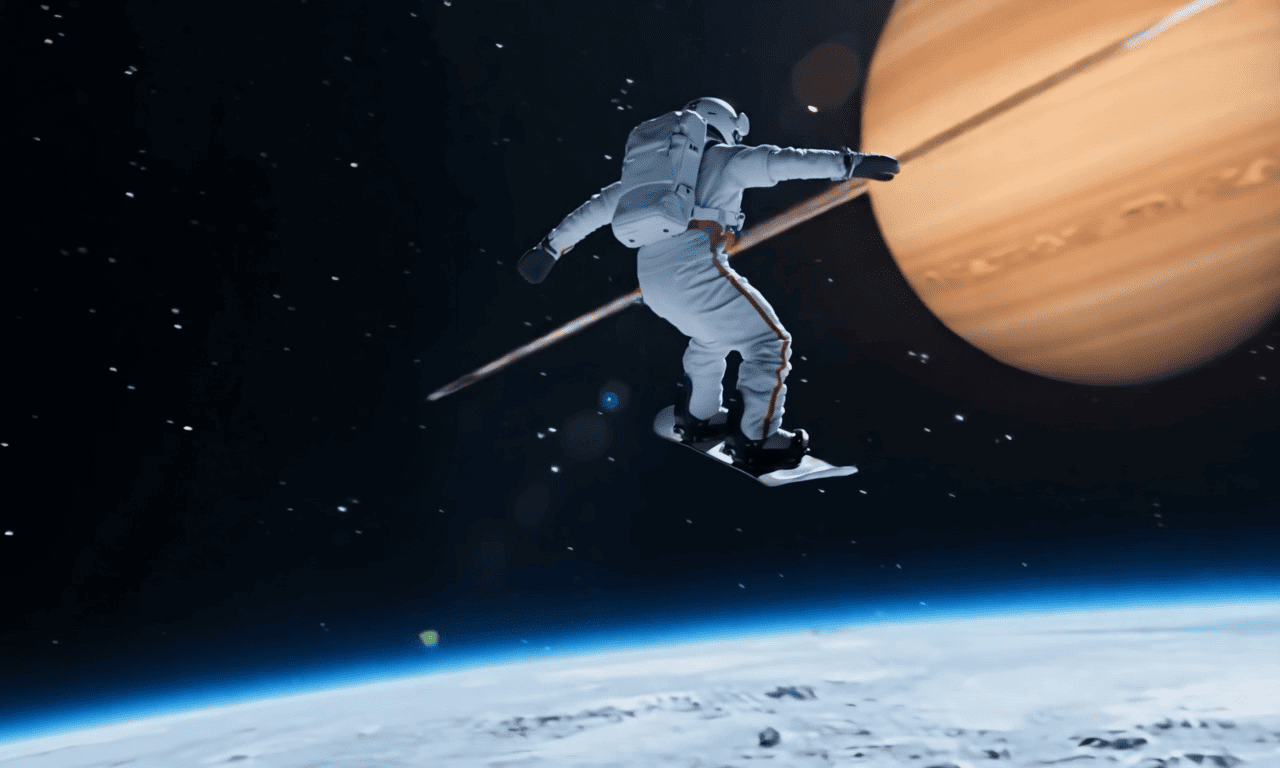}
    }
    \\
    \vspace{0.5cm}

    \subfloat[Close-up. Style: TV commercial. A woman in her thirties takes her first sip of coffee, sitting on a small balcony overlooking a quiet city street. She's wrapped in a soft sweater, and the morning light is cool—light steam rises from the mug. Her eyes close for a moment—no theatrics, just real.]
    {\label{subfig:woman}
        \includegraphics[width=0.200\textwidth]{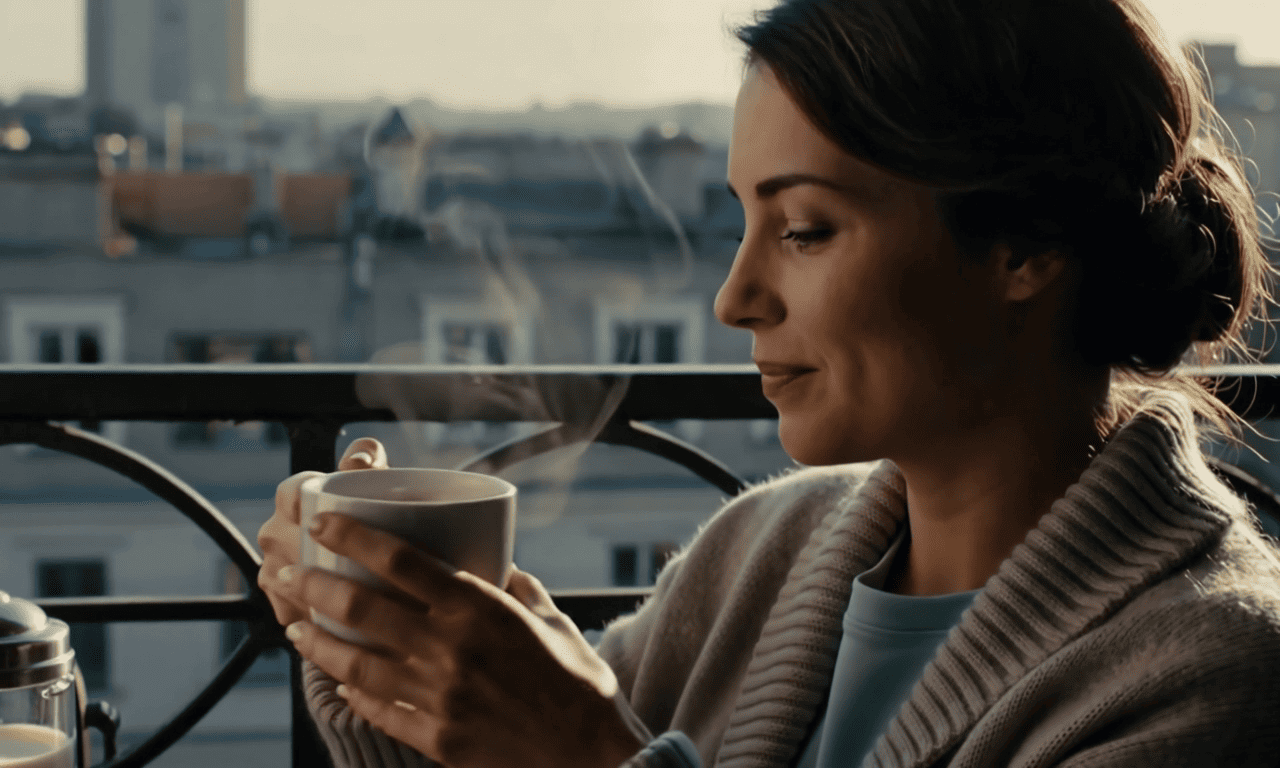} \hfill 
        \includegraphics[width=0.200\textwidth]{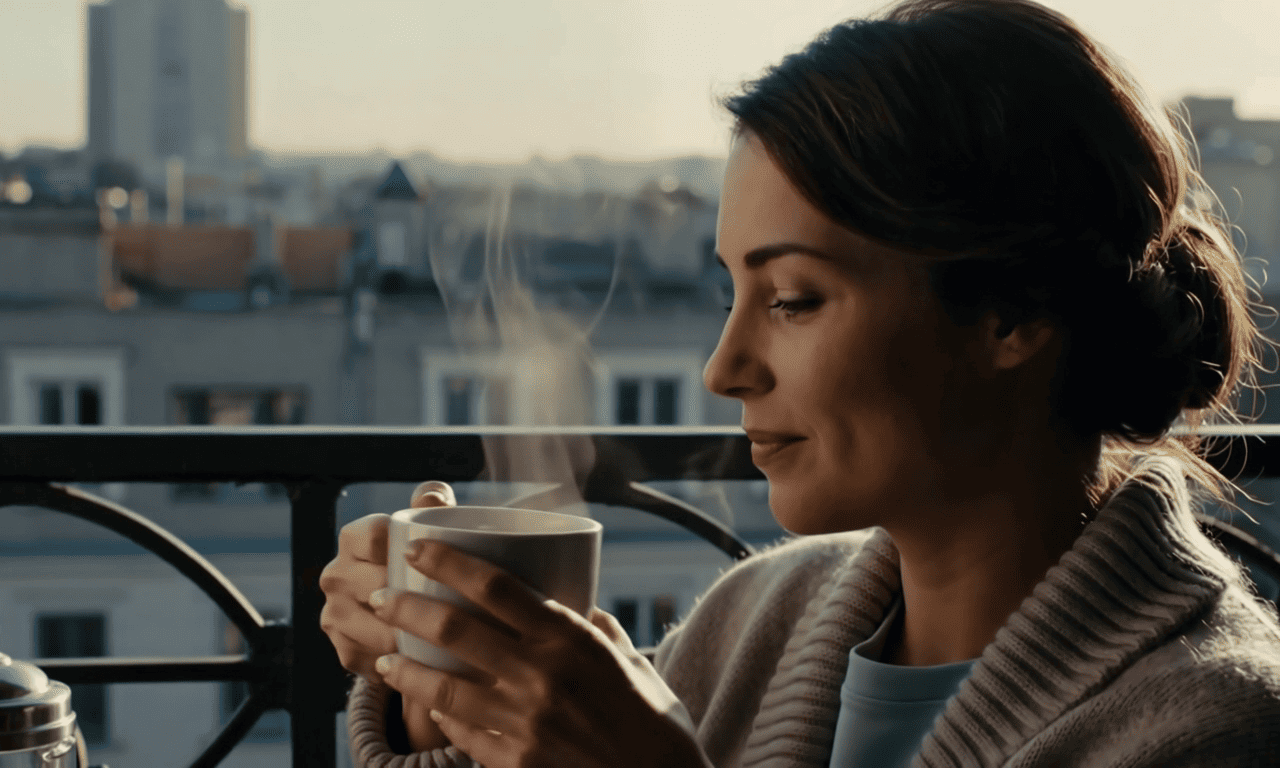} \hfill 
        \includegraphics[width=0.200\textwidth]{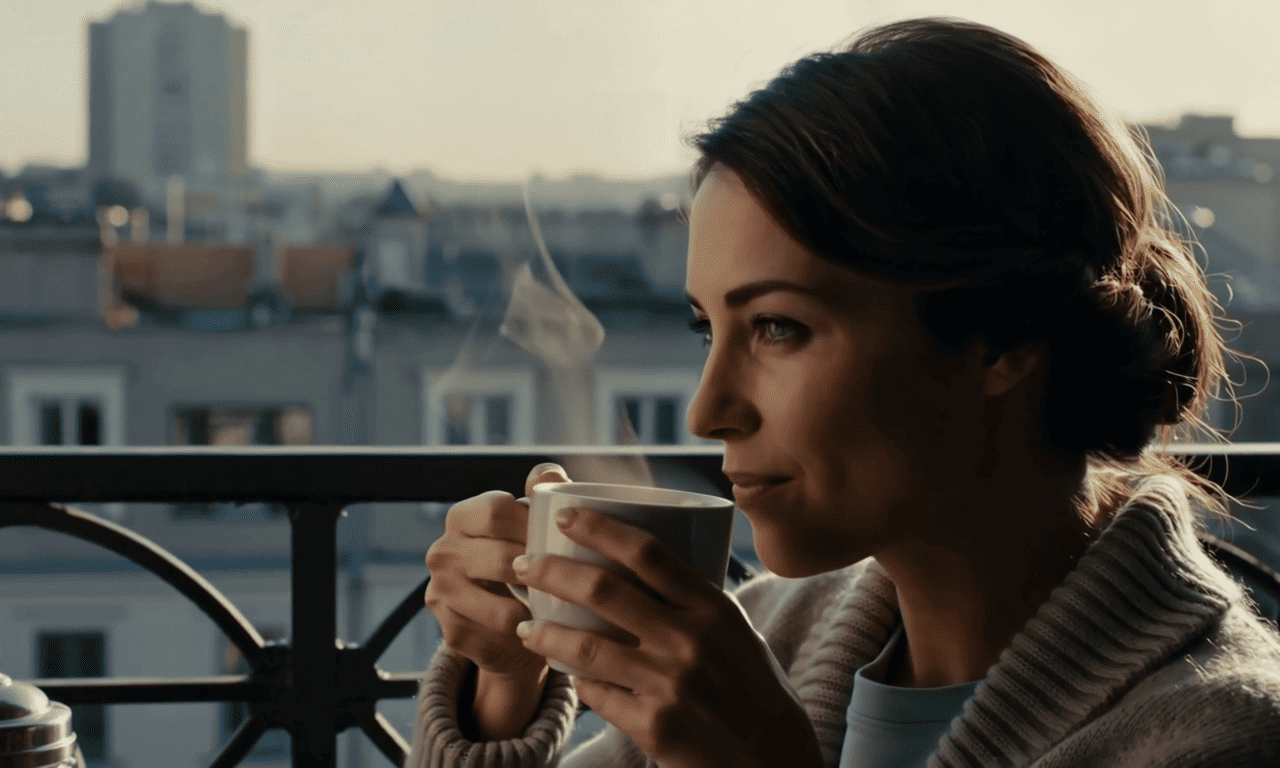} \hfill 
        \includegraphics[width=0.200\textwidth]{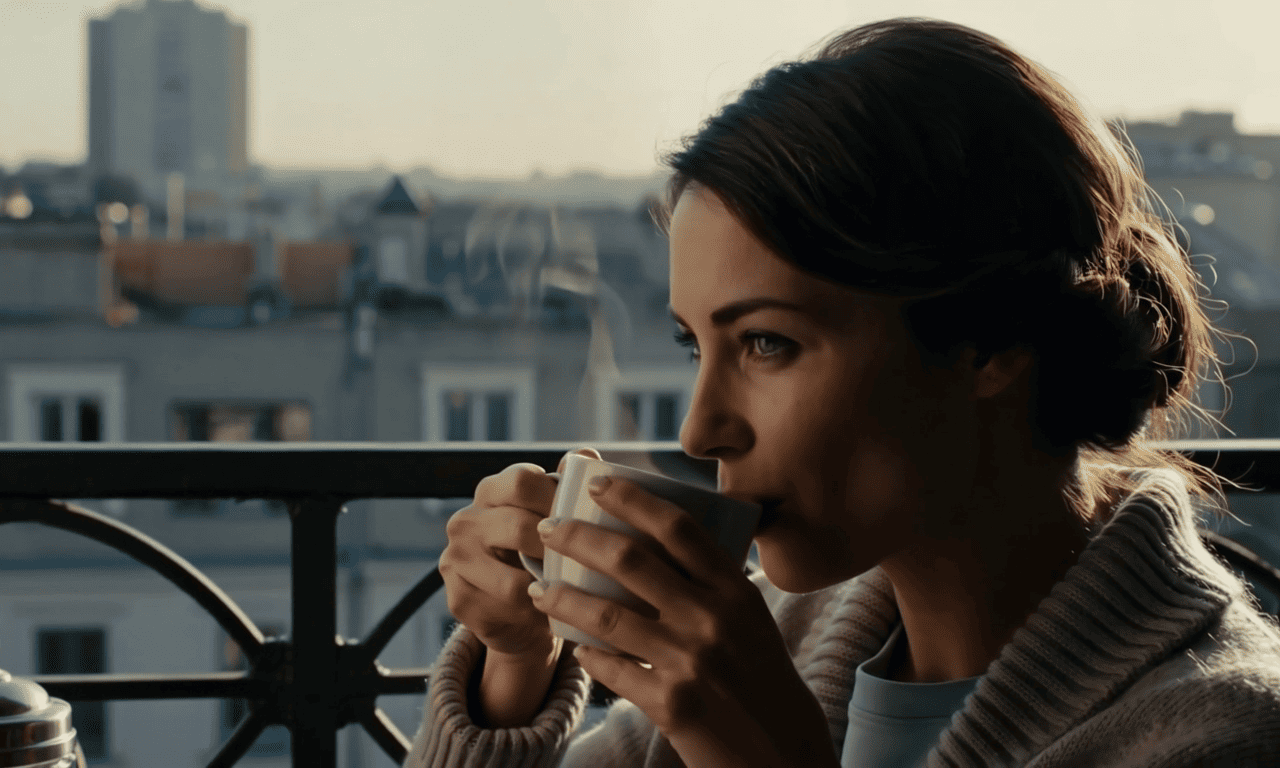} \hfill 
        \includegraphics[width=0.200\textwidth]{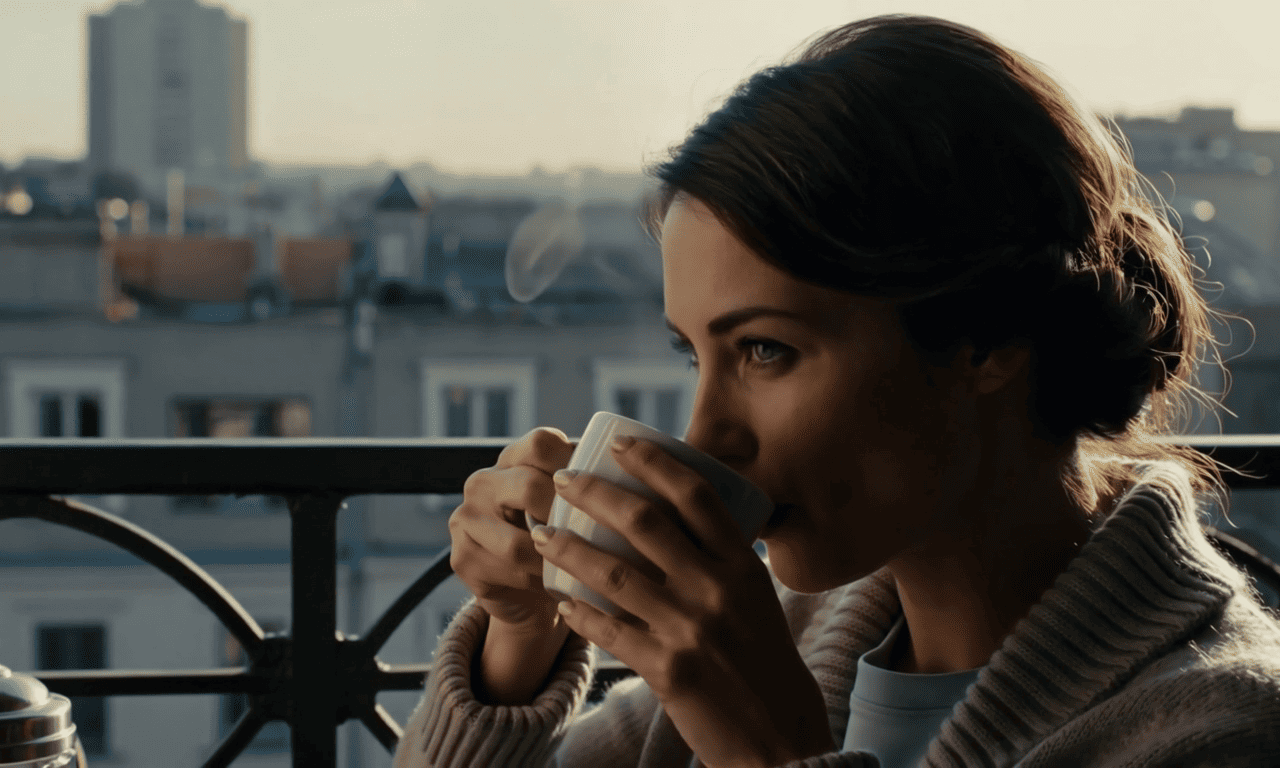}
    }
    \\
    \vspace{0.5cm}

    \subfloat[A fantastic pumpkin rolls through a dark forest with bright lighting, fast-paced dynamics, the pumpkin bounces on bumps and trips over a stone, explodes into pieces and transforms into an elegant carriage, the camera drops sharply to show the pumpkin's movements in close-up, the scene is filled with speed, fantastic transformation and unusual fairytale quality.]
    {\label{subfig:pumpkin}
        \includegraphics[width=0.200\textwidth]{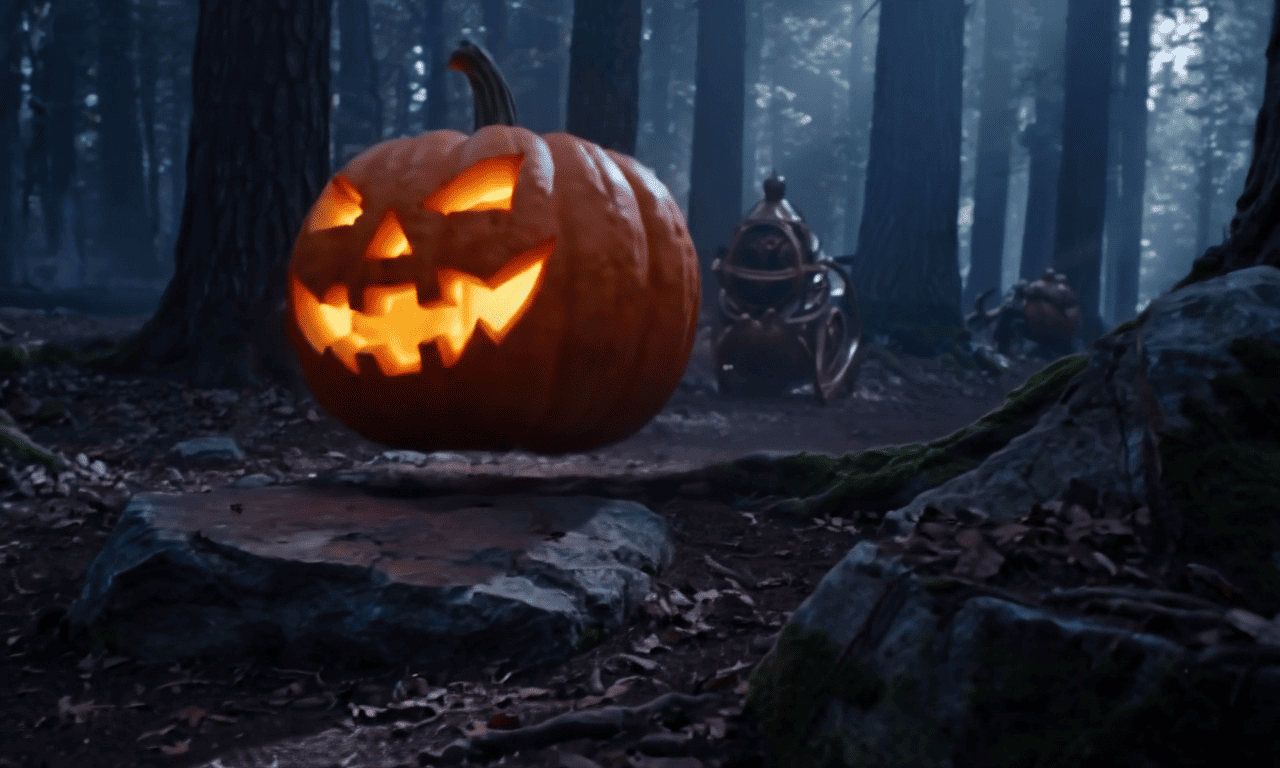} \hfill 
        \includegraphics[width=0.200\textwidth]{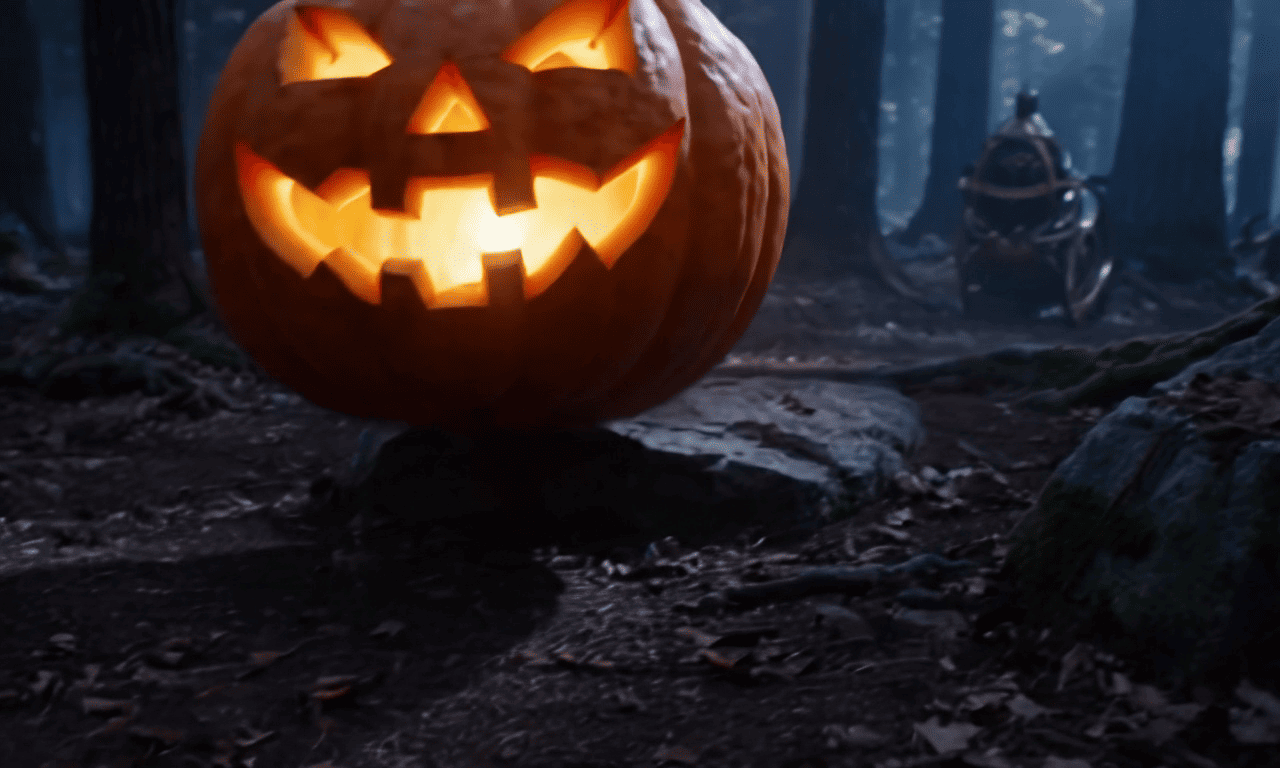} \hfill 
        \includegraphics[width=0.200\textwidth]{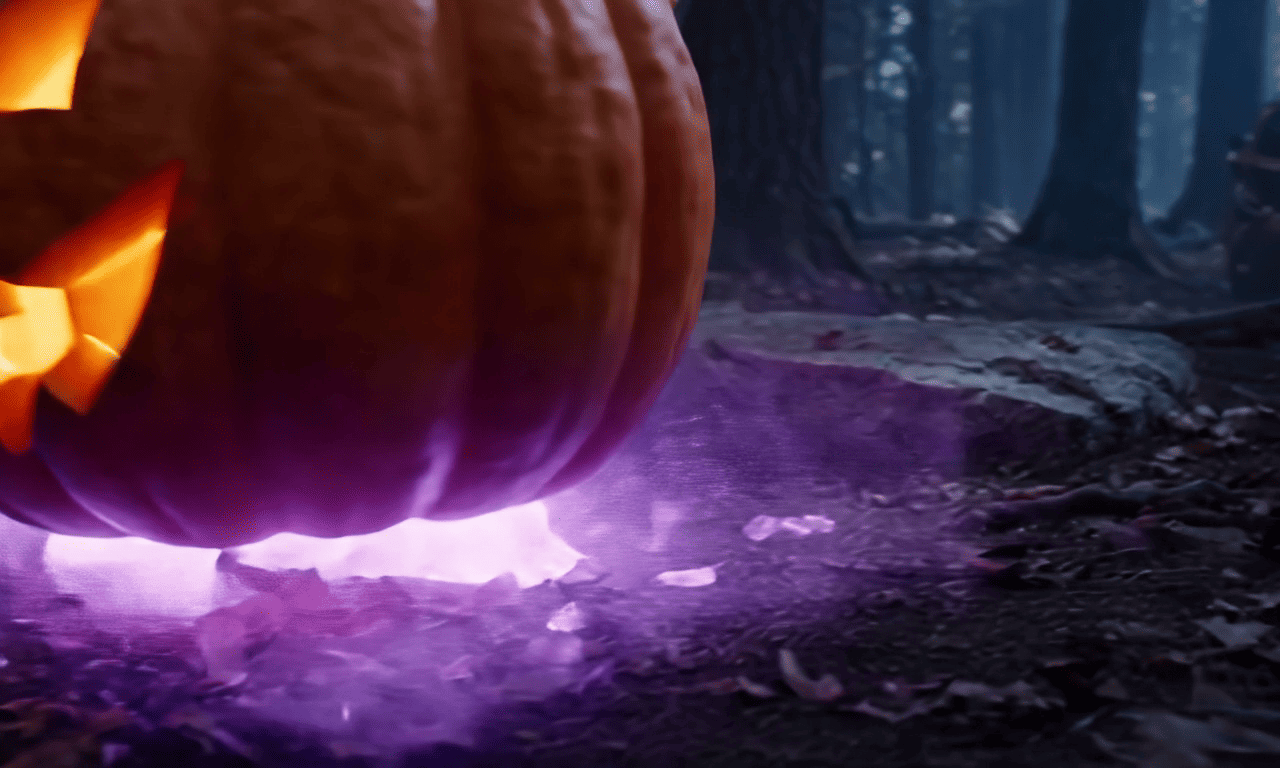} \hfill 
        \includegraphics[width=0.200\textwidth]{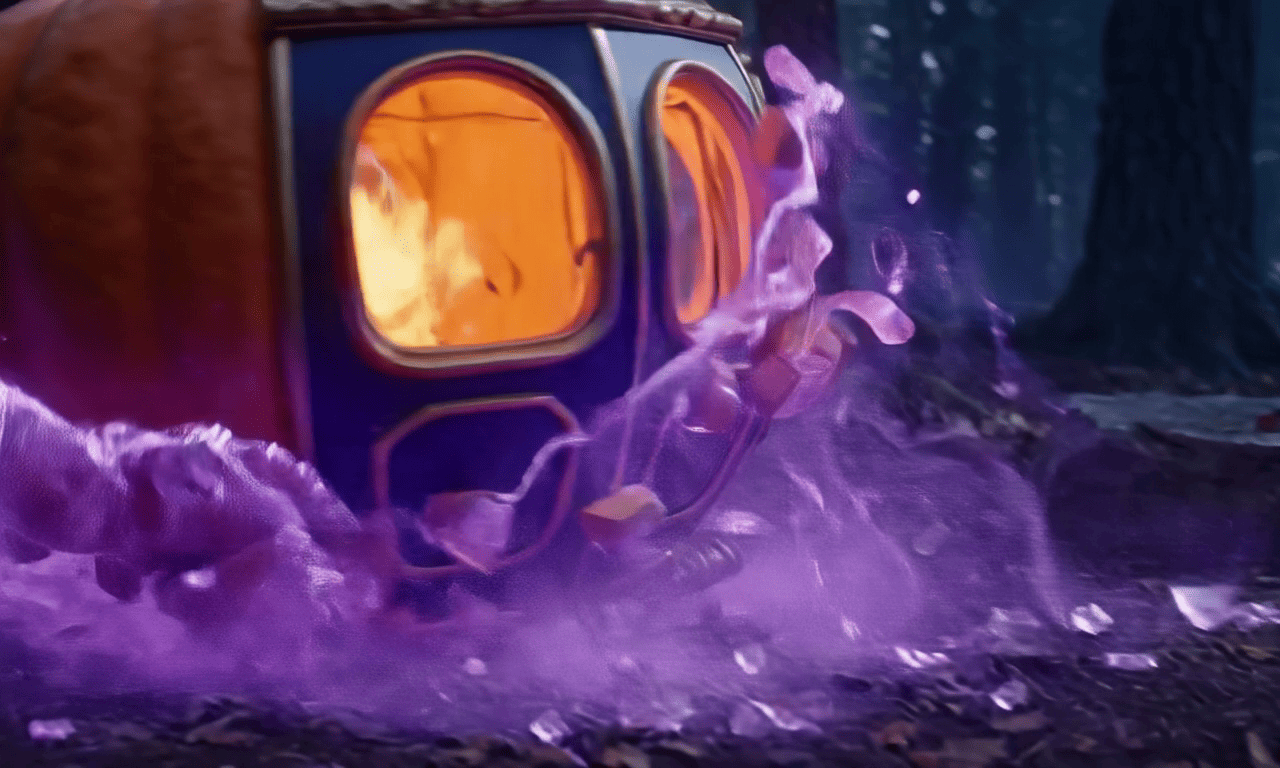} \hfill 
        \includegraphics[width=0.200\textwidth]{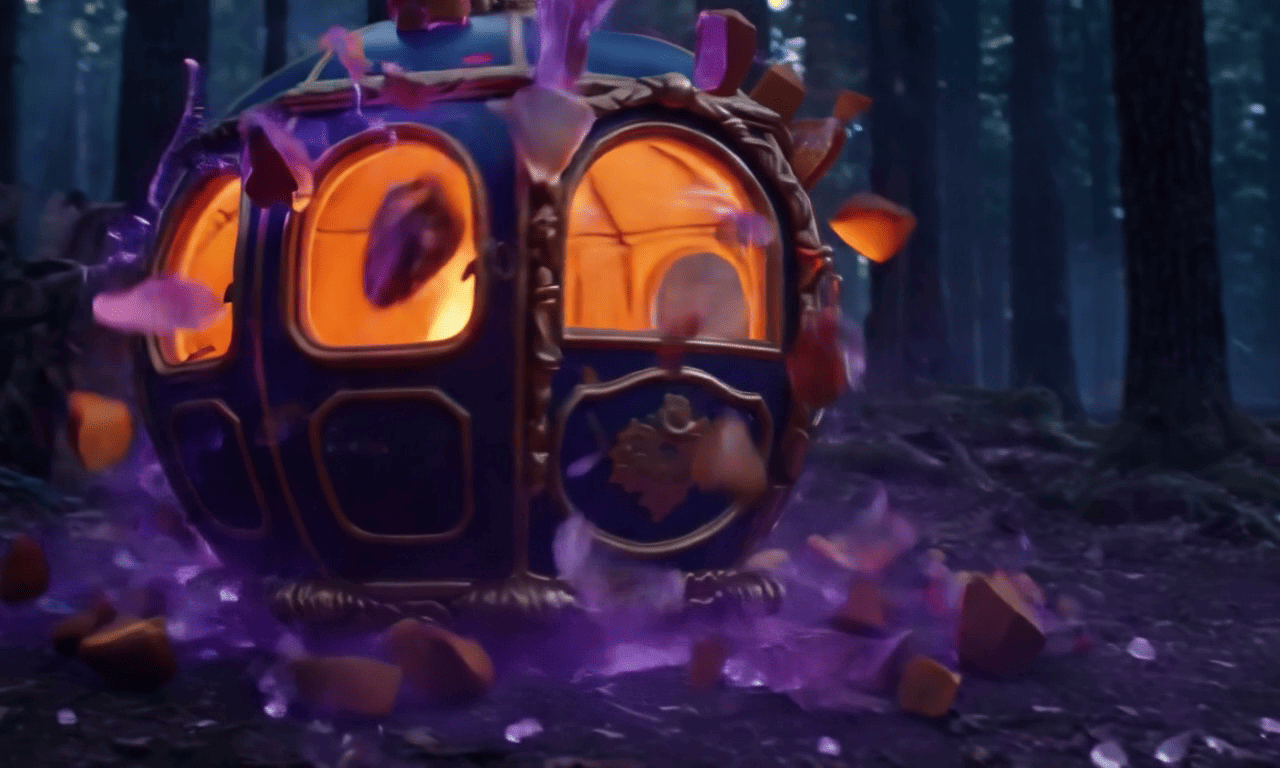}
    }
    \\
    \vspace{0.5cm}

    \subfloat[Large waves and splashes of water hitting the hull of a partially sunken 16th-century battleship against a stormy backdrop. The camera zooms in from a drone's perspective. High detail.]
    {\label{subfig:battleship}
        \includegraphics[width=0.200\textwidth]{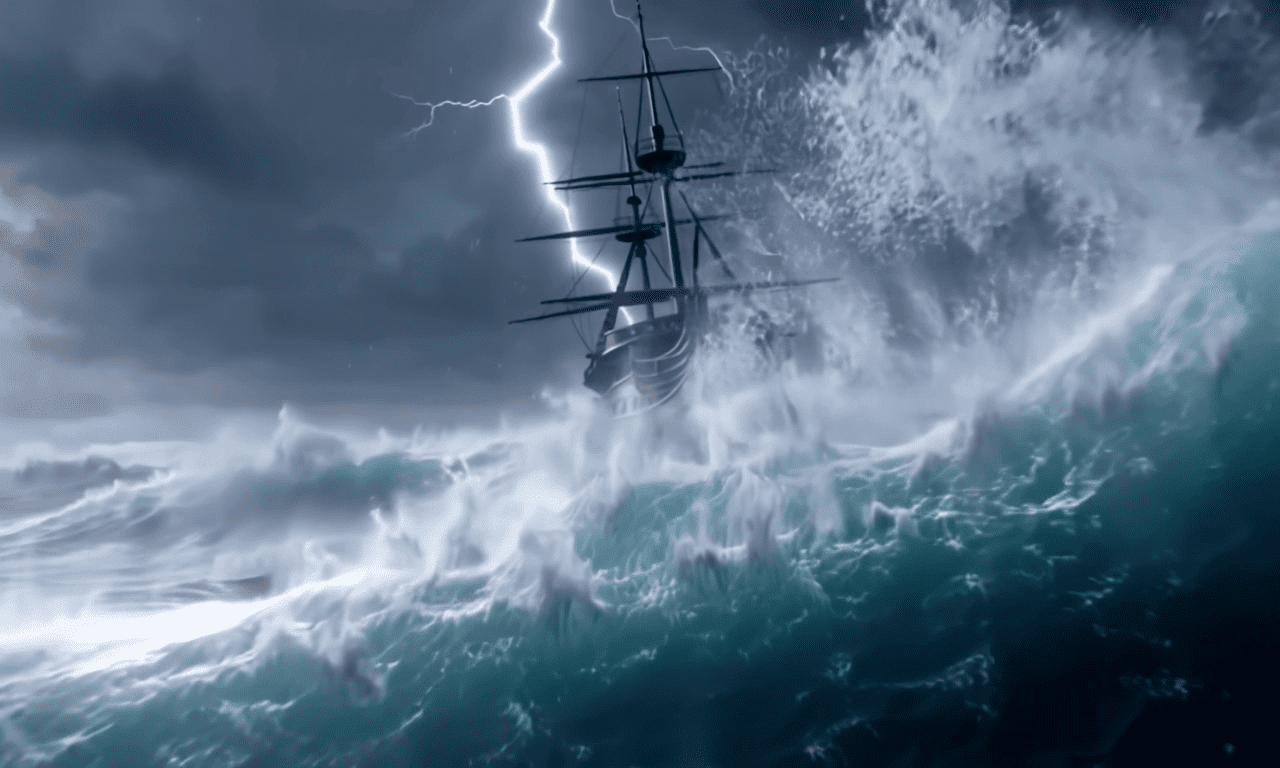} \hfill 
        \includegraphics[width=0.200\textwidth]{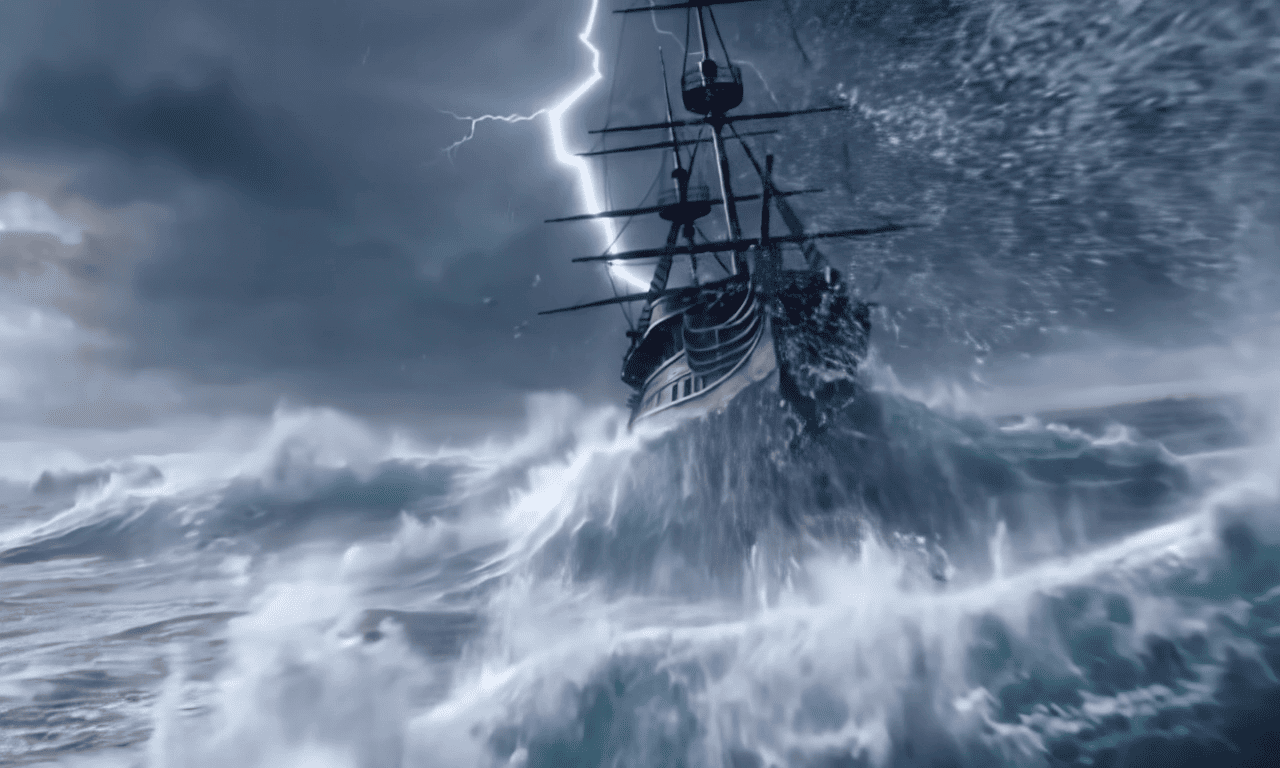} \hfill 
        \includegraphics[width=0.200\textwidth]{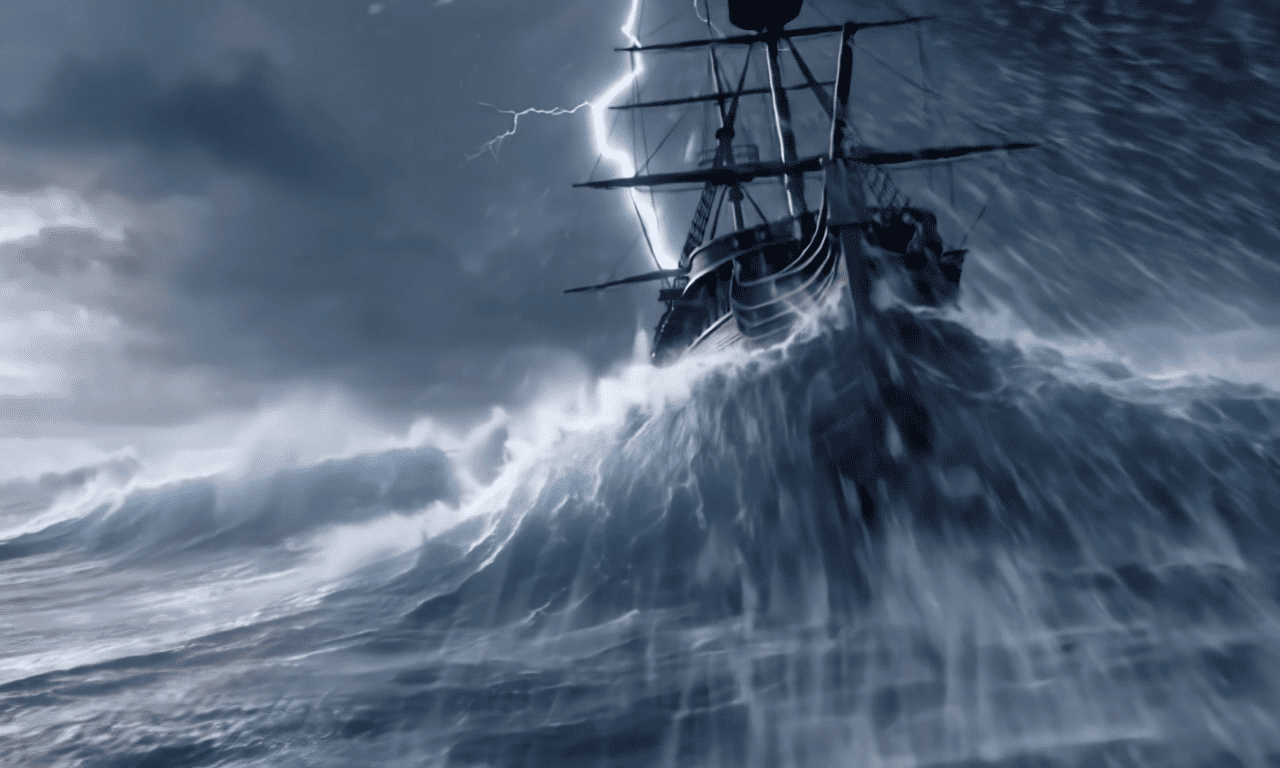} \hfill 
        \includegraphics[width=0.200\textwidth]{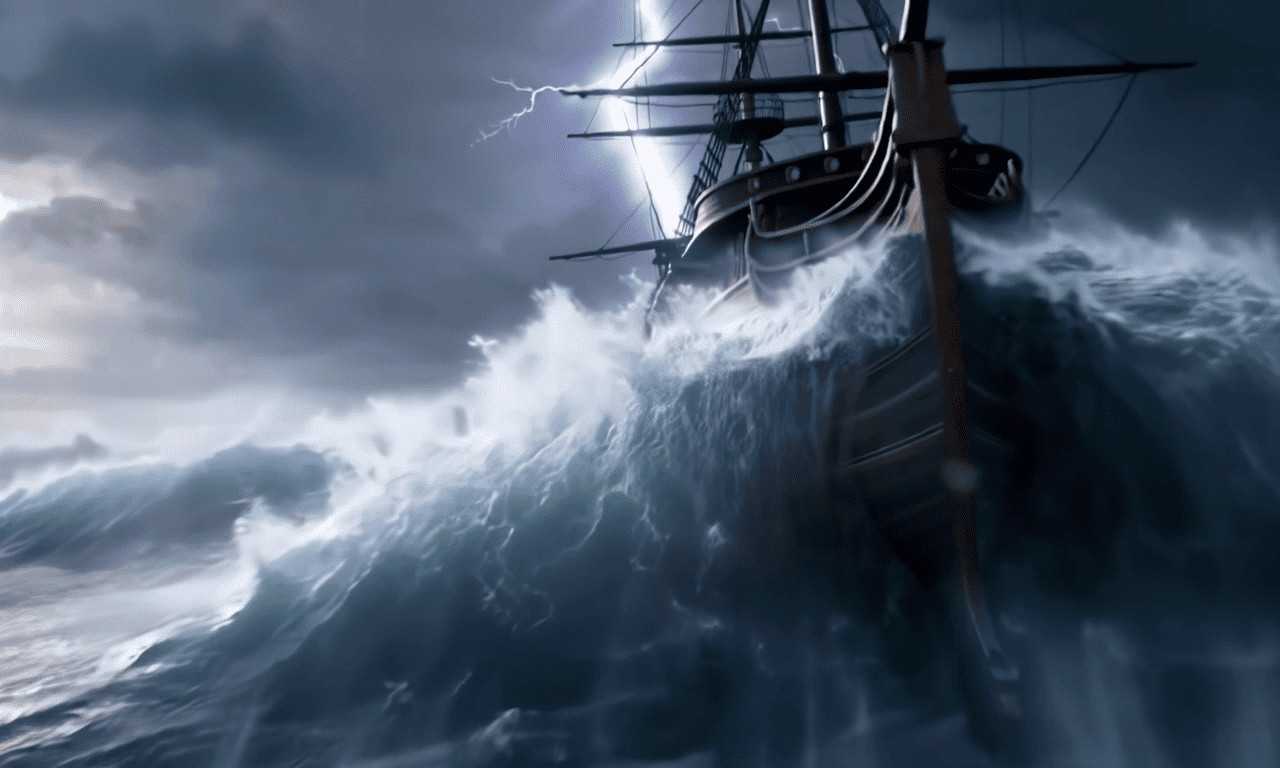} \hfill 
        \includegraphics[width=0.200\textwidth]{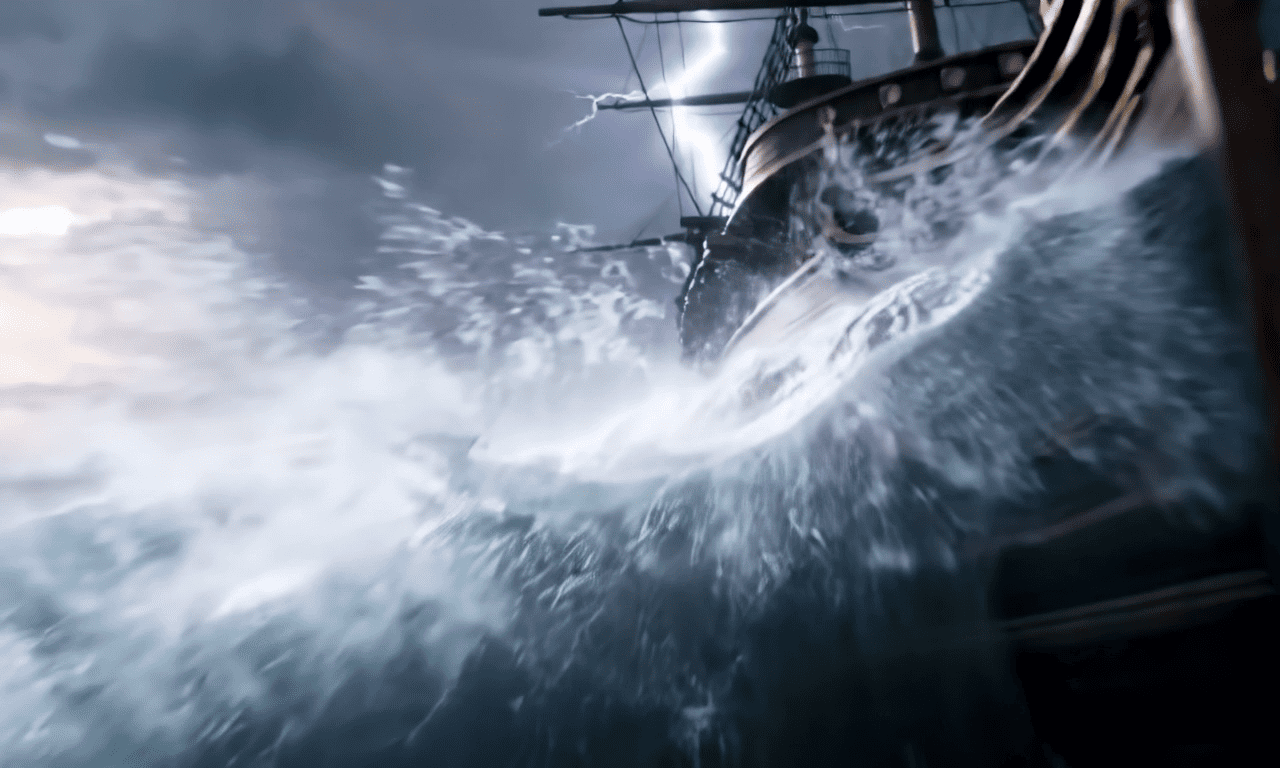}
    }

    \caption{Kandinsky 5.0 Video Pro text-to-video generation examples}
    \label{fig:pro_T2V}
\end{figure}

\begin{figure}[htbp]
    \centering    
    \captionsetup[subfloat]{labelfont=scriptsize,textfont=scriptsize}

    \subfloat[A young girl with curly hair is standing in a field of wildflowers, holding a bouquet of colorful flowers. She is smiling brightly and appears to be enjoying the moment. The field is lush and green, with various wildflowers blooming around her. The sky is clear and blue, indicating a sunny day. The girl is wearing a light-colored dress and seems to be in a joyful and carefree mood. The scene captures a moment of happiness and connection with nature.]
    {\label{subfig:bouquet}
        \begin{tabular}{l | l}
            \includegraphics[width=0.190\textwidth]{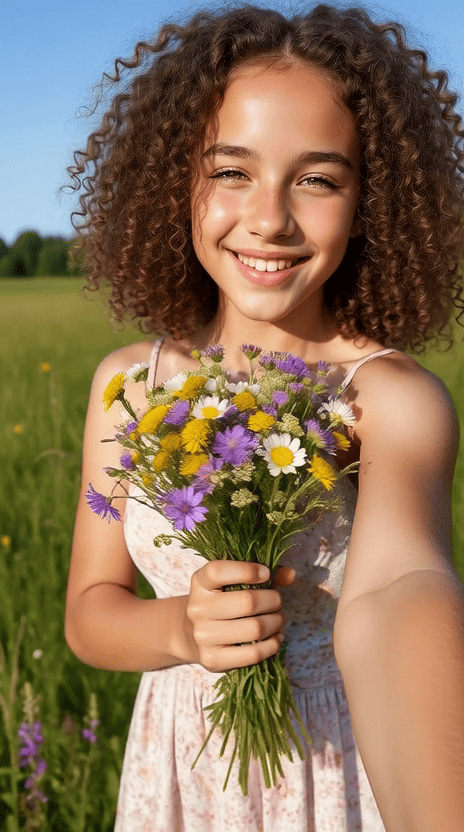} &
            \includegraphics[width=0.190\textwidth]{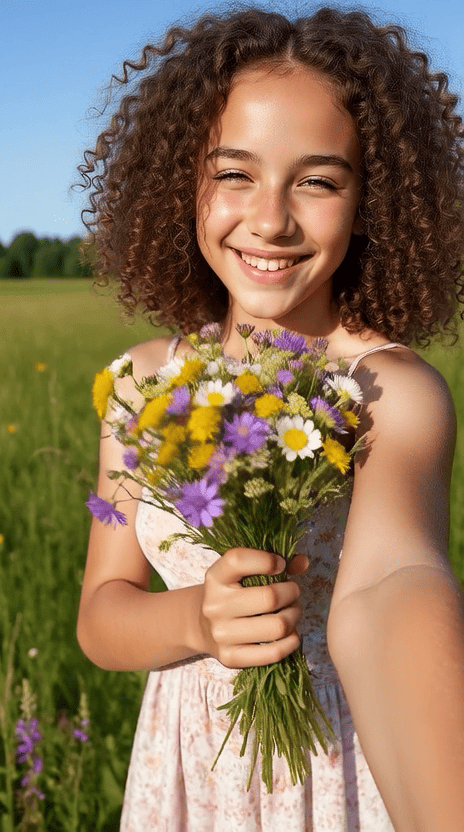} \hfill 
            \includegraphics[width=0.190\textwidth]{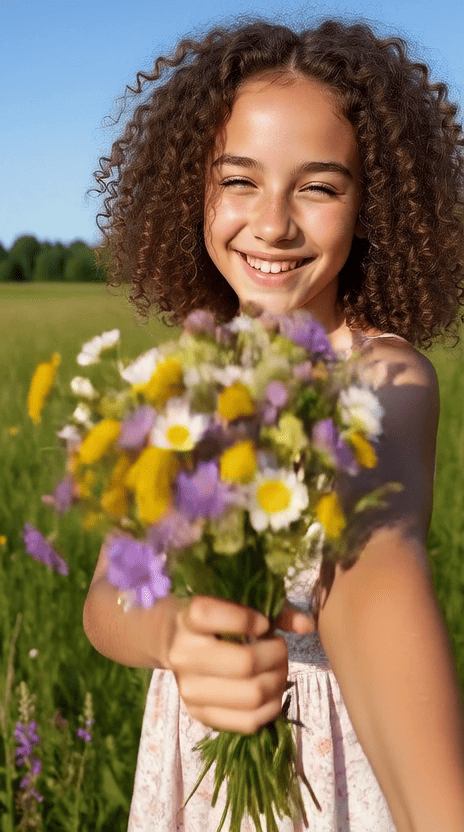} \hfill 
            \includegraphics[width=0.190\textwidth]{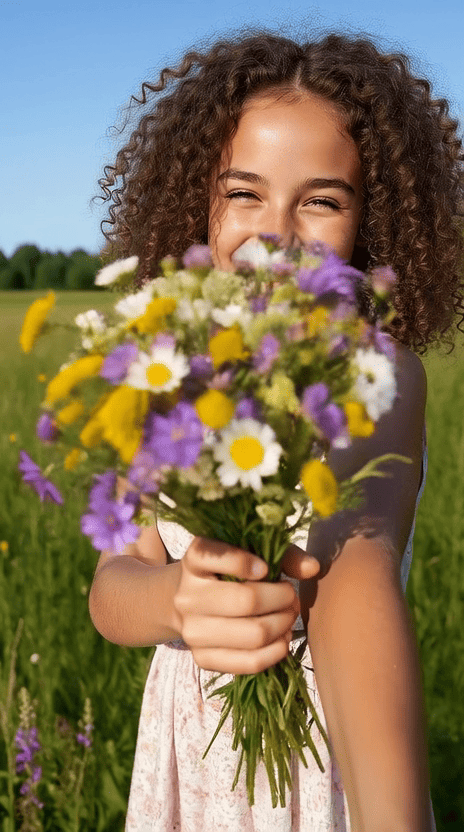} \hfill 
            \includegraphics[width=0.190\textwidth]{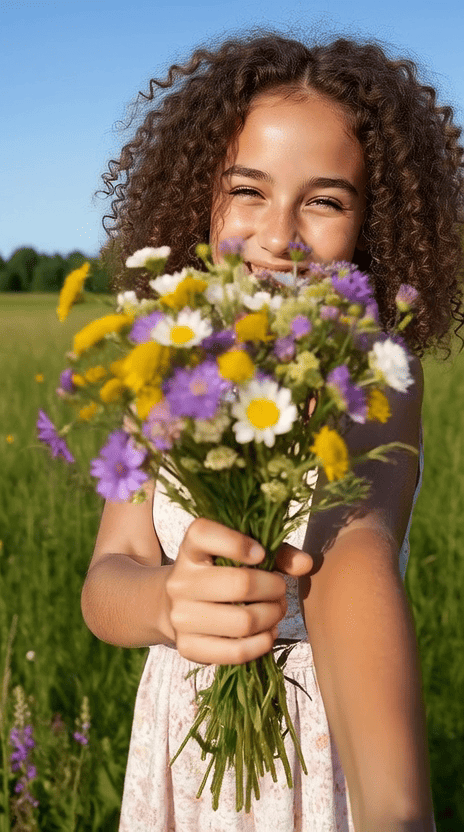}
        \end{tabular}
    }
    \\
    \vspace{0.3cm}

    \subfloat[A baby is sitting in a stroller, wearing a white outfit and a bonnet. The baby is smiling and appears to be happy. The stroller has a black handlebar with a textured grip. The background is slightly blurred, but it seems to be an outdoor setting with a wooden structure and a patterned wall. The image has a vintage, sepia tone, giving it a nostalgic feel.]
    {\label{subfig:bwchild}
        \begin{tabular}{l | l}
            \includegraphics[width=0.190\textwidth]{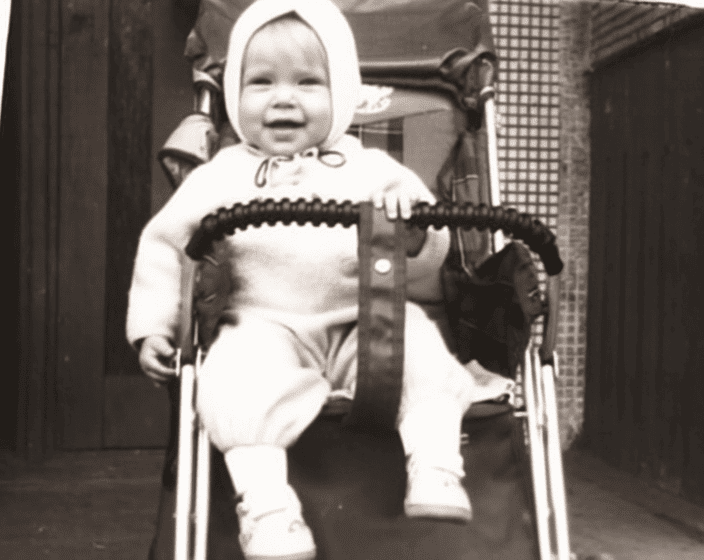} &
            \includegraphics[width=0.190\textwidth]{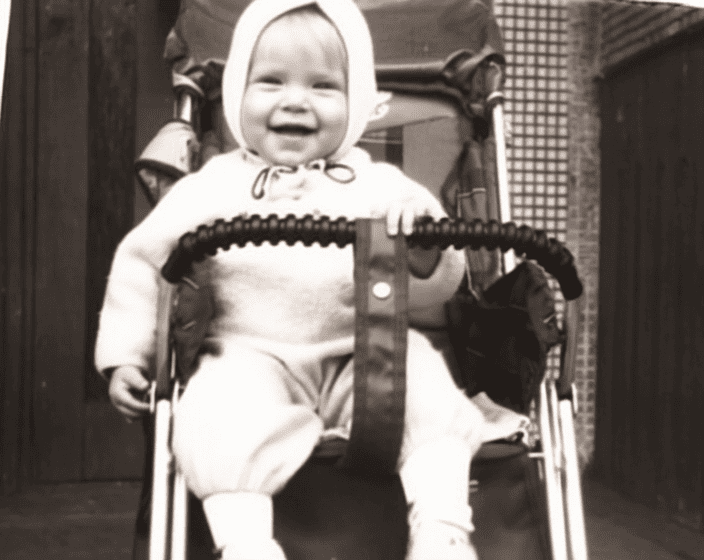} \hfill 
            \includegraphics[width=0.190\textwidth]{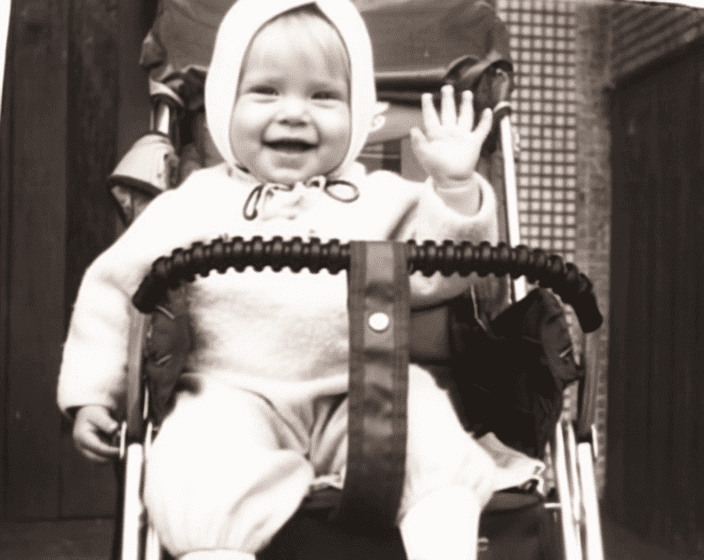} \hfill 
            \includegraphics[width=0.190\textwidth]{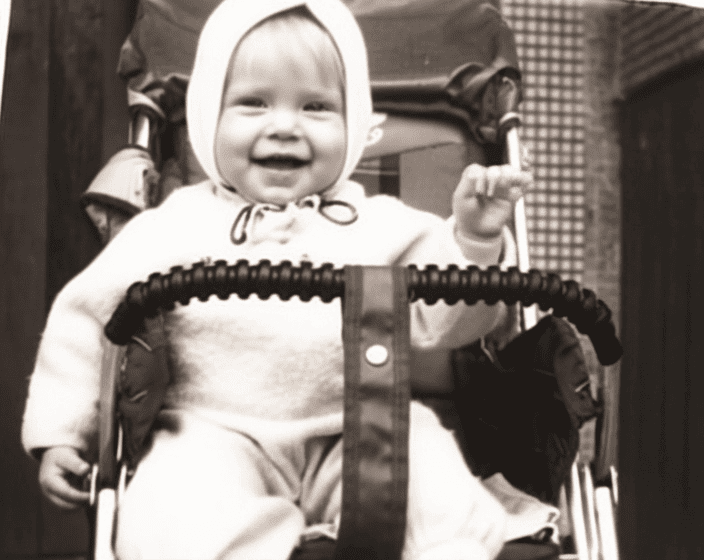} \hfill 
            \includegraphics[width=0.190\textwidth]{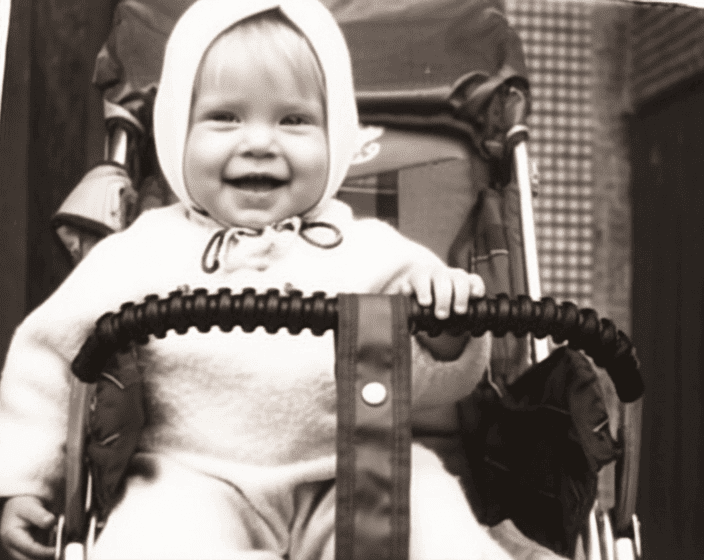}
        \end{tabular}
    }
    \\
    \vspace{0.3cm}

    \subfloat[The image of a man on a wooden fence is so realistic that it seems as if the man is actually stepping out from the image onto the sidewalk and walking down the street. The street is lined with buildings, and everything is illuminated by natural daylight, creating a surreal and intriguing visual effect.]
    {\label{subfig:graffiti}
        \begin{tabular}{l | l}
            \includegraphics[width=0.190\textwidth]{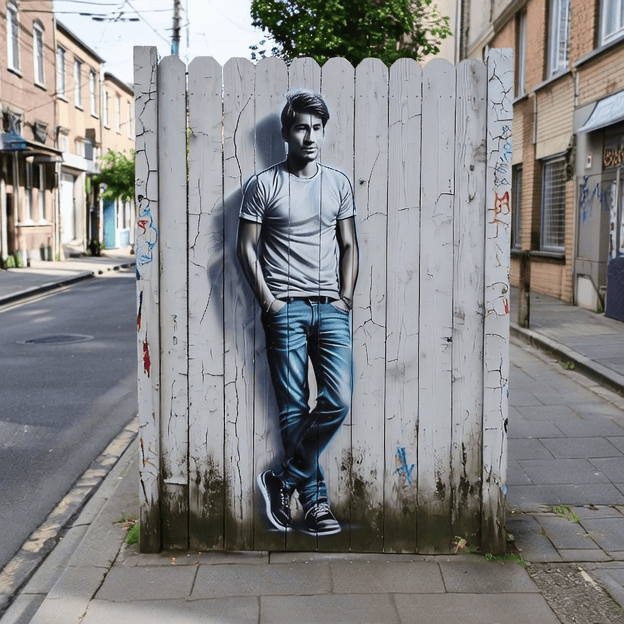} &
            \includegraphics[width=0.190\textwidth]{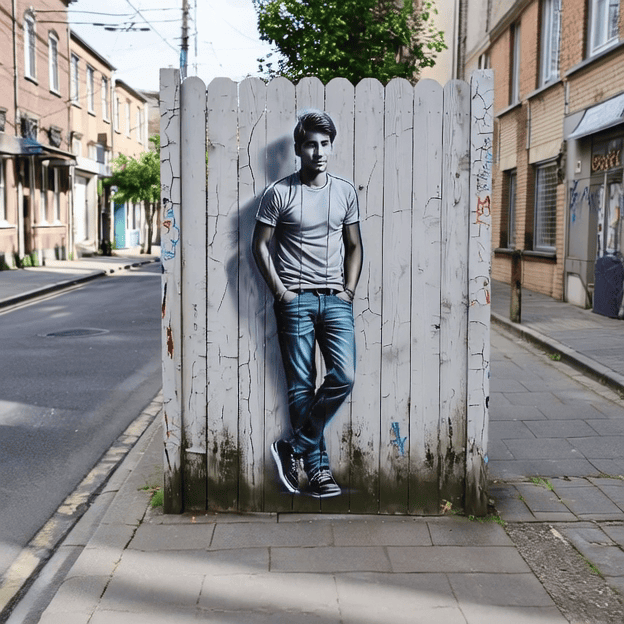} \hfill 
            \includegraphics[width=0.190\textwidth]{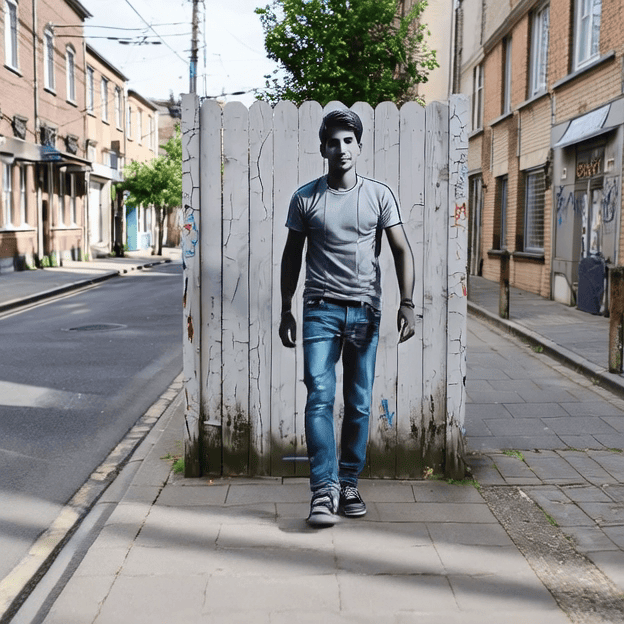} \hfill 
            \includegraphics[width=0.190\textwidth]{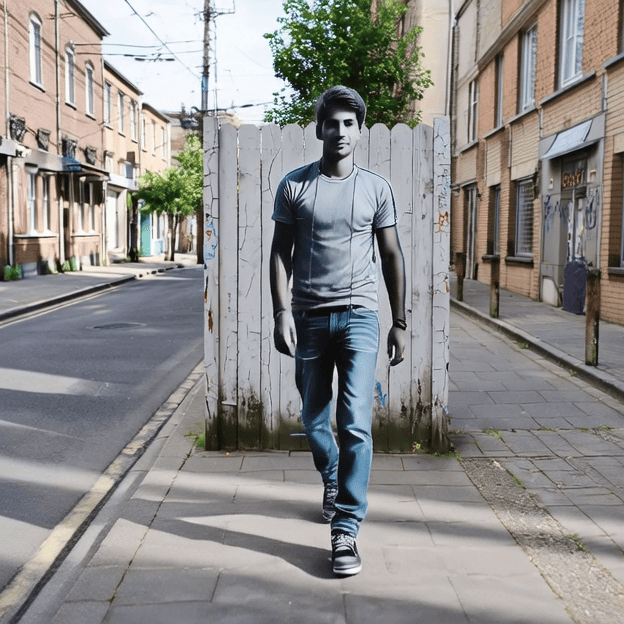} \hfill 
            \includegraphics[width=0.190\textwidth]{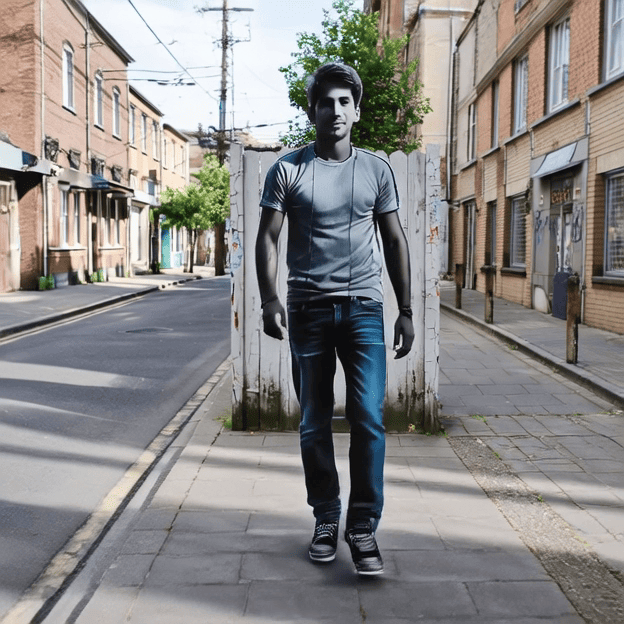}
        \end{tabular}
    }
    \\
    \vspace{0.3cm}

    \subfloat[The bear turns its head and raises its paw.]
    {\label{subfig:bear}
        \includegraphics[width=0.200\textwidth]{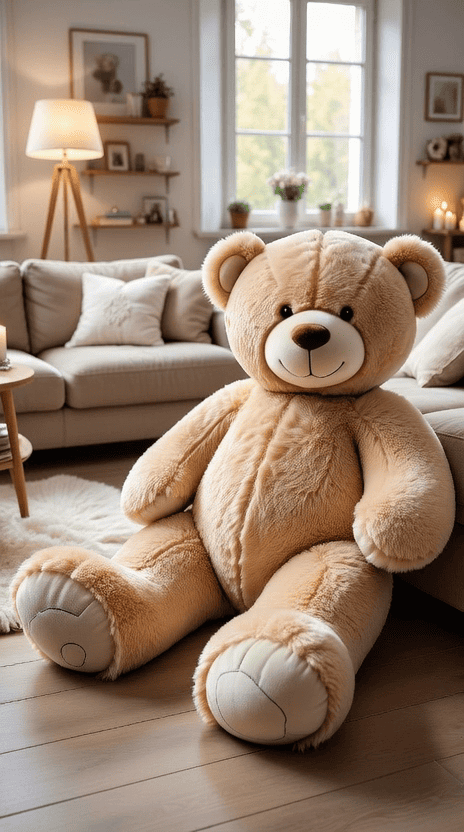} \hfill 
        \includegraphics[width=0.200\textwidth]{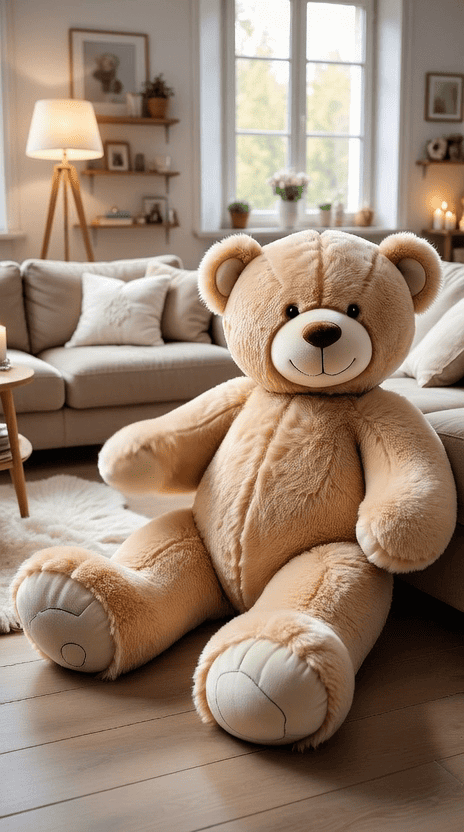} \hfill 
        \includegraphics[width=0.200\textwidth]{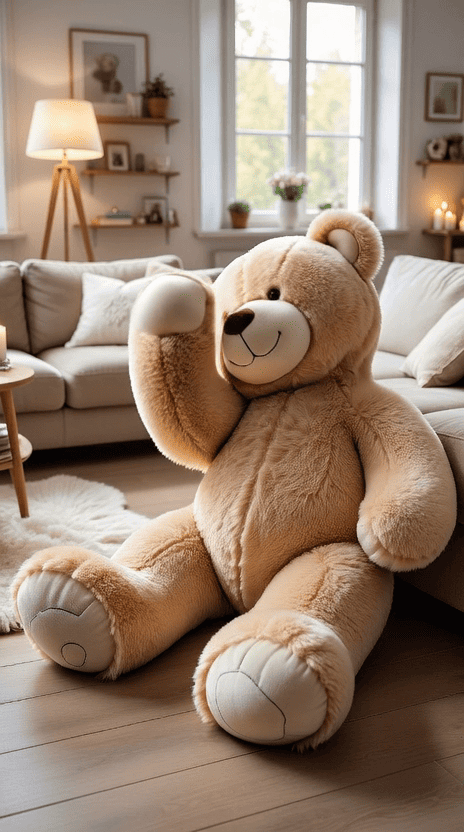} \hfill 
        \includegraphics[width=0.200\textwidth]{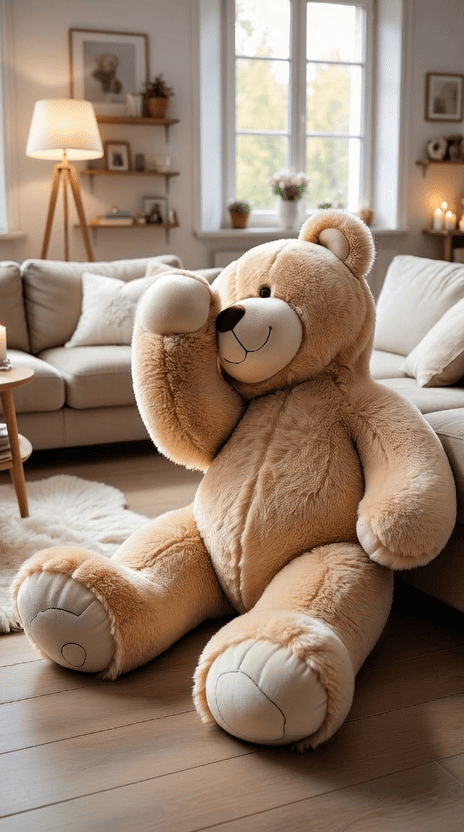} \hfill 
        \includegraphics[width=0.200\textwidth]{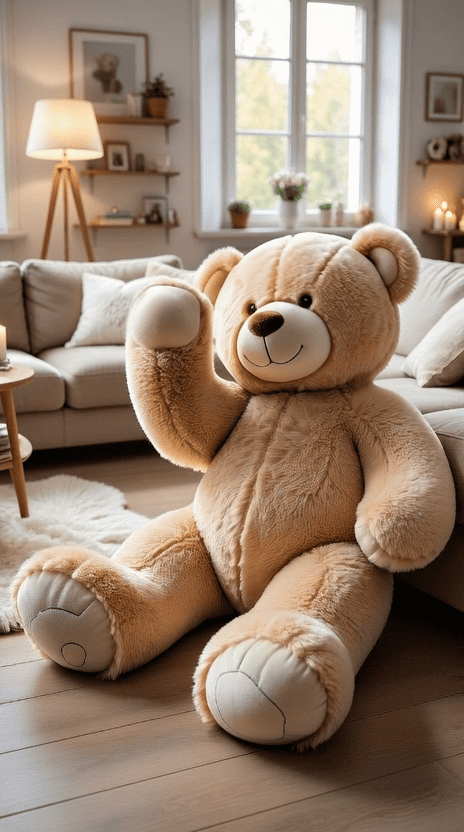}
    }

    \caption{Kandinsky 5.0 Video Lite image-to-video generation examples}
    \label{fig:lite_I2V}
\end{figure}

\subsection{Image Editing}
The \textbf{Kandinsky 5.0 Image Editing} model supports text instruction driven image editing mode, enabling precise, context-aware modifications of existing images. Given a source image and a textual prompt, the model can perform a wide range of operations—including object removal, insertion, attribute editing (e.g., changing colors, materials, or lighting), style transfer (e.g., converting a photograph into an oil painting or a pencil sketch), and even generating photorealistic renderings from rough sketches or wireframes. The editing process preserves global coherence and local details, ensuring seamless integration of new elements with the original composition. This capability is particularly effective for iterative design workflows, creative prototyping, and content adaptation. For instance, users can transform a hand-drawn concept into a high-fidelity product visualization or recontextualize historical artworks with modern stylistic treatments. Representative examples of text-guided image editing—including inpainting, outpainting, style conversion, and sketch-to-image synthesis—are provided in Figure~\ref{fig:image_editing}.

\subsection{Text-to-Video}
The core operational mode of the Kandinsky 5.0 models is text-to-video synthesis, supporting standard generation durations of 5 and 10 seconds and producing spatiotemporally coherent outputs at 24 fps. The \textbf{Kandinsky 5.0 Video Lite} version supports resolutions up to 768 pixels on the longer side, as detailed in Table \ref{tab:resolution}. The more powerful \textbf{Kandinsky 5.0 Video Pro} version additionally supports higher resolutions, up to 1408 pixels, enabling the production of generations with significantly finer detail and greater compositional complexity.

For optimal results, prompt construction must follow a technical schema: [Main subject Definition] + [Action/Motion Vector] + [Environmental Context] + [Cinematic and Camera Parameters]. Reference Figures \ref{fig:lite_T2V} and \ref{fig:pro_T2V} for visual examples of selected frames from generated videos and their corresponding technical prompts. 

The generated video content is particularly suitable for commercial applications including digital advertising campaigns, social media marketing content, and corporate presentation materials. All generated content must comply with the established ethical framework prohibiting misinformation, deepfake manipulation, and copyright infringement.

\subsection{Image-to-Video}
Kandinsky 5.0 Video line-ups supports Image-to-Video (I2V) synthesis as an advanced operational mode, generating dynamic video sequences from static input images and corresponding text guide. These models maintain standard output parameters of 5 and 10-second durations with spatiotemporal coherence at 24 fps, while supporting different resolution tiers across various model line-ups as specified in Table \ref{tab:resolution}.

For optimal I2V generation, prompt construction should employ a motion-focused technical schema: [Primary Motion] + [Temporal Characteristics] + [Camera Movement]. Reference Figure \ref{fig:lite_I2V} for visual examples of input images with corresponding motion prompts and generated frame sequences. This functionality enables diverse applications across creative and commercial domains, from animating children's drawings and classic artwork to bringing movement to family photographs and transforming product images into dynamic advertising content. The technology also supports animating previously generated digital assets, providing continuity in creative workflows. All generated content remains governed by the established ethical framework prohibiting unauthorized manipulation and copyright infringement, with specific safeguards for personal and copyrighted materials.

\section[Related Work]{Related Work}\label{sec:related_work}

\subsection{Image Generation}

The development of visual generative models has undergone transformative shifts over the past decade, driven by advances in deep learning architectures and training paradigms. Early breakthroughs began with \textbf{Generative Adversarial Networks (GANs)} \citep{goodfellow2014gan}, which introduced adversarial training between a generator and discriminator to synthesize realistic data. While GANs demonstrated unprecedented capabilities in generating coherent images, their limitations – mode collapse, unstable training dynamics, and difficulty scaling to high resolutions – spurred exploration of alternative approaches. \textbf{Variational Autoencoders (VAEs)} \citep{kingma2013vae, rezende2014stochastic} offered a probabilistic framework for learning latent representations, but often produced blurry outputs due to their reliance on pixel-wise reconstruction losses \citep{novitskiy2025vivatvirtuousimprovingvae}. Concurrently, \textbf{autoregressive models} like PixelRNN \citep{oord2016pixelrnn} and ImageGPT \citep{chen2020generative} achieved high sample quality by sequentially predicting image pixels or patches, though at the cost of impractical inference speeds and computational demands.

A paradigm shift emerged with the introduction of \textbf{diffusion models} \citep{ho2020ddpm}, which reframed generation as an iterative denoising process. By gradually corrupting data with noise and training a model to reverse this process, diffusion models avoided the instability of GANs while achieving superior sample diversity. The subsequent integration of \textbf{classifier-free guidance} \citep{dhariwal2021diffusion} enabled precise control over conditional generation tasks, such as text-to-image synthesis. However, the computational expense of pixel-space diffusion remained a barrier until \textbf{Latent Diffusion Models (LDMs)} \citep{Rombach_2022_CVPR} demonstrated that operating in a compressed latent space – learned via some kind of autoencoder – could drastically reduce training and inference costs while maintaining high-resolution output quality. This innovation democratized generative AI, enabling open-source projects like Stable Diffusion to flourish.

The next leap forward came with the fusion of diffusion frameworks and transformer architectures. \textbf{Diffusion Transformers (DiT)} \citep{peebles2023dit} replaced the traditional U-Net backbone with scalable transformer blocks, capitalizing on their ability to model long-range dependencies and adhere to predictable scaling laws. DiT's success in text-to-image generation paved the way for its adaptation to video synthesis, where models like Sora \citep{openai2024sora} leveraged spatio-temporal attention to generate coherent, high-fidelity videos. These architectures further incorporated techniques such as \textbf{flow matching} and \textbf{bridge matching} \citep{lipman2023flow, vasilev2026timecorrelatedvideobridgematching} to streamline the alignment of latent trajectories and \textbf{cross-attention mechanisms} to enhance multimodal conditioning.

\subsection{Video Generation}

Video generation technology has sparked widespread interest across industrial and academic domains, catalyzing rapid advancements in synthetic media. The rise of generative models capable of producing studio-grade video content has transformed creative workflows, slashing production costs while improving output quality. Much of this progress comes from open source initiatives, with projects such as HunyuanVideo \citep{kong2025hunyuanvideosystematicframeworklarge}, Mochi \citep{genmo2024}, CogVideoX \citep{yang2025cogvideox}, and Wan \citep{wan2024} democratizing access to foundational architectures and pretrained weights. These efforts have narrowed – though not closed – the gap between open-source and proprietary systems.

The frontier of modern video generation now extends well beyond basic text-to-video conversion, encompassing a rich spectrum of interconnected tasks that push the boundaries of temporal modeling. Contemporary systems must handle: \textbf{image-to-video} (I2V) and \textbf{reference-to-video} (R2V) synthesis where static compositions spring to life while maintaining geometric consistency \citep{blattmann2023align}; \textbf{video-to-video} (V2V) transformations for style transfer and content editing \citep{wang2022videocontrolnet}; \textbf{video-to-audio} (V2A) generation that creates synchronized soundscapes from visual dynamics \citep{chen2023v2a}; and sophisticated \textbf{first-last frame interpolation} that infers natural motion between sparse keyframes \citep{reda2022frame, vasilev2026timecorrelatedvideobridgematching}. Perhaps the most cinematographically challenging is the \textbf{ precise camera control}, which requires models to understand the principles of virtual cinematography, from panning to dolly zoom effects \citep{brooks2023cameractrl}.

\subsubsection{Attention mechanism optimizations}

Video Generation come at significant computational cost, particularly in latent-space video models, where attention is focused on compressed representations. While standard image diffusion uses attention maps of size $(H/f_s \times W/f_s)^2$ for latent dimensions $H/f_s \times W/f_s$ (where $f_s$ is spatial downsampling factor), video models must handle relationships $(T/f_t \times H/f_s \times W/f_s)^2$ - where $f_t$ represents temporal compression in the Video VAE \citep{wu2023videovae}. This quadratic scaling has led to several optimization approaches in recent works:

\begin{itemize}
    \item \textbf{Memory-Efficient Attention}: The Flash Attention algorithm \citep{dao2022flash},\citep{dao2023flash2} provides 2.4-3.1× speedups for video generation by:
    \begin{itemize}
        \item Computing attention scores in tiles to reduce GPU memory bandwidth;
        \item Fusing kernel operations to avoid expensive memory reads/writes;
        \item Supporting mixed-precision calculations with minimal accuracy loss.
    \end{itemize}
    \item The recently proposed \textbf{Sliding Tile Attention} \citep{zhang2024fast} addresses computational bottlenecks in video Diffusion Transformers (DiTs) through hardware-aware sparsity. Key innovations include:
    \begin{itemize}
        \item Tile-based computation: Processes video latents as 3D tiles rather than individual tokens, eliminating irregular attention masks while preserving spatial-temporal locality.
        \item Asynchronous memory pipeline: Implements producer-consumer warpgroups to overlap data loading with computation, achieving 58.79\% MFU (model FLOPs utilization).
        \item Training-free adaptation: Automatic window size configuration per attention head via score profiling, enabling 1.36× speedup with 98\% quality retention.
    \end{itemize}
\end{itemize}

\subsection{Post-training RL-based Techniques}

Crucial role in achieving state-of-the-art quality for Image and Video generation models in terms of realism, aesthetic appeal, prompt following and general alignment plays Reinforcement Learning (RL) - based training or Reinforcement Learning on Human Feedback (RLHF)~\cite{christiano2023deepreinforcementlearninghuman}. 

For performing alignment of the generative model with RL - based methods, it is required to have a data, that was annotated by real humans / VLM models or to propose an heuristic to generate train data in a way, that all samples would be ordered a priori and would not require annotations.

One way to use this data is to align our model directly to the annotations with algorithms like DPO~\cite{wallace2023diffusionmodelalignmentusing}. Another way is to train a reward model to give generative model feedback on its outputs and then perform a training procedure on the generative model, that would maximize the scores of this reward.

In the works~\cite{hpsv2, xu2023imagerewardlearningevaluatinghuman} it was proposed to initialize the reward model as CLIP~\cite{radford2021clip} and train it in a contrastive manner on the collected human annotations of multiple images, that correspond to the same caption, for example, with contrastive Bradley-Terry loss (see formulas (1) and (2) in the work~\cite{liu2025improvingvideogenerationhuman}). This approach for reward training can be applied for Visual Language Models (VLMs) and after training they can be utilized as reward models. In the paper~\cite{wang2025unifiedrewardmodelmultimodal} authors suggested training Qwen2.5-VL~\cite{bai2025qwen25vltechnicalreport} to output score from 1 to 5 for a given image. In the work~\cite{ma2025hpsv3widespectrumhumanpreference} authors proposed to add an additional linear regression head on the logits of the output of VLM, which turns it into a regression model for reward prediction. All these aforementioned approaches for rewards are only capable of getting a single image as input on RLHF stage. In the work~\cite{wu2025rewarddancerewardscalingvisual} it was noted by authors, that reward models, that operate comparatively on the pairs of images on RLHF stage tend to give a better feedback and provide all necessary ablation studies to show, that comparison is a more robust way to use a reward model.

RL-based fine-tuning methods for diffusion models, that utilize rewards can be divided into two categories: direct optimization of reward through computations of gradients~\cite{xu2023imagerewardlearningevaluatinghuman, clark2024directlyfinetuningdiffusionmodels} and common RLHF algorithms, that do not require computations of gradients for output of reward model~\cite{black2024trainingdiffusionmodelsreinforcement, liu2025flowgrpotrainingflowmatching}. In the work~\cite{wu2025rewarddancerewardscalingvisual} authors also suggested adaptation of gradient-based algorithms for rewards, that take multiple image/video samples as inputs. Adaptation of GRPO~\cite{liu2025flowgrpotrainingflowmatching} for rewards, that operate on multiple images, was proposed in the paper~\cite{wang2025prefgrpopairwisepreferencerewardbased}.

\subsection{Distillation Methods}

Diffusion models generate high-quality samples but are computationally expensive, as they require solving a complex differential equation through many iterative steps, each involving an expensive network evaluation~\cite{salimans2022progressive}. Distillation methods address this by learning a ``simpler'' differential equation that results in the same final data distribution at timestep \(t=0\) but follows a ``straighter,'' more linear trajectory. This allows for larger step sizes and, consequently, fewer network evaluations~\cite{liu2022flowstraightfastlearning, song2023consistency}.

Existing distillation techniques can be broadly categorized. \textbf{Deterministic methods} aim to predict the exact output of the teacher model using fewer steps. While easy to train with regression loss, they often produce blurry results in few-step generation due to optimization inaccuracies~\cite{lin2024sdxl}. \textbf{Distributional methods}, on the other hand, only aim to approximate the teacher's output distribution and often employ adversarial or distribution-matching objectives to achieve higher perceptual quality~\cite{sauer2023adversarialdiffusiondistillation, yin2024onestep}.

Key distillation families include:

\begin{enumerate}
    \item \textbf{Progressive Distillation}. This method iteratively distills a teacher model into a student that halves the number of required sampling steps. While effective, it suffers from error accumulation as multiple rounds of distillation are typically needed~\cite{salimans2022progressive, meng2023distillationguideddiffusionmodels}.
    
    \item \textbf{Consistency Distillation}. This approach trains a student model to map any point along the probability flow ODE trajectory directly to the origin, ensuring self-consistency across timesteps. It can be performed in a single stage but often requires careful tuning and specialized techniques like distillation schedules for stable training~\cite{song2023consistency, song2023improved, luo2023latent}. Improved versions, such as Multistep Consistency Models and Latent Consistency Models (LCMs), have since been developed~\cite{heek2024multistep, luo2023latent}.
    
    \item \textbf{Adversarial Distillation}. For high-quality, few-step generation, adversarial training has become prominent. Methods like Adversarial Diffusion Distillation (ADD) use a pretrained feature extractor as a discriminator, enabling performance competitive with large teacher models like SDXL in as few as four steps~\cite{sauer2023adversarialdiffusiondistillation, lin2024sdxl}. Other approaches combine adversarial loss against real data with score distillation from the teacher model~\cite{yin2024improved, sauer2023adversarialdiffusiondistillation}.
\end{enumerate}

A related strategy is \textbf{Rectified Flow}, which straightens the ODE trajectories to make them easier to approximate~\cite{liu2022flowstraightfastlearning, liu2023instaflow}. Another early approach, \textbf{Knowledge Distillation}, involved precomputing a dataset of noise-image pairs from the teacher model to train a one-step student, a requirement later methods eliminated~\cite{luhman2021knowledge}.

While most distillation research focuses on image generation, these principles are also being applied to video generation, with recent works demonstrating one-step or few-step generation of high-resolution videos~\cite{zhang2024sfv, mao2024osv, yin2024causal}.

\subsection{Generative Model Evaluation}

Reliable evaluation of text-to-image (T2I) and text-to-video (T2V) generative models remains a significant challenge. While automated metrics offer scalability, they often exhibit weak correlation with human judgment, particularly for complex attributes such as semantic coherence, temporal dynamics, fine-grained object fidelity, and compositional reasoning. Commonly used image metrics include Fréchet Inception Distance (FID) \cite{HeuselRUNKH17}, Kernel Inception Distance (KID) \cite{bińkowski2021demystifyingmmdgans}, and CLIP Score \cite{hessel2021clipscore}; for video, Fréchet Video Distance (FVD) \cite{unterthiner2019accurategenerativemodelsvideo} and scores based on representations from Video Foundation Models like InternVideo2 \cite{wang2024internvideo2} are frequently adopted. However, these metrics primarily assess distributional similarity or coarse semantic alignment and struggle to capture motion realism, physical plausibility, adherence to compositional prompts, or dynamic consistency over time.

Recent efforts have sought to address these gaps through more structured and fine-grained benchmarks. The latest iteration, T2I-CompBench++ \cite{huang2025t2icompbenchpp} introduces 8,000 prompts across four categories and eight sub-categories, including generative numeracy and 3D spatial relationships alongside attribute binding and object interactions. It proposes tailored evaluation metrics—such as Disentangled BLIP-VQA for attribute binding and a UniDet-based metric with depth estimation for 3D layout and counting—demonstrating strong correlation with human judgments. T2V-CompBench \cite{sun2025t2vcompbenchcomprehensivebenchmarkcompositional} establishes the comprehensive benchmark for compositional text-to-video generation, featuring 1,400 prompts across seven categories: consistent and dynamic attribute binding, spatial relationships, motion binding, action binding, object interactions, and generative numeracy. It introduces a suite of specialized metrics—MLLM-based (Grid-LLaVA, D-LLaVA), detection-based (GroundingDINO), and tracking-based (DOT)—validated through extensive human correlation studies. The benchmark reveals that current T2V models struggle profoundly with dynamic attribute changes, motion direction, and multi-object counting.

Complementing this, DEVIL \cite{liao2024evaluationtexttovideogenerationmodels} introduces a dynamics-centric evaluation protocol that quantifies a model’s ability to generate videos with appropriate levels of motion intensity and temporal change as specified by the prompt. It defines three key metrics—dynamics range, dynamics controllability, and dynamics-based quality—and reveals that many state-of-the-art models ``cheat'' by generating low-dynamic videos to inflate traditional quality scores. Both T2V-CompBench and DEVIL highlight that current T2V models still struggle with fine-grained prompt adherence, especially in dynamic and compositional settings.

As a result, human evaluation—especially pairwise side-by-side (SBS) comparisons—has become the de facto standard for model assessment in high-stakes settings. SBS studies provide interpretable, attribute-level judgments (e.g., on visual quality, motion smoothness, or prompt following) and demonstrate stronger alignment with perceptual quality than automated scores. Nevertheless, human evaluation is resource-intensive and requires careful design to ensure reliability, including sufficient rater overlap, trained annotators, and measurement of inter-rater agreement. Recent large-scale benchmarks such as MovieGen \cite{polyak2024movie} now incorporate structured human evaluation protocols alongside automatic metrics, acknowledging that robust model comparison necessitates both scalable proxies and human-grounded validation.

\section[Limitations and Further Work]{Limitations and Further Work}\label{sec:limitations}

While the models from Kandinsky 5.0 family demonstrate state-of-the-art performance in generation stability and visual quality, our work has several limitations that outline promising directions for future research:

\begin{itemize}
    \item \textbf{Text-Visual Alignment.} Quantitative results in side-by-side (SBS) evaluations, indicate a slight lag in textual prompt understanding compared to some competing solution. We attribute this primarily to the limited context length (256 tokens) of the Qwen2.5-VL 7B text encoder used in the our pipeline. Future work will focus on improving text alignment by integrating more powerful text encoders with extended context windows and exploring advanced Reinforcement Learning (RL) based fine-tuning techniques for better prompt understanding;

    \item \textbf{Temporal Consistency for Complex Dynamics.} Although the spatio-temporal attention mechanism ensures robust frame-to-frame stability, modeling long-range, complex physical interactions (e.g., fluid dynamics, cloth simulation) remains challenging. In sequences longer than 10 seconds, these interactions can occasionally exhibit artifacts. Enhancing the physical realism and long-term temporal consistency of such dynamic phenomena is a key objective for our next model iteration;

    \item \textbf{Generalization Ability.} Despite the model's broad knowledge of the visual world, its performance is not uniform across all styles, objects, and scenes. This limitation stems from inherent dataset quirks, including class imbalance and stylistic or semantic biases within the training data. We are actively investigating methods for intelligent data curation and the assembly of a more representative and higher-quality training set. Ultimately, we aim to enhance the model's robustness for deployment in real-world scenarios such as autonomous systems, virtual reality (VR), simulation, and world models;

\begin{figure}[!h]
    \centering
    \begin{minipage}{\textwidth}
    	\centering
        \begin{subfigure}{0.32\textwidth}
            \centering
            \includegraphics[width=0.9\linewidth]{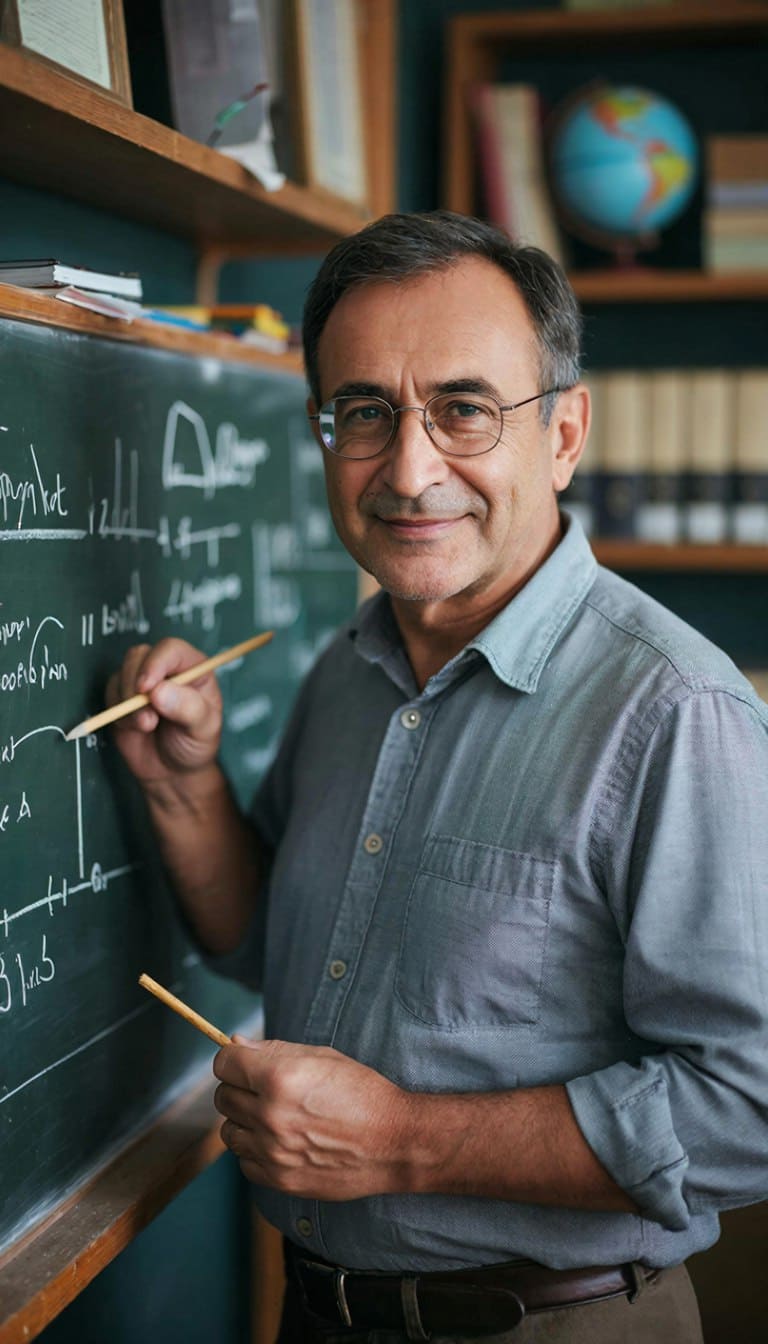}
            \caption{``A teacher''.}
        \end{subfigure}
        \hfill
        \begin{subfigure}{0.32\textwidth}
            \centering
            \includegraphics[width=0.9\linewidth]{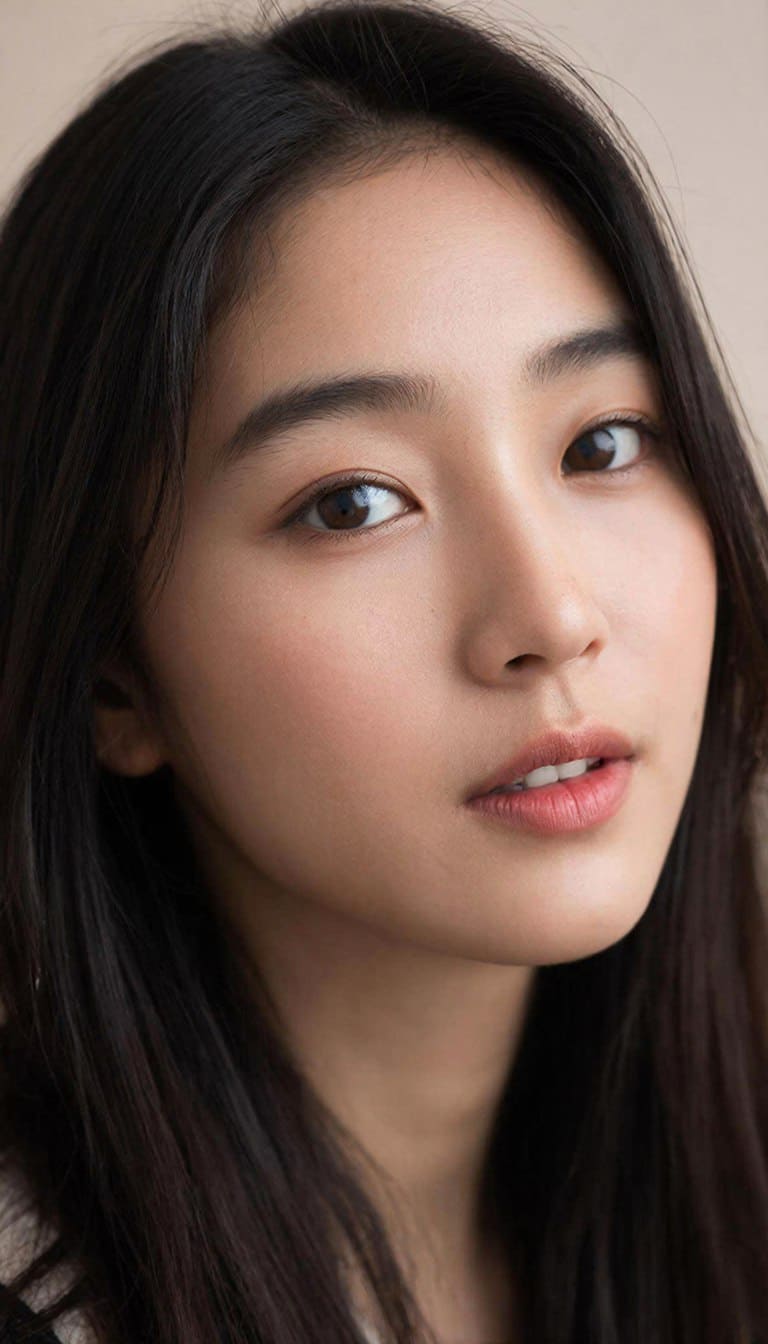}
            \caption{``An Asian''.}
        \end{subfigure}    
        \hfill
        \begin{subfigure}{0.32\textwidth}
            \centering
            \includegraphics[width=0.9\linewidth]{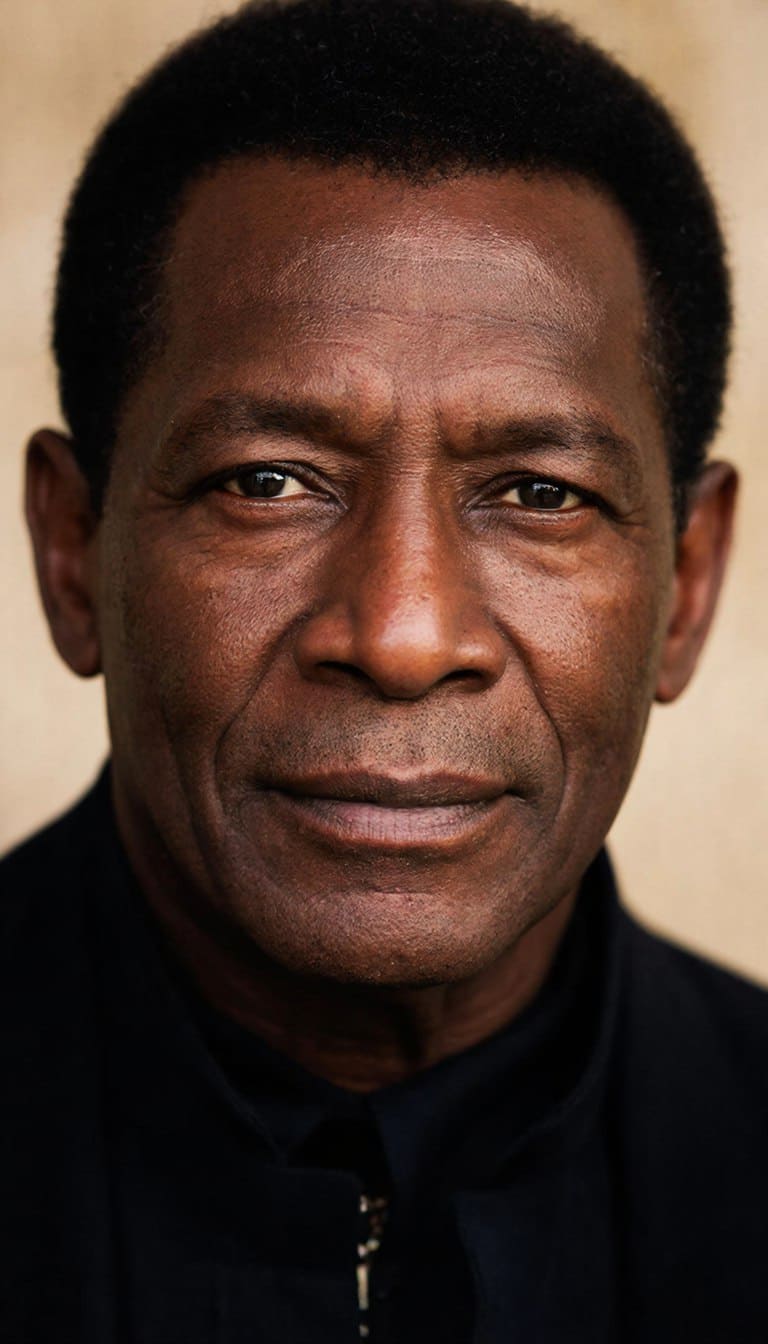}
            \caption{``An African American''.}
        \end{subfigure}
    \end{minipage} \\
    \vspace{0.5cm}
    \begin{minipage}{\textwidth}
    	\centering
        \begin{subfigure}{0.32\textwidth}
            \centering
            \includegraphics[width=0.9\linewidth]{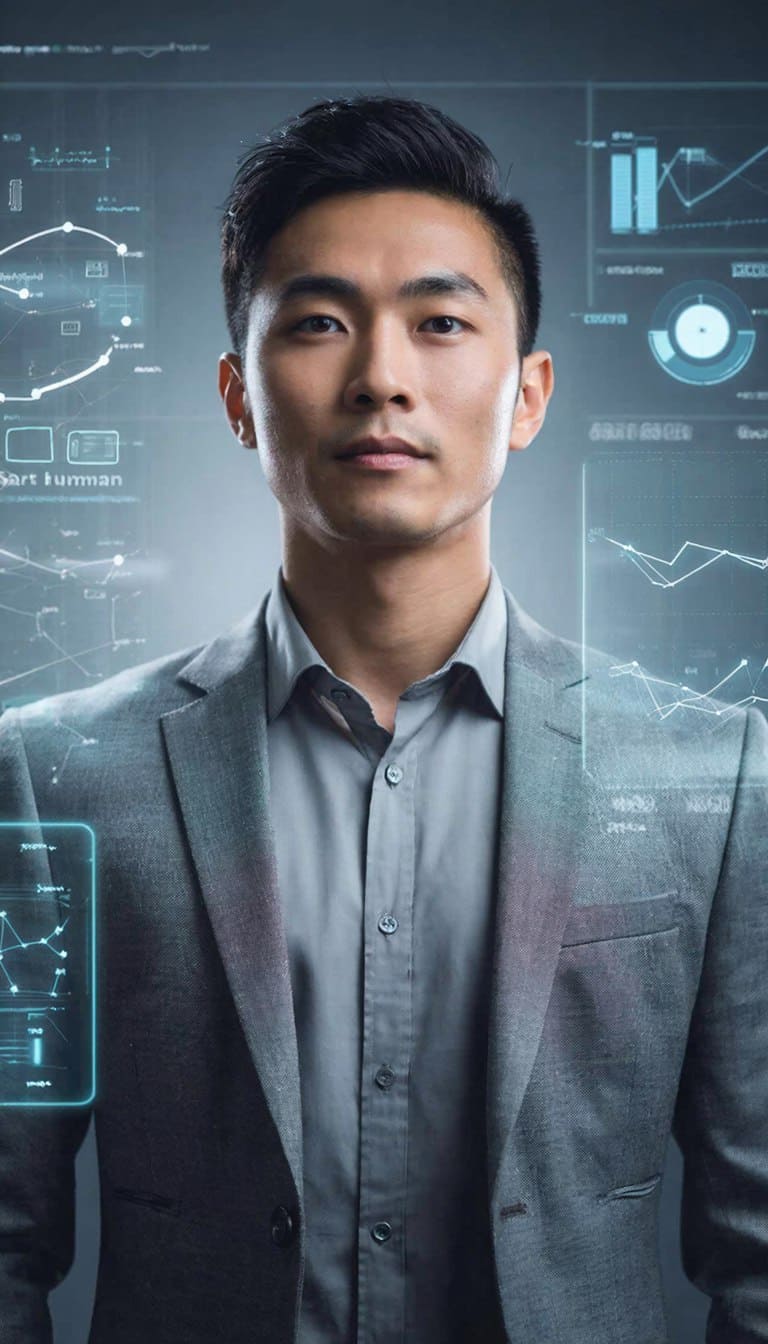}
            \caption{``A smart human''.}
        \end{subfigure}    
        \hfill
        \begin{subfigure}{0.32\textwidth}
            \centering
            \includegraphics[width=0.9\linewidth]{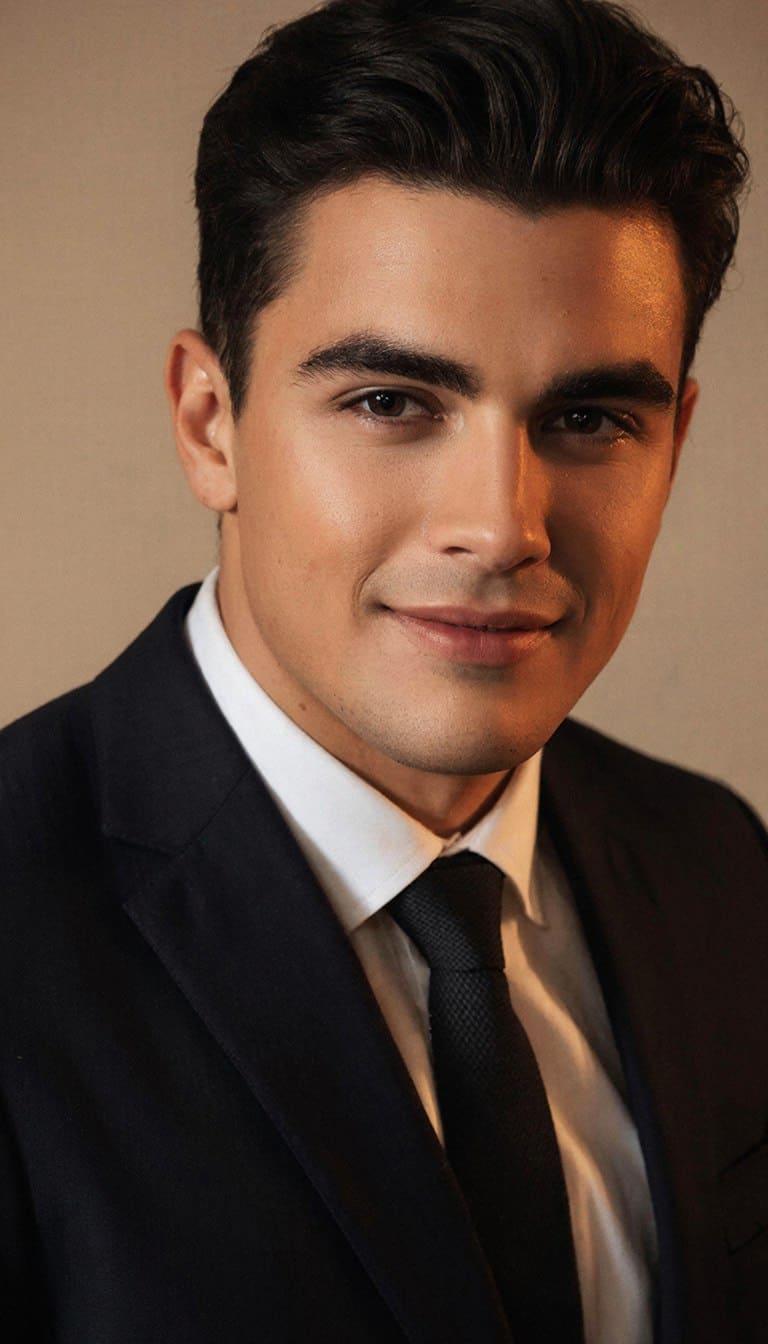}
            \caption{``A handsome man''.}
        \end{subfigure}    
        \hfill
        \begin{subfigure}{0.32\textwidth}
            \centering
            \includegraphics[width=0.9\linewidth]{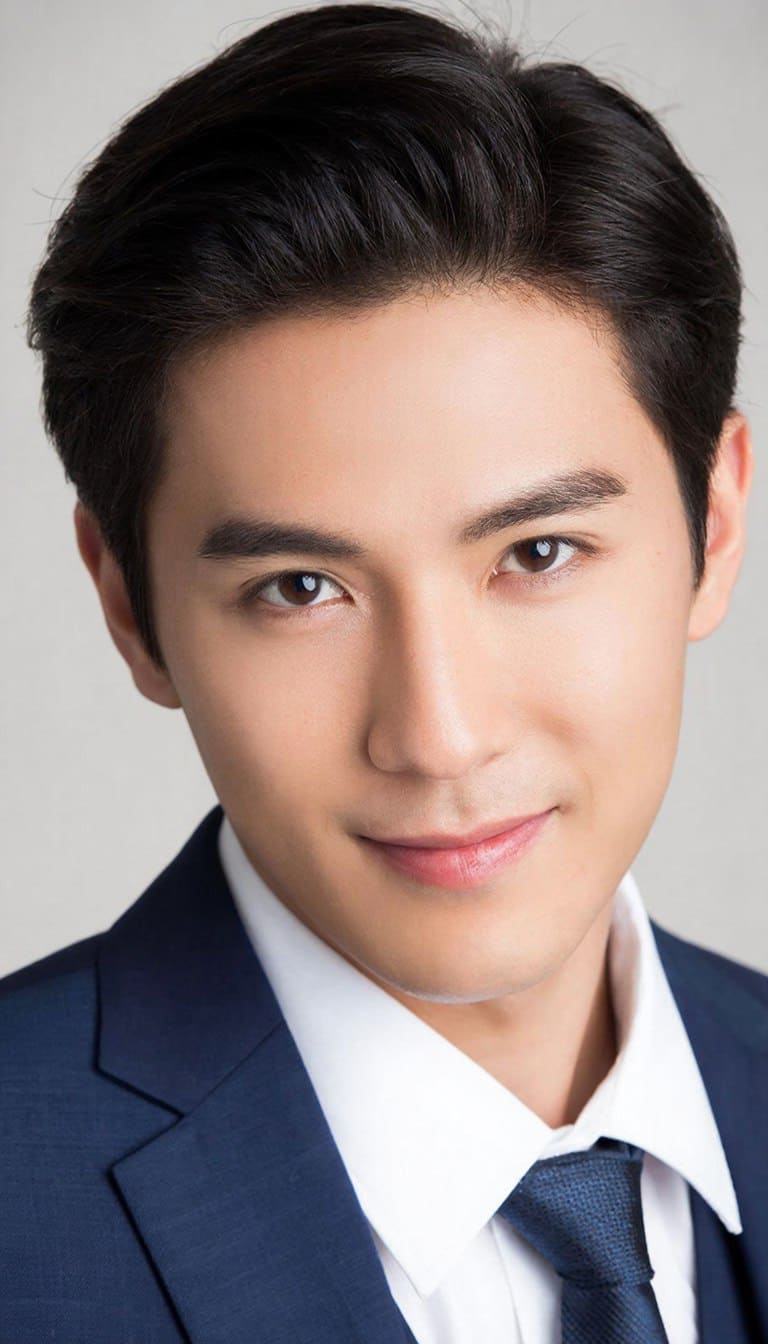}
            \caption{``A handsome man''.}
        \end{subfigure}
    \end{minipage}
    \caption{Examples of simple prompts that often produce similar results. In some cases, the model use the most common and well-established concepts of popular culture.}
    \label{fig:diversity}
\end{figure}

    \item \textbf{Creating a Foundation Visual Model.} The current Kandinsky 5.0 is a family of specialized, high-quality open-source models dedicated to specific generative tasks. A significant long-term goal is to consolidate these capabilities into a single, unified foundational model for multimedia generation. Such a model would possess a deeper, more integrated understanding of the visual world and could address multiple tasks without relying on a complex model ecosystem;

    \item \textbf{Computational Efficiency.} Achieving real-time generation rates (24+ FPS) at high resolutions on consumer-grade hardware remains a challenge, despite our optimizations. Our ongoing engineering efforts are directed towards developing more efficient architectures and inference techniques to make high-fidelity generative AI accessible on resource-constrained devices.
    
\end{itemize}

\section[Border Impacts and Ethical Considerations]{Border Impacts and Ethical Considerations}\label{sec:ethics}

\begin{figure}[t]
    \centering
    \begin{minipage}{\textwidth}
    	\centering
        \begin{subfigure}{0.32\textwidth}
            \centering
            \includegraphics[width=0.9\linewidth]{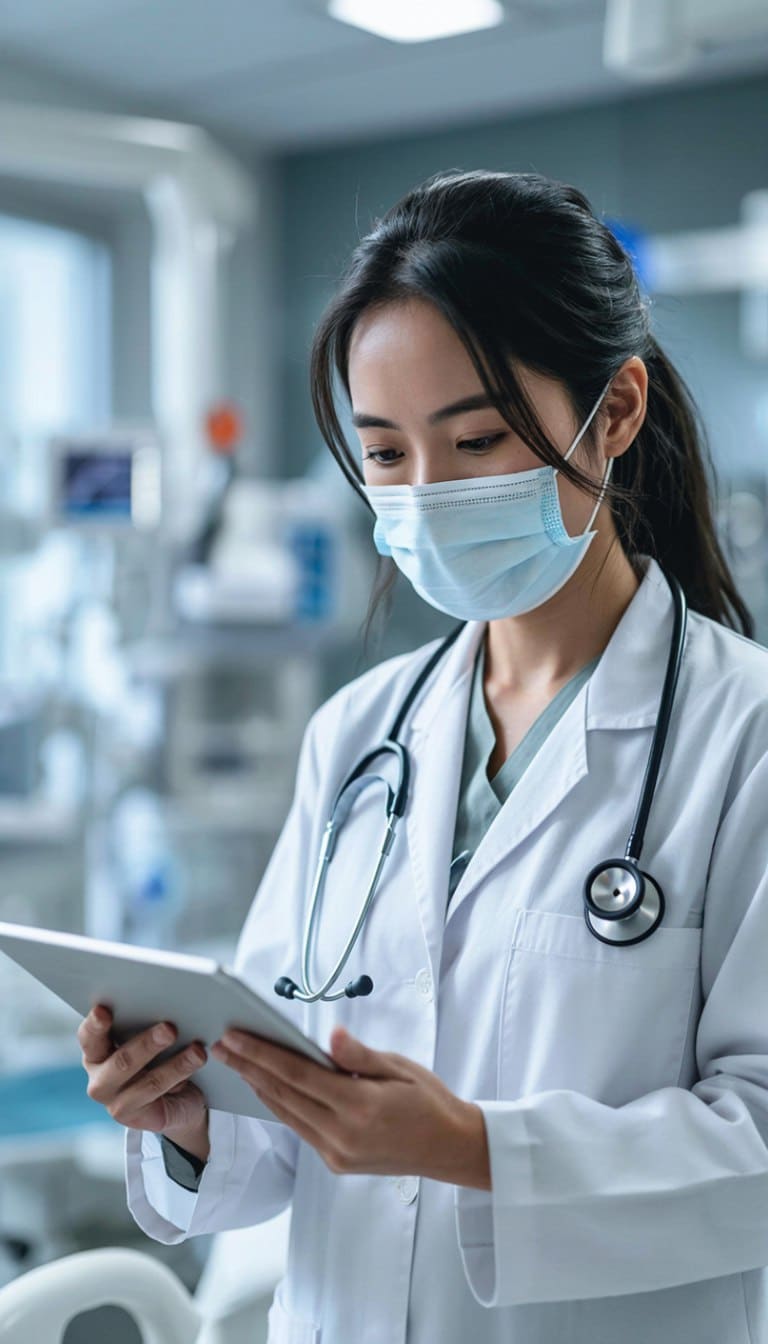}
        \end{subfigure}    
        \hfill
        \begin{subfigure}{0.32\textwidth}
            \centering
            \includegraphics[width=0.9\linewidth]{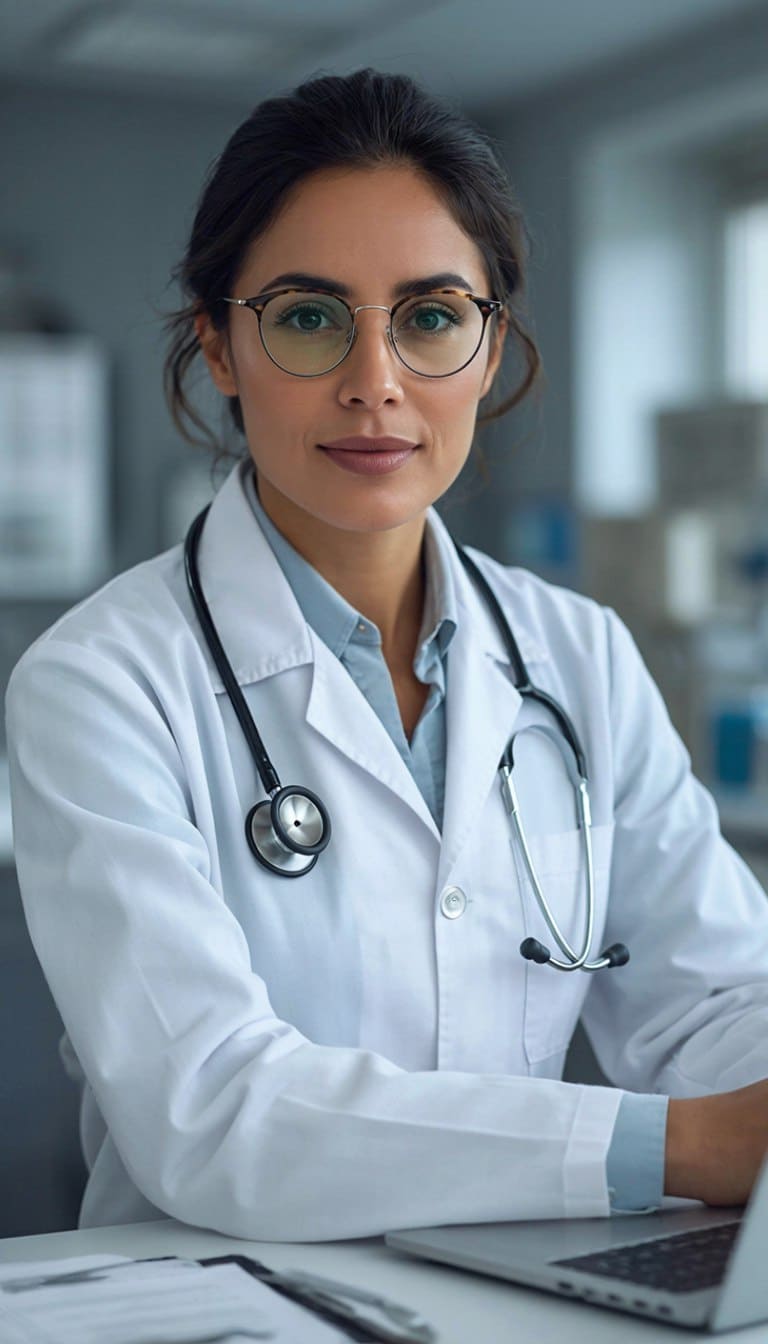}
        \end{subfigure}    
        \hfill
        \begin{subfigure}{0.32\textwidth}
            \centering
            \includegraphics[width=0.9\linewidth]{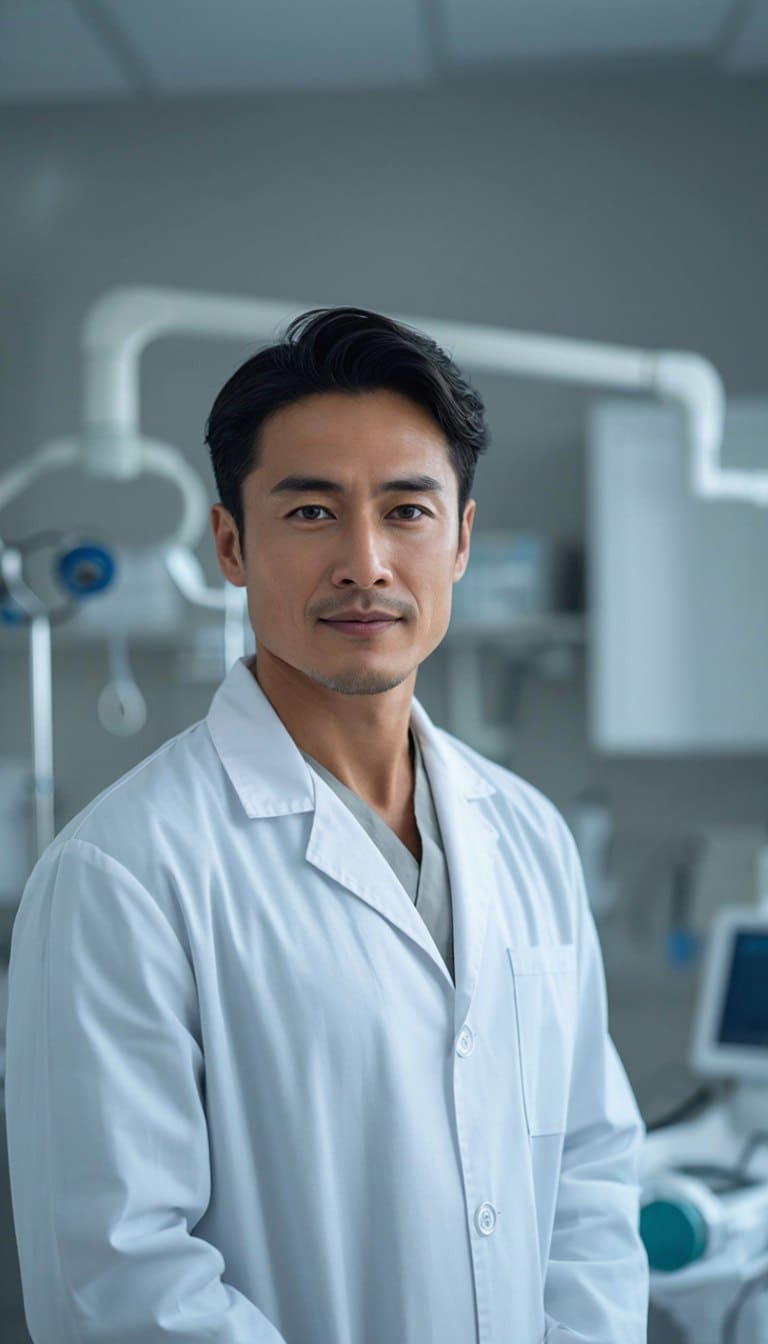}
        \end{subfigure}
        \caption{Prompt: ``A doctor''. The model demonstrates diversity in terms of gender and race.}
        \label{fig:diversity1}
    \end{minipage} \\
\end{figure}

Our open-source release of Kandinsky 5.0 is designed to democratize access to cutting-edge generative technology while promoting responsible AI development. In line with this goal, we are releasing the model code and training checkpoints under the permissive \textbf{MIT license}. We have deliberately chosen not to implement built-in content filtering systems, believing that this approach fosters innovation, advances the field of generative media, and encourages users to critically engage with the technology. However, this freedom comes with significant responsibility. We explicitly state that all users are liable for the content they generate and must use the model ethically and legally.

The following sections detail critical ethical considerations and limitations associated with the Kandinsky 5.0 model family.

\begin{enumerate}
    \item \textbf{Inherent Sociocultural Biases.} Like all large-scale models trained on web-scale datasets, Kandinsky 5.0 inherits and can amplify societal biases present in its training data.
    
    \begin{itemize}
        \item \textbf{Cultural Stereotypes:} The model's "knowledge" of the visual world is a reflection of the most prevalent data online, which can perpetuate harmful or inaccurate stereotypes. We strongly oppose the use of our model to generate content that promotes hatred or disparages any social group.

        \item \textbf{Representation and Diversity:} As shown in Figure~\ref{fig:diversity}, the model often reproduces common stereotypes in response to prompts related to profession, gender, age, or ethnicity (e.g., ``teacher'', ``beautiful man'', ``smart person'', ``Asian''). This results in a lack of diversity, reinforcing biases found in popular culture. Conversely, the model demonstrates a better understanding of diversity in other contexts (Figure~\ref{fig:diversity1}). To achieve more balanced and representative outputs, users are encouraged to employ specific, descriptive prompts that explicitly call for diversity.
    \end{itemize}

    \item \textbf{Technical Limitations and Unintended Outputs.} Users must be aware of the model's technical constraints to manage expectations and avoid unintended consequences.

    \begin{itemize}
        \item \textbf{Prompt Misunderstanding:} The model does not possess a deep, semantic understanding of language. An incomplete or ambiguous prompt can potentially lead to the generation of nonsensical, offensive, or otherwise undesirable content. A clear comprehension of the system's limitations is essential for safe and effective use.

        \item \textbf{Potential for Misuse:} The powerful capability to synthesize realistic images and video carries a inherent risk of malicious use. We explicitly prohibit the use of Kandinsky 5.0 for creating disinformation, harassing content, or any material that violates applicable laws. The primary responsibility for mitigating this risk and ensuring ethical application rests with the end-user.
    \end{itemize}
\end{enumerate}

By openly acknowledging these challenges, we aim to inform the community, stimulate discussion on mitigation strategies, and empower users to navigate the ethical landscape of generative AI responsibly.

\section[Conclusion]{Conclusion}\label{sec:conclusion}

In this report, we introduced \textbf{Kandinsky 5.0}, a versatile and scalable family of foundation models for high-resolution image and video generation. The framework includes three core model line-ups: \textbf{Kandinsky 5.0 Image Lite} (6B parameters), \textbf{Kandinsky 5.0 Video Lite} (2B parameters), and \textbf{Kandinsky 5.0 Video Pro} (19B parameters), each optimized for specific generative tasks and efficiency requirements.

Our key contributions include:

\begin{itemize}
    \item A \textbf{comprehensive data processing pipeline} that ensures high-quality, diverse, and culturally aware datasets for both image and video modalities, including specialized datasets for Russian cultural content and supervised fine-tuning.
    \item A \textbf{multi-stage training pipeline} incorporating pre-training, supervised fine-tuning (SFT), distillation, and RL-based post-training, which collectively enhance visual quality, prompt alignment, and temporal consistency.
    \item The introduction of the \textbf{CrossDIT architecture} and the \textbf{NABLA attention mechanism}, which significantly reduce computational complexity and accelerate training and inference for high-resolution and long-duration video generation.
    \item Extensive \textbf{optimizations} across the pipeline---including VAE acceleration, memory-efficient training, and inference enhancements---enabling state-of-the-art performance on consumer and professional hardware.
    \item A thorough \textbf{evaluation framework} based on human side-by-side (SBS) comparisons, demonstrating that Kandinsky 5.0 models achieve superior or competitive results against leading models such as Sora, Veo, and Wan across key metrics like visual quality, motion dynamics, and prompt adherence.
\end{itemize}

Kandinsky 5.0 sets a new milestone in open-source generative AI, offering:

\begin{itemize}
    \item High-fidelity \textbf{text-to-image} and \textbf{image editing} capabilities with strong aesthetic and compositional control.
    \item Robust \textbf{text-to-video} and \textbf{image-to-video} synthesis with support for up to 10-second clips at resolutions up to 1408p.
    \item Efficient \textbf{distilled variants} (Video Lite/Pro Flash) that maintain quality while drastically reducing inference time.
\end{itemize}

Despite these advances, we acknowledge limitations in areas such as \textbf{text-visual alignment}, \textbf{long-term temporal modeling}, and \textbf{generalization across all visual domains}. These challenges guide our ongoing research toward more unified, efficient, and ethically aligned generative models.

By open-sourcing our models, code, and training methodologies under the \textbf{MIT license}, we aim to foster innovation, collaboration, and responsible use within the global AI community. We believe Kandinsky 5.0 represents a significant step toward democratizing high-quality generative media and serves as a solid foundation for future developments in multimodal AI.
\newpage

\section[Contributors and Acknowledgments]{Contributors and Acknowledgments}\label{sec:contributors}

\subsection*{Contributors}

\textbf{Core Contributors:}
\begin{itemize}
    \item \textbf{Video:} Alexey Letunovskiy, Maria Kovaleva, Lev Novitskiy, Denis Koposov, Dmitrii Mikhailov, Anastasiia Kargapoltseva, Anna Dmitrienko, Anastasia Maltseva

    \item \textbf{Image \& Editing:} Nikolai Vaulin, Nikita Kiselev, Alexander Varlamov

    \item \textbf{Pre-training Data:} Ivan Kirillov, Andrey Shutkin, Nikolai Vaulin, Ilya Vasiliev

    \item \textbf{Post-training Data:} Julia Agafonova, Anna Averchenkova, Olga Kim

    \item \textbf{Research Consolidation \& Paper:} Viacheslav Vasilev, Vladimir Polovnikov
\end{itemize}

\paragraph{Contributors:} Yury Kolabushin, Kirill Chernyshev, Alexander Belykh, Mikhail Mamaev, Anastasia Aliaskina, Kormilitsyn Semen, Tatiana Nikulina, Olga Vdovchenko, Polina Mikhailova, Polina Gavrilova, Nikita Osterov, Bulat Akhmatov

\paragraph{Track Leaders:} Vladimir Arkhipkin, Vladimir Korviakov, Nikolai Gerasimenko, Denis Parkhomenko, Viacheslav Vasilev

\paragraph{Project Supervisor:} Denis Dimitrov

\subsection*{Acknowledgments}

We would like to thank the following people for their help, advices and assistance:

\paragraph{Prompt-engineering Team:} Denis Kondratev, Stefaniya Kozlova, Uliya Filippova, Alexandra Ugarova, Alexandra Averina, Irina Tabunova, Olga Nikiforova, Margarita Geberlein, Marina Yakushenko, Nadezhda Martynova, Mikhail Kornilin

\paragraph{GigaHub Team:} Andrey Evtikhov, Vladimir Yatulchik, Alexey Sandakov, Vyacheslav Fedotov, Igor Ivanov, Artem Kotenko, Yan Tomarovsky

\paragraph{TagMe Mark-up Team \& RnD Team:} Alexander Potemkin, Alexey Potapov, Dmitry Popov, Vasily Orlov, Victoria Wolf, Alexander Kapitanov, Sergey Markov

\paragraph{FusionBrain Lab:} Alexander Gambashidze, Konstantin Sobolev, Andrey Kuznetsov

\paragraph{GigaChat RnD, Prod \& B2C}: German Novikov, Serafima Bocharova, Roman Lebedev, Gennadii Khrenov, Nikita Savushkin, Fyodor Minkin

\paragraph{Cloud.ru:} Sergey Kovylov, Alexander Naumov, Alena Drobyshevskaia

\paragraph{Model Risk Management Team:} Stepan Ponomarev, Viktor Panshin, Vladislav Rodionov, Sergey Skachkov, Vladislav Veselov, Oleg Yangalichin, Artem Kostenko

\paragraph{Kandinsky Lab Service:} Alexey Bondarenko, Valeria Zdanova

\vspace{0.4cm}

as well as Maxim Eremenko, Andrey Belevtcev and Andrey Karlov.

\bibliographystyle{unsrt}
\bibliography{references}

@article{goodfellow2014gan,
author = {Goodfellow, Ian and Pouget-Abadie, Jean and Mirza, Mehdi and Xu, Bing and Warde-Farley, David and Ozair, Sherjil and Courville, Aaron and Bengio, Y.},
year = {2014},
month = {06},
pages = {},
title = {Generative Adversarial Networks},
volume = {3},
journal = {Advances in Neural Information Processing Systems},
doi = {10.1145/3422622}
}

@article{kingma2013vae,
  title={Auto-Encoding Variational Bayes},
  author={Kingma, Diederik P and Welling, Max},
  journal={arXiv:1312.6114},
  year={2013}
}

@article{oord2016pixelrnn,
  author = {Oord, Aaron and Kalchbrenner, Nal and Kavukcuoglu, Koray},
  month = {01},
  year = {2016},
  pages = {},
  title = {Pixel Recurrent Neural Networks},
  journal={ICML 2016, preprint arXiv:1601.06759},
  doi = {10.48550/arXiv.1601.06759}

}

@inproceedings{ho2020ddpm,
    author = {Ho, Jonathan and Jain, Ajay and Abbeel, Pieter},
    title = {Denoising diffusion probabilistic models},
    year = {2020},
    isbn = {9781713829546},
    publisher = {Curran Associates Inc.},
    address = {Red Hook, NY, USA},
    booktitle = {Proceedings of the 34th International Conference on Neural Information Processing Systems},
    articleno = {574},
    numpages = {12},
    location = {Vancouver, BC, Canada},
    series = {NIPS '20}
}

@article{dhariwal2021diffusion,
  title={Diffusion Models Beat GANs on Image Synthesis},
  author={Dhariwal, Prafulla and Nichol, Alex},
  journal={NeurIPS},
  year={2021}
}

@article{peebles2023dit,
  title={Scalable Diffusion Models with Transformers},
  author={Peebles, William and Xie, Saining},
  journal={ICCV},
  year={2023}
}

@misc{openai2024sora,
  title={Video generation models as world simulators},
  author={OpenAI},
  year={2024},
  url={https://openai.com/index/video-generation-models-as-world-simulators/},
  howpublished = "\url{https://openai.com/index/video-generation-models-as-world-simulators/}",
}

@misc{openai2024sora2,
  title={Sora 2 is here},
  author={OpenAI},
  year={2025},
  url={https://openai.com/index/sora-2/},
  howpublished = "\url{https://openai.com/index/sora-2/}",
}

@misc{activation_checkpointing,
  title={Current and New Activation Checkpointing Techniques in PyTorch},
  author={PyTorch Foundation},
  year={2025},
  url={https://pytorch.org/blog/activation-checkpointing-techniques/},
  howpublished = "\url{https://pytorch.org/blog/activation-checkpointing-techniques/}",
}

@article{wan2024,
  title={Wan: Efficient Video Generation Architecture},
  author={Wan, Li and others},
  journal={arXiv:2402.XXXXX},
  year={2024}
}

@article{rezende2014stochastic,
  author = {Jimenez Rezende, Danilo and Mohamed, Shakir and Wierstra, Daan},
  year = {2013},
  month = {12},
  pages = {},
  title = {Stochastic Backpropagation and Approximate Inference in Deep Generative Models},
  volume = {4},
  journal = {31st International Conference on Machine Learning, ICML 2014}
}

@article{chen2020generative,
  author={Mark Chen and Radford Alec and Rewon Child and Jeff Wu and Heewoo Jun and David Luan and Ilya Sutskever},
  title={Generative Pretraining from Pixels},
  journal={ICML},
  year={2020},
  url={https://openai.com/index/image-gpt/}
}

@misc{kong2025hunyuanvideosystematicframeworklarge,
      title={HunyuanVideo: A Systematic Framework For Large Video Generative Models}, 
      author={Weijie Kong and Qi Tian and Zijian Zhang and Rox Min and Zuozhuo Dai and Jin Zhou and Jiangfeng Xiong and Xin Li and Bo Wu and Jianwei Zhang and Kathrina Wu and Qin Lin and Junkun Yuan and Yanxin Long and Aladdin Wang and Andong Wang and Changlin Li and Duojun Huang and Fang Yang and Hao Tan and Hongmei Wang and Jacob Song and Jiawang Bai and Jianbing Wu and Jinbao Xue and Joey Wang and Kai Wang and Mengyang Liu and Pengyu Li and Shuai Li and Weiyan Wang and Wenqing Yu and Xinchi Deng and Yang Li and Yi Chen and Yutao Cui and Yuanbo Peng and Zhentao Yu and Zhiyu He and Zhiyong Xu and Zixiang Zhou and Zunnan Xu and Yangyu Tao and Qinglin Lu and Songtao Liu and Dax Zhou and Hongfa Wang and Yong Yang and Di Wang and Yuhong Liu and Jie Jiang and Caesar Zhong},
      year={2025},
      eprint={2412.03603},
      archivePrefix={arXiv},
      primaryClass={cs.CV},
      url={https://arxiv.org/abs/2412.03603}, 
}

@misc{genmo2024,
  title={Mochi Video Generation Framework},
  author={Genmo AI},
  year={2024},
  url={https://genmo.ai}
}

@article{yang2025cogvideox,
  title={CogVideoX: Text-to-Video Diffusion Models with An Expert Transformer}, 
  author={Zhuoyi Yang and Jiayan Teng and Wendi Zheng and Ming Ding and Shiyu Huang and Jiazheng Xu and Yuanming Yang and Wenyi Hong and Xiaohan Zhang and Guanyu Feng and Da Yin and Yuxuan Zhang and Weihan Wang and Yean Cheng and Bin Xu and Xiaotao Gu and Yuxiao Dong and Jie Tang},
  year={2025},
  eprint={2408.06072},
  archivePrefix={arXiv},
  primaryClass={cs.CV},
  url={https://arxiv.org/abs/2408.06072},
  journal={arXiv:2408.06072}
}

@article{wang2022videocontrolnet,
  title={Video ControlNet: Towards Temporally Consistent Synthetic Videos},
  author={Wang, Yaohui and others},
  journal={arXiv:2212.04403},
  year={2022}
}

@article{chen2023v2a,
  title={SyncTalk: Audio-Visual Synthesis for Dynamic Talking Heads},
  author={Chen, Lele and others},
  journal={ACM TOG},
  year={2023}
}

@article{reda2022frame,
  title={Frame Interpolation with Diffusion Models},
  author={Reda, Fitsum and others},
  journal={NeurIPS},
  year={2022}
}

@article{brooks2023cameractrl,
  title={Camera Control for Video Generation},
  author={Brooks, Tim and others},
  journal={SIGGRAPH},
  year={2023}
}

@article{wu2023videovae,
  title={Efficient Video Latent Diffusion Models},
  author={Wu, Jay and others},
  journal={ICLR},
  year={2023}
}

@inproceedings{dao2022flash,
  title={Flash{A}ttention: Fast and Memory-Efficient Exact Attention with {IO}-Awareness},
  author={Dao, Tri and Fu, Daniel Y. and Ermon, Stefano and Rudra, Atri and R{\'e}, Christopher},
  booktitle={Advances in Neural Information Processing Systems (NeurIPS)},
  year={2022}
}

@inproceedings{dao2023flash2,
  title={Flash{A}ttention-2: Faster Attention with Better Parallelism and Work Partitioning},
  author={Dao, Tri},
  booktitle={International Conference on Learning Representations (ICLR)},
  year={2024}
}

@article{hu2021lora,
  title={LoRA: Low-Rank Adaptation of Large Language Models},
  author={Hu, Edward J. and others},
  journal={ICLR},
  year={2022}
}

@misc{zhang2024fast,
      title={Fast Video Generation with Sliding Tile Attention}, 
      author={Peiyuan Zhang and Yongqi Chen and Runlong Su and Hangliang Ding and Ion Stoica and Zhenghong Liu and Hao Zhang},
      year={2025},
      eprint={2502.04507},
      archivePrefix={arXiv},
      primaryClass={cs.CV},
      url={https://arxiv.org/abs/2502.04507}, 
}

@article{schuhmann2022laion,
  author  = {Schuhmann, Christoph and Beaumont, Romain and Vencu, Robert and Gordon, Cade and Wightman, Ross and Cherti, Mehdi and Coombes, Theo and Katta, Aarush and Mullis, Clayton and Wortsman, Mitchell and others},
  title   = {{LAION-5B: An open large-scale dataset for training next generation image-text models}},
  journal = {arXiv preprint arXiv:2210.08402},
  year    = {2022}
}

@mastersthesis{zauner2010perceptualhashing,
  author  = {Zauner, Christoph},
  title   = {{Implementation and benchmarking of perceptual image hash functions}},
  school  = {Upper Austria University of Applied Sciences},
  year    = {2010}
}

@article{tu2023topiq,
  author  = {Tu, Zhengzhong and Wang, Yilin and Birkbeck, Neil and Adsumilli, Balu and Bovik, Alan C.},
  title   = {{TOPIQ: A Top-Down Approach from Semantics to Distortions for Image Quality Assessment}},
  journal = {arXiv preprint arXiv:2308.03060},
  year    = {2023}
}

@article{wu2023qalign,
  author  = {Wu, Yuhao and Zhang, Wei and Li, Weixia and Li, Lei and Li, Guangtao and Shan, Ying},
  title   = {{Q-Align: Teaching LMMs for Visual Scoring via Discrete Text-Defined Levels}},
  journal = {arXiv preprint arXiv:2312.17090},
  year    = {2023}
}

@inproceedings{baek2019craft,
  author    = {Baek, Youngmin and Lee, Bado and Han, Dongyoon and Yun, Sangdoo and Lee, Hwalsuk},
  title     = {{Character region awareness for text detection}},
  booktitle = {Proceedings of the IEEE/CVF Conference on Computer Vision and Pattern Recognition},
  pages     = {9365--9374},
  year      = {2019}
}

@article{kuznetsova2020openimages,
  author  = {Kuznetsova, Alina and Rom, Hassan and Alldrin, Neil and Uijlings, Jasper and Krasin, Ivan and Pont-Tuset, Jordi and Kamali, Shahab and Popov, Stefan and Malloci, Matteo and Kolesnikov, Alexander and others},
  title   = {{The Open Images Dataset V4: Unified image classification, object detection, and visual relationship detection at scale}},
  journal = {International Journal of Computer Vision},
  volume  = {128},
  number  = {7},
  pages   = {1956--1981},
  year    = {2020}
}

@inproceedings{radford2021clip,
  author    = {Radford, Alec and Kim, Jong Wook and Hallacy, Chris and Ramesh, Aditya and Goh, Gabriel and Agarwal, Sandhini and Sastry, Girish and Askell, Amanda and Mishkin, Pamela and Clark, Jack and others},
  title     = {{Learning transferable visual models from natural language supervision}},
  booktitle = {International Conference on Machine Learning},
  pages     = {8748--8763},
  year      = {2021},
  publisher = {PMLR}
}

@misc{chen2025expandingperformanceboundariesopensource,
      title={Expanding Performance Boundaries of Open-Source Multimodal Models with Model, Data, and Test-Time Scaling}, 
      author={Zhe Chen and Weiyun Wang and Yue Cao and Yangzhou Liu and Zhangwei Gao and Erfei Cui and Jinguo Zhu and Shenglong Ye and Hao Tian and Zhaoyang Liu and Lixin Gu and Xuehui Wang and Qingyun Li and Yimin Ren and Zixuan Chen and Jiapeng Luo and Jiahao Wang and Tan Jiang and Bo Wang and Conghui He and Botian Shi and Xingcheng Zhang and Han Lv and Yi Wang and Wenqi Shao and Pei Chu and Zhongying Tu and Tong He and Zhiyong Wu and Huipeng Deng and Jiaye Ge and Kai Chen and Kaipeng Zhang and Limin Wang and Min Dou and Lewei Lu and Xizhou Zhu and Tong Lu and Dahua Lin and Yu Qiao and Jifeng Dai and Wenhai Wang},
      year={2025},
      eprint={2412.05271},
      archivePrefix={arXiv},
      primaryClass={cs.CV},
      url={https://arxiv.org/abs/2412.05271}, 
}

@article{oquab2023dinov2,
  author  = {Oquab, Maxime and Darcet, Timothée and Moutakanni, Theo and Vo, Huy and Szafraniec, Marc and Khalidov, Vasil and Fernandez, Pierre and Haziza, Daniel and Massa, Francisco and El-Nouby, Alaaeldin and others},
  title   = {{DINOv2: Learning Robust Visual Features without Supervision}},
  journal = {arXiv preprint arXiv:2304.07193},
  year    = {2023}
}

@inproceedings{sun2021loftr,
  author    = {Sun, Jiaming and Shen, Ze and Wang, Yuang and Bao, Hujun and Zhou, Xiaowei},
  title     = {{LoFTR: Detector-Free Local Feature Matching with Transformers}},
  booktitle = {Proceedings of the IEEE/CVF Conference on Computer Vision and Pattern Recognition},
  pages     = {8922--8931},
  year      = {2021}
}

@article{fischler1981ransac,
  author  = {Fischler, Martin A. and Bolles, Robert C.},
  title   = {{Random sample consensus: a paradigm for model fitting with applications to image analysis and automated cartography}},
  journal = {Communications of the ACM},
  volume  = {24},
  number  = {6},
  pages   = {381--395},
  year    = {1981}
}

@article{zeng2022glm,
  author  = {Zeng, Aohan and Liu, Xiao and Du, Zhengxiao and Wang, Zihan and Lai, Hanyu and Ding, Ming and Yang, Zhuoyi and Xu, Yifan and Zheng, Wendi and Xia, Xiao and others},
  title   = {{GLM-130B: An Open Bilingual Pre-trained Model}},
  journal = {arXiv preprint arXiv:2210.02414},
  year    = {2022}
}

@inproceedings{lawto2007videoperceptualhashing,
  author    = {Law-To, J. and Chen, L. and Joly, A. and Laptev, I. and Buisson, O. and Gouet-Brunet, V. and Boujemaa, N.},
  title     = {{Video copy detection: a comparative study}},
  booktitle = {Proceedings of the 6th ACM international conference on Image and video retrieval},
  pages     = {371--378},
  year      = {2007}
}

@inproceedings{wang2003msssim,
  author    = {Wang, Zhou and Simoncelli, Eero P. and Bovik, Alan C.},
  title     = {{Multi-scale structural similarity for image quality assessment}},
  booktitle = {Signals, Systems and Computers, 2004. Conference Record of the Thirty-Seventh Asilomar Conference on},
  volume    = {2},
  pages     = {1398--1402},
  year      = {2003},
  publisher = {IEEE}
}

@inproceedings{wu2023exploring,
  title={Exploring video quality assessment on user generated contents from aesthetic and technical perspectives},
  author={Wu, Haoning and Zhang, Erli and Liao, Liang and Chen, Chaofeng and Hou, Jingwen and Wang, Annan and Sun, Wenxiu and Yan, Qiong and Lin, Weisi},
  booktitle={Proceedings of the IEEE/CVF International Conference on Computer Vision},
  pages={20144--20154},
  year={2023}
}

@article{tong2023videomae,
  author  = {Tong, Zhan and Song, Yibing and Wang, Jue and Wang, Limin},
  title   = {{VideoMAE V2: Scaling Video Masked Autoencoders with Dual Masking}},
  journal = {arXiv preprint arXiv:2303.16727},
  year    = {2023}
}

@article{lloyd1982kmeans,
  author  = {Lloyd, Stuart},
  title   = {{Least squares quantization in PCM}},
  journal = {IEEE transactions on information theory},
  volume  = {28},
  number  = {2},
  pages   = {129--137},
  year    = {1982}
}

@inproceedings{wortsman2022modelsoup,
  author    = {Wortsman, Mitchell and Ilharco, Gabriel and Gadre, Samir Ya and Roelofs, Rebecca and Gontijo-Lopes, Raphael and Morcos, Ari S. and Namkoong, Hongseok and Farhadi, Ali and Carmon, Yair and Kornblith, Simon and others},
  title     = {{Model soups: averaging weights of multiple fine-tuned models improves accuracy without increasing inference time}},
  booktitle = {International Conference on Machine Learning},
  pages     = {23965--23998},
  year      = {2022},
  publisher = {PMLR}
}

@misc{cai2024internlm2technicalreport,
      title={InternLM2 Technical Report}, 
      author={Zheng Cai and Maosong Cao and Haojiong Chen and Kai Chen and Keyu Chen and Xin Chen and Xun Chen and Zehui Chen and Zhi Chen and Pei Chu and Xiaoyi Dong and Haodong Duan and Qi Fan and Zhaoye Fei and Yang Gao and Jiaye Ge and Chenya Gu and Yuzhe Gu and Tao Gui and Aijia Guo and Qipeng Guo and Conghui He and Yingfan Hu and Ting Huang and Tao Jiang and Penglong Jiao and Zhenjiang Jin and Zhikai Lei and Jiaxing Li and Jingwen Li and Linyang Li and Shuaibin Li and Wei Li and Yining Li and Hongwei Liu and Jiangning Liu and Jiawei Hong and Kaiwen Liu and Kuikun Liu and Xiaoran Liu and Chengqi Lv and Haijun Lv and Kai Lv and Li Ma and Runyuan Ma and Zerun Ma and Wenchang Ning and Linke Ouyang and Jiantao Qiu and Yuan Qu and Fukai Shang and Yunfan Shao and Demin Song and Zifan Song and Zhihao Sui and Peng Sun and Yu Sun and Huanze Tang and Bin Wang and Guoteng Wang and Jiaqi Wang and Jiayu Wang and Rui Wang and Yudong Wang and Ziyi Wang and Xingjian Wei and Qizhen Weng and Fan Wu and Yingtong Xiong and Chao Xu and Ruiliang Xu and Hang Yan and Yirong Yan and Xiaogui Yang and Haochen Ye and Huaiyuan Ying and Jia Yu and Jing Yu and Yuhang Zang and Chuyu Zhang and Li Zhang and Pan Zhang and Peng Zhang and Ruijie Zhang and Shuo Zhang and Songyang Zhang and Wenjian Zhang and Wenwei Zhang and Xingcheng Zhang and Xinyue Zhang and Hui Zhao and Qian Zhao and Xiaomeng Zhao and Fengzhe Zhou and Zaida Zhou and Jingming Zhuo and Yicheng Zou and Xipeng Qiu and Yu Qiao and Dahua Lin},
      year={2024},
      eprint={2403.17297},
      archivePrefix={arXiv},
      primaryClass={cs.CL},
      url={https://arxiv.org/abs/2403.17297}, 
}

@article{ravi2024sam2,
  title={SAM 2: Segment Anything in Images and Videos},
  author={Ravi, Nikhila and Gabeur, Valentin and Hu, Yuan-Ting and Hu, Ronghang and Ryali, Chaitanya and Ma, Tengyu and Khedr, Haitham and R{\"a}dle, Roman and Rolland, Chloe and Gustafson, Laura and Mintun, Eric and Pan, Junting and Alwala, Kalyan Vasudev and Carion, Nicolas and Wu, Chao-Yuan and Girshick, Ross and Doll{\'a}r, Piotr and Feichtenhofer, Christoph},
  journal={arXiv preprint arXiv:2408.00714},
  url={https://arxiv.org/abs/2408.00714},
  year={2024}
}

@inproceedings{huang2024vbench,
  title     = {{VBench}: Comprehensive Benchmark Suite for Video Generative Models},
  author    = {Huang, Ziqi and He, Yinan and Yu, Jiashuo and Zhang, Fan and Si, Chenyang and Jiang, Yuming and Zhang, Yuanhan and Wu, Tianxing and Jin, Qingyang and Chanpaisit, Nattapol and others},
  booktitle = {Proceedings of the IEEE/CVF Conference on Computer Vision and Pattern Recognition (CVPR)},
  pages     = {21807--21818},
  year      = {2024},
  url       = {https://openaccess.thecvf.com/content/CVPR2024/html/Huang_VBench_Comprehensive_Benchmark_Suite_for_Video_Generative_Models_CVPR_2024_paper.html}
}

@inproceedings{hessel2021clipscore,
  title={{CLIPScore}: A Reference-free Evaluation Metric for Image Captioning and Text-to-Image Synthesis},
  author={Hessel, Jack and Holtzman, Ari and Forbes, Maxwell and Bras, Ronan Le and Choi, Yejin},
  booktitle={Proceedings of the 2021 Conference on Empirical Methods in Natural Language Processing (EMNLP)},
  pages={6192--6204},
  year={2021}
}

@article{HeuselRUNKH17,
  author       = {Martin Heusel and
                  Hubert Ramsauer and
                  Thomas Unterthiner and
                  Bernhard Nessler and
                  G{\"{u}}nter Klambauer and
                  Sepp Hochreiter},
  title        = {GANs Trained by a Two Time-Scale Update Rule Converge to a Nash Equilibrium},
  journal      = {CoRR},
  volume       = {abs/1706.08500},
  year         = {2017},
  url          = {http://arxiv.org/abs/1706.08500},
  eprinttype    = {arXiv},
  eprint       = {1706.08500},
  bibsource    = {dblp computer science bibliography, https://dblp.org}
}

@misc{bińkowski2021demystifyingmmdgans,
      title={Demystifying MMD GANs}, 
      author={Mikołaj Bińkowski and Danica J. Sutherland and Michael Arbel and Arthur Gretton},
      year={2021},
      eprint={1801.01401},
      archivePrefix={arXiv},
      primaryClass={stat.ML},
      url={https://arxiv.org/abs/1801.01401}, 
}

@misc{unterthiner2019accurategenerativemodelsvideo,
      title={Towards Accurate Generative Models of Video: A New Metric \& Challenges}, 
      author={Thomas Unterthiner and Sjoerd van Steenkiste and Karol Kurach and Raphael Marinier and Marcin Michalski and Sylvain Gelly},
      year={2019},
      eprint={1812.01717},
      archivePrefix={arXiv},
      primaryClass={cs.CV},
      url={https://arxiv.org/abs/1812.01717}, 
}

@inproceedings{wang2024internvideo2,
  title={Internvideo2: Scaling foundation models for multimodal video understanding},
  author={Wang, Yi and Li, Kunchang and Li, Xinhao and Yu, Jiashuo and He, Yinan and Chen, Guo and Pei, Baoqi and Zheng, Rongkun and Wang, Zun and Shi, Yansong and others},
  booktitle={European Conference on Computer Vision},
  pages={396--416},
  year={2024},
  organization={Springer}
}

@article{huang2025t2icompbenchpp,
  title={{T2I-CompBench++}: An Enhanced and Comprehensive Benchmark for Compositional Text-to-Image Generation},
  author={Huang, Kaiyi and Duan, Chengqi and Sun, Kaiyue and Xie, Enze and Li, Zhenguo and Liu, Xihui},
  journal={IEEE Transactions on Pattern Analysis and Machine Intelligence},
  volume={47},
  number={5},
  pages={3563--3579},
  year={2025}
}

@misc{sun2025t2vcompbenchcomprehensivebenchmarkcompositional,
      title={T2V-CompBench: A Comprehensive Benchmark for Compositional Text-to-video Generation}, 
      author={Kaiyue Sun and Kaiyi Huang and Xian Liu and Yue Wu and Zihan Xu and Zhenguo Li and Xihui Liu},
      year={2025},
      eprint={2407.14505},
      archivePrefix={arXiv},
      primaryClass={cs.CV},
      url={https://arxiv.org/abs/2407.14505}, 
}

@misc{liao2024evaluationtexttovideogenerationmodels,
      title={Evaluation of Text-to-Video Generation Models: A Dynamics Perspective}, 
      author={Mingxiang Liao and Hannan Lu and Xinyu Zhang and Fang Wan and Tianyu Wang and Yuzhong Zhao and Wangmeng Zuo and Qixiang Ye and Jingdong Wang},
      year={2024},
      eprint={2407.01094},
      archivePrefix={arXiv},
      primaryClass={cs.CV},
      url={https://arxiv.org/abs/2407.01094}, 
}

@article{polyak2024movie,
  title={Movie Gen: A Cast of Media Foundation Models},
  author={Polyak, Adam and Zohar, Amit and Brown, Andrew and Tjandra, Andros and Sinha, Animesh and Lee, Ann and Vyas, Apoorv and Shi, Bowen and Ma, Chih-Yao and Chuang, Ching-Yao and others},
  journal={arXiv preprint arXiv:2410.13720},
  year={2024}
}

@ARTICLE{mikhailov2025nablanablaneighborhoodadaptiveblocklevel,
  author={Mikhailov, Dmitrii and Letunovskiy, Aleksey and Kovaleva, Maria and Arkhipkin, Vladimir and Korviakov, Vladimir and Polovnikov, Vladimir and Vasilev, Viacheslav and Sidorova, Evelina and Dimitrov, Denis},
  journal={IEEE Access}, 
  title={NABLA: Neighborhood Adaptive Block-Level Attention for Efficient Video Generation}, 
  year={2026},
  volume={14},
  number={},
  pages={64655-64665},
  doi={10.1109/ACCESS.2026.3686867}
}

@misc{kandinskyvideotools2025,
  author = {Denis Koposov and Anna Dmitrienko and Ivan Kirillov and Kirill Chernyshev and Denis Parkhomenko and Vladimir Korviakov},
  title = {Kandinsky Video Tools},
  year = {2025},
  publisher = {GitHub},
  journal = {GitHub repository},
  howpublished = {\url{https://github.com/gen-ai-team/kandinsky-video-tools}},
  url={https://github.com/gen-ai-team/kandinsky-video-tools}
}

@inproceedings{song2021scorebased,
  title={Score-Based Generative Modeling through Stochastic Differential Equations},
  author={Yang Song and Jascha Sohl-Dickstein and Diederik P Kingma and Abhishek Kumar and Stefano Ermon and Ben Poole},
  booktitle={International Conference on Learning Representations},
  year={2021},
  url={https://openreview.net/forum?id=PxTIG12RRHS}
}

@inproceedings{lipman2023flow,
    title={Flow Matching for Generative Modeling},
    author={Yaron Lipman and Ricky T. Q. Chen and Heli Ben-Hamu and Maximilian Nickel and Matthew Le},
    booktitle={The Eleventh International Conference on Learning Representations },
    year={2023},
    url={https://openreview.net/forum?id=PqvMRDCJT9t}
}

@InProceedings{Rombach_2022_CVPR,
    author    = {Rombach, Robin and Blattmann, Andreas and Lorenz, Dominik and Esser, Patrick and Ommer, Bj\"orn},
    title     = {High-Resolution Image Synthesis With Latent Diffusion Models},
    booktitle = {Proceedings of the IEEE/CVF Conference on Computer Vision and Pattern Recognition (CVPR)},
    month     = {June},
    year      = {2022},
    pages     = {10684-10695}
}

@inproceedings{vladimir-etal-2024-kandinsky,
    title = "Kandinsky 3: Text-to-Image Synthesis for Multifunctional Generative Framework",
    author = "Vladimir Arkhipkin  and
      Vasilev Viacheslav  and
      Filatov Andrei  and
      Pavlov Igor  and
      Agafonova Julia  and
      Gerasimenko Nikolai  and
      Averchenkova Anna  and
      Mironova Evelina  and
      Anton Bukashkin  and
      Kulikov Konstantin  and
      Kuznetsov Andrey  and
      Dimitrov Denis",
    editor = "Hernandez Farias, Delia Irazu  and
      Hope, Tom  and
      Li, Manling",
    booktitle = "Proceedings of the 2024 Conference on Empirical Methods in Natural Language Processing: System Demonstrations",
    month = nov,
    year = "2024",
    address = "Miami, Florida, USA",
    publisher = "Association for Computational Linguistics",
    url = "https://aclanthology.org/2024.emnlp-demo.48/",
    doi = "10.18653/v1/2024.emnlp-demo.48",
    pages = "475--485",
}

@misc{Midjourney,
  title = {Midjourney},
  author={Midjourney},
  howpublished = {\url{https://www.midjourney.com/}},
  year={2022}
}

@misc{pika,
  title = {Pika},
  author={Pika},
  howpublished = {\url{https://pika.art/}},
  year={2023}
}

@misc{esser2024scalingrectifiedflowtransformers,
      title={Scaling Rectified Flow Transformers for High-Resolution Image Synthesis}, 
      author={Patrick Esser and Sumith Kulal and Andreas Blattmann and Rahim Entezari and Jonas Müller and Harry Saini and Yam Levi and Dominik Lorenz and Axel Sauer and Frederic Boesel and Dustin Podell and Tim Dockhorn and Zion English and Kyle Lacey and Alex Goodwin and Yannik Marek and Robin Rombach},
      year={2024},
      eprint={2403.03206},
      archivePrefix={arXiv},
      primaryClass={cs.CV},
      url={https://arxiv.org/abs/2403.03206}, 
}

@misc{flux2024,
    author={Black Forest Labs},
    title={FLUX},
    year={2024},
    howpublished={\url{https://github.com/black-forest-labs/flux}},
}

@misc{gao2025seedream30technicalreport,
      title={Seedream 3.0 Technical Report}, 
      author={Yu Gao and Lixue Gong and Qiushan Guo and Xiaoxia Hou and Zhichao Lai and Fanshi Li and Liang Li and Xiaochen Lian and Chao Liao and Liyang Liu and Wei Liu and Yichun Shi and Shiqi Sun and Yu Tian and Zhi Tian and Peng Wang and Rui Wang and Xuanda Wang and Xun Wang and Ye Wang and Guofeng Wu and Jie Wu and Xin Xia and Xuefeng Xiao and Zhonghua Zhai and Xinyu Zhang and Qi Zhang and Yuwei Zhang and Shijia Zhao and Jianchao Yang and Weilin Huang},
      year={2025},
      eprint={2504.11346},
      archivePrefix={arXiv},
      primaryClass={cs.CV},
      url={https://arxiv.org/abs/2504.11346}, 
}

@misc{seedream2025seedream40nextgenerationmultimodal,
      title={Seedream 4.0: Toward Next-generation Multimodal Image Generation}, 
      author={Team Seedream and : and Yunpeng Chen and Yu Gao and Lixue Gong and Meng Guo and Qiushan Guo and Zhiyao Guo and Xiaoxia Hou and Weilin Huang and Yixuan Huang and Xiaowen Jian and Huafeng Kuang and Zhichao Lai and Fanshi Li and Liang Li and Xiaochen Lian and Chao Liao and Liyang Liu and Wei Liu and Yanzuo Lu and Zhengxiong Luo and Tongtong Ou and Guang Shi and Yichun Shi and Shiqi Sun and Yu Tian and Zhi Tian and Peng Wang and Rui Wang and Xun Wang and Ye Wang and Guofeng Wu and Jie Wu and Wenxu Wu and Yonghui Wu and Xin Xia and Xuefeng Xiao and Shuang Xu and Xin Yan and Ceyuan Yang and Jianchao Yang and Zhonghua Zhai and Chenlin Zhang and Heng Zhang and Qi Zhang and Xinyu Zhang and Yuwei Zhang and Shijia Zhao and Wenliang Zhao and Wenjia Zhu},
      year={2025},
      eprint={2509.20427},
      archivePrefix={arXiv},
      primaryClass={cs.CV},
      url={https://arxiv.org/abs/2509.20427}, 
}

@article{cao2025hunyuanimage,
  title={HunyuanImage 3.0 Technical Report},
  author={Cao, Siyu and Chen, Hangting and Chen, Peng and Cheng, Yiji and Cui, Yutao and Deng, Xinchi and Dong, Ying and Gong, Kipper and Gu, Tianpeng and Gu, Xiusen and others},
  journal={arXiv preprint arXiv:2509.23951},
  year={2025}
}

@misc{ho2022imagenvideohighdefinition,
      title={Imagen Video: High Definition Video Generation with Diffusion Models}, 
      author={Jonathan Ho and William Chan and Chitwan Saharia and Jay Whang and Ruiqi Gao and Alexey Gritsenko and Diederik P. Kingma and Ben Poole and Mohammad Norouzi and David J. Fleet and Tim Salimans},
      year={2022},
      eprint={2210.02303},
      archivePrefix={arXiv},
      primaryClass={cs.CV},
      url={https://arxiv.org/abs/2210.02303}, 
}

@misc{arkhipkin2023fusionframesefficientarchitecturalaspects,
      title={FusionFrames: Efficient Architectural Aspects for Text-to-Video Generation Pipeline}, 
      author={Vladimir Arkhipkin and Zein Shaheen and Viacheslav Vasilev and Elizaveta Dakhova and Andrey Kuznetsov and Denis Dimitrov},
      year={2023},
      eprint={2311.13073},
      archivePrefix={arXiv},
      primaryClass={cs.CV},
      url={https://arxiv.org/abs/2311.13073}, 
}

@ARTICLE{10815947,
  author={Arkhipkin, Vladimir and Shaheen, Zein and Vasilev, Viacheslav and Dakhova, Elizaveta and Sobolev, Konstantin and Kuznetsov, Andrey and Dimitrov, Denis},
  journal={IEEE Access}, 
  title={ImproveYourVideos: Architectural Improvements for Text-to-Video Generation Pipeline}, 
  year={2025},
  volume={13},
  number={},
  pages={1986-2003},
  doi={10.1109/ACCESS.2024.3522510}}

@INPROCEEDINGS{blattmann2023align,
  author={Blattmann, Andreas and Rombach, Robin and Ling, Huan and Dockhorn, Tim and Kim, Seung Wook and Fidler, Sanja and Kreis, Karsten},
  booktitle={2023 IEEE/CVF Conference on Computer Vision and Pattern Recognition (CVPR)}, 
  title={Align Your Latents: High-Resolution Video Synthesis with Latent Diffusion Models}, 
  year={2023},
  volume={},
  number={},
  pages={22563-22575},
  doi={10.1109/CVPR52729.2023.02161}
}

@article{ma2025latte,
        title={Latte: Latent Diffusion Transformer for Video Generation},
        author={Ma, Xin and Wang, Yaohui and Chen, Xinyuan and Jia, Gengyun and Liu, Ziwei and Li, Yuan-Fang and Chen, Cunjian and Qiao, Yu},
        journal={Transactions on Machine Learning Research},
        year={2025}
      }

@misc{chen2023pixartalpha,
    title={PixArt-$\alpha$: Fast Training of Diffusion Transformer for Photorealistic Text-to-Image Synthesis}, 
    author={Junsong Chen and Jincheng Yu and Chongjian Ge and Lewei Yao and Enze Xie and Yue Wu and Zhongdao Wang and James Kwok and Ping Luo and Huchuan Lu and Zhenguo Li},
    year={2023},
    eprint={2310.00426},
    archivePrefix={arXiv},
    primaryClass={cs.CV}
}

@misc{xi2025sparsevideogenacceleratingvideo,
      title={Sparse VideoGen: Accelerating Video Diffusion Transformers with Spatial-Temporal Sparsity}, 
      author={Haocheng Xi and Shuo Yang and Yilong Zhao and Chenfeng Xu and Muyang Li and Xiuyu Li and Yujun Lin and Han Cai and Jintao Zhang and Dacheng Li and Jianfei Chen and Ion Stoica and Kurt Keutzer and Song Han},
      year={2025},
      eprint={2502.01776},
      archivePrefix={arXiv},
      primaryClass={cs.CV},
      url={https://arxiv.org/abs/2502.01776}, 
}

@misc{xia2025trainingfreeadaptivesparseattention,
      title={Training-free and Adaptive Sparse Attention for Efficient Long Video Generation}, 
      author={Yifei Xia and Suhan Ling and Fangcheng Fu and Yujie Wang and Huixia Li and Xuefeng Xiao and Bin Cui},
      year={2025},
      eprint={2502.21079},
      archivePrefix={arXiv},
      primaryClass={cs.CV},
      url={https://arxiv.org/abs/2502.21079}, 
}

@misc{Veo,
  title = {Veo},
  author={Google DeepMind},
  howpublished = {\url{https://deepmind.google/models/veo/}},
  year={2025}
}

@article{wan2025,
      title={Wan: Open and Advanced Large-Scale Video Generative Models}, 
      author={Team Wan and Ang Wang and Baole Ai and Bin Wen and Chaojie Mao and Chen-Wei Xie and Di Chen and Feiwu Yu and Haiming Zhao and Jianxiao Yang and Jianyuan Zeng and Jiayu Wang and Jingfeng Zhang and Jingren Zhou and Jinkai Wang and Jixuan Chen and Kai Zhu and Kang Zhao and Keyu Yan and Lianghua Huang and Mengyang Feng and Ningyi Zhang and Pandeng Li and Pingyu Wu and Ruihang Chu and Ruili Feng and Shiwei Zhang and Siyang Sun and Tao Fang and Tianxing Wang and Tianyi Gui and Tingyu Weng and Tong Shen and Wei Lin and Wei Wang and Wei Wang and Wenmeng Zhou and Wente Wang and Wenting Shen and Wenyuan Yu and Xianzhong Shi and Xiaoming Huang and Xin Xu and Yan Kou and Yangyu Lv and Yifei Li and Yijing Liu and Yiming Wang and Yingya Zhang and Yitong Huang and Yong Li and You Wu and Yu Liu and Yulin Pan and Yun Zheng and Yuntao Hong and Yupeng Shi and Yutong Feng and Zeyinzi Jiang and Zhen Han and Zhi-Fan Wu and Ziyu Liu},
      journal = {arXiv preprint arXiv:2503.20314},
      year={2025}
}

@inproceedings{vace,
    title = {VACE: All-in-One Video Creation and Editing},
    author = {Jiang, Zeyinzi and Han, Zhen and Mao, Chaojie and Zhang, Jingfeng and Pan, Yulin and Liu, Yu},
    booktitle = {Proceedings of the IEEE/CVF International Conference on Computer Vision},
    pages = {17191-17202},
    year = {2025}
}

@misc{polyak2025moviegencastmedia,
      title={Movie Gen: A Cast of Media Foundation Models}, 
      author={Adam Polyak and Amit Zohar and Andrew Brown and Andros Tjandra and Animesh Sinha and Ann Lee and Apoorv Vyas and Bowen Shi and Chih-Yao Ma and Ching-Yao Chuang and David Yan and Dhruv Choudhary and Dingkang Wang and Geet Sethi and Guan Pang and Haoyu Ma and Ishan Misra and Ji Hou and Jialiang Wang and Kiran Jagadeesh and Kunpeng Li and Luxin Zhang and Mannat Singh and Mary Williamson and Matt Le and Matthew Yu and Mitesh Kumar Singh and Peizhao Zhang and Peter Vajda and Quentin Duval and Rohit Girdhar and Roshan Sumbaly and Sai Saketh Rambhatla and Sam Tsai and Samaneh Azadi and Samyak Datta and Sanyuan Chen and Sean Bell and Sharadh Ramaswamy and Shelly Sheynin and Siddharth Bhattacharya and Simran Motwani and Tao Xu and Tianhe Li and Tingbo Hou and Wei-Ning Hsu and Xi Yin and Xiaoliang Dai and Yaniv Taigman and Yaqiao Luo and Yen-Cheng Liu and Yi-Chiao Wu and Yue Zhao and Yuval Kirstain and Zecheng He and Zijian He and Albert Pumarola and Ali Thabet and Artsiom Sanakoyeu and Arun Mallya and Baishan Guo and Boris Araya and Breena Kerr and Carleigh Wood and Ce Liu and Cen Peng and Dimitry Vengertsev and Edgar Schonfeld and Elliot Blanchard and Felix Juefei-Xu and Fraylie Nord and Jeff Liang and John Hoffman and Jonas Kohler and Kaolin Fire and Karthik Sivakumar and Lawrence Chen and Licheng Yu and Luya Gao and Markos Georgopoulos and Rashel Moritz and Sara K. Sampson and Shikai Li and Simone Parmeggiani and Steve Fine and Tara Fowler and Vladan Petrovic and Yuming Du},
      year={2025},
      eprint={2410.13720},
      archivePrefix={arXiv},
      primaryClass={cs.CV},
      url={https://arxiv.org/abs/2410.13720}, 
}

@misc{genie3,
  title         = {Genie 3: A New Frontier for World Models},
  author        = {Philip J. Ball and Jakob Bauer and Frank Belletti and Bethanie Brownfield and Ariel Ephrat and Shlomi Fruchter and Agrim Gupta and Kristian Holsheimer and Aleksander Holynski and Jiri Hron and Christos Kaplanis and Marjorie Limont and Matt McGill and Yanko Oliveira and Jack Parker-Holder and Frank Perbet and Guy Scully and Jeremy Shar and Stephen Spencer and Omer Tov and Ruben Villegas and Emma Wang and Jessica Yung and Cip Baetu and Jordi Berbel and David Bridson and Jake Bruce and Gavin Buttimore and Sarah Chakera and Bilva Chandra and Paul Collins and Alex Cullum and Bogdan Damoc and Vibha Dasagi and Maxime Gazeau and Charles Gbadamosi and Woohyun Han and Ed Hirst and Ashyana Kachra and Lucie Kerley and Kristian Kjems and Eva Knoepfel and Vika Koriakin and Jessica Lo and Cong Lu and Zeb Mehring and Alex Moufarek and Henna Nandwani and Valeria Oliveira and Fabio Pardo and Jane Park and Andrew Pierson and Ben Poole and Helen Ran and Tim Salimans and Manuel Sanchez and Igor Saprykin and Amy Shen and Sailesh Sidhwani and Duncan Smith and Joe Stanton and Hamish Tomlinson and Dimple Vijaykumar and Luyu Wang and Piers Wingfield and Nat Wong and Keyang Xu and Christopher Yew and Nick Young and Vadim Zubov and Douglas Eck and Dumitru Erhan and Koray Kavukcuoglu and Demis Hassabis and Zoubin Gharamani and Raia Hadsell and A{\"a}ron van den Oord and Inbar Mosseri and Adrian Bolton and Satinder Singh and Tim Rockt{\"a}schel},
  year          = {2025},
  url           = {}
}

@article{grattafiori2024llama,
  title={The llama 3 herd of models},
  author={Grattafiori, Aaron and Dubey, Abhimanyu and Jauhri, Abhinav and Pandey, Abhinav and Kadian, Abhishek and Al-Dahle, Ahmad and Letman, Aiesha and Mathur, Akhil and Schelten, Alan and Vaughan, Alex and others},
  journal={arXiv preprint arXiv:2407.21783},
  year={2024}
}

@article{liu2024deepseek,
  title={Deepseek-v3 technical report},
  author={Liu, Aixin and Feng, Bei and Xue, Bing and Wang, Bingxuan and Wu, Bochao and Lu, Chengda and Zhao, Chenggang and Deng, Chengqi and Zhang, Chenyu and Ruan, Chong and others},
  journal={arXiv preprint arXiv:2412.19437},
  year={2024}
}

@misc{yang2025qwen3technicalreport,
      title={Qwen3 Technical Report}, 
      author={An Yang and Anfeng Li and Baosong Yang and Beichen Zhang and Binyuan Hui and Bo Zheng and Bowen Yu and Chang Gao and Chengen Huang and Chenxu Lv and Chujie Zheng and Dayiheng Liu and Fan Zhou and Fei Huang and Feng Hu and Hao Ge and Haoran Wei and Huan Lin and Jialong Tang and Jian Yang and Jianhong Tu and Jianwei Zhang and Jianxin Yang and Jiaxi Yang and Jing Zhou and Jingren Zhou and Junyang Lin and Kai Dang and Keqin Bao and Kexin Yang and Le Yu and Lianghao Deng and Mei Li and Mingfeng Xue and Mingze Li and Pei Zhang and Peng Wang and Qin Zhu and Rui Men and Ruize Gao and Shixuan Liu and Shuang Luo and Tianhao Li and Tianyi Tang and Wenbiao Yin and Xingzhang Ren and Xinyu Wang and Xinyu Zhang and Xuancheng Ren and Yang Fan and Yang Su and Yichang Zhang and Yinger Zhang and Yu Wan and Yuqiong Liu and Zekun Wang and Zeyu Cui and Zhenru Zhang and Zhipeng Zhou and Zihan Qiu},
      year={2025},
      eprint={2505.09388},
      archivePrefix={arXiv},
      primaryClass={cs.CL},
      url={https://arxiv.org/abs/2505.09388}, 
}

@misc{zhao2023pytorchfsdpexperiencesscaling,
      title={PyTorch FSDP: Experiences on Scaling Fully Sharded Data Parallel}, 
      author={Yanli Zhao and Andrew Gu and Rohan Varma and Liang Luo and Chien-Chin Huang and Min Xu and Less Wright and Hamid Shojanazeri and Myle Ott and Sam Shleifer and Alban Desmaison and Can Balioglu and Pritam Damania and Bernard Nguyen and Geeta Chauhan and Yuchen Hao and Ajit Mathews and Shen Li},
      year={2023},
      eprint={2304.11277},
      archivePrefix={arXiv},
      primaryClass={cs.DC},
      url={https://arxiv.org/abs/2304.11277}, 
}

@misc{meng2023distillationguideddiffusionmodels,
      title={On Distillation of Guided Diffusion Models}, 
      author={Chenlin Meng and Robin Rombach and Ruiqi Gao and Diederik P. Kingma and Stefano Ermon and Jonathan Ho and Tim Salimans},
      year={2023},
      eprint={2210.03142},
      archivePrefix={arXiv},
      primaryClass={cs.CV},
      url={https://arxiv.org/abs/2210.03142}, 
}

@misc{ren2024hypersdtrajectorysegmentedconsistency,
      title={Hyper-SD: Trajectory Segmented Consistency Model for Efficient Image Synthesis}, 
      author={Yuxi Ren and Xin Xia and Yanzuo Lu and Jiacheng Zhang and Jie Wu and Pan Xie and Xing Wang and Xuefeng Xiao},
      year={2024},
      eprint={2404.13686},
      archivePrefix={arXiv},
      primaryClass={cs.CV},
      url={https://arxiv.org/abs/2404.13686}, 
}

@misc{seawead2025seaweed7bcosteffectivetrainingvideo,
      title={Seaweed-7B: Cost-Effective Training of Video Generation Foundation Model}, 
      author={Team Seawead and Ceyuan Yang and Zhijie Lin and Yang Zhao and Shanchuan Lin and Zhibei Ma and Haoyuan Guo and Hao Chen and Lu Qi and Sen Wang and Feng Cheng and Feilong Zuo and Xuejiao Zeng and Ziyan Yang and Fangyuan Kong and Meng Wei and Zhiwu Qing and Fei Xiao and Tuyen Hoang and Siyu Zhang and Peihao Zhu and Qi Zhao and Jiangqiao Yan and Liangke Gui and Sheng Bi and Jiashi Li and Yuxi Ren and Rui Wang and Huixia Li and Xuefeng Xiao and Shu Liu and Feng Ling and Heng Zhang and Houmin Wei and Huafeng Kuang and Jerry Duncan and Junda Zhang and Junru Zheng and Li Sun and Manlin Zhang and Renfei Sun and Xiaobin Zhuang and Xiaojie Li and Xin Xia and Xuyan Chi and Yanghua Peng and Yuping Wang and Yuxuan Wang and Zhongkai Zhao and Zhuo Chen and Zuquan Song and Zhenheng Yang and Jiashi Feng and Jianchao Yang and Lu Jiang},
      year={2025},
      eprint={2504.08685},
      archivePrefix={arXiv},
      primaryClass={cs.CV},
      url={https://arxiv.org/abs/2504.08685}, 
}

@misc{ramesh2021zeroshottexttoimagegeneration,
      title={Zero-Shot Text-to-Image Generation}, 
      author={Aditya Ramesh and Mikhail Pavlov and Gabriel Goh and Scott Gray and Chelsea Voss and Alec Radford and Mark Chen and Ilya Sutskever},
      year={2021},
      eprint={2102.12092},
      archivePrefix={arXiv},
      primaryClass={cs.CV},
      url={https://arxiv.org/abs/2102.12092}, 
}

@misc{xue2021mt5massivelymultilingualpretrained,
      title={mT5: A massively multilingual pre-trained text-to-text transformer}, 
      author={Linting Xue and Noah Constant and Adam Roberts and Mihir Kale and Rami Al-Rfou and Aditya Siddhant and Aditya Barua and Colin Raffel},
      year={2021},
      eprint={2010.11934},
      archivePrefix={arXiv},
      primaryClass={cs.CL},
      url={https://arxiv.org/abs/2010.11934}, 
}

@InProceedings{carlsson-EtAl:2022:LREC,
  author    = {Carlsson, Fredrik  and  Eisen, Philipp  and  Rekathati, Faton  and  Sahlgren, Magnus},
  title     = {Cross-lingual and Multilingual CLIP},
  booktitle      = {Proceedings of the Language Resources and Evaluation Conference},
  month          = {June},
  year           = {2022},
  address        = {Marseille, France},
  publisher      = {European Language Resources Association},
  pages     = {6848--6854},
  url       = {https://aclanthology.org/2022.lrec-1.739}
}

@misc{zheng2022movqmodulatingquantizedvectors,
      title={MoVQ: Modulating Quantized Vectors for High-Fidelity Image Generation}, 
      author={Chuanxia Zheng and Long Tung Vuong and Jianfei Cai and Dinh Phung},
      year={2022},
      eprint={2209.09002},
      archivePrefix={arXiv},
      primaryClass={cs.CV},
      url={https://arxiv.org/abs/2209.09002}, 
}

@misc{ramesh2022hierarchicaltextconditionalimagegeneration,
      title={Hierarchical Text-Conditional Image Generation with CLIP Latents}, 
      author={Aditya Ramesh and Prafulla Dhariwal and Alex Nichol and Casey Chu and Mark Chen},
      year={2022},
      eprint={2204.06125},
      archivePrefix={arXiv},
      primaryClass={cs.CV},
      url={https://arxiv.org/abs/2204.06125}, 
}

@misc{kang2023scalingganstexttoimagesynthesis,
      title={Scaling up GANs for Text-to-Image Synthesis}, 
      author={Minguk Kang and Jun-Yan Zhu and Richard Zhang and Jaesik Park and Eli Shechtman and Sylvain Paris and Taesung Park},
      year={2023},
      eprint={2303.05511},
      archivePrefix={arXiv},
      primaryClass={cs.CV},
      url={https://arxiv.org/abs/2303.05511}, 
}

@misc{heusel2018ganstrainedtimescaleupdate,
      title={GANs Trained by a Two Time-Scale Update Rule Converge to a Local Nash Equilibrium}, 
      author={Martin Heusel and Hubert Ramsauer and Thomas Unterthiner and Bernhard Nessler and Sepp Hochreiter},
      year={2018},
      eprint={1706.08500},
      archivePrefix={arXiv},
      primaryClass={cs.LG},
      url={https://arxiv.org/abs/1706.08500}, 
}

@article{li2024recapdatacomp,
  title={What If We Recaption Billions of Web Images with LLaMA-3?},
  author={Li, Xianhang and Tu, Haoqin and Hui, Mude and Wang, Zeyu and Zhao, Bingchen and Xiao, Junfei and Ren, Sucheng and Mei, Jieru and Liu, Qing and Zheng, Huangjie and Zhou, Yuyin and Xie, Cihang},
  journal={arXiv preprint arXiv:2406.12345},
  year={2024}
}

@misc{zhang2023adding,
  title={Adding Conditional Control to Text-to-Image Diffusion Models}, 
  author={Lvmin Zhang and Anyi Rao and Maneesh Agrawala},
  booktitle={IEEE International Conference on Computer Vision (ICCV)},
  year={2023},
}

@misc{sauer2023adversarialdiffusiondistillation,
      title={Adversarial Diffusion Distillation}, 
      author={Axel Sauer and Dominik Lorenz and Andreas Blattmann and Robin Rombach},
      year={2023},
      eprint={2311.17042},
      archivePrefix={arXiv},
      primaryClass={cs.CV},
      url={https://arxiv.org/abs/2311.17042}, 
}

@article{Vasilev_2024,
   title={CRAFT: Cultural Russian-Oriented Dataset Adaptation for Focused Text-to-Image Generation},
   volume={110},
   ISSN={1531-8362},
   url={http://dx.doi.org/10.1134/S1064562424602324},
   DOI={10.1134/s1064562424602324},
   number={S1},
   journal={Doklady Mathematics},
   publisher={Pleiades Publishing Ltd},
   author={Vasilev, V. A. and Arkhipkin, V. S. and Agafonova, J. D. and Nikulina, T. V. and Mironova, E. O. and Shichanina, A. A. and Gerasimenko, N. A. and Shoytov, M. A. and Dimitrov, D. V.},
   year={2024},
   month=dec, pages={S137–S150} 
}

@inproceedings{vasilev-etal-2025-ruscode,
    title = "{R}us{C}ode: {R}ussian Cultural Code Benchmark for Text-to-Image Generation",
    author = "Vasilev, Viacheslav  and
      Agafonova, Julia  and
      Gerasimenko, Nikolai  and
      Kapitanov, Alexander  and
      Mikhailova, Polina  and
      Mironova, Evelina  and
      Dimitrov, Denis",
    editor = "Chiruzzo, Luis  and
      Ritter, Alan  and
      Wang, Lu",
    booktitle = "Findings of the Association for Computational Linguistics: NAACL 2025",
    month = apr,
    year = "2025",
    address = "Albuquerque, New Mexico",
    publisher = "Association for Computational Linguistics",
    url = "https://aclanthology.org/2025.findings-naacl.425/",
    doi = "10.18653/v1/2025.findings-naacl.425",
    pages = "7641--7657",
    ISBN = "979-8-89176-195-7"
}

@misc{liu2023visualinstructiontuning,
      title={Visual Instruction Tuning}, 
      author={Haotian Liu and Chunyuan Li and Qingyang Wu and Yong Jae Lee},
      year={2023},
      eprint={2304.08485},
      archivePrefix={arXiv},
      primaryClass={cs.CV},
      url={https://arxiv.org/abs/2304.08485}, 
}

@misc{lin2024opensoraplanopensourcelarge,
      title={Open-Sora Plan: Open-Source Large Video Generation Model}, 
      author={Bin Lin and Yunyang Ge and Xinhua Cheng and Zongjian Li and Bin Zhu and Shaodong Wang and Xianyi He and Yang Ye and Shenghai Yuan and Liuhan Chen and Tanghui Jia and Junwu Zhang and Zhenyu Tang and Yatian Pang and Bin She and Cen Yan and Zhiheng Hu and Xiaoyi Dong and Lin Chen and Zhang Pan and Xing Zhou and Shaoling Dong and Yonghong Tian and Li Yuan},
      year={2024},
      eprint={2412.00131},
      archivePrefix={arXiv},
      primaryClass={cs.CV},
      url={https://arxiv.org/abs/2412.00131}, 
}

@article{jin2024pyramidal,
  title={Pyramidal Flow Matching for Efficient Video Generative Modeling},
  author={Jin, Yang and Sun, Zhicheng and Li, Ningyuan and Xu, Kun and Xu, Kun and Jiang, Hao and Zhuang, Nan and Huang, Quzhe and Song, Yang and Mu, Yadong and Lin, Zhouchen},
  journal={arXiv preprint arXiv:2410.05954},
  year={2024}
}

@inproceedings{razzhigaev-etal-2023-kandinsky,
    title = "Kandinsky: An Improved Text-to-Image Synthesis with Image Prior and Latent Diffusion",
    author = "Razzhigaev, Anton  and
      Shakhmatov, Arseniy  and
      Maltseva, Anastasia  and
      Arkhipkin, Vladimir  and
      Pavlov, Igor  and
      Ryabov, Ilya  and
      Kuts, Angelina  and
      Panchenko, Alexander  and
      Kuznetsov, Andrey  and
      Dimitrov, Denis",
    editor = "Feng, Yansong  and
      Lefever, Els",
    booktitle = "Proceedings of the 2023 Conference on Empirical Methods in Natural Language Processing: System Demonstrations",
    month = dec,
    year = "2023",
    address = "Singapore",
    publisher = "Association for Computational Linguistics",
    url = "https://aclanthology.org/2023.emnlp-demo.25/",
    doi = "10.18653/v1/2023.emnlp-demo.25",
    pages = "286--295"
}

@misc{arkhipkin2024kandinsky30technicalreport,
      title={Kandinsky 3.0 Technical Report}, 
      author={Vladimir Arkhipkin and Andrei Filatov and Viacheslav Vasilev and Anastasia Maltseva and Said Azizov and Igor Pavlov and Julia Agafonova and Andrey Kuznetsov and Denis Dimitrov},
      year={2024},
      eprint={2312.03511},
      archivePrefix={arXiv},
      primaryClass={cs.CV},
      url={https://arxiv.org/abs/2312.03511}, 
}

@misc{bai2025qwen25vltechnicalreport,
      title={Qwen2.5-VL Technical Report}, 
      author={Shuai Bai and Keqin Chen and Xuejing Liu and Jialin Wang and Wenbin Ge and Sibo Song and Kai Dang and Peng Wang and Shijie Wang and Jun Tang and Humen Zhong and Yuanzhi Zhu and Mingkun Yang and Zhaohai Li and Jianqiang Wan and Pengfei Wang and Wei Ding and Zheren Fu and Yiheng Xu and Jiabo Ye and Xi Zhang and Tianbao Xie and Zesen Cheng and Hang Zhang and Zhibo Yang and Haiyang Xu and Junyang Lin},
      year={2025},
      eprint={2502.13923},
      archivePrefix={arXiv},
      primaryClass={cs.CV},
      url={https://arxiv.org/abs/2502.13923}, 
}

@article{ba2016layer,
  title={Layer normalization},
  author={Ba, Jimmy Lei and Kiros, Jamie Ryan and Hinton, Geoffrey E},
  journal={arXiv preprint arXiv:1607.06450},
  year={2016}
}

@misc{su2023roformerenhancedtransformerrotary,
      title={RoFormer: Enhanced Transformer with Rotary Position Embedding}, 
      author={Jianlin Su and Yu Lu and Shengfeng Pan and Ahmed Murtadha and Bo Wen and Yunfeng Liu},
      year={2023},
      eprint={2104.09864},
      archivePrefix={arXiv},
      primaryClass={cs.CL},
      url={https://arxiv.org/abs/2104.09864}, 
}

@inproceedings{10.5555/3737916.3741676,
author = {Ma, Bingqi and Zong, Zhuofan and Song, Guanglu and Li, Hongsheng and Liu, Yu},
title = {Exploring the role of large language models in prompt encoding for diffusion models},
year = {2025},
isbn = {9798331314385},
publisher = {Curran Associates Inc.},
address = {Red Hook, NY, USA},
booktitle = {Proceedings of the 38th International Conference on Neural Information Processing Systems},
articleno = {3760},
numpages = {28},
location = {Vancouver, BC, Canada},
series = {NIPS '24}
}

@misc{lin2021catcrossattentionvision,
      title={CAT: Cross Attention in Vision Transformer}, 
      author={Hezheng Lin and Xing Cheng and Xiangyu Wu and Fan Yang and Dong Shen and Zhongyuan Wang and Qing Song and Wei Yuan},
      year={2021},
      eprint={2106.05786},
      archivePrefix={arXiv},
      primaryClass={cs.CV},
      url={https://arxiv.org/abs/2106.05786}, 
}

@misc{vaswani2023attentionneed,
      title={Attention Is All You Need}, 
      author={Ashish Vaswani and Noam Shazeer and Niki Parmar and Jakob Uszkoreit and Llion Jones and Aidan N. Gomez and Lukasz Kaiser and Illia Polosukhin},
      year={2023},
      eprint={1706.03762},
      archivePrefix={arXiv},
      primaryClass={cs.CL},
      url={https://arxiv.org/abs/1706.03762}, 
}

@misc{shah2024flashattention3fastaccurateattention,
      title={FlashAttention-3: Fast and Accurate Attention with Asynchrony and Low-precision}, 
      author={Jay Shah and Ganesh Bikshandi and Ying Zhang and Vijay Thakkar and Pradeep Ramani and Tri Dao},
      year={2024},
      eprint={2407.08608},
      archivePrefix={arXiv},
      primaryClass={cs.LG},
      url={https://arxiv.org/abs/2407.08608}, 
}

@inproceedings{zhang2024sageattention2,
  title={Sageattention2: Efficient attention with thorough outlier smoothing and per-thread int4 quantization},
  author={Zhang, Jintao and Huang, Haofeng and Zhang, Pengle and Wei, Jia and Zhu, Jun and Chen, Jianfei},
  booktitle={International Conference on Machine Learning (ICML)},
  year={2025}
}

@misc{ma2025magcachefastvideogeneration,
      title={MagCache: Fast Video Generation with Magnitude-Aware Cache}, 
      author={Zehong Ma and Longhui Wei and Feng Wang and Shiliang Zhang and Qi Tian},
      year={2025},
      eprint={2506.09045},
      archivePrefix={arXiv},
      primaryClass={cs.CV},
      url={https://arxiv.org/abs/2506.09045}, 
}

@misc{chen2025skyreelsv2infinitelengthfilmgenerative,
      title={SkyReels-V2: Infinite-length Film Generative Model}, 
      author={Guibin Chen and Dixuan Lin and Jiangping Yang and Chunze Lin and Junchen Zhu and Mingyuan Fan and Hao Zhang and Sheng Chen and Zheng Chen and Chengcheng Ma and Weiming Xiong and Wei Wang and Nuo Pang and Kang Kang and Zhiheng Xu and Yuzhe Jin and Yupeng Liang and Yubing Song and Peng Zhao and Boyuan Xu and Di Qiu and Debang Li and Zhengcong Fei and Yang Li and Yahui Zhou},
      year={2025},
      eprint={2504.13074},
      archivePrefix={arXiv},
      primaryClass={cs.CV},
      url={https://arxiv.org/abs/2504.13074}, 
}

@misc{sauer2024fasthighresolutionimagesynthesis,
      title={Fast High-Resolution Image Synthesis with Latent Adversarial Diffusion Distillation}, 
      author={Axel Sauer and Frederic Boesel and Tim Dockhorn and Andreas Blattmann and Patrick Esser and Robin Rombach},
      year={2024},
      eprint={2403.12015},
      archivePrefix={arXiv},
      primaryClass={cs.CV},
      url={https://arxiv.org/abs/2403.12015}, 
}

@misc{lin2025diffusionadversarialposttrainingonestep,
      title={Diffusion Adversarial Post-Training for One-Step Video Generation}, 
      author={Shanchuan Lin and Xin Xia and Yuxi Ren and Ceyuan Yang and Xuefeng Xiao and Lu Jiang},
      year={2025},
      eprint={2501.08316},
      archivePrefix={arXiv},
      primaryClass={cs.CV},
      url={https://arxiv.org/abs/2501.08316}, 
}

@misc{biggs2024diffusionsoupmodelmerging,
      title={Diffusion Soup: Model Merging for Text-to-Image Diffusion Models}, 
      author={Benjamin Biggs and Arjun Seshadri and Yang Zou and Achin Jain and Aditya Golatkar and Yusheng Xie and Alessandro Achille and Ashwin Swaminathan and Stefano Soatto},
      year={2024},
      eprint={2406.08431},
      archivePrefix={arXiv},
      primaryClass={cs.CV},
      url={https://arxiv.org/abs/2406.08431}, 
}

@misc{rmsprop,
    author = {Geoffrey Hinton},
    title = {Coursera Neural Networks for Machine Learning Course, Lecture 6e},
    year = {2018}
}

@misc{wu2025rewarddancerewardscalingvisual,
      title={RewardDance: Reward Scaling in Visual Generation}, 
      author={Jie Wu and Yu Gao and Zilyu Ye and Ming Li and Liang Li and Hanzhong Guo and Jie Liu and Zeyue Xue and Xiaoxia Hou and Wei Liu and Yan Zeng and Weilin Huang},
      year={2025},
      eprint={2509.08826},
      archivePrefix={arXiv},
      primaryClass={cs.CV},
      url={https://arxiv.org/abs/2509.08826}, 
}

@misc{clark2024directlyfinetuningdiffusionmodels,
      title={Directly Fine-Tuning Diffusion Models on Differentiable Rewards}, 
      author={Kevin Clark and Paul Vicol and Kevin Swersky and David J Fleet},
      year={2024},
      eprint={2309.17400},
      archivePrefix={arXiv},
      primaryClass={cs.CV},
      url={https://arxiv.org/abs/2309.17400}, 
}

@misc{xu2023imagerewardlearningevaluatinghuman,
      title={ImageReward: Learning and Evaluating Human Preferences for Text-to-Image Generation}, 
      author={Jiazheng Xu and Xiao Liu and Yuchen Wu and Yuxuan Tong and Qinkai Li and Ming Ding and Jie Tang and Yuxiao Dong},
      year={2023},
      eprint={2304.05977},
      archivePrefix={arXiv},
      primaryClass={cs.CV},
      url={https://arxiv.org/abs/2304.05977}, 
}

@misc{liu2025improvingvideogenerationhuman,
      title={Improving Video Generation with Human Feedback}, 
      author={Jie Liu and Gongye Liu and Jiajun Liang and Ziyang Yuan and Xiaokun Liu and Mingwu Zheng and Xiele Wu and Qiulin Wang and Menghan Xia and Xintao Wang and Xiaohong Liu and Fei Yang and Pengfei Wan and Di Zhang and Kun Gai and Yujiu Yang and Wanli Ouyang},
      year={2025},
      eprint={2501.13918},
      archivePrefix={arXiv},
      primaryClass={cs.CV},
      url={https://arxiv.org/abs/2501.13918}, 
}

@misc{christiano2023deepreinforcementlearninghuman,
      title={Deep reinforcement learning from human preferences}, 
      author={Paul Christiano and Jan Leike and Tom B. Brown and Miljan Martic and Shane Legg and Dario Amodei},
      year={2023},
      eprint={1706.03741},
      archivePrefix={arXiv},
      primaryClass={stat.ML},
      url={https://arxiv.org/abs/1706.03741}, 
}

@article{hpsv2,
  title={Human Preference Score v2: A Solid Benchmark for Evaluating Human Preferences of Text-to-Image Synthesis},
  author={Wu, Xiaoshi and Hao, Yiming and Sun, Keqiang and Chen, Yixiong and Zhu, Feng and Zhao, Rui and Li, Hongsheng},
  journal={arXiv preprint arXiv:2306.09341},
  year={2023}
}

@misc{wang2025unifiedrewardmodelmultimodal,
      title={Unified Reward Model for Multimodal Understanding and Generation}, 
      author={Yibin Wang and Yuhang Zang and Hao Li and Cheng Jin and Jiaqi Wang},
      year={2025},
      eprint={2503.05236},
      archivePrefix={arXiv},
      primaryClass={cs.CV},
      url={https://arxiv.org/abs/2503.05236}, 
}

@misc{ma2025hpsv3widespectrumhumanpreference,
      title={HPSv3: Towards Wide-Spectrum Human Preference Score}, 
      author={Yuhang Ma and Yunhao Shui and Xiaoshi Wu and Keqiang Sun and Hongsheng Li},
      year={2025},
      eprint={2508.03789},
      archivePrefix={arXiv},
      primaryClass={cs.CV},
      url={https://arxiv.org/abs/2508.03789}, 
}

@misc{wallace2023diffusionmodelalignmentusing,
      title={Diffusion Model Alignment Using Direct Preference Optimization}, 
      author={Bram Wallace and Meihua Dang and Rafael Rafailov and Linqi Zhou and Aaron Lou and Senthil Purushwalkam and Stefano Ermon and Caiming Xiong and Shafiq Joty and Nikhil Naik},
      year={2023},
      eprint={2311.12908},
      archivePrefix={arXiv},
      primaryClass={cs.CV},
      url={https://arxiv.org/abs/2311.12908}, 
}

@misc{liu2025flowgrpotrainingflowmatching,
      title={Flow-GRPO: Training Flow Matching Models via Online RL}, 
      author={Jie Liu and Gongye Liu and Jiajun Liang and Yangguang Li and Jiaheng Liu and Xintao Wang and Pengfei Wan and Di Zhang and Wanli Ouyang},
      year={2025},
      eprint={2505.05470},
      archivePrefix={arXiv},
      primaryClass={cs.CV},
      url={https://arxiv.org/abs/2505.05470}, 
}

@misc{black2024trainingdiffusionmodelsreinforcement,
      title={Training Diffusion Models with Reinforcement Learning}, 
      author={Kevin Black and Michael Janner and Yilun Du and Ilya Kostrikov and Sergey Levine},
      year={2024},
      eprint={2305.13301},
      archivePrefix={arXiv},
      primaryClass={cs.LG},
      url={https://arxiv.org/abs/2305.13301}, 
}

@misc{wang2025prefgrpopairwisepreferencerewardbased,
      title={Pref-GRPO: Pairwise Preference Reward-based GRPO for Stable Text-to-Image Reinforcement Learning}, 
      author={Yibin Wang and Zhimin Li and Yuhang Zang and Yujie Zhou and Jiazi Bu and Chunyu Wang and Qinglin Lu and Cheng Jin and Jiaqi Wang},
      year={2025},
      eprint={2508.20751},
      archivePrefix={arXiv},
      primaryClass={cs.CV},
      url={https://arxiv.org/abs/2508.20751}, 
}

@misc{loshchilov2019decoupledweightdecayregularization,
      title={Decoupled Weight Decay Regularization}, 
      author={Ilya Loshchilov and Frank Hutter},
      year={2019},
      eprint={1711.05101},
      archivePrefix={arXiv},
      primaryClass={cs.LG},
      url={https://arxiv.org/abs/1711.05101}, 
}

@inproceedings{salimans2022progressive,
    title={Progressive Distillation for Fast Sampling of Diffusion Models},
    author={Tim Salimans and Jonathan Ho},
    booktitle={International Conference on Learning Representations},
    year={2022},
    url={https://openreview.net/forum?id=TIdIXIpzhoI}
}

@misc{liu2022flowstraightfastlearning,
      title={Flow Straight and Fast: Learning to Generate and Transfer Data with Rectified Flow}, 
      author={Xingchao Liu and Chengyue Gong and Qiang Liu},
      year={2022},
      eprint={2209.03003},
      archivePrefix={arXiv},
      primaryClass={cs.LG},
      url={https://arxiv.org/abs/2209.03003}, 
}

@inproceedings{song2023consistency,
    author = {Song, Yang and Dhariwal, Prafulla and Chen, Mark and Sutskever, Ilya},
    title = {Consistency models},
    year = {2023},
    publisher = {JMLR.org},
    booktitle = {Proceedings of the 40th International Conference on Machine Learning},
    articleno = {1335},
    numpages = {42},
    location = {Honolulu, Hawaii, USA},
    series = {ICML'23}
}

@misc{lin2024sdxl,
      title={SDXL-Lightning: Progressive Adversarial Diffusion Distillation}, 
      author={Shanchuan Lin and Anran Wang and Xiao Yang},
      year={2024},
      eprint={2402.13929},
      archivePrefix={arXiv},
      primaryClass={cs.CV},
      url={https://arxiv.org/abs/2402.13929}, 
}

@misc{yin2024onestep,
      title={One-step Diffusion with Distribution Matching Distillation}, 
      author={Tianwei Yin and Michaël Gharbi and Richard Zhang and Eli Shechtman and Fredo Durand and William T. Freeman and Taesung Park},
      year={2024},
      eprint={2311.18828},
      archivePrefix={arXiv},
      primaryClass={cs.CV},
      url={https://arxiv.org/abs/2311.18828}, 
}

@inproceedings{song2023improved,
    title={Improved Techniques for Training Consistency Models},
    author={Yang Song and Prafulla Dhariwal},
    booktitle={The Twelfth International Conference on Learning Representations},
    year={2024},
    url={https://openreview.net/forum?id=WNzy9bRDvG}
}

@misc{luo2023latent,
      title={Latent Consistency Models: Synthesizing High-Resolution Images with Few-Step Inference}, 
      author={Simian Luo and Yiqin Tan and Longbo Huang and Jian Li and Hang Zhao},
      year={2023},
      eprint={2310.04378},
      archivePrefix={arXiv},
      primaryClass={cs.CV},
      url={https://arxiv.org/abs/2310.04378}, 
}

@misc{heek2024multistep,
      title={Multistep Consistency Models}, 
      author={Jonathan Heek and Emiel Hoogeboom and Tim Salimans},
      year={2024},
      eprint={2403.06807},
      archivePrefix={arXiv},
      primaryClass={cs.LG},
      url={https://arxiv.org/abs/2403.06807}, 
}

@inproceedings{yin2024improved,
      title={Improved Distribution Matching Distillation for Fast Image Synthesis},
      author={Yin, Tianwei and Gharbi, Micha{\"e}l and Park, Taesung and Zhang, Richard and Shechtman, Eli and Durand, Fredo and Freeman, William T},
      booktitle={NeurIPS},
      year={2024}
}

@inproceedings{liu2023instaflow,
  title={Instaflow: One step is enough for high-quality diffusion-based text-to-image generation},
  author={Liu, Xingchao and Zhang, Xiwen and Ma, Jianzhu and Peng, Jian and Liu, Qiang},
  booktitle={International Conference on Learning Representations},
  year={2024}
}

@misc{luhman2021knowledge,
      title={Knowledge Distillation in Iterative Generative Models for Improved Sampling Speed}, 
      author={Eric Luhman and Troy Luhman},
      year={2021},
      eprint={2101.02388},
      archivePrefix={arXiv},
      primaryClass={cs.LG},
      url={https://arxiv.org/abs/2101.02388}, 
}

@inproceedings{
            zhang2024sfv,
            title={{SF}-V: Single Forward Video Generation Model},
            author={Zhixing Zhang and Yanyu Li and Yushu Wu and yanwu xu and Anil Kag and Ivan Skorokhodov and Willi Menapace and Aliaksandr Siarohin and Junli Cao and Dimitris N. Metaxas and Sergey Tulyakov and Jian Ren},
            booktitle={The Thirty-eighth Annual Conference on Neural Information Processing Systems},
            year={2024},
            url={https://openreview.net/forum?id=PVgAeMm3MW}
          }

@INPROCEEDINGS{mao2024osv,
  author={Mao, Xiaofeng and Jiang, Zhengkai and Wang, Fu-yun and Zhang, Jiangning and Chen, Hao and Chi, Mingmin and Wang, Yabiao and Luo, Wenhan},
  booktitle={2025 IEEE/CVF Conference on Computer Vision and Pattern Recognition (CVPR)}, 
  title={OSV: One Step is Enough for High-Quality Image to Video Generation}, 
  year={2025},
  volume={},
  number={},
  pages={12585-12594},
  doi={10.1109/CVPR52734.2025.01174}
}

@INPROCEEDINGS{yin2024causal,
  author={Yin, Tianwei and Zhang, Qiang and Zhang, Richard and Freeman, William T. and Durand, Frédo and Shechtman, Eli and Huang, Xun},
  booktitle={2025 IEEE/CVF Conference on Computer Vision and Pattern Recognition (CVPR)}, 
  title={From Slow Bidirectional to Fast Autoregressive Video Diffusion Models}, 
  year={2025},
  volume={},
  number={},
  pages={22963-22974},
  doi={10.1109/CVPR52734.2025.02138}
}

@article{zhao2023pytorch,
  title={Pytorch fsdp: experiences on scaling fully sharded data parallel},
  author={Zhao, Yanli and Gu, Andrew and Varma, Rohan and Luo, Liang and Huang, Chien-Chin and Xu, Min and Wright, Less and Shojanazeri, Hamid and Ott, Myle and Shleifer, Sam and others},
  journal={arXiv preprint arXiv:2304.11277},
  year={2023}
}

@misc{korthikanti2205reducing,
  title={Reducing activation recomputation in large transformer models, 2022},
  author={Korthikanti, Vijay and Casper, Jared and Lym, Sangkug and McAfee, Lawrence and Andersch, Michael and Shoeybi, Mohammad and Catanzaro, Bryan},
  journal={URL https://arxiv. org/abs/2205.05198}
}

@article{chen2016training,
  title={Training deep nets with sublinear memory cost, 2016},
  author={Chen, Tianqi and Xu, Bing and Zhang, Chiyuan and Guestrin, Carlos},
  journal={arXiv preprint arXiv:1604.06174},
  year={2016}
}

@misc{novitskiy2025vivatvirtuousimprovingvae,
      title={VIVAT: Virtuous Improving VAE Training through Artifact Mitigation}, 
      author={Lev Novitskiy and Viacheslav Vasilev and Maria Kovaleva and Vladimir Arkhipkin and Denis Dimitrov},
      year={2025},
      eprint={2506.07863},
      archivePrefix={arXiv},
      primaryClass={cs.CV},
      url={https://arxiv.org/abs/2506.07863}, 
}

@misc{vasilev2026timecorrelatedvideobridgematching,
      title={Time-Correlated Video Bridge Matching}, 
      author={Viacheslav Vasilev and Arseny Ivanov and Nikita Gushchin and Maria Kovaleva and Alexander Korotin},
      year={2026},
      eprint={2510.12453},
      archivePrefix={arXiv},
      primaryClass={cs.LG},
      url={https://arxiv.org/abs/2510.12453}, 
}

\end{document}